\documentclass[10pt,twocolumn,letterpaper]{article}

\usepackage[pagenumbers]{cvpr} %

\def\eg{\emph{e.g.}\@\xspace} 

\def\ie{\emph{i.e.}\@\xspace}

\def\etc{\emph{etc.}\@\xspace} 
\def\vs{\emph{vs.}\@\xspace}

\usepackage[table]{xcolor}
\usepackage{colortbl}
\usepackage{booktabs}
\usepackage{float}
\usepackage{multirow}

\usepackage{stfloats}

\usepackage{pifont}
\usepackage[toc,page,header]{appendix}

\newcommand{\name}{{LitePT}}
\newcommand{\rope}{{PointROPE}}

\newcommand{\circlenumblack}[1]{\textcolor{black}{\textcircled{\scriptsize{#1}}}}

\definecolor{xcpe}{RGB}{86, 180, 233}
\definecolor{transformer}{RGB}{213, 94, 0}

\definecolor{waymoDetVehicle}{RGB}{185,203,242}
\definecolor{waymoDetPedestrian}{RGB}{230, 25, 75}
\definecolor{waymoDetCyclist}{RGB}{245, 158, 37}

\definecolor{scannet_wall}{RGB}{166,192,226}
\definecolor{scannet_floor}{RGB}{151,213,135}
\definecolor{scannet_cabinet}{RGB}{42,108,168}
\definecolor{scannet_bed}{RGB}{249,180,117}
\definecolor{scannet_chair}{RGB}{181,179,57}
\definecolor{scannet_sofa}{RGB}{125,79,67}
\definecolor{scannet_table}{RGB}{247,148,141}
\definecolor{scannet_door}{RGB}{199,53,41}
\definecolor{scannet_window}{RGB}{186,169,206}
\definecolor{scannet_bookshelf}{RGB}{133,96,176}
\definecolor{scannet_picture}{RGB}{185,147,138}
\definecolor{scannet_counter}{RGB}{58,180,198}
\definecolor{scannet_desk}{RGB}{240,176,201}
\definecolor{scannet_curtain}{RGB}{214,212,137}
\definecolor{scannet_refrigerator}{RGB}{246,124,41}
\definecolor{scannet_shower}{RGB}{154,210,225}
\definecolor{scannet_toilet}{RGB}{59,146,54}
\definecolor{scannet_sink}{RGB}{102,116,131}
\definecolor{scannet_bathtub}{RGB}{215,117,182}
\definecolor{scannet_otherfurniture}{RGB}{72,77,148}
\definecolor{scannet_unlabeled}{RGB}{0,0,0}

\definecolor{stru3d_wall}{RGB}{166,192,226}
\definecolor{stru3d_floor}{RGB}{151,213,135}
\definecolor{stru3d_cabinet}{RGB}{42,108,168}
\definecolor{stru3d_bed}{RGB}{249,180,117}
\definecolor{stru3d_chair}{RGB}{181,179,57}
\definecolor{stru3d_sofa}{RGB}{125,79,67}
\definecolor{stru3d_table}{RGB}{247,148,141}
\definecolor{stru3d_door}{RGB}{199,53,41}
\definecolor{stru3d_window}{RGB}{186,169,206}
\definecolor{stru3d_picture}{RGB}{185,147,138}
\definecolor{stru3d_desk}{RGB}{240,176,201}
\definecolor{stru3d_shelves}{RGB}{133,96,176}
\definecolor{stru3d_curtain}{RGB}{214,212,137}
\definecolor{stru3d_dresser}{RGB}{58,180,198}
\definecolor{stru3d_pillow}{RGB}{255,83,177}
\definecolor{stru3d_mirror}{RGB}{58,146,53}
\definecolor{stru3d_ceiling}{RGB}{255,213,88}
\definecolor{stru3d_refrigerator}{RGB}{246,124,41}
\definecolor{stru3d_television}{RGB}{162,0,255}
\definecolor{stru3d_nightstand}{RGB}{0,131,255}
\definecolor{stru3d_sink}{RGB}{102,116,131}
\definecolor{stru3d_lamp}{RGB}{188,238,225}
\definecolor{stru3d_otherstructure}{RGB}{215,117,182}
\definecolor{stru3d_otherfurniture}{RGB}{72,77,148}
\definecolor{stru3d_otherprop}{RGB}{176,196,178}

\definecolor{nuscenes_barrier}{RGB}{128, 128, 0}
\definecolor{nuscenes_bicycle}{RGB}{245, 130, 48}
\definecolor{nuscenes_bus}{RGB}{250, 190, 190}
\definecolor{nuscenes_car}{RGB}{185,203,242}
\definecolor{nuscenes_construction_veh}{RGB}{0, 0, 128}
\definecolor{nuscenes_motorcycle}{RGB}{220, 190, 255}
\definecolor{nuscenes_pedestrian}{RGB}{230, 25, 75}
\definecolor{nuscenes_traffic_cone}{RGB}{0, 130, 200}
\definecolor{nuscenes_trailer}{RGB}{70, 240, 240}
\definecolor{nuscenes_truck}{RGB}{210, 245, 60}
\definecolor{nuscenes_driveable_surface}{RGB}{112,62,114}
\definecolor{nuscenes_other_flat}{RGB}{255, 250, 200}
\definecolor{nuscenes_sidewalk}{RGB}{228,64,219}
\definecolor{nuscenes_terrain}{RGB}{0, 128, 128}
\definecolor{nuscenes_manmade}{RGB}{201,156,0}
\definecolor{nuscenes_vegetation}{RGB}{60, 180, 75}

\definecolor{waymo_car}{RGB}{185,203,242}
\definecolor{waymo_truck}{RGB}{210, 245, 60}
\definecolor{waymo_bus}{RGB}{250, 190, 190}
\definecolor{waymo_other_veh}{RGB}{0, 0, 128}
\definecolor{waymo_motorcyclist}{RGB}{70, 240, 240}
\definecolor{waymo_bicyclist}{RGB}{221,128,112}
\definecolor{waymo_pedestrian}{RGB}{230, 25, 75}
\definecolor{waymo_sign}{RGB}{255,213,88}
\definecolor{waymo_traffic_light}{RGB}{0, 130, 200}
\definecolor{waymo_traffic_pole}{RGB}{166,0,255}
\definecolor{waymo_construction_cone}{RGB}{187,231,252}
\definecolor{waymo_bicycle}{RGB}{245, 130, 48}
\definecolor{waymo_motorcycle}{RGB}{220, 190, 255}
\definecolor{waymo_building}{RGB}{201,156,0}
\definecolor{waymo_vegetation}{RGB}{60, 180, 75}
\definecolor{waymo_tree_trunk}{RGB}{168,75,0}
\definecolor{waymo_curb}{RGB}{128, 128, 0}
\definecolor{waymo_road}{RGB}{185,140,176}
\definecolor{waymo_lane_marker}{RGB}{170,170,170}
\definecolor{waymo_other_ground}{RGB}{255, 250, 200}
\definecolor{waymo_walkable}{RGB}{94, 193, 185}
\definecolor{waymo_sidewalk}{RGB}{228,64,219}

\newcommand{\gapp}[1]{%
  \raisebox{1.5ex}{\scriptsize \textit{{\textcolor{OliveGreen}{#1}}}}%
}

\usepackage{siunitx}
\definecolor{cvprblue}{rgb}{0.21,0.49,0.74}
\usepackage[pagebackref,breaklinks,colorlinks,allcolors=cvprblue]{hyperref}
\usepackage[accsupp]{axessibility}
\usepackage{times}

\def\paperID{61} %
\def\confName{CVPR}
\def\confYear{2026}

\title{\name: Lighter Yet Stronger Point Transformer}

\author{
Yuanwen Yue\textsuperscript{1,2} \quad  Damien Robert\textsuperscript{3} \quad Jianyuan Wang\textsuperscript{2} \quad  Sunghwan Hong\textsuperscript{1} \quad  Jan Dirk Wegner\textsuperscript{3} \\  Christian Rupprecht\textsuperscript{2} \quad  Konrad Schindler\textsuperscript{1}
\vspace{5px}
 \\
\textsuperscript{1}ETH Zurich \quad \textsuperscript{2}University of Oxford \quad \textsuperscript{3}University of Zurich
}

\begin{document}

\twocolumn[{%
\renewcommand\twocolumn[1][]{#1}%
\maketitle
\vspace{-15px}
\includegraphics[width=.44\linewidth]{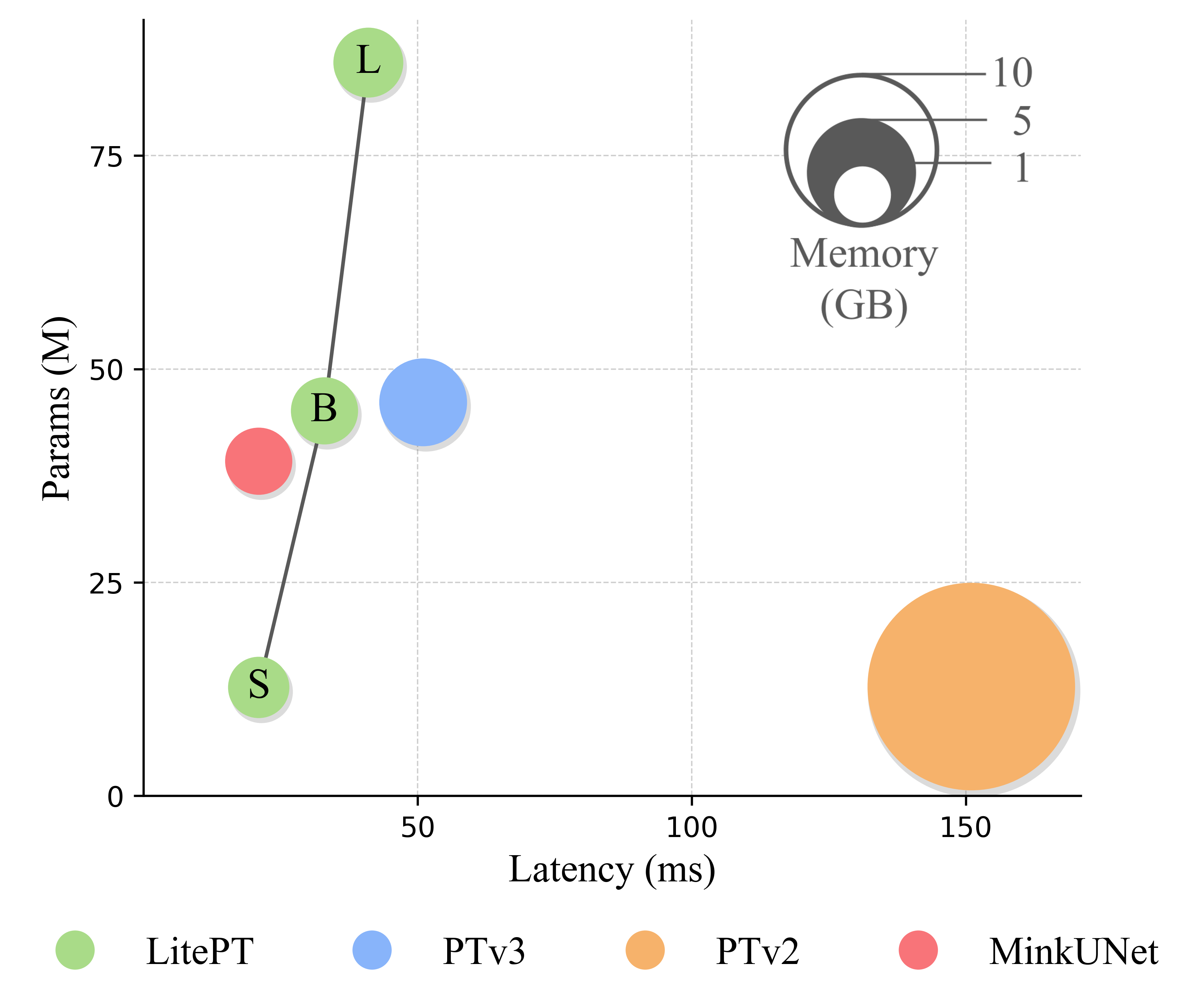}
\includegraphics[width=.49\linewidth]{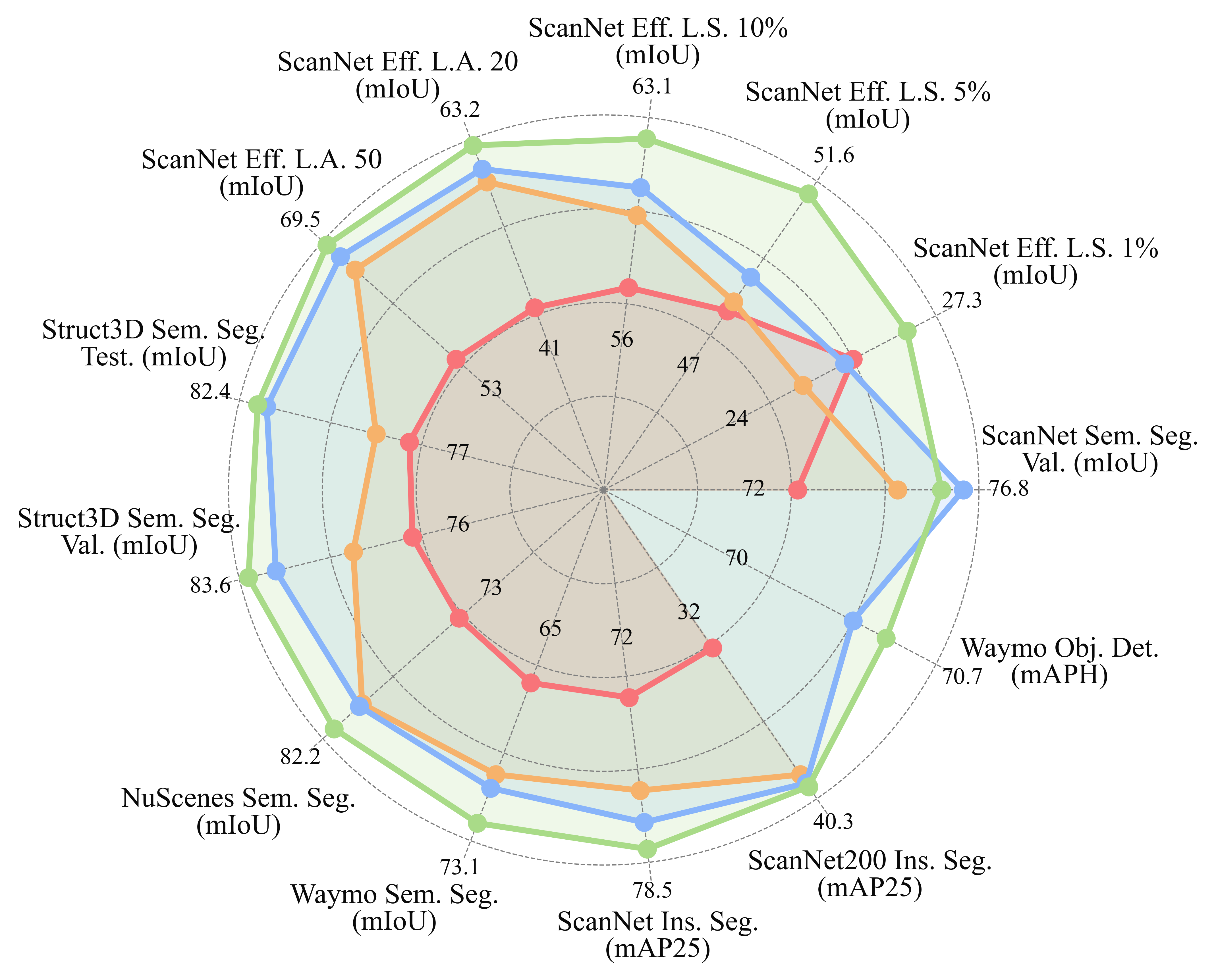}
\vspace{-0.5em}
\captionof{figure}{\textbf{\name{} is a lightweight, high-performance 3D point cloud architecture.} 
\underline{Left}: \name{}-S has $3.6\times$ fewer parameters, $2\times$ faster runtime and $2\times$ lower memory footprint than the state-of-the-art Point Transformer V3, and is even more memory-efficient than classical convolutional backbones. Moreover, it remains fast and memory-efficient even when scaled up to 86M parameters (\name{}-L). 
\underline{Right}: Already the smallest variant, \name{}-S, matches or outperforms state-of-the-art point cloud backbones across a range of benchmarks. \vspace{3em}}
\label{fig:teaser}
}]

\begin{abstract}
Modern neural architectures for 3D point cloud processing contain both convolutional layers and attention blocks, but the best way to assemble them remains unclear. 
We analyse the role of different computational blocks in 3D point cloud networks and find an intuitive behaviour: convolution is adequate to extract low-level geometry at high-resolution in early layers, where attention is expensive without bringing any benefits; attention captures high-level semantics and context in low-resolution, deep layers more efficiently, where convolution inflates the parameter count.
Guided by this design principle, we propose a new, improved 3D point cloud backbone that employs convolutions in early stages and switches to attention for deeper layers.
To avoid the loss of spatial layout information when discarding redundant convolution layers, we introduce a novel, parameter-free 3D positional encoding, \rope. 
The resulting \name{} model has $3.6\times$ fewer parameters, runs $2\times$ faster, and uses $2\times$ less memory
than the state-of-the-art Point Transformer V3, but nonetheless matches or outperforms it on a range of tasks and datasets. Code and models are available at: \href{https://github.com/prs-eth/LitePT}{https://github.com/prs-eth/LitePT}.
\end{abstract}
    
\section{Introduction}
\label{sec:intro}

Visual understanding of 3D point clouds is central to a wide range of applications, including
robotics~\cite{wurm2010occupancy, zhang2014loam, xu2015real, busch2025one},
autonomous driving~\cite{geiger2012kitti,sun2020scalability},
localisation~\cite{luo20243d},
mapping~\cite{pfaff2007towards, wang2018lidar, varney2020dales},
and 
environmental monitoring~\cite{iglhaut2019structure, song2025comprehensive}.
A variety of deep learning architectures and neural processing layers for unstructured point clouds have been proposed, yet the field still lacks a detailed understanding of their relative strengths and weaknesses, and principled guidelines on how to most efficiently combine them into versatile, high-performance architectures.

Lately, Transformer-based models have dominated 3D benchmarks. In particular, their most recent incarnation \emph{Point Transformer V3} (PTv3)~\cite{wu2024point} has been shown to outperform earlier sparse convolutional~\cite{graham20183d, choy20194d} and attention-based models~\cite{guo2021pct, zhao2021point, wu2022point}, and is considered the state of the art.
Importantly, PTv3 is in fact \emph{not} a pure Transformer architecture: $67\%$ of its parameters are allocated to (residual) sparse convolution layers. These are interleaved with the Transformer-style attention+MLP blocks and, among others, serve as a form of positional encoding.
That design, with both convolution and attention operations at all hierarchy levels (resp., depths) of a U-Net-like encoder-decoder scheme~\cite{ronneberger2015u}, is common in modern 3D point cloud architectures, which naturally leads to the question: \emph{what are the respective roles of convolution and attention?} 

Here, we analyse the contribution and interplay of these layers in more detail. We find a clear division of labour along the feature hierarchy.
Early, high-resolution stages are dominated by the encoding of \textit{local} geometry. Convolution or attention perform similarly well for that purpose, as the locality of convolutions is the right inductive bias. However, attention is substantially more expensive for early layers with high spatial resolutions (\ie, a large number of tokens).
Later, at lower-resolution stages, semantics and global context emerge. To capture the associated long-range interactions, the highly expressive attention mechanism is more suitable and also more parameter-efficient.
As mentioned, in PTv3 and related architectures, the SparseConv~\cite{graham20183d} layer was primarily included to encode positional information. It turns out that, for that particular purpose, convolution is a possible solution, but not a necessity. 
We find that a RoPE-inspired~\cite{su2024roformer} query-key positional encoding, which we call \rope, fulfils the role more effectively, while being more efficient and introducing no learnable parameters.
Overall, our analysis points to a clear design principle: apply convolution when the focus is on local geometry, and attention when reasoning about semantics and global layout.

Building on these insights, we design \name, a hybrid network architecture for 3D point cloud analysis that leverages the computational tools in the most efficient manner; \ie, sparse convolutions in the early stages and \rope-enhanced attention in the later stages.
By tailoring the information processing to the level of abstraction, \name{}  requires $3.6\times$ fewer parameters than PTv3.  Our architecture cuts memory consumption by 
$60.3\%$ during training and by $51.2\%$ during inference, and reduces latency by $34.5\%$ during training and by $58.8\%$ during inference.
Remarkably, \name{} also improves performance compared to PTv3 across a range of benchmarks on 3D semantic segmentation, 3D instance segmentation, and 3D object detection.

\section{Related Work}
\label{sec:related_work}

In line with the purpose of \name{}, we review deep learning-based point cloud representations, with a specific focus on Transformer architectures and hybrid approaches. 

\noindent\textbf{Deep Point Cloud Understanding.} 
To take advantage of mature image-based networks, early approaches used to project 3D point clouds into 2D image planes and then leverage standard 2D CNNs to extract features~\cite{su2015multi, chen2017multi, boulch2018snapnet, lang2019pointpillars, kalogerakis20173d, wu2018squeezeseg}. 
These projection-based methods tend to work well only when several implicit assumptions are met, \eg, relatively uniform point density, sufficient coverage, opaque surfaces, \etc 
Voxel-based methods transform irregular point clouds to regular voxel grids and then apply 3D convolution operations~\cite{maturana2015voxnet, song2017semantic, huang2016point, liu2019point, han2020occuseg}. 
However, voxel representations are both computationally expensive and memory-intensive, motivating follow-up works to develop efficient sparse convolution frameworks~\cite{graham20183d, choy20194d, tang2020searching, chen2023largekernel3d, peng2024oa}.
Instead of projecting or quantising irregular point clouds into regular grids in 2D or 3D, point-based methods design operators that work directly on raw point coordinates, better preserving geometric information.
Point operators have progressed from early MLP-based designs~\cite{qi2017pointnet,qi2017pointnet++,ma2022rethinking,qian2022pointnext,deng2024linnet,zeng2025deepla} to point convolutions~\cite{thomas2019kpconv,hua2018pointwise,xu2018spidercnn,atzmon2018point,groh2018flex,wu2019pointconv,li2018pointcnn}, graph-based networks~\cite{wang2019dynamic,landrieu2018large}, and, more recently, attention-based mechanisms~\cite{zhao2021point,guo2021pct,wu2022point,robert2023efficient,robert2024scalable,wu2024point,chen2025flash3d,gitmerge3d2025}. 
Among modern point cloud backbones, Transformer-based architectures represent the state of the art.

\noindent\textbf{Point Cloud Transformers.} Transformer-based architectures employ the attention mechanism as their core feature extractor. 
To mitigate the quadratic complexity of global self-attention, most approaches adopt some form of windowed attention, restricted to a local spatial neighbourhood. 
Point cloud Transformers mainly differ in how these localised attention patches are constructed to best balance performance and efficiency.
Common strategies include $k$-nearest neighbour search~\cite{zhao2021point,wu2022point,yang2022unified}, window or voxel partitioning~\cite{park2022fast,wang2023octformer,yang2025swin3d,yang2025swin3d++,liu2023flatformer,sun2022swformer,zhang2022patchformer,fan2022embracing}, superpoints~\cite{robert2023efficient, robert2024scalable}, and 1D sorting with space-filling curves~\cite{chen2022efficient,wu2024point}. 
Such local attention mechanisms are often integrated with shifted patch grouping~\cite{yang2025swin3d} and hierarchical architectures in the spirit of U-Net~\cite{ronneberger2015u}, so as to aggregate global context. 
Existing works typically apply attention at all stages of the hierarchical network. We argue that attention in shallow stages, where the number of tokens is large and local patterns dominate, is computationally inefficient and unnecessary, as seen in~\cref{sec:method_revisit_ptv3,ablation_studies}. 

\noindent\textbf{Positional Encoding in Point Cloud Transformers.} 
Attention does not take into account spatial layout; therefore, positional encoding plays an important role in Transformers. PTv1~\cite{zhao2021point} and PTv2~\cite{wu2022point} employ relative positional encoding (RPE), where an MLP encodes relative positions between points. 
Stratified Transformer~\cite{lai2022stratified} and Swin3D~\cite{yang2025swin3d} use contextual relative positional encoding (cRPE), which maintains three learnable look-up tables for the $(x,y,z)$ axes that are computationally rather inefficient. 
OctFormer~\cite{wang2023octformer} and PTv3~\cite{wu2024point} employ conditional positional encoding (CPE)~\cite{chu2021conditional}, which is implemented via a convolutional layer preceding each attention module. 
CPE improves efficiency, but introduces a substantial number of learnable parameters. 
Here, we adapt rotary positional embedding (RoPE)~\cite{su2024roformer} to point cloud learning, a parameter-free module that offers both efficiency and strong empirical performance.

\noindent\textbf{Hybrid Models.} 
Convolution is by design capable of capturing local features, whereas Transformers excel at modelling long-range dependencies. 
In the vision domain, since the introduction of the Vision Transformer~\cite{dosovitskiy2020image}, numerous studies have explored the integration of convolutional operators with attention for efficient image analysis~\cite{wu2021cvt,mehta2021mobilevit,tu2022maxvit,xu2021vitae,guo2022cmt,dai2021coatnet}. 
Similarly, in the 3D point cloud field, several works have investigated hybrid architectures that combine the strengths of convolution and attention. 
DyCo3D~\cite{he2021dyco3d} augments Sparse U-Net with a bottleneck Transformer to capture long-range context.
Stratified Transformer~\cite{lai2022stratified} reports that a KPConv~\cite{thomas2019kpconv} block provides substantially stronger local features than attention.
Superpoint Transformer~\cite{robert2023efficient} leverages a lightweight PointNet~\cite{qi2017pointnet} to encode geometrically-homogeneous superpoints.
PointConvFormer~\cite{wu2023pointconvformer} and KPConvX~\cite{thomas2024kpconvx} augment convolution kernels with attention to improve feature modelling. 
ConDaFormer~\cite{duan2023condaformer} adds two sparse convolution blocks before and after each attention module to better capture local structure. 
We note that PTv3~\cite{wu2024point} is also arguably a hybrid model, as it utilises sparse convolutions as positional encoding, which account for the majority of its trainable parameters. 
Another common design paradigm applies a sparse U-Net for feature extraction followed by a task-specific Transformer decoder~\cite{yang2023pvt,schult2023mask3d}.
In contrast to prior works, which typically employ a uniform hybrid block repeated across the hierarchy, we rethink hybrid design from a multi-scale perspective and decouple convolution and attention, allowing for the selective use of each at different hierarchy levels to exploit their complementary advantages.

\section{Methodology}
\label{sec:method}

\begin{figure}[h]
    \centering
    \includegraphics[width=0.8\linewidth]{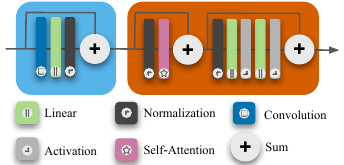}
    \vspace{-0.5em}
    \caption{\textbf{PTv3 block.} The block is composed of a convolutional conditional positional encoding module followed by an attention module.}\vspace{-5pt}
    \label{fig:ptv3_block}
\end{figure}

\begin{figure*}[h]
    \centering
    \begin{subfigure}{\textwidth}
        \centering
        \includegraphics[width=0.99\textwidth, trim={0 90 0 0}, clip]{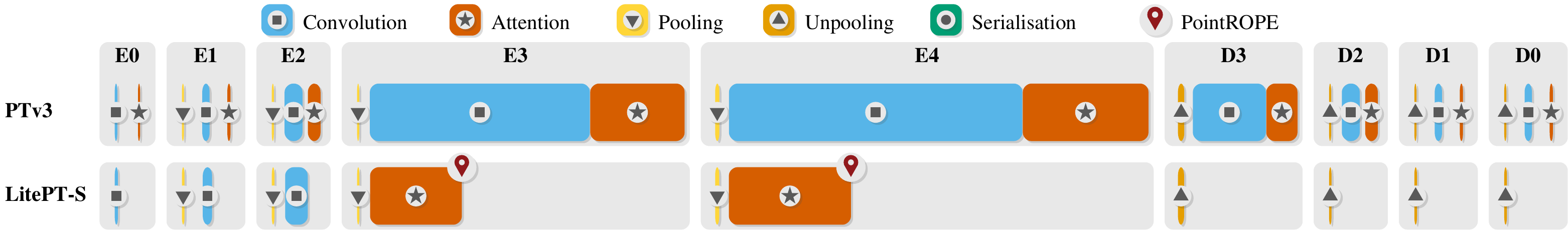}
        \caption{Breakdown of trainable parameters}
        \label{fig:parameter_treemap}
    \end{subfigure}

    \vspace{0.5em} %

    \begin{subfigure}{\textwidth}
        \centering
        \includegraphics[width=0.99\textwidth, trim={0 25 0 25}, clip]{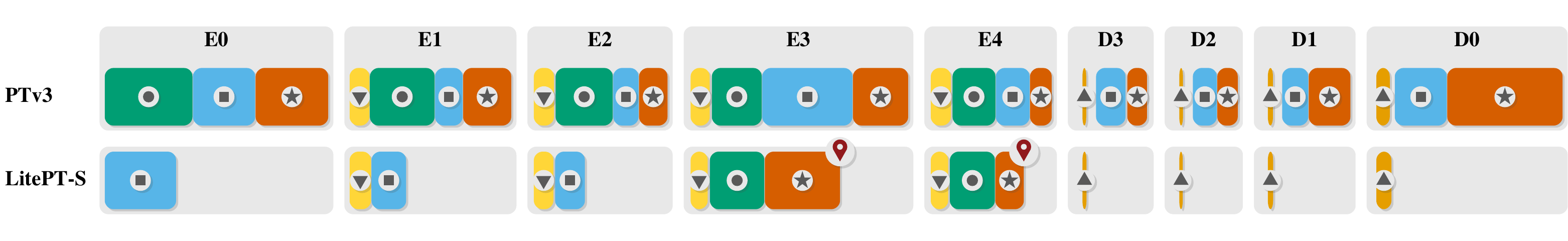}
        \caption{Breakdown of latencies}
        \label{fig:latency_treemap}
    \end{subfigure}
    \vspace{-2em}
    \caption{
    \textbf{Parameter count and latency.} E0-E4 denote encoder stages from shallow to deep, and D3-D0 denote decoder stages from deep to shallow. The length of each bar reflects the relative parameter count or latency of the corresponding module. \underline{Top:} In PTv3, the positional encoding implemented via a convolution block accounts for the majority of its parameters, particularly in the later stages. In contrast, our PointROPE is parameter-free. \underline{Bottom:} The PTv3 latency map reveals the significant cost of early-stage attention. \name{} restricts attention to late stages, where it is most effective and less costly.
    }%
    \vspace{-5pt}
    \label{fig:ptv3_treemap}
\end{figure*}

\begin{figure}[!bh]
    \centering
    \includegraphics[width=\linewidth]{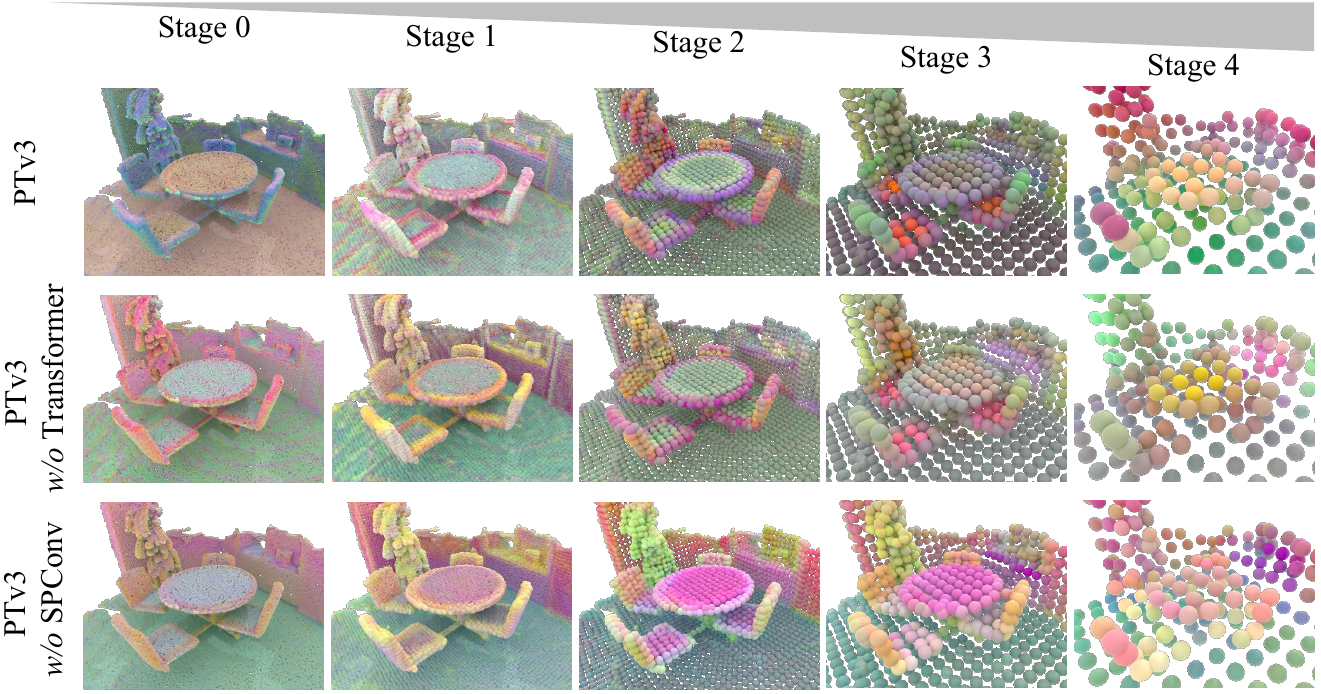}
    \caption{\textbf{Representations learnt by the hierarchical U-Net encoder.} 
    The hierarchical U-Net encoder exhibits an operator-agnostic feature hierarchy:
    shallow stages consistently encode local geometric structure, while  semantics emerge in deeper stages. 
    }%
    \label{fig:pca_map}
\end{figure}

To motivate our network design, we begin with an empirical study that investigates the respective roles of convolution and attention in PTv3~\cite{wu2024point}. We then introduce the components of \name{}: computational blocks that are reduced to the essentials and tailored to different processing stages (\cref{sec:method_stage_tailored}); and an alternative, learning-free positional encoding for the simplified blocks (\cref{sec:method_point_rope}). Finally, we describe the overall architecture in \cref{sec:method_architecture}. 

\subsection{Revisiting PTv3: Convolution vs.\ Attention}
\label{sec:method_revisit_ptv3}

\noindent\textbf{Preliminaries.}  PTv3~\cite{wu2024point} represents the current state-of-the-art architecture for point cloud understanding. 
Similar to earlier point cloud backbones~\cite{choy20194d, zhao2021point, wu2022point, robert2023efficient}, it adopts a U-Net architecture~\cite{ronneberger2015u} composed of multiple encoder and decoder stages with skip connections. 
Between consecutive encoding (or decoding) stages, pooling (or unpooling) operations are applied to downsample (or upsample) the point cloud and its associated features. 
Each encoder and decoder stage consists of several blocks. 
\cref{fig:ptv3_block} depicts a single block as used in PTv3, consisting of a convolutional positional encoding module \textcolor{xcpe}{\rule{0.25cm}{0.25cm}} and an attention module \textcolor{transformer}{\rule{0.25cm}{0.25cm}}. 
Inspired by~\cite{chu2021conditional}, PTv3 adopts conditional positional encoding, implemented by prepending a sparse convolution layer, a linear projection, and a LayerNorm, with a skip connection, before each attention module. 
The attention module follows a standard pre-norm structure~\cite{xiong2020layer}, where self-attention is applied between local groups of points obtained via serialisation sorting, followed by a multilayer perceptron (MLP).

Conditional positional encoding, and in particular its sparse convolution layer, has proved to be an important part of the overall architecture, but its precise role remains somewhat unclear. Does it indeed just serve to encode the spatial layout of the tokens that flow through the attention layer, or does it actually act as a local feature extractor in the spirit of classical convolutional networks?
In the following, we analyse the parameter efficiency and the computational cost of different components along the U-Net hierarchy, revealing striking differences between the stages.

\begin{table}[ht]
\centering
\setlength{\tabcolsep}{10pt}
\resizebox{\columnwidth}{!}{
\begin{tabular}{llccc}
\toprule
\cmidrule{1-5} 
& &  &
\multicolumn{1}{c}{ScanNet~\cite{dai2017scannet}} & \multicolumn{1}{c}{nuScenes~\cite{caesar2020nuscenes}} \\
\cmidrule(r){4-4}	\cmidrule(r){5-5} 
 & Model & \#Params & mIoU & mIoU  \\
\midrule	

 & PTv3~\cite{wu2024point} & 46.1M & 77.5 & 80.4\\
\circlenumblack{1} & PTv3 \emph{w/o} Transformer & 32.4M & 73.4  & 76.1 \\
\circlenumblack{2} &  PTv3 \emph{w/o} SPConv & 15.4M & 70.7 & 74.9 \\
\bottomrule
\end{tabular}
}
\vspace{-6px}
\caption{
\textbf{Revisiting PTv3.} We evaluate two PTv3 variants: in \circlenumblack{1}, the attention and MLP modules are removed, and in \circlenumblack{2}, only the sparse convolution layers are removed.
}
\label{tab:deconstruction}
\end{table}

\noindent\textbf{Number of parameters.} An often overlooked, yet important fact is that $67\%$ of the total parameter budget in PTv3 is spent on the sparse convolution layers of the positional encoding, while the Transformer part (\ie, attention and MLP) only accounts for $30\%$ of the learnable parameters.
Furthermore, the parameter count of the sparse convolution layers increases substantially with depth and is largest near the bottleneck, due to the high feature dimension of the late encoder and early decoder stages. See \cref{fig:parameter_treemap}.
 
\noindent\textbf{Latency.} \cref{fig:latency_treemap} graphically depicts the computational latency of attention and convolution across different network stages. 
Attention, with its quadratic computational complexity, accounts for the majority of the computational cost. 
Importantly, that cost decreases as one progresses towards deeper stages near the bottleneck, because hierarchical downsampling quadratically reduces the number of point tokens.

\noindent\textbf{Convolution vs.\ attention.} So far, we have clarified that convolution accounts for the majority of trainable parameters, whereas attention dominates the computational cost, and that both vary strongly along the U-Net hierarchy.
To separate the contributions of the two modules, we design two reduced variants of the PTv3 block. In the first one, we remove the attention modules.
Using exclusively this variant degenerates to a classical sparse U-Net structure~\cite{choy20194d,graham20183d}.
In the second variant, we remove only the sparse convolution layer to obtain a ``pure'' Transformer.
\Cref{tab:deconstruction} contrasts the semantic segmentation performance of the two variants for ScanNet~\cite{dai2017scannet} and nuScenes~\cite{caesar2020nuscenes}.
It turns out that removing convolutions causes a larger performance drop than removing the attention modules, suggesting that the ``positional encoding'' actually does much of the heavy lifting.
We visualise the learnt embeddings at each encoding stage for the three variants using PCA (\cref{fig:pca_map}) and find that a distinct division of labour emerges along the hierarchy, regardless of whether convolution, attention, or both are used. 
Early stages primarily encode local geometry, later stages capture high-level semantics.

\begin{figure*}[h]
    \centering
    \includegraphics[width=0.9\textwidth]{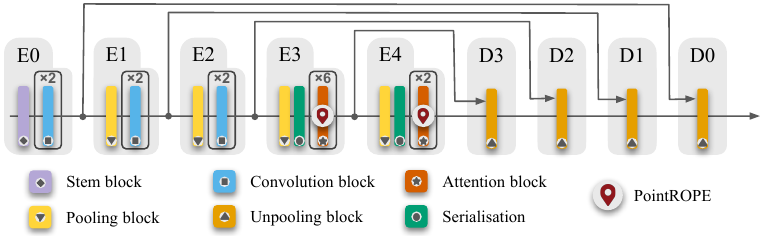}
    \vspace{-0.5em}
    \caption{\textbf{\name{}-S architecture.} Our model comprises five stages, employing convolution blocks in the early stages and PointROPE augmented attention blocks in the later ones. \name{}-S uses a lightweight decoder. Alternatively, adding convolution or attention blocks symmetrically in the decoder produces \name{}-S*.}
    \label{fig:pipeline}
    \vspace{-5pt}
\end{figure*}

\noindent\textbf{Discussion.} The above analysis leads us to the following hypotheses:

\begin{enumerate}
\item
It may not be necessary to use both convolution \emph{and} attention at every stage. In the early stages, which prioritise local feature extraction, convolution is adequate. In deep stages, where the focus is on long-range context and semantic concepts, attention is key.
\item
It would be a sweet spot in terms of efficiency if one could indeed avoid attention at early stages, where it is most expensive, and convolution at late stages, where it inflates the parameter count.
\item
Pure attention blocks will require an alternative positional encoding---but storing spatial layout is apparently \emph{not} the main function of the convolution, so a more parameter-efficient replacement should be possible.
\end{enumerate}

\subsection{Tailored Blocks for Different Network Stages}
\label{sec:method_stage_tailored}

Driven by the insights from the study described above, we propose a simple yet effective design that retains only the essential operations in each stage. Convolutions are allocated to earlier stages with high spatial resolution and low channel depth, and attention is reserved for deep stages with only few, but high-dimensional tokens.

Formally, let the hierarchical encoder consist of $L$ stages, where the $i$-th stage transforms the feature representation $f_{i-1}$ into $f_{i}$ via a function $\mathcal{B}_i(\cdot)$:
\begin{equation}
    f_{i}=\mathcal{B}_{i}(f_{i-1}), \quad i =1,...,L
\end{equation}
Depending on the stage index, each block  $\mathcal{B}_i$ is instantiated as either pure convolution or pure attention:
\begin{equation}
\mathcal{B}_{i} = 
\left\{
\begin{aligned}
  &\text{ConvBlock}_{i}, & \text{if}   \ \ i \le L_{c} \\
  &\text{AttnBlock}_{i}, & \text{if}   \ \ i > L_{c} \\
\end{aligned}
\right.
\end{equation}
Early stages ($i\!\leq\!L_c$) operate on point sets with high spatial resolution and density, where local geometric reasoning is critical. Employing convolution layers in these stages efficiently aggregates information over local receptive fields, with minimal parameter overhead. As one progresses to deeper stages ($i\!>\!L_c$), the number of point tokens is greatly reduced and semantic abstraction becomes more important, hence one switches to attention-based blocks. Optionally, one can also include a ``hand-over'' stage $i$ with both $\text{ConvBlock}_{i}$ and $\text{AttnBlock}_{i}$. See ablation studies in \cref{ablation_studies}.
More gradual transitions between the two mechanisms are, in principle, possible, but unnecessarily complicate the design.

Our \name{} follows a different philosophy than PTv3 and other hybrid point cloud Transformers: \cite{wu2023pointconvformer,thomas2024kpconvx,duan2023condaformer} all uniformly repeat the same computational block at all stages; as a consequence, that unit must include both attention and convolution. In contrast, we prefer to simplify individual blocks as much as possible, which then requires different forms of simplification depending on the network stage.
Empirically, we find that strategically distributing custom blocks along the hierarchy yields higher performance with significantly lower memory footprint and computational cost.

\subsection{Point Rotary Positional Embedding}
\label{sec:method_point_rope}

Discarding the expensive convolution layer at deep hierarchy levels has an undesired side effect: one loses the positional encoding. Hence, a more parameter-efficient replacement is needed.

Rotary Positional Embedding (RoPE)~\cite{su2024roformer} has proven to be remarkably effective in natural language processing.
In RoPE, relative positional awareness is introduced into the attention mechanism through rotations of the feature space.
Originally, the method is designed for 1D sequence data. It does not have a direct generalisation to irregular point clouds in 3D point space.

We adapt RoPE to 3D in a straightforward manner to obtain Point Rotary Positional Embedding (PointROPE). Given a point feature vector $\mathbf{f}_i \in \mathbb{R}^d$ at position $\mathbf{p}_i = (x_i,y_i,z_i)$, we divide the embedding dimension $d$ into three equal subspaces corresponding to the $x$, $y$, and $z$ axes:
\begin{equation}
    \mathbf{f}_i =  [\mathbf{f}^{x}_i; \mathbf{f}^{y}_i; \mathbf{f}^{z}_i], \ \ \ \ \ \mathbf{f}^{x}_i, \mathbf{f}^{y}_i, \mathbf{f}^{z}_i \in  \mathbb{R}^{d/3}\;.
\end{equation}
We then independently apply the standard 1D RoPE embedding to each subspace, using the respective point coordinate, and concatenate the axis-wise embeddings to form the final point representation:
\begin{equation}
    \tilde{\mathbf{f}_i} =
    \begin{bmatrix}
    \tilde{\mathbf{f}^{x}_i} \\
    \tilde{\mathbf{f}^{y}_i} \\
    \tilde{\mathbf{f}^{z}_i}
    \end{bmatrix}
    =
    \begin{bmatrix}
        \text{RoPE}_{1D}(\mathbf{f}^{x}_i, x_i) \\
        \text{RoPE}_{1D}(\mathbf{f}^{y}_i, y_i) \\
        \text{RoPE}_{1D}(\mathbf{f}^{z}_i, z_i)
    \end{bmatrix}\;.
\end{equation}
For each point with coordinates $(x_i,y_i,z_i)$, we directly use its grid coordinates as input, which are already correctly scaled during the pooling operation. 

The embedding scheme preserves the directional separability of 3D points while jointly encoding the feature's positional phase rotation, effectively capturing relative geometry.
Compared to the learnt convolutional positional encoding of PTv3~\cite{wu2024point}, PointROPE is parameter-free, lightweight, and, by construction, rotation-friendly. As part of our open source code, we provide an optimised CUDA implementation.

\subsection{Architecture}
\label{sec:method_architecture}

Our model follows the conventional U-Net~\cite{ronneberger2015u} structure, with five stages. 
We build three variants of the encoder, with varying number $C$ of channels in each stage and $B$ blocks per stage. Note that $C$ must be divisible by 6 in stages that include PointROPE.

\begin{itemize}
  \item[] $\!\!$\name{}-S: \mbox{\small $C=(36, 72, 144, 252, 504),B=(2, 2, 2, 6, 2)$}
  \item[] $\!\!$\name{}-B: \mbox{\small $C\!=\!(54, 108, 216, 432, 576),B\!=\!(3, 3, 3, 12, 3)$}
  \item[] $\!\!$\name{}-L: \mbox{\small $C\!=\!(72, 144, 288, 576, 864),B\!=\!(3, 3, 3, 12, 3)$}
\end{itemize}

We use \name{}-S as the main variant for the experiments, since it already delivers excellent performance across all benchmarks. Model scaling is examined in \cref{tab:model_scaling}. Per default, we set $L_c\!=\!3$, meaning that stages 1, 2, 3 use $\text{ConvBlock}_{i}$, while stages 4, 5 use $\text{AttnBlock}_{i}$. Each $\text{ConvBlock}_{i}$ consists of a sparse convolution layer, a linear layer and LayerNorm, and has a residual connection. Each $\text{AttnBlock}_{i}$ consists of a PointROPE embedding followed by attention, where the latter is computed locally within groups of points, found with the same serialisation sorting as in PTv3~\cite{wu2024point}.
For semantic segmentation, we simplify the decoder to only the linear projection layer and LayerNorm in each stage.
For instance segmentation, we apply the stage-specific design also in the decoder and symmetrically assign $\text{ConvBlock}_{i}$ and $\text{AttnBlock}_{i}$, in reverse order of the encoder.

\section{Experiments}
\label{sec:experiments}

\begin{table}[ht]
\centering
\setlength{\tabcolsep}{5pt}
\resizebox{\columnwidth}{!}{
\begin{tabular}{lccccc}
\toprule
\cmidrule{1-6} 
& & 
\multicolumn{2}{c}{Training} & \multicolumn{2}{c}{Inference} \\
\cmidrule(r){3-4}	\cmidrule(r){5-6} 
Method & \#Params & Latency & Memory & Latency & Memory  \\
\midrule	
MinkUNet~\cite{choy20194d} & 39.2M & 60ms & 1.9G & 21ms & 2.4G  \\
PTv2~\cite{wu2022point} & 12.8M & $\!\!$188ms & $\!\!\!$22.8G & $\!\!$151ms &  $\!\!\!$22.9G \\
PTv3~\cite{wu2024point} & 46.1M & $\!\!$110ms & 5.8G  & 51ms & 4.1G  \\
\arrayrulecolor{black!10}\midrule\arrayrulecolor{black}

\name{}-S (Ours) & 12.7M & 72ms & 2.3G & 21ms & 2.0G\\
\name{}-S* (Ours) & 16.0M & 81ms & 3.3G & 26ms & 2.0G\\
\name{}-B (Ours) & 45.1M & 93ms & 5.5G & 33ms & 2.4G \\
\name{}-L (Ours) & 85.9M & 97ms & 8.4G & 41ms & 2.6G \\
\bottomrule
\end{tabular}
}
\vspace{-4px}
\caption{
\textbf{Efficiency comparison.} Results are reported as average over the full ScanNet dataset using a single RTX 4090 GPU. Automatic Mixed Precision (AMP) is enabled for all models during training and disabled during inference. * denotes our variant with a heavier decoder that includes attention or convolutional blocks.
}\vspace{-5pt}
\label{tab:efficiency}
\end{table}

We begin with a series of ablation studies to analyse different configurations of our hybrid design, the model's scaling behaviour, and PointROPE (\cref{ablation_studies}). We then present comparisons with state-of-the-art methods for 3D semantic segmentation (\cref{exp:sem_seg}), 3D instance segmentation (\cref{exp:ins_seg}) and 3D object detection (\cref{exp:obj_det}).

\subsection{Ablation Studies and Analysis}
\label{ablation_studies}

\begin{figure}[h]
    \centering
    \begin{minipage}{0.49\columnwidth}
        \centering
        \includegraphics[width=\linewidth]{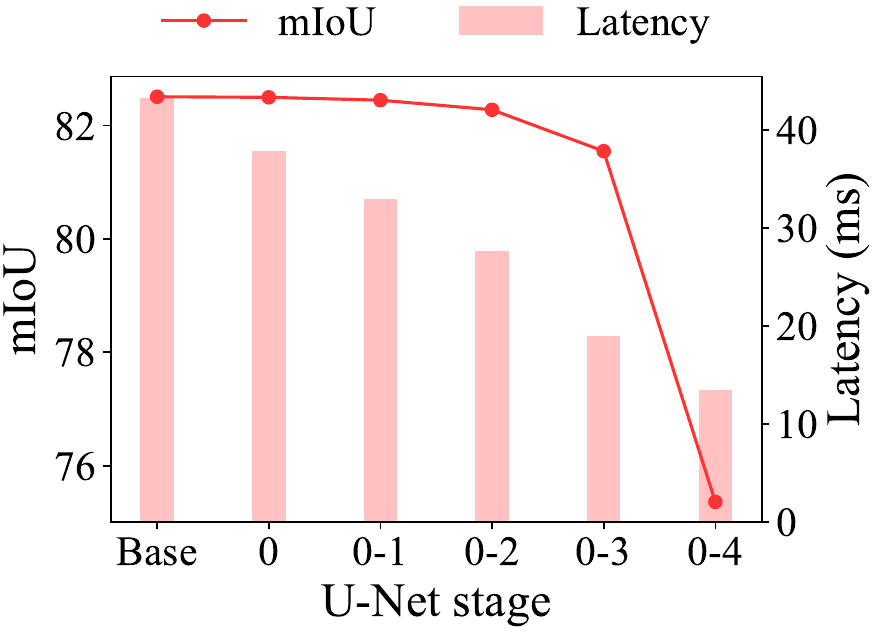}
    \end{minipage}
    \hfill
    \begin{minipage}{0.49\columnwidth}
        \centering
        \includegraphics[width=\linewidth]{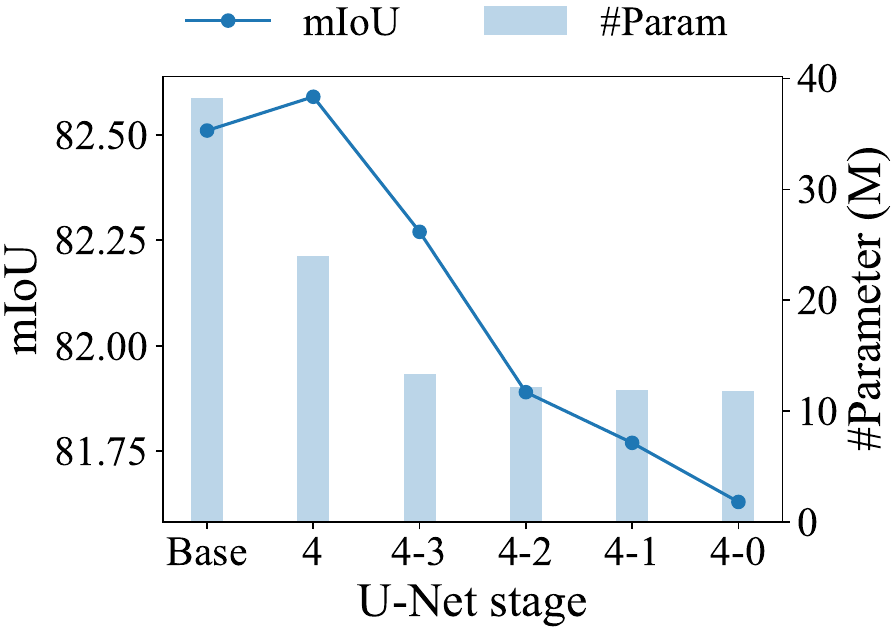}
    \end{minipage}
    \vspace{-0.5em}
    \caption{\textbf{Performance-efficiency trade-off.} \underline{Left:} Progressively dropping attention in more of the early stages. \underline{Right:} Progressively dropping convolution in more of the late stages.}\vspace{-5pt}
    \label{fig:vary_conv_attn}
\end{figure}

\noindent\textbf{Are both convolution and attention needed at every stage?} To verify our first hypothesis from \cref{sec:method_revisit_ptv3}, we design two sets of experiments on nuScenes. We begin with a baseline model that incorporates both convolution and PointROPE attention at all stages. In Experiment 1, we progressively remove \emph{attention}, first from stage 0, then from stages 0 and 1, \etc In Experiment 2, we progressively remove \emph{convolution}, first from stage 4, then from stages 4 and 3, \etc We then plot the mIoU of those configurations against latency (resp.\ parameter count).

As shown in \cref{fig:vary_conv_attn} (\emph{left}), removing attention in early stages boosts efficiency with almost no drop in mIoU, whereas removing attention in later stages harms performance. On the other hand, \cref{fig:vary_conv_attn} (\emph{right}) shows that removing convolution in later stages greatly reduces the parameter count with a negligible change in mIoU, whereas removing convolution in early stages only marginally improves efficiency but adversely affects performance.
The analysis confirms that one need \emph{not} include both convolution and attention at every stage. Their contribution and their cost highly depend on the hierarchy level.

\noindent\textbf{Where is the sweet spot in terms of efficiency and performance?} To determine the optimal transition point $L_c$ between convolution and attention, we conduct an ablation study on nuScenes as shown in \cref{tab:ablation_attn_vs_conv}. Optionally, we include a ``hand-over'' stage, denoted by ``X'', that includes both convolution and attention. Setting $L_{c}\!=\!3$, \ie, convolution in the first three stages and attention in the last two, achieves the best trade-off between parameter count, latency, and mIoU. We adopt $L_{c}\!=\!3$ as our default setting for all experiments. 

\begin{table}[ht]
\centering
\setlength{\tabcolsep}{18pt}
\resizebox{\columnwidth}{!}{
\begin{tabular}{lcccc}
\toprule
\cmidrule{1-5}  
$L_c$ & Setting & \#Params & Latency  & mIoU  \\
\midrule	
0 & A-A-A-A-A & 11.8M & 35.1ms &   82.1\\
1 & C-A-A-A-A & 11.9M & 30.4ms  &81.7 \\
2 & C-C-A-A-A & 12.0M &  25.8ms & 82.0 \\
\rowcolor{gray!15}
3 & C-C-C-A-A & 12.7M & 21.5ms & 82.2\\
4 & C-C-C-C-A & 18.8M &  16.2ms & 80.9\\
5 & C-C-C-C-C & 26.9M & 13.5ms &75.4 \\
\midrule	
 & C-X-A-A-A & 12.2M &  30.9ms & 81.9\\
 & C-C-X-A-A & 13.2M & 26.7ms & 82.3\\
 & C-C-C-X-A & 23.4M  & 24.9ms  & 82.4\\

\bottomrule
\end{tabular}
}
\vspace{-4px}
\caption{
\textbf{Effect of $L_c$ and ``hand-over'' stage.}  C: convolutional block; A: attention block; X: both convolution and attention are used at that stage. 
We compare model variants and report parameter count, latency, and validation mIoU on the nuScenes dataset. The grey-shaded row is our recommended setting.
}\vspace{-15pt}
\label{tab:ablation_attn_vs_conv}
\end{table}

\noindent\textbf{Decoder design.} The mixed design with blocks tailored to the layer depth is always used in the U-Net \textbf{en}coder. On the contrary, we propose two design variants for the U-Net \textbf{de}coder. In \name{}-S*, the same mixed design is used in the decoder, in reverse order. In \name{}-S, we further strip down the architecture and keep only a linear projection layer per stage (as needed to integrate skip connections), making the method even more efficient. We find empirically that the optimal choice is task-dependent, as shown in \cref{tab:ablation_decoder}. For semantic segmentation, the simple decoder is the best choice. For instance segmentation, the variant with convolution and attention blocks has a noticeable edge. We point out that even the slightly heavier \name{}-S* is still a lot more efficient than other Point Transformers (see \cref{tab:efficiency}), and leave the choice of decoder to the user.

\begin{table}[ht]
\centering
\setlength{\tabcolsep}{2pt}
\resizebox{\columnwidth}{!}{
\begin{tabular}{lccccc}
\toprule
\cmidrule{1-6} 
& 
\multicolumn{4}{c}{Semantic Segmentation (mIoU)} & \multicolumn{1}{c}{Instance Segmentation ($\text{mAP}_{50}$)} \\
\cmidrule(r){2-5}	\cmidrule(r){6-6} 
Decoder &  ScanNet~\cite{dai2017scannet} & Structured3D~\cite{zheng2020structured3d} & nuScenes~\cite{caesar2020nuscenes} & Waymo~\cite{sun2020scalability} & ScanNet~\cite{dai2017scannet} \\
\midrule	
\name{}-S & 76.5 & \textbf{83.6}  & \textbf{82.2} & \textbf{73.1} & 62.2 \\
\name{}-S* & \textbf{76.8} & 83.0  & 81.8 & 72.7 &  \textbf{64.9}  \\

\bottomrule
\end{tabular}
}
\vspace{-4px}
\caption{
\textbf{Decoder design.}  We compare two decoder variants: in \name{}-S*, we apply our stage-tailored design symmetrically to the decoder stages, while in \name{}-S, we retain only linear projection layers in all decoder stages.
}\vspace{-5pt}
\label{tab:ablation_decoder}
\end{table}

\noindent\textbf{Model scaling.} Due to the parameter-free PointROPE encoding, our model has substantially fewer trainable weights. This offers the possibility to repurpose the saved capacity and scale up \name{}. We assess scaling behaviour on Structured3D, the largest dataset in our evaluation suite. As shown in \cref{tab:model_scaling}, the model scales favourably: increasing the model size from \name{}-S to \name{}-L continuously improves performance, with only a modest increase in test-time latency and memory usage. Notably, even \name{}-L, with a parameter count twice that of PTv3, still runs faster than PTv3 and has a lower memory footprint.

\begin{table}[ht]
\centering
\setlength{\tabcolsep}{12pt}
\resizebox{\columnwidth}{!}{
\begin{tabular}{lcccc}
\toprule
\cmidrule{1-5}  
Method & \#Params & Latency & Memory & mIoU  \\
\midrule	
PTv3~\cite{wu2024point} & 46.1M & 57ms &  5.83G & 82.4   \\
\arrayrulecolor{black!10}\midrule\arrayrulecolor{black}

\name{}-S (Ours) & 12.7M & 23ms & 2.56G & 83.6 \\
\name{}-B (Ours) & 45.1M & 36ms & 2.60G & 85.1 \\
\name{}-L (Ours) & 85.9M & 44ms & 3.58G & 85.4 \\
\bottomrule
\end{tabular}
}
\vspace{-4px}
\caption{
\textbf{Model scaling on Structured3D dataset.} Our model scales efficiently, achieving consistent performance gains from small to large variants with modest increases in latency and memory. Even when scaled to twice the parameters of PTv3, \name{}-L remains more efficient.
}\vspace{-5pt}
\label{tab:model_scaling}
\end{table}

\noindent\textbf{PointROPE.} In \cref{tab:point_rope} we ablate the effectiveness of the proposed PointROPE, on nuScenes. Removing PointROPE leads to a significant performance drop of 2.6 percentage points in mIoU. We additionally ablate the base frequency $b$, which controls how \emph{fast} each embedding dimension ``rotates'' as the position increases (uniformly for the three axes). PointROPE is fairly robust to the choice of frequency. Setting $b\!=\!100$ yields the best score; we fix that value for all datasets to avoid excessive hyperparameter tuning.

\begin{table}[ht]
\centering
\setlength{\tabcolsep}{12pt}
\resizebox{\columnwidth}{!}{
\begin{tabular}{lc|c>{\columncolor{gray!15}}ccc}
\toprule
\cmidrule{1-6}  
 & $w/o$ PointROPE & $b=10$ & $b=100$ & $b=1000$ & $b=10000$ \\
\midrule	
mIoU &  79.6 & 81.7 & 82.2 & 81.8 & 81.3 \\
\bottomrule
\end{tabular}
}
\vspace{-4px}
\caption{
\textbf{PointROPE.} Dedicated positional encoding is needed---dropping PointROPE leads to a significant performance drop. PointROPE works similarly well with a wide range of base frequencies, the grey-shaded column is our recommended setting.
}\vspace{-15pt}
\label{tab:point_rope}
\end{table}

\subsection{Semantic Segmentation}
\label{exp:sem_seg}

\begin{table}[ht]
\vspace{-0.5em}
\centering
\setlength{\tabcolsep}{12pt}
\resizebox{\columnwidth}{!}{
\begin{tabular}{lccccc}
\toprule
\cmidrule{1-6} 
& & 
\multicolumn{2}{c}{nuScenes~\cite{caesar2020nuscenes}} & \multicolumn{2}{c}{Waymo~\cite{sun2020scalability}} \\
\cmidrule(r){3-4}	\cmidrule(r){5-6} 
Method & \#Param & mIoU & mAcc & mIoU &  mAcc  \\
\midrule		
MinkUNet~\cite{choy20194d} &  39.2M & 73.3 & - & 65.9 & 76.6  \\
SPVNAS~\cite{tang2020searching} & - & 77.4 & - & - & - \\
Cylinder3D~\cite{zhu2021cylindrical} & - & 76.1 & - & - & -\\
AF2S3Net~\cite{cheng20212} & - & 62.2 & - & - & -   \\
SphereFormer~\cite{lai2023spherical} & - & 78.4 & - & 69.9 & - \\
PTv2~\cite{wu2022point} & 12.8M & 80.2 & - & 70.6 & 80.2 \\
PTv3~\cite{wu2024point} & 46.1M & \underline{80.4} & \underline{87.2} & \underline{71.3} & \underline{80.5} \\

\arrayrulecolor{black!10}\midrule\arrayrulecolor{black}

\name{}-S (Ours) & 12.7M & \textbf{82.2} & \textbf{88.1} & \textbf{73.1} & \textbf{83.8} \\
\bottomrule
\end{tabular}
}
\vspace{-0.6em}
\caption{
\textbf{Outdoor semantic segmentation on nuScenes and Waymo validation set.}
Scores of prior work courtesy of~\cite{wu2024point,wu2025sonata}.
}\vspace{-10pt}
\label{tab:outdoor_sem}
\end{table}

\begin{table}[ht]
\vspace{-0.5em}
\centering
\setlength{\tabcolsep}{5pt}
\resizebox{\columnwidth}{!}{
\begin{tabular}{lcccccccccc}
\toprule
\cmidrule{1-11} 
& & & 
\multicolumn{4}{c}{Limited Scenes (Pct.)} & \multicolumn{4}{c}{Limited Annotations (Pts.)}  
\\
\cmidrule(r){4-7}	\cmidrule(r){8-11} 
Method & \#Params &  Full & 1$\%$ & 5$\%$  & 10$\%$ & 20$\%$ & 20 & 50 & 100 & 200 \\
\midrule	
MinkUNet~\cite{choy20194d} & 39.2M & 72.2 & 26.0 & 47.8 & 56.7 & 62.9 & 41.9 & 53.9 & 62.2 & 65.5 \\
PTv2~\cite{wu2022point} & 12.8M & 75.4 & 24.8 & 48.1 & 59.8 & 66.3 & 58.4 & 66.1 & 70.3 & 71.2   \\
PTv3~\cite{wu2024point} & 46.1M & \textbf{77.5} & 25.8 & 48.9 & 61.0 & 67.0 & 60.1 & 67.9 & \underline{71.4} & 72.7    \\
\arrayrulecolor{black!10}\midrule\arrayrulecolor{black}

\name{}-S (Ours) & 12.7M & 76.5 & \textbf{27.3} & \underline{50.6} & \textbf{63.1} & \textbf{67.3} & \underline{62.5} & \underline{68.4} & 70.9 & \underline{72.8} \\
\name{}-S* (Ours) & 16.0M & \underline{76.8} & \underline{27.2} & \textbf{51.6} & \underline{63.0} & \underline{67.1} & \textbf{63.2} & \textbf{69.5} & \textbf{72.0} & \textbf{74.2} \\
\bottomrule
\end{tabular}
}
\vspace{-0.6em}
\caption{
\textbf{Indoor semantic segmentation on ScanNet validation set.}
In mean IoU.
Scores of prior work courtesy of~\cite{wu2024point}. 
}\vspace{-5pt}
\label{tab:scannet_sem}
\end{table}

\begin{table}[ht]
\vspace{-0.75em}
\centering
\setlength{\tabcolsep}{12pt}
\resizebox{\columnwidth}{!}{
\begin{tabular}{lccccc}
\toprule
\cmidrule{1-6} 
& & 
\multicolumn{2}{c}{Val} & \multicolumn{2}{c}{Test} \\
\cmidrule(r){3-4}	\cmidrule(r){5-6} 
Method & \#Params & mIoU & mAcc & mIoU & mAcc  \\
\midrule	
MinkUNet~\cite{choy20194d} & 39.2M & 76.4 & 84.3  & 77.4 & 85.5 \\
PTv2~\cite{wu2022point} & 12.8M &  79.0 & 86.8 & 78.5 & 86.6\\
PTv3~\cite{wu2024point} & 46.1M & \underline{82.4} & \underline{90.3} & \underline{82.1} & 90.3  \\
\arrayrulecolor{black!10}\midrule\arrayrulecolor{black}

\name{}-S (Ours) & 12.7M & \textbf{83.6} & \textbf{90.7}  & \textbf{82.4} & \textbf{90.3}\\
\bottomrule
\end{tabular}
}
\vspace{-0.6em}
\caption{
\textbf{Indoor semantic segmentation on Structured3D.}\vspace{-5pt}
}
\label{tab:stru3d_sem}
\end{table}

\noindent\textbf{Setup.} We perform semantic segmentation for four different datasets. 
nuScenes~\cite{caesar2020nuscenes} and Waymo~\cite{sun2020scalability} are two outdoor datasets of first-person driving scenes, captured with vehicle-mounted LiDAR. ScanNet~\cite{dai2017scannet} and Structured3D~\cite{zheng2020structured3d} show indoor settings. The former was captured using an RGB-D camera. It is relatively small by today's standards, comprising 1,201 training scenes. Structured3D is a synthetic dataset and the largest public collection of 3D scenes with semantic annotations, and contains 18,348 training scenes. We follow PTv3 and use test time augmentation (TTA). 
Results without TTA can be found in the appendix.

\noindent\textbf{Results.} \cref{tab:outdoor_sem} reports semantic segmentation results on the nuScenes and Waymo validation sets. \name{} achieves marked improvements over competing architectures, in both cases +1.8 mIoU. We note that automotive LiDAR has different, more challenging properties compared with indoor datasets: the model must learn to handle massive differences in point density due to the large range, and highly anisotropic point distributions due to the scan line pattern and frequent specular reflections and ray drops.

\Cref{tab:scannet_sem} shows IoU scores for the ScanNet validation set. Following the literature~\cite{hou2021exploring}, we also report results with limited training, obtained either by restricting the number of available training scenes or by reducing the number of annotated points per scene. The performance of \name{} is comparable to PTv3, which has $\approx$4$\times$ more parameters---in data-constrained settings, even slightly better---and clearly superior to PTv2, which has a similar parameter count.
On the more than 10$\times$ larger Structured3D dataset, \name{} consistently outperforms all competing methods, including the much larger state-of-the-art PTv3.

\subsection{Instance Segmentation}
\label{exp:ins_seg}

\begin{table}[ht]
\vspace{-0.5em}
\centering
\setlength{\tabcolsep}{5pt}
\resizebox{\columnwidth}{!}{
\begin{tabular}{lccccccc}
\toprule
\cmidrule{1-8} 
&  &
\multicolumn{3}{c}{ScanNet~\cite{dai2017scannet}} & \multicolumn{3}{c}{ScanNet200~\cite{rozenberszki2022language}} \\
\cmidrule(r){3-5}	\cmidrule(r){6-8} 
PointGroup~\cite{jiang2020pointgroup} & \#Params & mAP$_{25}$ & mAP$_{50}$ & mAP &  mAP$_{25}$ & mAP$_{50}$ & mAP \\
\midrule	
MinkUNet~\cite{choy20194d} & 39.2M & 72.8 & 56.9 & 36.0 & 32.2 & 24.5 & 15.8\\
PTv2~\cite{wu2022point} & 12.8M & 76.3 & 60.0 & 38.3 & 39.6 & 31.9 & 21.4 \\
PTv3~\cite{wu2024point} & 46.2M & \underline{77.5} & \underline{61.7} & \underline{40.9} & \underline{40.1} & \textbf{33.2} & \textbf{23.1}\\
\arrayrulecolor{black!10}\midrule\arrayrulecolor{black}

\name{}-S* (Ours) & 16.0M & \textbf{78.5} & \textbf{64.9} & \textbf{41.7} &  \textbf{40.3} & \underline{33.1} & \underline{22.2}  \\
\bottomrule
\end{tabular}
}
\vspace{-0.5em}
\caption{
\textbf{Indoor instance segmentation on ScanNet and ScanNet200 validation set.}
Scores of prior work courtesy of~\cite{wu2024point}. 
}%
\label{tab:indoor_ins}
\end{table}

\noindent\textbf{Setup.}
We evaluate our method for instance segmentation on ScanNet~\cite{dai2017scannet} and ScanNet200~\cite{rozenberszki2022language}. Following the protocol of prior work, we employ PointGroup~\cite{jiang2020pointgroup} as instance segmentation head on top of the decoder to achieve a fair comparison. 

\noindent\textbf{Results.} 
\cref{tab:indoor_ins} summarises the results. On ScanNet, \name{} again outperforms all prior backbones and sets a new state of the art, with 64.9 $\text{mAP}_{50}$, a 3.2 percentage point improvement over PTv3. On ScanNet200, which includes a long tail of rare categories, the results are comparable to PTv3 and significantly better than all other previous methods. For example, our method achieves 1.2 percentage points higher $\text{mAP}_{50}$ than PTv2, which has a similar parameter count, but an 11$\times$ larger memory footprint and a 6$\times$ longer runtime.

\subsection{Object Detection}
\label{exp:obj_det}

\begin{table}[ht]
\centering
\setlength{\tabcolsep}{5pt}
\resizebox{\columnwidth}{!}{
\begin{tabular}{lccccccc}
\toprule
\cmidrule{1-8} 
 & 
\multicolumn{2}{c}{Vehicle L2} & \multicolumn{2}{c}{Pedestrian L2} & 
\multicolumn{2}{c}{Cyclist L2}  & 
\multicolumn{1}{c}{Mean L2} \\
\cmidrule(r){2-3}	\cmidrule(r){4-5} \cmidrule(r){6-7}  \cmidrule(r){8-8}
Method & mAP  & APH & mAP  & APH & mAP  & APH &  mAPH \\
\midrule	
PointPillars~\cite{lang2019pointpillars} & 63.6 & 63.1 & 62.8 & 50.3 & 61.9 & 59.9 & 57.8\\
CenterPoint~\cite{yin2021center} & 66.7 & 66.2 & 68.3 & 62.6 & 68.7 & 67.6 & 65.5\\
SST~\cite{fan2022embracing} & 64.8 & 64.4 & 71.7 & 63.0 & 68.0 & 66.9 & 64.8\\
SST-Center~\cite{fan2022embracing} & 66.6 & 66.2 & 72.4 & 65.0 & 68.9 & 67.6 & 66.3 \\
VoxSet~\cite{he2022voxel} & 66.0 & 65.6 & 72.5 & 65.4 & 69.0 & 67.7 & 66.2 \\
PillarNet~\cite{shi2022pillarnet} & 70.4 & 69.9 & 71.6 & 64.9 & 67.8 & 66.7 & 67.2 \\
FlatFormer~\cite{liu2023flatformer} & 69.0 & 68.6 & 71.5 & 65.3 & 68.6 & 67.5 & 67.2 \\
PTv3~\cite{wu2024point} & \underline{71.2} & \underline{70.8} & \textbf{76.3} & \textbf{70.4} & \underline{71.5} & \underline{70.4} & \underline{70.5} \\
\arrayrulecolor{black!10}\midrule\arrayrulecolor{black}

\name{} (Ours) & \textbf{71.6} & \textbf{71.2} & \underline{76.1} & \underline{70.1} & \textbf{71.8} & \textbf{70.7} & \textbf{70.7} \\

\bottomrule
\end{tabular}
}
\vspace{-4px}
\caption{
\textbf{Outdoor object detection on Waymo with single frames input.}
Scores of prior work courtesy of~\cite{wu2024point}. 
}%
\vspace{-7pt}
\label{tab:outdoor_det}
\end{table}

\noindent\textbf{Setup.} We evaluate 3D object detection on Waymo. For a fair comparison with prior work~\cite{wu2024point,liu2023flatformer}, we employ the same 3D object detection framework, CenterPoint-Pillar~\cite{yin2021center}. Consistent with~\cite{wu2024point,liu2023flatformer, fan2022embracing}, we avoid spatial downsampling, thus turning \name{} into a single-stage network with 8 blocks, to allow detection of small objects. Objects are divided into two difficulty levels, and we report level-2 metrics.

\noindent\textbf{Results.} \cref{tab:outdoor_det} reports scores based on single-scan LiDAR inputs. Also in this application, \name{} reaches the highest score overall and on two out of three object categories, and comfortably matches the performance of the closest competitor, PTv3.

\section{Conclusion and Discussion}
\label{sec:conclusion}

We have introduced \name{}, a lighter yet stronger point Transformer for various point cloud processing tasks. Our starting point was the question, which distinct roles and impacts different operators have along the processing hierarchy.
Experiments confirm that (sparse) convolutions are adequate, and more efficient, at early hierarchy levels, whereas attention comes into its own at higher levels, where semantic abstraction and global context over a comparatively small token set are key.
In itself, these observations are not unexpected, but surprisingly, they have not been leveraged in contemporary point cloud architectures. 
\name{} embodies the simple principle ``convolutions for low-level geometry, attention for high-level relations'' and strategically places only the required operations at each hierarchy level, avoiding wasted computations. To achieve this, we equip our method with parameter-free PointROPE positional encoding to compensate for the loss of spatial layout information that occurs when discarding convolutional layers. We hope that \name{} will be useful as a generic high-performance backbone for 3D point cloud processing, and that our analysis can serve as  practical guidance for architecture design beyond our current version.

In our architecture, attention is applied only in the later stages, where the reduced token count is small. It would therefore be affordable to compute self-attention globally across all tokens, rather than locally.
In future work, it may be interesting to eliminate the local grouping operation, which could on the one hand strengthen long-range context modelling, and on the other hand further reduce the computation time at inference.

\noindent\textbf{Acknowledgments.} Part of the compute is supported by the Swiss AI Initiative under projects a144 and a154 on Alps. 
We thank Xiaoyang Wu, Liyan Chen and Liyuan Zhu for their help with the comparison to PTv3. 
The project is supported by the Circular Bio-based Europe Joint Undertaking and its members under grant agreement No 101157488 (SingleTree).
Embed2Scale is co-funded by the EU Horizon Europe program under grant agreement No 101131841. Additional funding for this project has been provided by the Swiss State Secretariat for Education, Research and Innovation (SERI) and UK Research and Innovation (UKRI). 
Funded by the European Union. Views and opinions expressed are however those of the author(s) only and do not necessarily reflect those of the European Union or CBE JU. Neither the European Union nor the CBE JU can be held responsible for them.

\begin{appendices}

In this Appendix, we provide detailed architecture of \name{} (\cref{app:architecture}), detailed experimental settings (\cref{app:experimental_settings}), additional experiments (\cref{app:experiments}), and visualisation of \name{}'s predictions for 3D semantic segmentation, 3D instance segmentation, and 3D object detection (\cref{app:visualization}).

\section{Detailed Architecture}
\label{app:architecture}

\begin{figure*}[h]
    \centering
    \includegraphics[width=\textwidth]{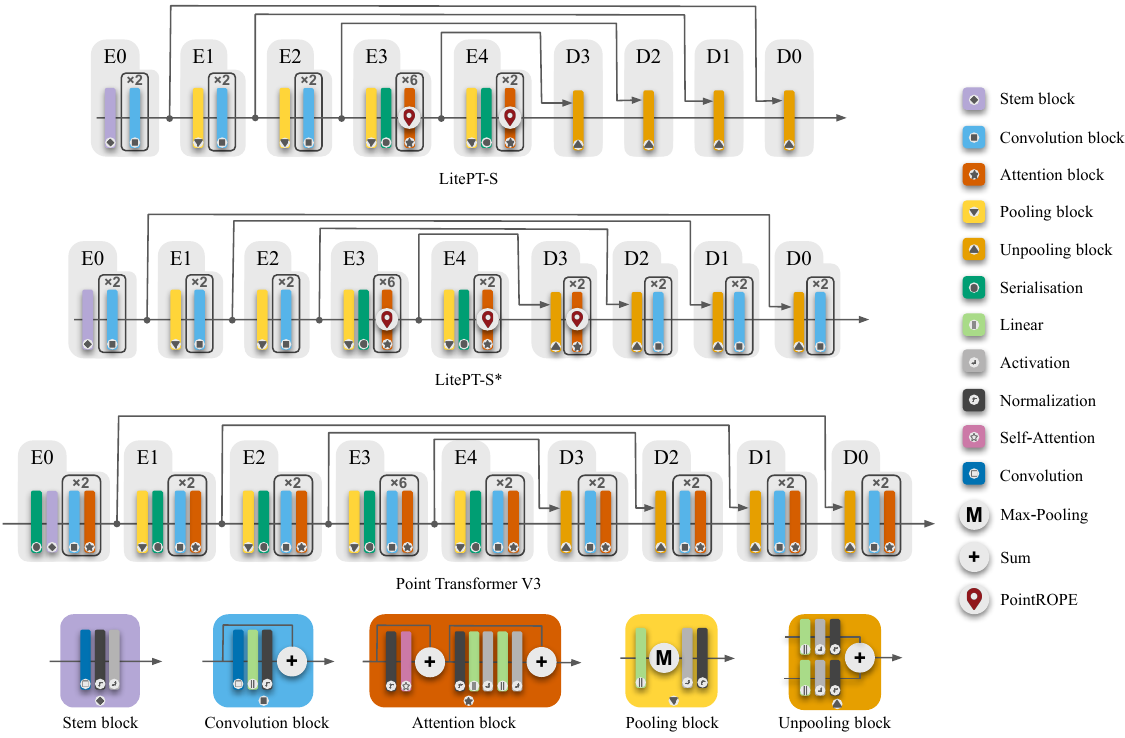}
    \caption{
    \textbf{Detailed architectures.} 
    We illustrate the full pipelines of \name{}-S, \name{}-S*, Point Transformer V3~\cite{wu2024point}, and the building blocks of each architecture.
    }
    \label{fig:detailed_architecture}\vspace{-10pt}
\end{figure*}

\begin{figure}[htbp]
    \centering
    \includegraphics[width=0.6\linewidth]{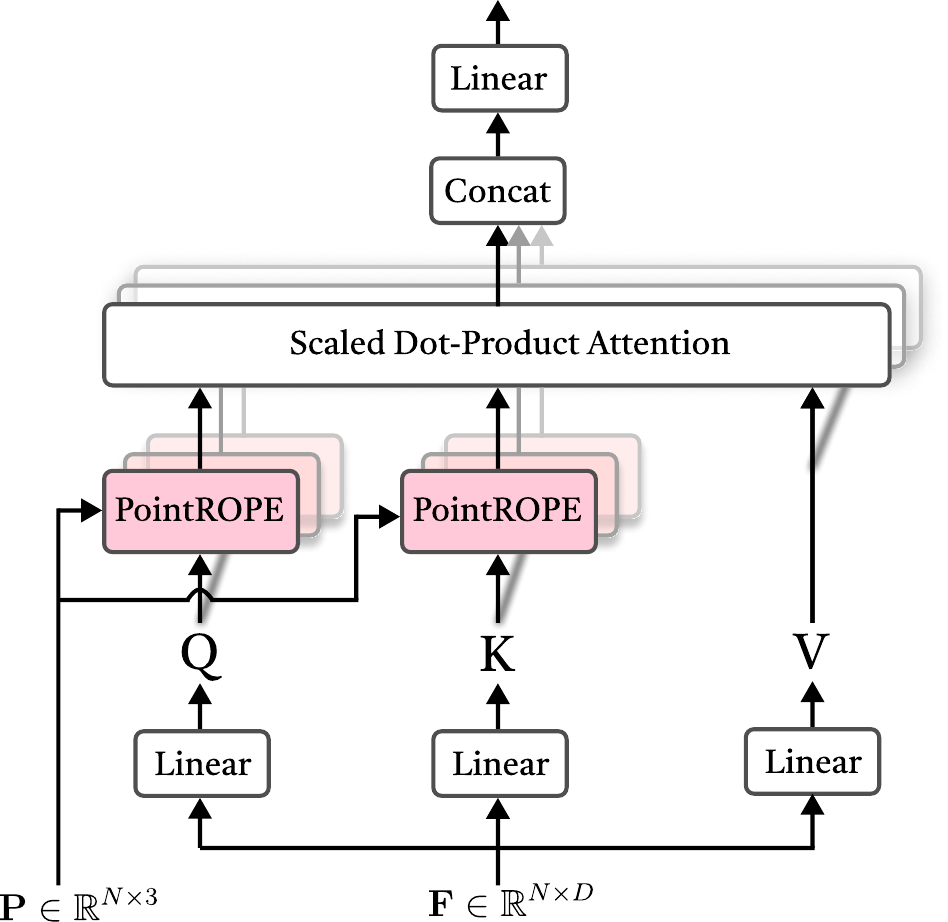}
    \caption{\textbf{PointROPE attention.} We apply PointROPE to query and key before standard scaled dot-product attention.}
    \vspace{-15pt}
    \label{fig:rope3d_attn}
\end{figure}

\begin{table*}[]
    \centering
    \renewcommand{\arraystretch}{1.4}
    \setlength{\tabcolsep}{2pt}
    \resizebox{\textwidth}{!}{
    \begin{tabular}{c|c|c|c|c}
        &  \textbf{\name{}-S} & \textbf{\name{}-S*} & \textbf{\name{}-B} & \textbf{\name{}-L} \\
    \hline

    \textbf{stem} & $\text{C}\!=\!36, \text{K}\!=\!5{\times}5{\times}5$ & $\text{C}\!=\!36, \text{K}\!=\!5{\times}5{\times}5$ & $\text{C}\!=\!36, \text{K}\!=\!5{\times}5{\times}5$ & $\text{C}\!=\!36, \text{K}\!=\!5{\times}5{\times}5$ \\
    \hline
    \textbf{E0} & 
    $\left[
    \begin{array}{c}
       \text{C}\!=\!36 \\
       \text{K}\!=\!3{\times}3{\times}3
    \end{array}
    \right] \!\times\! 2$ & 

      $\left[
    \begin{array}{c}
       \text{C}\!=\!36 \\
       \text{K}\!=\!3{\times}3{\times}3
    \end{array}
    \right] \!\times\! 2$ &
    $\left[
    \begin{array}{c}
       \text{C}\!=\!54 \\
       \text{K}\!=\!3{\times}3{\times}3
    \end{array}
    \right] \!\times\! 3$ &
    $\left[
    \begin{array}{c}
       \text{C}\!=\!72 \\
       \text{K}\!=\!3{\times}3{\times}3
    \end{array}
    \right] \!\times\! 3$

    \\
    \hline

    \multirow{3}{*}{
    \begin{tabular}{c}
       \centering \textbf{E1}
    \end{tabular}
    } & pool stride 2 & pool stride 2 & pool stride 2 & pool stride 2 \\
    \cline{2-5}
    & $\left[
    \begin{array}{c}
       \text{C}\!=\!72 \\
       \text{K}\!=\!3{\times}3{\times}3
    \end{array}
    \right] \!\times\! 2$ &
   $\left[
    \begin{array}{c}
       \text{C}\!=\!72 \\
       \text{K}\!=\!3{\times}3{\times}3
    \end{array}
    \right] \!\times\! 2$ & 
    $\left[
    \begin{array}{c}
       \text{C}\!=\!108 \\
       \text{K}\!=\!3{\times}3{\times}3
    \end{array}
    \right] \!\times\! 3$ &
    $\left[
    \begin{array}{c}
       \text{C}\!=\!144 \\
       \text{K}\!=\!3{\times}3{\times}3
    \end{array}
    \right] \!\times\! 3$ \\
    \hline

    \multirow{3}{*}{
    \begin{tabular}{c}
       \centering \textbf{E2}
    \end{tabular}
    } & pool stride 2 & pool stride 2 & pool stride 2 & pool stride 2\\
    \cline{2-5}
    & $\left[
    \begin{array}{c}
       \text{C}\!=\!144 \\
       \text{K}\!=\!3{\times}3{\times}3
    \end{array}
    \right] \!\times\! 2$ & 
    $\left[
    \begin{array}{c}
       \text{C}\!=\!144 \\
       \text{K}\!=\!3{\times}3{\times}3
    \end{array}
    \right] \!\times\! 2$ & 
    $\left[
    \begin{array}{c}
       \text{C}\!=\!216 \\
       \text{K}\!=\!3{\times}3{\times}3
    \end{array}
    \right] \!\times\! 3$ &
    $\left[
    \begin{array}{c}
       \text{C}\!=\!288 \\
       \text{K}\!=\!3{\times}3{\times}3
    \end{array}
    \right] \!\times\! 3$ \\
    \hline

    \multirow{3.75}{*}{
    \begin{tabular}{c}
       \centering \textbf{E3}
    \end{tabular}
    } & pool stride 2 & pool stride 2 & pool stride 2 & pool stride 2\\
    \cline{2-5}
    & $\left[
    \begin{array}{c}
       \text{C}\!=\!252, \text{H}\!=\!14 \\
       b\!=\!100, \text{F}\!=\!4 \\
       \text{N}\!=\!1024
    \end{array}
    \right] \!\times\! 6$ & 
    $\left[
    \begin{array}{c}
       \text{C}\!=\!252, \text{H}\!=\!14 \\
       b\!=\!100, \text{F}\!=\!4 \\
       \text{N}\!=\!1024
    \end{array}
    \right] \!\times\! 6$&
    $\left[
    \begin{array}{c}
       \text{C}\!=\!432, \text{H}\!=\!24 \\
       b\!=\!100, \text{F}\!=\!4 \\
       \text{N}\!=\!1024
    \end{array}
    \right] \!\times\! 12$ &
    $\left[
    \begin{array}{c}
       \text{C}\!=\!576, \text{H}\!=\!32 \\
       b\!=\!100, \text{F}\!=\!4 \\
       \text{N}\!=\!1024
    \end{array}
    \right] \!\times\! 12$\\
    \hline
    \multirow{3.75}{*}{
    \begin{tabular}{c}
       \centering \textbf{E4}
    \end{tabular}
    } & pool stride 2 & pool stride 2 & pool stride 2 & pool stride 2\\
    \cline{2-5}
    & $\left[
    \begin{array}{c}
       \text{C}\!=\!504, \text{H}\!=\!28 \\
       b\!=\!100, \text{F}\!=\!4 \\
       \text{N}\!=\!1024
    \end{array}
    \right] \!\times\! 2$ & 
     $\left[
    \begin{array}{c}
       \text{C}\!=\!504, \text{H}\!=\!28 \\
       b\!=\!100, \text{F}\!=\!4 \\
       \text{N}\!=\!1024
    \end{array}
    \right] \!\times\! 2$ & 
     $\left[
    \begin{array}{c}
       \text{C}\!=\!576, \text{H}\!=\!32 \\
       b\!=\!100, \text{F}\!=\!4 \\
       \text{N}\!=\!1024
    \end{array}
    \right] \!\times\! 3$ & 
     $\left[
    \begin{array}{c}
       \text{C}\!=\!864, \text{H}\!=\!48 \\
       b\!=\!100, \text{F}\!=\!4 \\
       \text{N}\!=\!1024
    \end{array}
    \right] \!\times\! 3$\\
    \hline

    \multirow{3.75}{*}{
    \begin{tabular}{c}
       \centering \textbf{D3}
    \end{tabular}
    } & \multirow{3.75}{*}{
    \begin{tabular}{c}
       \centering unpool $\text{C}\!=\!252$
    \end{tabular}
    }  & unpool $\text{C}\!=\!252$ & \multirow{3.75}{*}{
    \begin{tabular}{c}
       \centering unpool $\text{C}\!=\!432$
    \end{tabular}
    } & \multirow{3.75}{*}{
    \begin{tabular}{c}
       \centering unpool $\text{C}\!=\!576$
    \end{tabular}
    }\\
    \cline{3-3}
    & &  
    $\left[
    \begin{array}{c}
       \text{C}\!=\!252, \text{H}\!=\!14 \\
       b\!=\!100, \text{F}\!=\!4 \\
       \text{N}\!=\!1024
    \end{array}
    \right] \!\times\! 2$ &
     & \\
    \hline

    \multirow{3.75}{*}{
    \begin{tabular}{c}
       \centering \textbf{D2}
    \end{tabular}
    } & \multirow{3}{*}{
    \begin{tabular}{c}
       \centering unpool $\text{C}\!=\!144$
    \end{tabular}
    }  & unpool $\text{C}\!=\!144$ & \multirow{3}{*}{
    \begin{tabular}{c}
       \centering unpool $\text{C}\!=\!216$
    \end{tabular}
    } & \multirow{3}{*}{
    \begin{tabular}{c}
       \centering unpool $\text{C}\!=\!288$
    \end{tabular}
    }\\
    \cline{3-3}
    & &  
    $\left[
    \begin{array}{c}
       \text{C}\!=\!144 \\
       \text{K}\!=\!3{\times}3{\times}3
    \end{array}
    \right] \!\times\! 2$ &
     & \\
    \hline

    \multirow{3.75}{*}{
    \begin{tabular}{c}
       \centering \textbf{D1}
    \end{tabular}
    } & \multirow{3}{*}{
    \begin{tabular}{c}
       \centering unpool $\text{C}\!=\!72$
    \end{tabular}
    }  & unpool $\text{C}\!=\!72$ & \multirow{3}{*}{
    \begin{tabular}{c}
       \centering unpool $\text{C}\!=\!108$
    \end{tabular}
    } & \multirow{3}{*}{
    \begin{tabular}{c}
       \centering unpool $\text{C}\!=\!144$
    \end{tabular}
    }\\
    \cline{3-3}
    & &  
    $\left[
    \begin{array}{c}
       \text{C}\!=\!72 \\
       \text{K}\!=\!3{\times}3{\times}3
    \end{array}
    \right] \!\times\! 2$ &
     & \\
    \hline
    
    \multirow{3}{*}{
    \begin{tabular}{c}
       \centering \textbf{D0}
    \end{tabular}
    } & \multirow{3}{*}{
    \begin{tabular}{c}
       \centering unpool $\text{C}\!=\!72$ 
    \end{tabular}
    }  & unpool $\text{C}\!=\!72$ & \multirow{3}{*}{
    \begin{tabular}{c}
       \centering unpool $\text{C}\!=\!72$
    \end{tabular}
    } & \multirow{3}{*}{
    \begin{tabular}{c}
       \centering unpool $\text{C}\!=\!72$
    \end{tabular}
    }\\
    \cline{3-3}
    & &  
    $\left[
    \begin{array}{c}
       \text{C}\!=\!72 \\
       \text{K}\!=\!3{\times}3{\times}3
    \end{array}
    \right] \!\times\! 2$ &
     & \\
    \hline
    \#Params & 12.7M & 16.0M & 45.1M & 85.9M
    
    \end{tabular}
    }
    \caption{\textbf{Detailed architecture specifications.} C: channel dimension, K: kernel size in the convolution block, H: number of heads, $b$: base frequency of PointROPE, F: MLP ratio in the FFN module,  N: number of points in local group.}
    \label{tab:architecture_specifications}
\end{table*}

Our full architecture is shown in \cref{fig:detailed_architecture}. It follows U-Net-style~\cite{ronneberger2015u} encoder-decoder design with skip connections, and is organised into five stages. Adjacent encoder (or decoder) stages are connected via pooling (or unpooling) blocks. We apply our stage-tailored design on the encoder: the first three stages use convolution blocks, while the final two use attention blocks. For \name{}-S/B/L, each stage in the decoder contains only an unpooling block. For \name{}-S*, we mirror the stage-tailored design in the decoder as well. Detailed architecture specifications can be found in \cref{tab:architecture_specifications}. Below, we describe each block type in detail.

\noindent\textbf{Attention block.} Each attention block consists of a \rope{} attention module and a feed-forward network (FFN) module. Following the pre-norm~\cite{xiong2020layer} convention, a LayerNorm~\cite{ba2016layer} is placed before both the attention and FFN modules. The FFN uses a hidden dimension four times larger than the channel dimension of its stage. 
We observe that adding an extra LayerNorm before the attention block further stabilises the training. 
In the PointROPE attention module (\cref{fig:rope3d_attn}), input point features are projected to query (Q), key (K), and value (V) representations. \rope{} is computed from point coordinates $\text{P}$ and applied to Q and K, leaving V unchanged. The resulting ``rotated'' $\text{Q}'$ and $\text{K}'$ are fed into a standard scaled dot-product multi-head attention together with V, followed by a linear projection to produce the final output embeddings. Our \rope{} implementation is compatible with FlashAttention~\cite{dao2022flashattention,dao2023flashattention,shah2024flashattention}, which we use in our model. We apply \rope{} to locally-aggregated groups of 1024 points, formed using the same serialisation sorting strategy as~\cite{wu2024point}.

\noindent\textbf{Convolution block.} The convolution block consists of a single sparse convolution layer~\cite{choy20194d, graham20183d} with a kernel size of $3\times3\times3$, followed by a linear projection layer and a LayerNorm~\cite{ba2016layer} layer. A residual connection~\cite{he2016deep} links the block's input and output.

\noindent\textbf{Pooling and unpooling blocks.} We adopt the grid pooling and unpooling operation from~\cite{wu2022point}. During pooling, points are divided into non-overlapping partitions. Point features are first projected by a linear layer, then points within the same partition are max-pooled, followed by a GELU~\cite{hendrycks2016gaussian} activation and a BatchNorm layer~\cite{ioffe2015batch}. The pooling stride is set to 2 at each stage, reducing the spatial resolution by a factor of 2 each time. During unpooling, point features from the current decoder stage and the corresponding encoder stage are each passed through their own linear layer, GELU activation, and BatchNorm. The resulting features are then merged through a skip connection via summation.

\section{Detailed Experimental Settings}
\label{app:experimental_settings}

For indoor datasets, we use RGB and surface normals as input features. For outdoor datasets, where RGB and normal information are unavailable, we use $xyz$ coordinates and intensity (plus elongation for object detection). Following common practice~\cite{wu2022point,wu2024point, choy20194d}, we first downsample the point cloud on a grid. For 3D segmentation tasks, we set the grid size to 0.02m for indoor scenes and 0.05m for outdoor scenes. For 3D object detection, we adopt grid sizes of 0.32m in the \emph{xy} plane and 6m along the \emph{z} axis, consistent with~\cite{wu2024point,liu2023flatformer}. 
Detailed training configurations for semantic segmentation, instance segmentation and object detection are provided in \cref{tab:experimental_settings_sem}, \cref{tab:experimental_settings_ins}, and \cref{tab:experimental_settings_det}, respectively.

\begin{table}[!t]
\centering
\setlength{\tabcolsep}{2pt}
\resizebox{\columnwidth}{!}{
\begin{tabular}{lcccc}
\toprule
 & nuScenes~\cite{caesar2020nuscenes} & 
Waymo~\cite{sun2020scalability} & ScanNet~\cite{dai2017scannet} & Structured3D~\cite{zheng2020structured3d}  \\
\midrule	

Input feature  & \multicolumn{2}{c}{XYZ+Intensity} & \multicolumn{2}{c}{RGB+Normal} \\ 
\hline
Grid size  & \multicolumn{2}{c}{0.05m} & \multicolumn{2}{c}{0.02m} \\
\hline
Head (framework) & \multicolumn{2}{c}{Linear segmentor} & \multicolumn{2}{c}{Linear segmentor}  \\
\hline
Loss & \multicolumn{2}{c}{CrossEntropy+Lovasz~\cite{berman2018lovasz}} & \multicolumn{2}{c}{CrossEntropy+Lovasz~\cite{berman2018lovasz}}  \\
\hline

Optimizer & \multicolumn{2}{c}{AdamW~\cite{loshchilov2017decoupled}} & \multicolumn{2}{c}{AdamW~\cite{loshchilov2017decoupled}}  \\
\hline
Weight decay  & \multicolumn{2}{c}{0.005} & \multicolumn{2}{c}{0.05} \\
\hline
Scheduler & \multicolumn{2}{c}{OneCycleLR~\cite{smith2019super}}  & \multicolumn{2}{c}{OneCycleLR~\cite{smith2019super}} \\
\hline
Learning rate &  \multicolumn{2}{c}{0.002} & 0.006 & 0.012   \\
\hline
Block lr rate &  \multicolumn{2}{c}{0.0002} & 0.0006 & 0.0012   \\

\hline
Batch size &  \multicolumn{2}{c}{12}  & 12 & 48  \\
\hline
Epochs &  \multicolumn{2}{c}{50} & 1200 (800) & 200  \\
\hline
Num GPUs &  \multicolumn{2}{c}{4} & 4 & 16  \\
\hline
Data augmentation  & 
    \multicolumn{2}{c}{
    $\begin{array}{c}
       \text{Random rotate, random scale} \\
        \text{random flip, random jitter} \\
    \end{array}$
    }   & 
\multicolumn{2}{c}{
    $\begin{array}{c}
       \text{Random shift, random dropout} \\
        \text{random rotate, random scale} \\
        \text{random flip, random jitter} \\
        \text{elastic distortion, color auto contrast} \\
        \text{color jitter, sphere crop} \\
        \text{color normalization}
    \end{array}$
    } 
  
    \\

\bottomrule
\end{tabular}
}
\vspace{-4px}
\caption{
\textbf{Detailed training settings for semantic segmentation.} 
}
\label{tab:experimental_settings_sem}
\end{table}

\begin{table}[!t]
\centering
\setlength{\tabcolsep}{20pt}
\resizebox{\columnwidth}{!}{
\begin{tabular}{lcc}
\toprule
 & 
ScanNet~\cite{dai2017scannet} & 
ScanNet200~\cite{rozenberszki2022language} \\
\midrule	

Input feature &    \multicolumn{2}{c}{RGB+Normal} \\ 
\hline
Grid size & \multicolumn{2}{c}{0.02m} \\
\hline
Head (framework) &  \multicolumn{2}{c}{PointGroup~\cite{jiang2020pointgroup}} \\
\hline
Loss &  \multicolumn{2}{c}{As in PointGroup~\cite{jiang2020pointgroup}}   \\
\hline

Optimizer &  \multicolumn{2}{c}{AdamW~\cite{loshchilov2017decoupled}} \\
\hline
Weight decay & \multicolumn{2}{c}{0.05} \\
\hline
Scheduler& \multicolumn{2}{c}{OneCycleLR~\cite{smith2019super}} \\
\hline
Learning rate & \multicolumn{2}{c}{0.006} \\
\hline
Block lr rate & \multicolumn{2}{c}{0.0006} \\

\hline
Batch size & \multicolumn{2}{c}{12}  \\
\hline
Epochs & \multicolumn{2}{c}{800} \\
\hline
Num GPUs  & \multicolumn{2}{c}{4} \\
\hline
Data augmentation    & 

    \multicolumn{2}{c}{
    $\begin{array}{c}
       \text{Random shift, random dropout} \\
        \text{random rotate, random scale} \\
        \text{random flip, random jitter} \\
        \text{elastic distortion, color auto contrast} \\
        \text{color jitter} \\
        \text{sphere crop, color normalization}
    \end{array}$
    } 
  
    \\

\bottomrule
\end{tabular}
}
\vspace{-4px}
\caption{
\textbf{Detailed training settings for instance segmentation.} 
}
\label{tab:experimental_settings_ins}
\end{table}

\begin{table}[!t]
\centering
\setlength{\tabcolsep}{20pt}
\resizebox{\columnwidth}{!}{
\begin{tabular}{lc}
\toprule
& 
Object Detection \\
 & 
Waymo~\cite{sun2020scalability} \\
\midrule	

Input feature & XYZ+Intensity+Elongation\\ 
\hline
Grid size & (0.32m, 0.32m, 6.0m)\\
\hline
Head (framework) & CenterPoint-Pillar~\cite{yin2021center} \\
\hline
Loss  & As in CenterPoint-Pillar~\cite{yin2021center}  \\
\hline

Optimizer & Adam~\cite{kingma2014adam} \\
\hline
Weight decay & 0.01\\
\hline
Scheduler &  OneCycleLR~\cite{smith2019super}\\
\hline
Learning rate & 0.006 \\
\hline
Block lr rate  & 0.006 \\

\hline
Batch size & 64 \\
\hline
Epochs & 40\\
\hline
Num GPUs & 16\\
\hline
Data augmentation   & 
   
    $\begin{array}{c}
       \text{Random flip, random rotate} \\
        \text{random scale} \\
    \end{array}$
  
    \\

\bottomrule
\end{tabular}
}
\vspace{-4px}
\caption{
\textbf{Detailed training settings for object detection.} 
}
\label{tab:experimental_settings_det}
\end{table}

\section{Additional Experiments}
\label{app:experiments}

\begin{table}[t]
\centering
\setlength{\tabcolsep}{30pt}
\resizebox{\columnwidth}{!}{
\begin{tabular}{lcc}
\toprule
\cmidrule{1-3}  
 & mIoU & mAcc\\
\midrule	
$w/o$ PointROPE &  79.6 & 86.5\\
\arrayrulecolor{black!10}\midrule\arrayrulecolor{black}
Spherical  &  80.7 & 87.1 \\
Cartesian &  \textbf{82.2}  & \textbf{88.1} \\
\arrayrulecolor{black!10}\midrule\arrayrulecolor{black}
$x\!:\!y\!:\!z\!=\!6\!:\!6\!:\!6$  &  \textbf{82.2}  & \textbf{88.1} \\
$x\!:\!y\!:\!z\!=\!4\!:\!4\!:\!10$ &  80.3  & 86.8 \\
$x\!:\!y\!:\!z\!=\!8\!:\!8\!:\!2$  &   80.3 & 86.7 \\

\bottomrule
\end{tabular}
}
\vspace{-4px}
\caption{
\textbf{Additional ablation on PointROPE on nuScenes.}
}
\label{tab:point_rope_supp}
\end{table}

\subsection{Further Ablation on PointROPE}

\noindent\textbf{Spherical \vs Cartesian coordinates.} In PointROPE, we divide each point's feature embedding into three equal subspaces and then apply the standard 1D RoPE~\cite{su2024roformer} embedding to each subspace using the respective Cartesian coordinates. Here, we investigate an alternative design that uses spherical coordinates. Specifically, we transform each point $(x_{i},y_{i},z_{i})$ into spherical coordinates ($r_{i}$, $\theta_{i}$, $\phi_{i}$), using the mean of all points as the origin. We then apply 1D RoPE using $r_{i}$, $\theta_{i}$ and $\phi_{i}$ separately and concatenate the resulting embeddings. The motivation is that spherical coordinates decouple radial distance and angular structure, which could potentially make positional relationships easier to learn. However, as shown in \cref{tab:point_rope_supp}, we empirically find that \rope{} in spherical coordinates is effective but offers no improvement over Cartesian coordinates, while adding additional computational overhead. Therefore, we retain our simpler per-axis Cartesian design.

\noindent\textbf{Subdivision of the input space.} For each attention head (with head dimension 18), we split the embedding evenly across three axes $(x_{i},y_{i},z_{i})$. Here we explore the impact of different subdivisions on each axis. In addition to equal split ($6\!:\!6\!:\!6$), we try emphasising the $z$ axis ($4\!:\!4\!:\!10$) and emphasising the $xy$ axes ($8\!:\!8\!:\!2$). As shown in \cref{tab:point_rope_supp}, uneven splits lead to suboptimal performance compared with equal weighting. This suggests that positional information along all three axes is similarly important, and manual reweighting is unnecessary.

\noindent\textbf{Other positional encodings.} We additionally compare PointROPE with absolute positional encoding (APE) using sinusoidal functions, and with relative positional encoding (RPE) on nuScenes, in \cref{tab:rebuttal_posenc_supp}.
PointROPE achieves the clearly highest mIoU while introducing the smallest increase in latency.
\begin{table}[ht]
\centering
\setlength{\tabcolsep}{15pt}
\resizebox{\columnwidth}{!}{
\begin{tabular}{lc|ccc}
\toprule
\cmidrule{1-5}  
  & $w/o$ PosEnc & APE (sinusoidal) & RPE & PointROPE  \\
\midrule	
mIoU & 79.6 &  80.2 & 80.6 & \textbf{82.2}  \\
Latency & 20.7$\,$ms$\!\!\!\!\!\!\!\!$ & 23.3$\,$ms$\!\!\!\!\!\!\!\!$ & 27.6$\,$ms$\!\!\!\!\!\!\!\!$ & \textbf{21.5$\,$ms$\!\!\!\!\!\!\!\!$}  \\
\bottomrule
\end{tabular}
}
\vspace{-4px}
\caption{
\textbf{Comparison with other positional encoding schemes on nuScenes.} 
}
\label{tab:rebuttal_posenc_supp}
\end{table}

\subsection{Cross-dataset Transfer}

To evaluate cross-dataset transfer performance, we first pretrain PTv3 and LitePT-S for semantic segmentation on Structured3D, the largest dataset in our benchmark suite. We then freeze the pretrained encoders and test on ScanNet with a linear probe on the multi-scale features. 
As shown in \cref{tab:rebuttal_transfer_supp}, LitePT outperforms PTv3 by $7.9\,$pp in mIoU, indicating better generalisation. We attribute the cross-dataset generalisation to PointROPE, which preserves relative spatial relationships across scenes with differing layouts and sensor characteristics. This property enables the model to learn more transferable representations, reducing overfitting to dataset-specific structures and improving robustness under domain shift.
\begin{table}[t]
\vspace{-0.5em}
\centering
\setlength{\tabcolsep}{25pt}
\resizebox{\columnwidth}{!}{
\begin{tabular}{lccc}
\toprule
\cmidrule{1-4} 
Method & \#Param & mIoU & mAcc  \\
\midrule		
PTv3~\cite{wu2024point} & 46.1M & 57.1 & 69.5  \\
\name{}-S & 12.7M & \textbf{65.0} & \textbf{77.2}  \\
\bottomrule
\end{tabular}
}
\vspace{-4px}
\caption{
\textbf{Cross-dataset transfer.} Models are pretrained on Structured3D and evaluated on ScanNet validation set using linear probing.
}
\label{tab:rebuttal_transfer_supp}
\end{table}

\subsection{Comparison with PointMamba}
We additionally compare LitePT with the Mamba-based point cloud backbone PointMamba (PM)~\cite{liang2024pointmamba}. For small, object-centric datasets, PM is indeed a viable alternative, see shape classification results for ModelNet40 in \textcolor{black}{\cref{tab:rebuttal_pointmamba_supp} \circlenumblack{1}}. Both models are trained from scratch and evaluated without voting.

\begin{figure}[h]
    \centering
    \begin{minipage}{0.49\columnwidth}
        \centering
        \includegraphics[width=\linewidth]{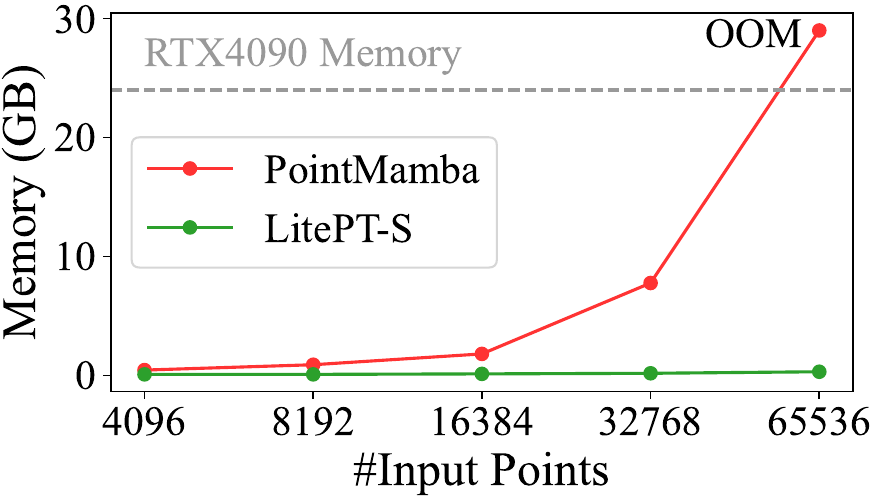}
    \end{minipage}
    \hfill
    \begin{minipage}{0.49\columnwidth}
        \centering
        \includegraphics[width=\linewidth]{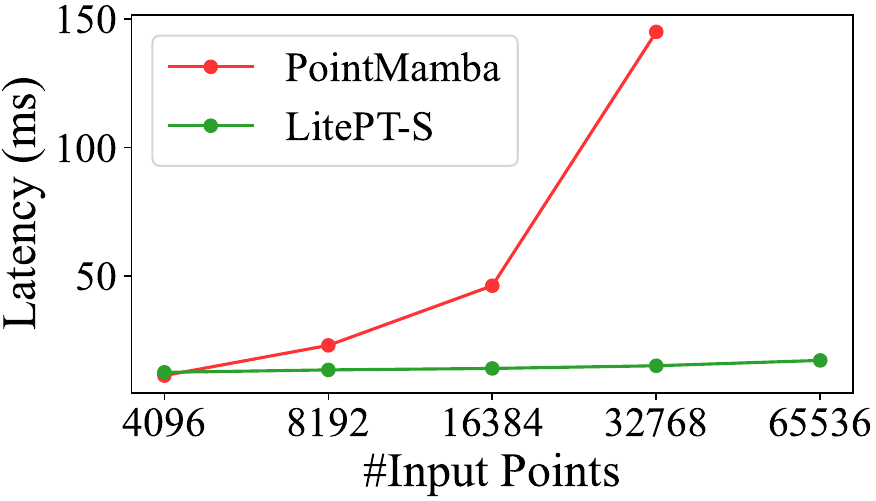}
    \end{minipage}
    \vspace{-4px}
    \caption{\textbf{LitePT-S vs.\ PointMamba efficiency on ScanNet.}}
    \label{fig:rebuttal_pointmamba_supp}
\end{figure}

To adapt PM to scene-level semantic segmentation, we fix the $\tfrac{\mathtt{\#group\_tokens}}{\mathtt{\#input\_points}}$ ratio of the best-reported model and progressively increase the number of input points (\cref{fig:rebuttal_pointmamba_supp}). 
We observe that PM’s reliance on farthest point sampling (FPS), $k$-NN, and PointNet-style feature propagation leads to rapid growth in compute and memory, limiting its scalability to larger scenes. 
In \textcolor{black}{\cref{tab:rebuttal_pointmamba_supp} \circlenumblack{2}}, we train PM on ScanNet with $640$ tokens per scene, the maximum feasible on an RTX 4090, resulting in a $16\,$pp mIoU drop compared to LitePT.
\begin{table}[ht]
\centering
\setlength{\tabcolsep}{12pt}
\resizebox{\columnwidth}{!}{
\begin{tabular}{lcccc|c}
\toprule
\cmidrule{1-6}  
& 
\multicolumn{4}{c}{\circlenumblack{1} ModelNet40} & \multicolumn{1}{c}{\circlenumblack{2} ScanNet} \\
\cmidrule(r){2-5}	\cmidrule(r){6-6} 
  & \#Params & Latency & Memory & OA (\%) & mIoU \\
\midrule	
PointMamba & 12.3$\,$M & \textbf{10.5$\,$ms} & 0.20$\,$G & 92.4 & 60.5 \\
\name{}-S (Ours) & 12.7$\,$M  & 11.7$\,$ms & \textbf{0.13$\,$G} & \textbf{92.5} & \textbf{76.5} \\
\bottomrule
\end{tabular}
}
\vspace{-4px}
\caption{
\textbf{Comparison with PointMamba.}}
\label{tab:rebuttal_pointmamba_supp}
\end{table}

\subsection{Additional Results for LitePT-L}

In \cref{tab:rebuttal_liteptl_supp}, we show LitePT-L results on key benchmarks. We note that, in accordance with established scaling laws, model scaling should be considered in conjunction with data volume. Upgrading from LitePT-S to LitePT-L consistently improves performance. As expected, the gain strongly depends on the dataset size (as denoted by \#Points). 
For instance, scaling up to LitePT-L yields a +1.8$\,$pp gain in mIoU for Structured3D (9.5 billion points) whereas the improvement is only +0.4$\,$pp for nuScenes (1 billion points), as there is not enough data to fully exploit the higher capacity.
\begin{table}[ht]
\centering
\setlength{\tabcolsep}{2pt}
\resizebox{\columnwidth}{!}{
\begin{tabular}{lcccc}
\toprule
\cmidrule{1-5} 
 & 
\multicolumn{3}{c}{Semantic Segmentation (mIoU)} & \multicolumn{1}{c}{Instance Segmentation ($\text{mAP}$)} \\
\cmidrule(r){2-4}	\cmidrule(r){5-5} 
 & Structured3D & Waymo  & nuScenes   & ScanNet \\
 \#Points (Billion) &  9.55   & 4.15 & 0.98  & 0.17 \\
\midrule	
PTv3 &  82.4  & 71.3 & 80.4 & 40.9 \\
\arrayrulecolor{black!10}\midrule\arrayrulecolor{black}
\name{}-S &  83.6  & 73.1 & 82.2 & 41.7 \\
\name{}-L &  \hphantom{1111}\textbf{85.4}  \gapp{(+1.8)} & \hphantom{1111}\textbf{74.0}  \gapp{(+0.9)} & \hphantom{1111}\textbf{82.6}  \gapp{(+0.4)} &  \hphantom{1111}\textbf{42.3}  \gapp{(+0.6)} \\

\bottomrule
\end{tabular}
}
\vspace{-4px}
\caption{
\textbf{More results for LitePT-L.} 
}
\label{tab:rebuttal_liteptl_supp}
\end{table}

\subsection{Benchmarking Outdoor Efficiency}

We evaluate the efficiency of semantic segmentation on Waymo, a large-scale outdoor LiDAR dataset, in \cref{tab:rebuttal_outdoor_efficiency_supp}. Similar to our observation on indoor datasets, LitePT-S remains more efficient than PTv3, with $>2\times$ faster runtime at inference.
\begin{table}[ht]
\vspace{-6px}
\centering
\setlength{\tabcolsep}{10pt}
\resizebox{\columnwidth}{!}{
\begin{tabular}{lcccccc}
\toprule
\cmidrule{1-7} 
 & & 
\multicolumn{2}{c}{Training} & \multicolumn{2}{c}{Inference} \\
\cmidrule(r){3-4}	\cmidrule(r){5-6} 
Method & \#Params & Latency & Memory & Latency & Memory & mIoU \\
\midrule	
PTv3 & 46.1$\,$M & $\!\!\!$153$\,$ms & $\!\!\!$14.1$\,$G & $\!\!\!$100$\,$ms & 7.1$\,$G & 71.3\\
\arrayrulecolor{black!10}\midrule\arrayrulecolor{black}
\name{}-S (Ours) & 12.7$\,$M & \textbf{86$\,$ms} & \textbf{9.4$\,$G} & \textbf{47$\,$ms} & \textbf{5.4$\,$G} & \textbf{73.1}\\
\bottomrule
\end{tabular}
}
\vspace{-4px}
\caption{
\textbf{Outdoor Efficiency comparison on Waymo.} Results are reported as average over the full Waymo dataset using a single RTX 4090 GPU. Automatic Mixed Precision (AMP) is enabled for all models during training and disabled during inference.
}
\label{tab:rebuttal_outdoor_efficiency_supp}
\end{table}

\subsection{Chunking and Test-Time Augmentation}

In the main paper, we report semantic segmentation results following the same evaluation protocol as prior works~\cite{wu2022point,wu2024point} to ensure a fair comparison. The testing pipeline applies chunking and test-time augmentations (TTA). 
Specifically, each augmented sample is partitioned into overlapping chunks, ensuring that every point is assigned to at least one chunk during grid sampling. 
The model is then run on each chunk individually, and the final label of each point is aggregated by voting across the predictions from all chunks it appears in. Although this multi-run and TTA protocol is common practice and is known to boost performance~\cite{simonyan2014very}, it obscures the intrinsic merits of the underlying backbone.
To communicate performance in a simpler single-pass setting useful for downstream users, we additionally report results for PTv3 and \name{}-S without TTA or chunking in \cref{tab:nuscenes_sem_wo_tta}.
Overall, removing chunking and TTA reduces performance by roughly 2 percentage points in mIoU for both methods.

\begin{table}[t]
\vspace{-0.5em}
\centering
\setlength{\tabcolsep}{15pt}
\resizebox{\columnwidth}{!}{
\begin{tabular}{lccc}
\toprule
\cmidrule{1-4} 
Method & \#Param & mIoU & mAcc  \\
\midrule		
PTv3~\cite{wu2024point} & 46.1M & 80.4 & 87.2 \\
\name{}-S & 12.7M & \textbf{82.2} & \textbf{88.1}  \\

\arrayrulecolor{black!10}\midrule\arrayrulecolor{black}

PTv3~\cite{wu2024point} ($w/o$ chunking and TTA) & 46.1M & 78.3 & 86.0 \\
\name{}-S ($w/o$ chunking and TTA)  & 12.7M & \textbf{80.4} & \textbf{86.9}  \\
\bottomrule
\end{tabular}
}
\vspace{-0.6em}
\caption{
\textbf{Semantic segmentation on nuScenes without chunking and TTA.}
}
\label{tab:nuscenes_sem_wo_tta}
\end{table}

\section{Visualisation}
\label{app:visualization}

\begin{figure*}
    \centering
    \begin{tabular}{@{}lccc@{}}

    \multicolumn{4}{c}{

    \begin{minipage}{\textwidth}
    \centering
        {\textcolor{nuscenes_barrier}{\ding{108}}}\,\,barrier\,\, 
        {\textcolor{nuscenes_bicycle}{\ding{108}}}\,\,bicycle\,\,
        {\textcolor{nuscenes_bus}{\ding{108}}}\,\,bus\,\, 
        {\textcolor{nuscenes_car}{\ding{108}}}\,\,car\,\, 
        {\textcolor{nuscenes_construction_veh}{\ding{108}}}\,\,construction vehicle\,\,
        {\textcolor{nuscenes_motorcycle}{\ding{108}}}\,\,motorcycle\,\, 
        {\textcolor{nuscenes_pedestrian}{\ding{108}}}\,\,pedestrian\,\, 
        {\textcolor{nuscenes_traffic_cone}{\ding{108}}}\,\,traffic cone\,\, 
        {\textcolor{nuscenes_trailer}{\ding{108}}}\,\,trailer\,\, 
        {\textcolor{nuscenes_truck}{\ding{108}}}\,\,truck\,\, 
        {\textcolor{nuscenes_driveable_surface}{\ding{108}}}\,\,driveable surface\,\, 
        {\textcolor{nuscenes_other_flat}{\ding{108}}}\,\,other flat surface\,\, 
        {\textcolor{nuscenes_sidewalk}{\ding{108}}}\,\,sidewalk\,\, 
        {\textcolor{nuscenes_terrain}{\ding{108}}}\,\,terrain\,\, 
        {\textcolor{nuscenes_manmade}{\ding{108}}}\,\,manmade\,\, 
        {\textcolor{nuscenes_vegetation}{\ding{108}}}\,\,vegetation\,\, 
        {\textcolor{scannet_unlabeled}{\ding{108}}}\,\,unlabelled
        
    \end{minipage}
    } 
    \vspace{15pt}
    
    \\ 
    
    \rotatebox{90}{\scriptsize \texttt{1ccdbec944bd4994...}}
    &
    \begin{subfigure}[b]{0.28\textwidth}
      \includegraphics[width=\linewidth]{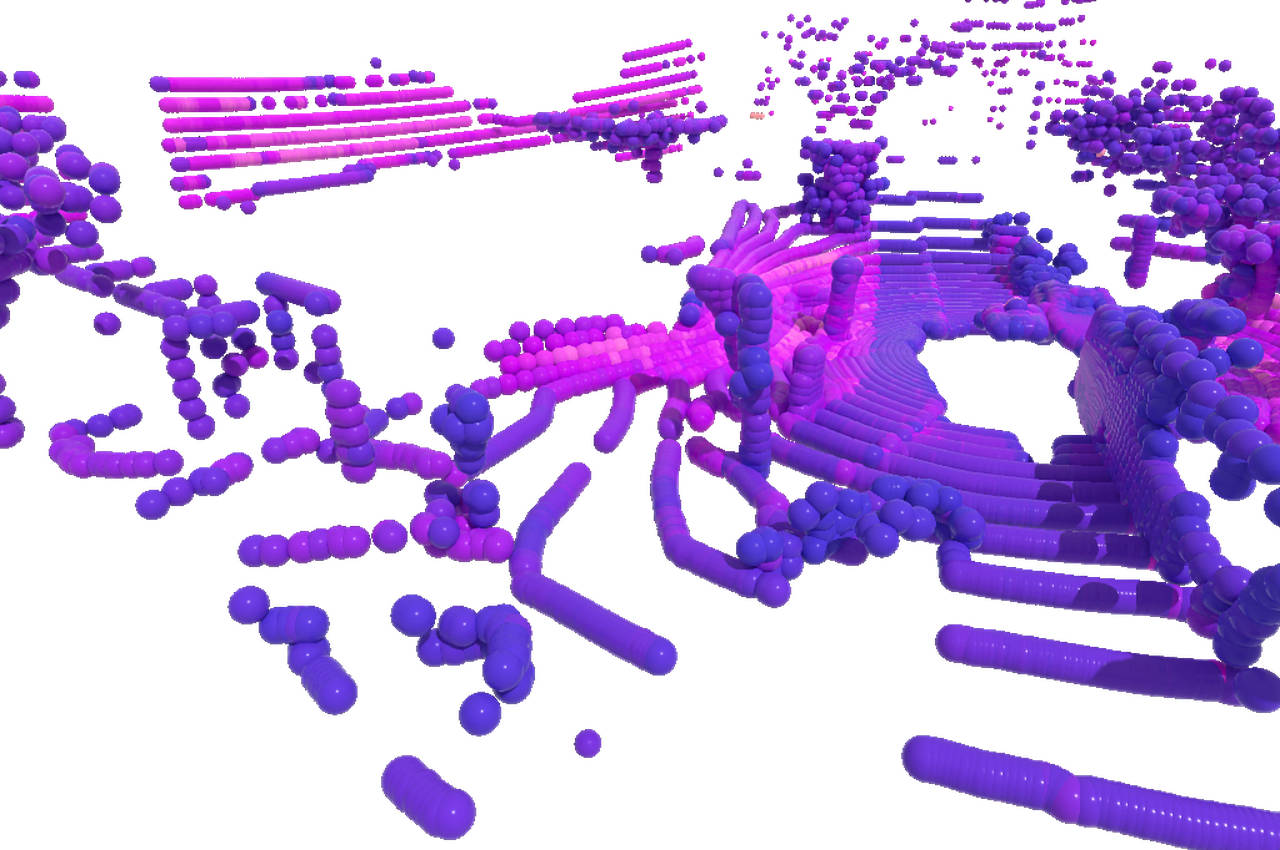}
    \end{subfigure}
         & 
    \begin{subfigure}[b]{0.28\textwidth}
      \includegraphics[width=\linewidth]{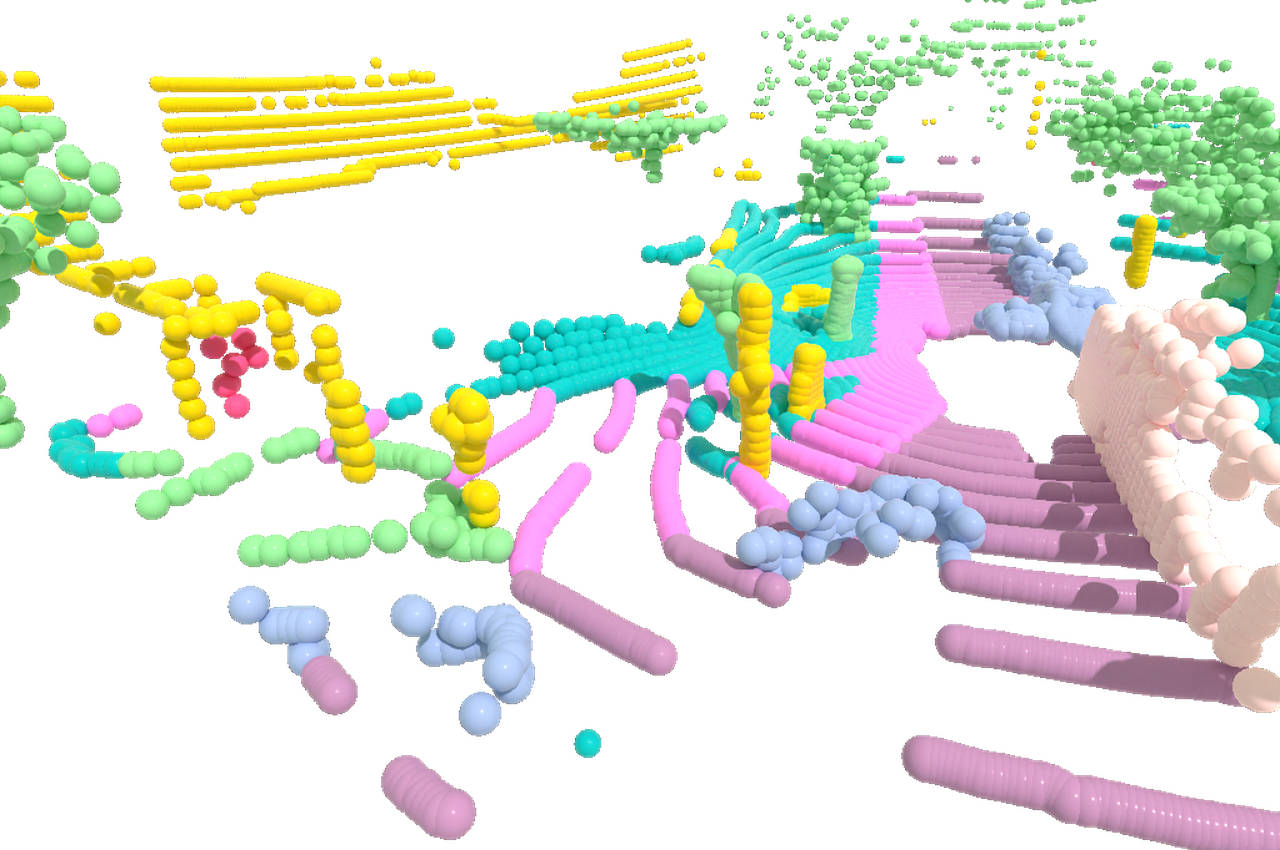}
    \end{subfigure}
         & 
    \begin{subfigure}[b]{0.28\textwidth}
      \includegraphics[width=\linewidth]{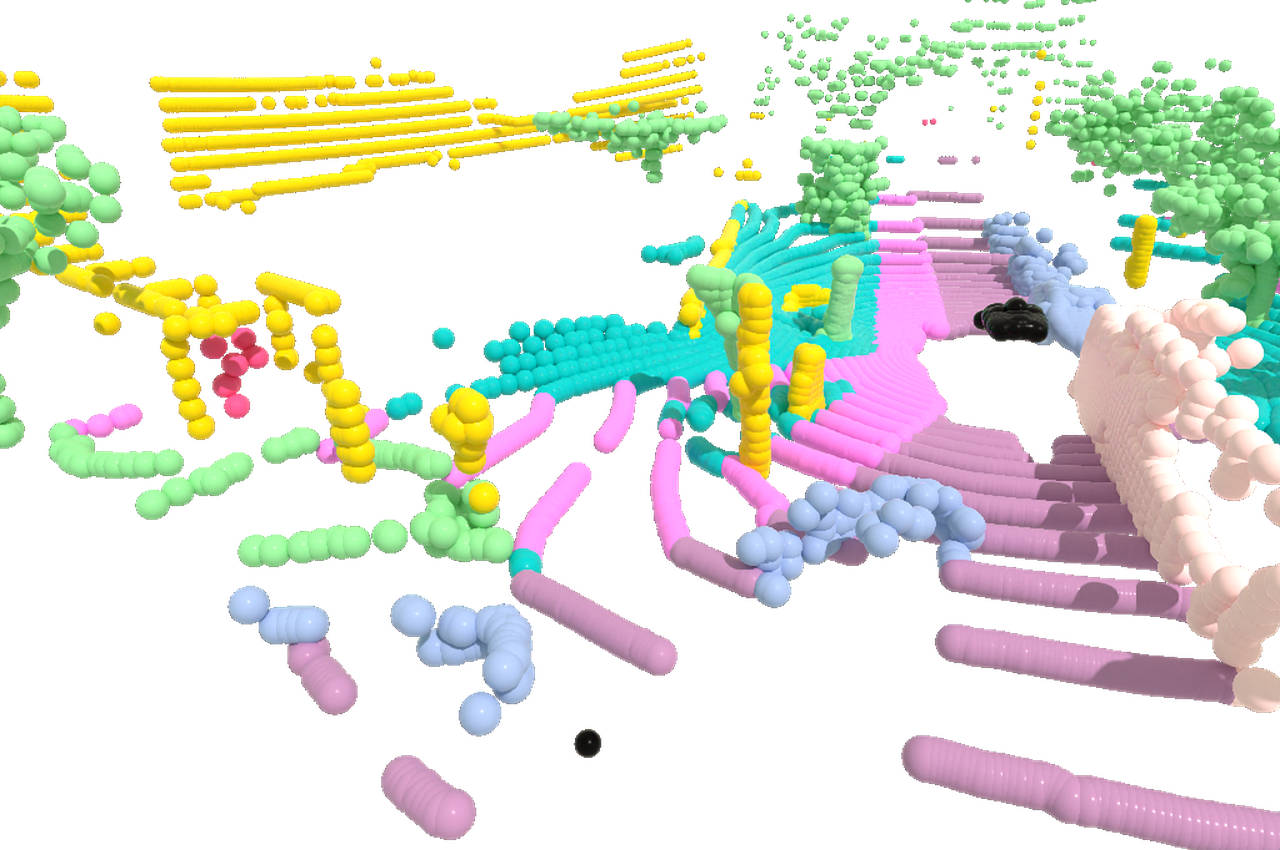}
    \end{subfigure}  \\

    \rotatebox{90}{\scriptsize \texttt{2f678cb1e67d42ae...}}
    &
    \begin{subfigure}[b]{0.28\textwidth}
      \includegraphics[width=\linewidth]{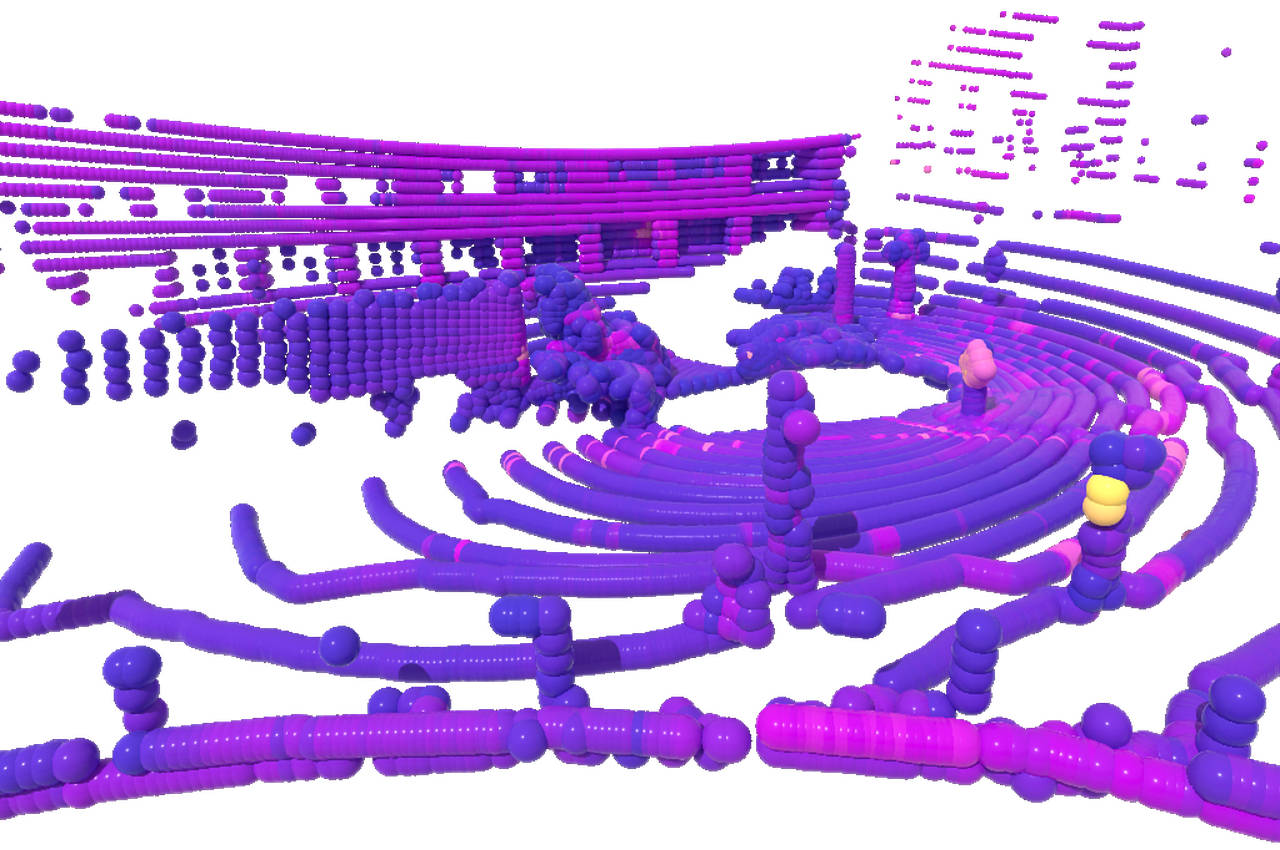}
    \end{subfigure}
         & 
    \begin{subfigure}[b]{0.28\textwidth}
      \includegraphics[width=\linewidth]{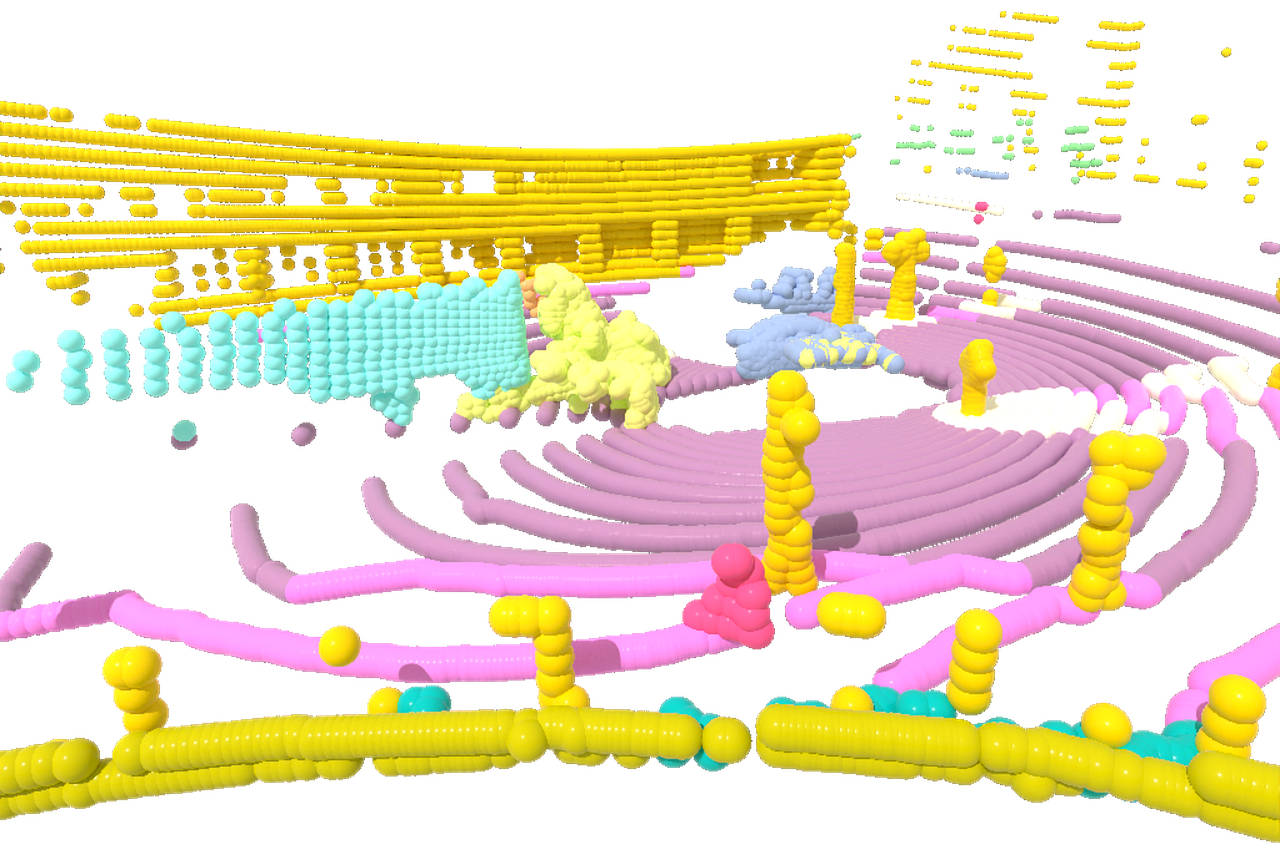}
    \end{subfigure}
         & 
    \begin{subfigure}[b]{0.28\textwidth}
      \includegraphics[width=\linewidth]{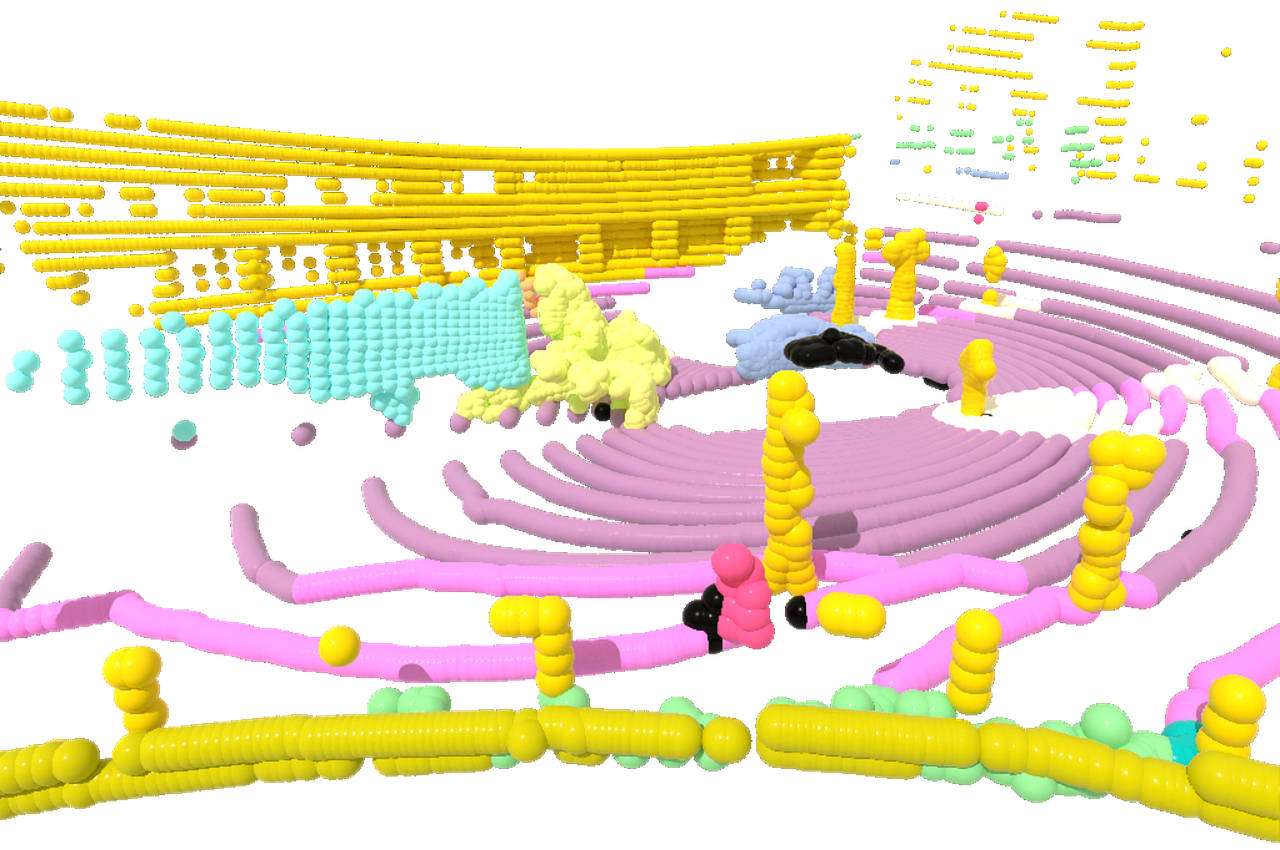}
    \end{subfigure}  \\

    \rotatebox{90}{\scriptsize \texttt{5f8393250fae4960...}}
    &
    \begin{subfigure}[b]{0.28\textwidth}
      \includegraphics[width=\linewidth]{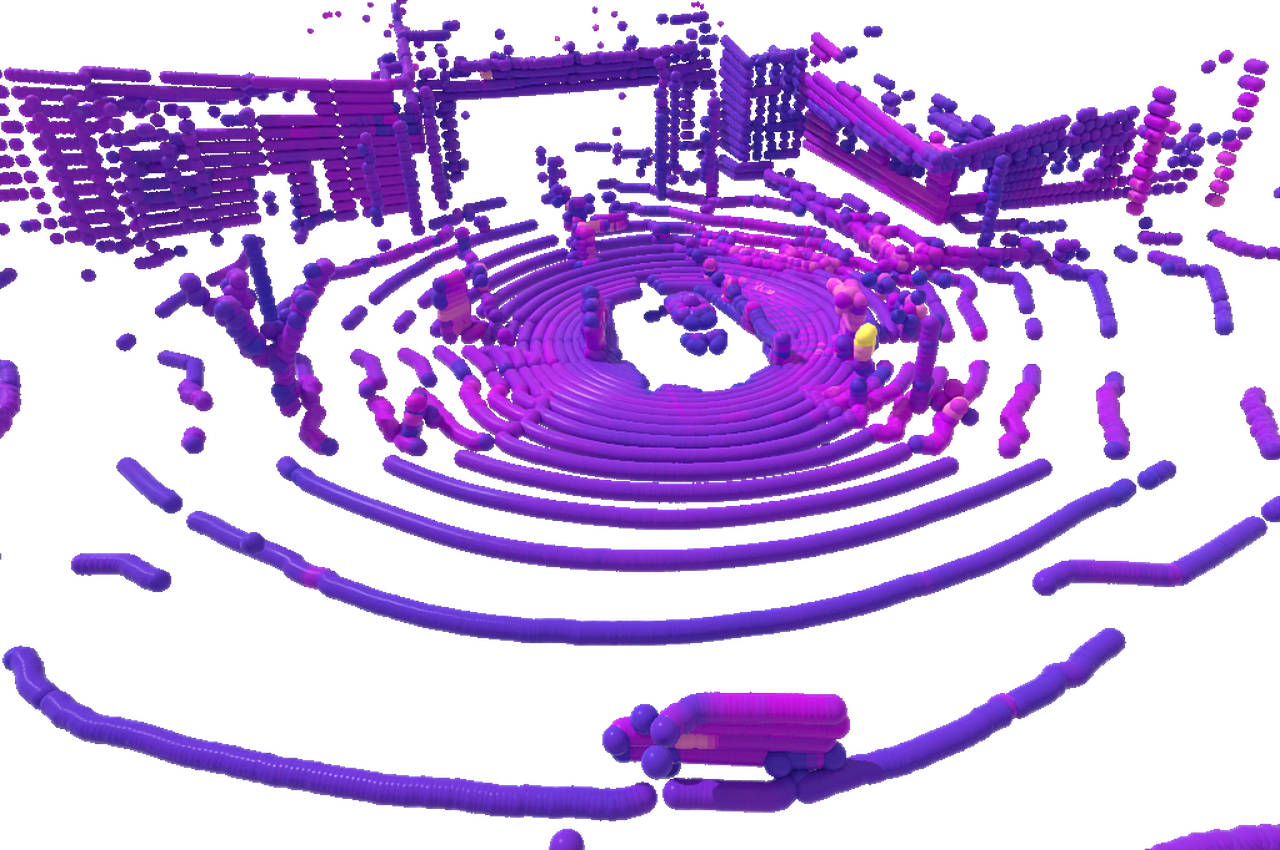}
    \end{subfigure}
         & 
    \begin{subfigure}[b]{0.28\textwidth}
      \includegraphics[width=\linewidth]{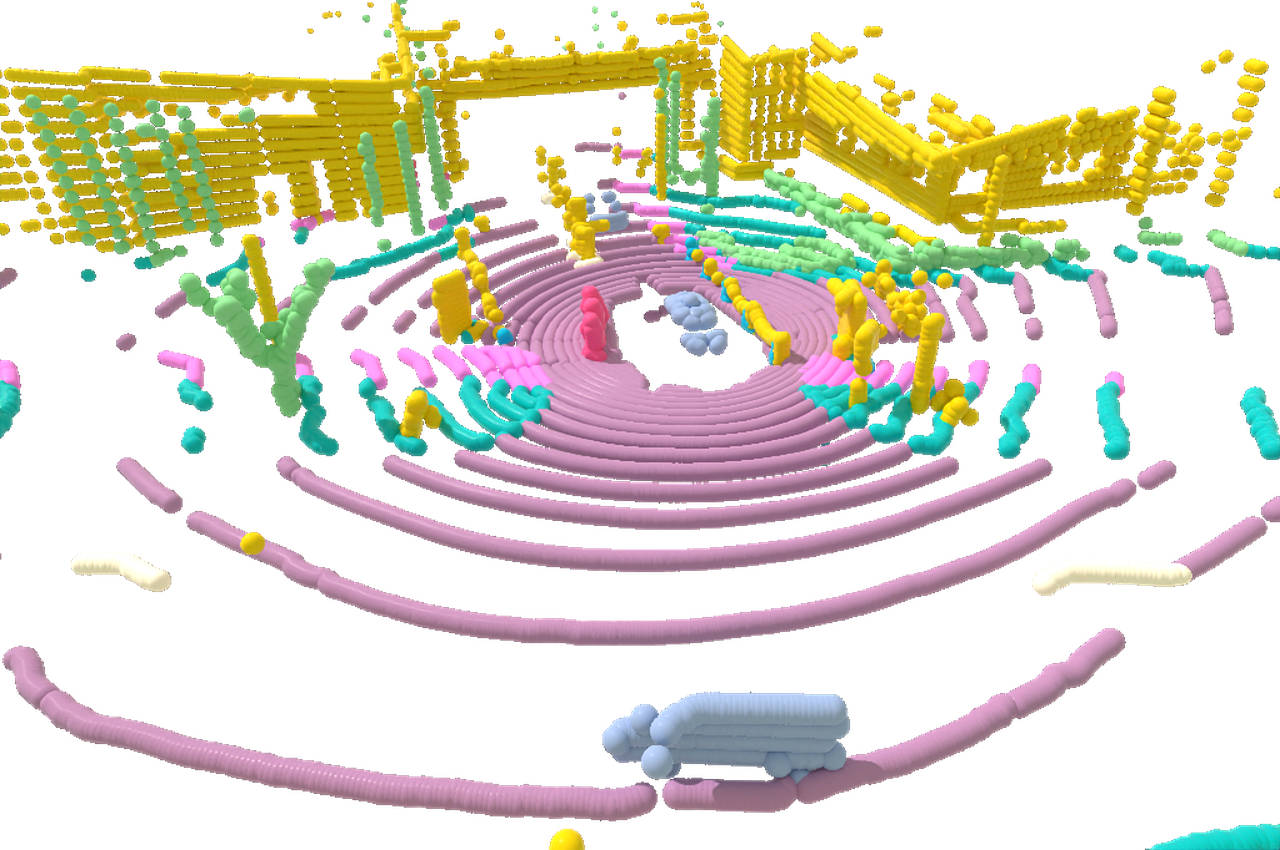}
    \end{subfigure}
         & 
    \begin{subfigure}[b]{0.28\textwidth}
      \includegraphics[width=\linewidth]{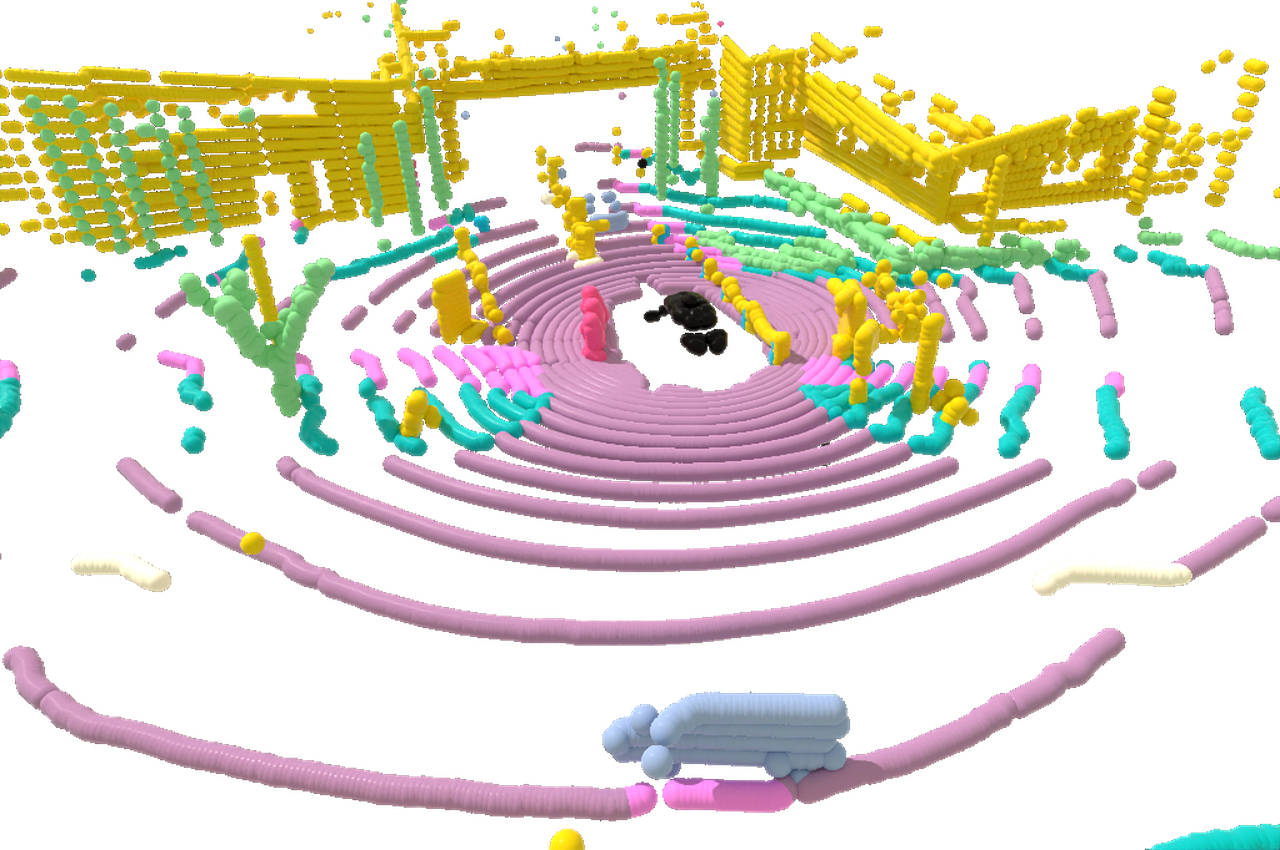}
    \end{subfigure}  \\

    \rotatebox{90}{\scriptsize \texttt{6bfd64d077884228...}}
    &
    \begin{subfigure}[b]{0.28\textwidth}
      \includegraphics[width=\linewidth]{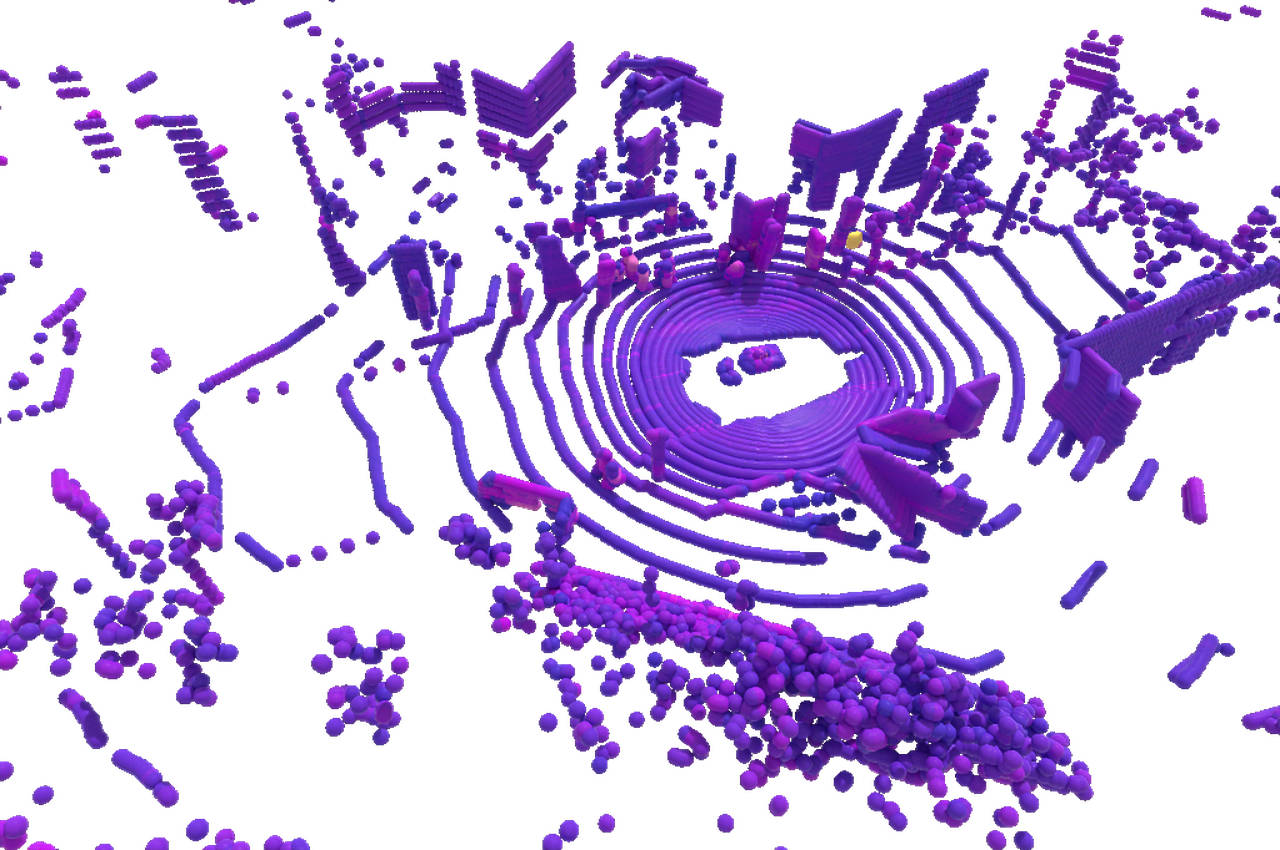}
    \end{subfigure}
         & 
    \begin{subfigure}[b]{0.28\textwidth}
      \includegraphics[width=\linewidth]{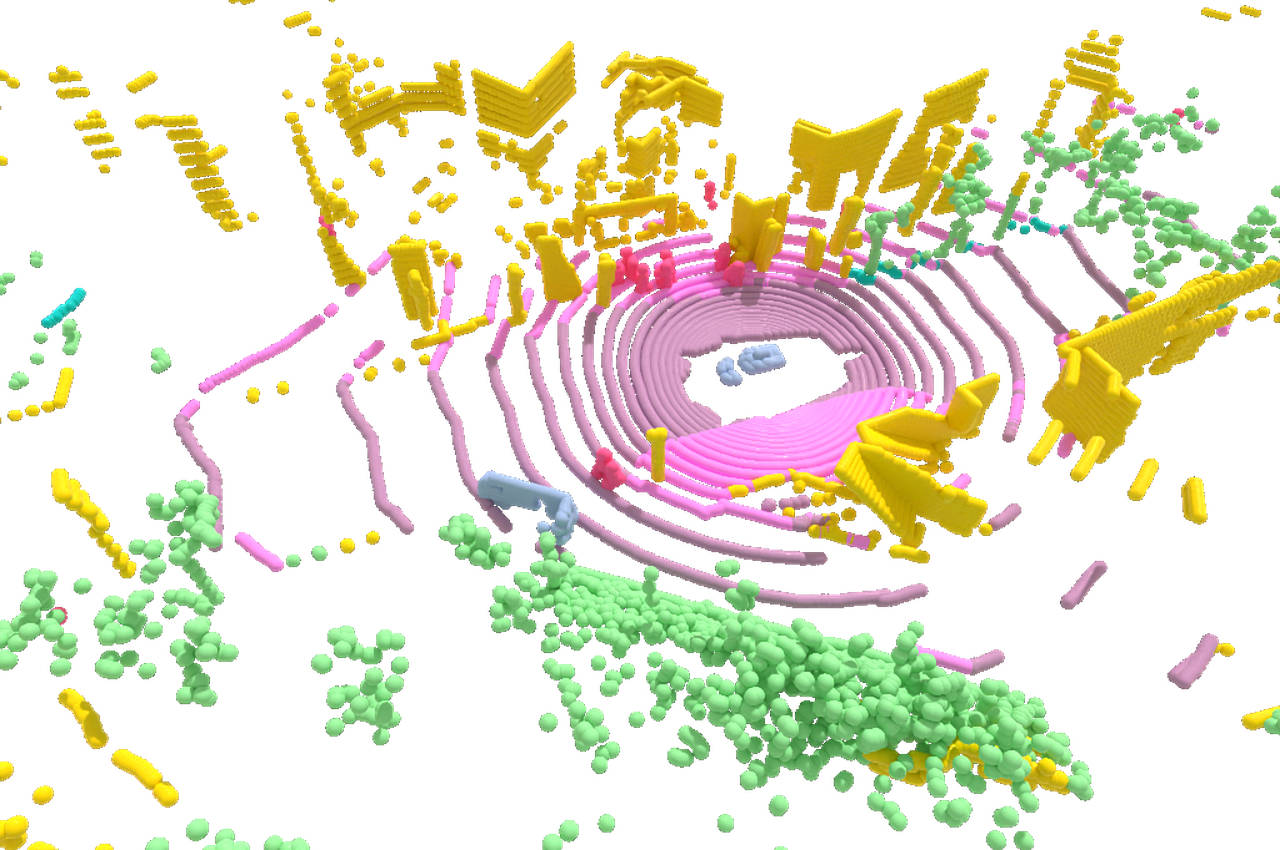}
    \end{subfigure}
         & 
    \begin{subfigure}[b]{0.28\textwidth}
      \includegraphics[width=\linewidth]{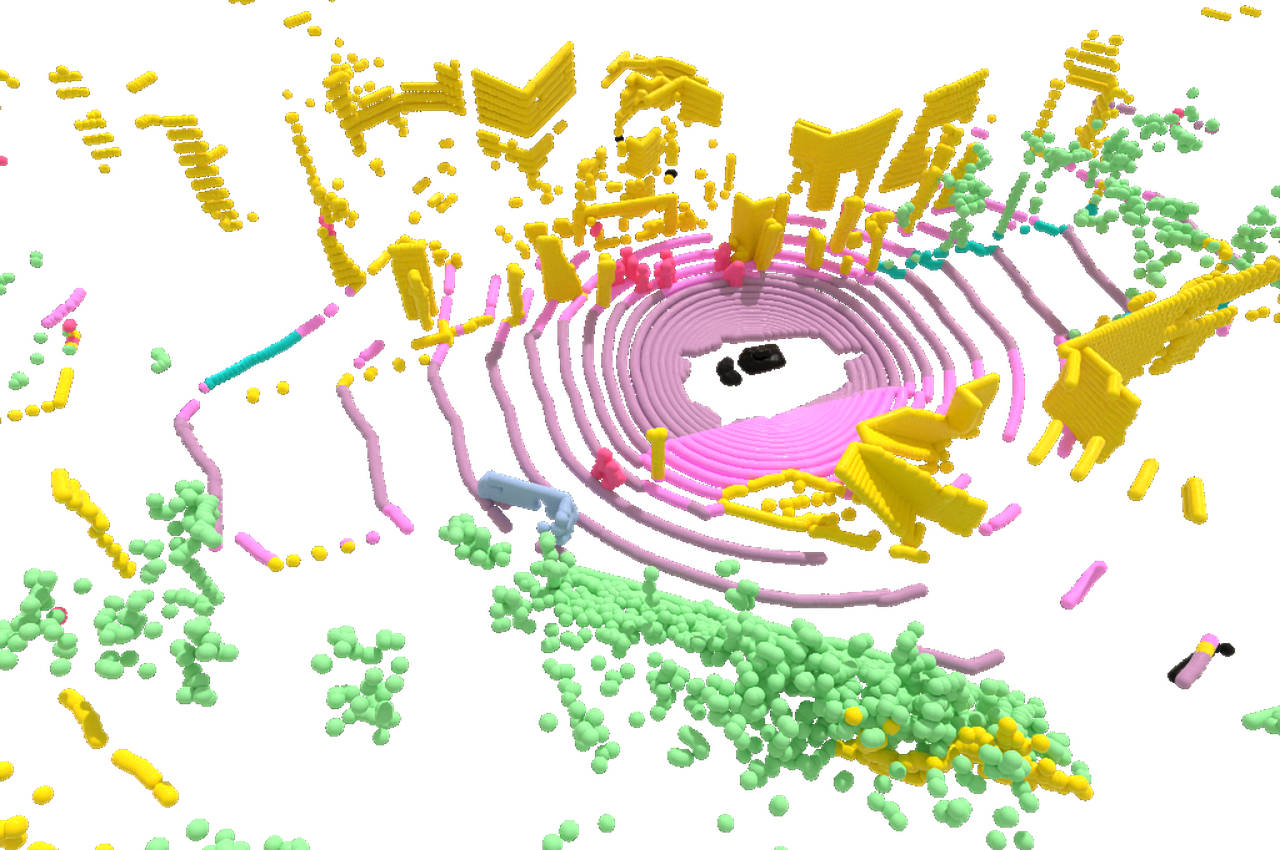}
    \end{subfigure}  \\

    \rotatebox{90}{\scriptsize \texttt{8f78c446a68d4854...}}
    &
    \begin{subfigure}[b]{0.28\textwidth}
      \includegraphics[width=\linewidth]{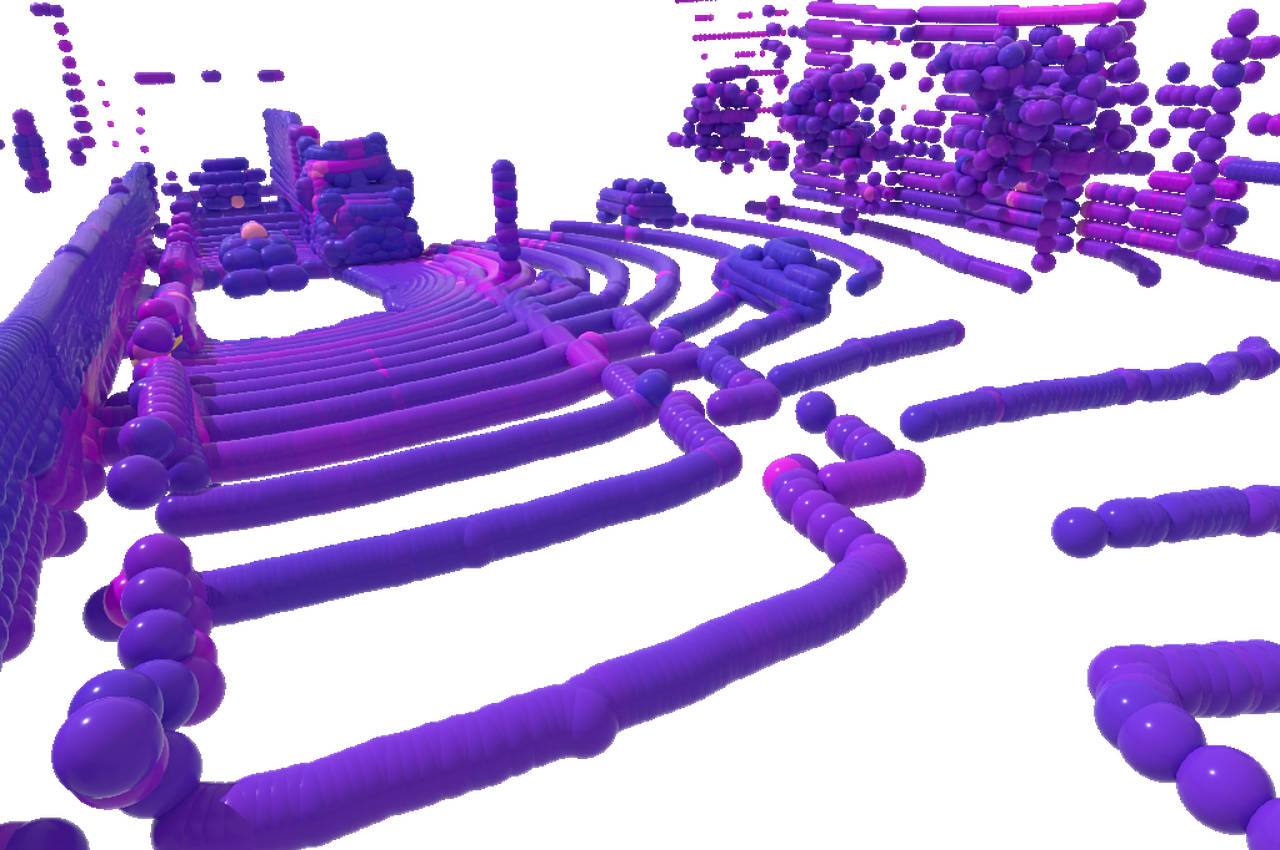}
    \end{subfigure}
         & 
    \begin{subfigure}[b]{0.28\textwidth}
      \includegraphics[width=\linewidth]{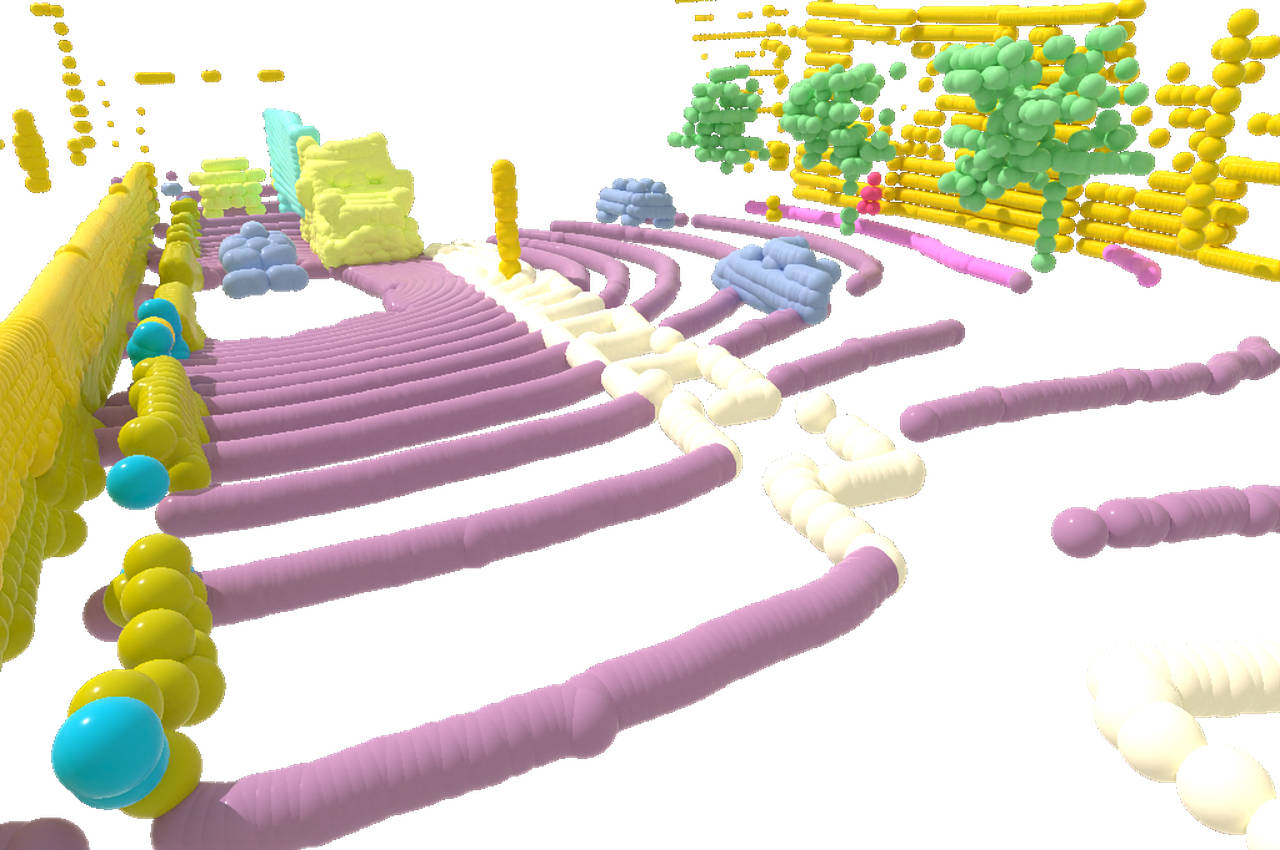}
    \end{subfigure}
         & 
    \begin{subfigure}[b]{0.28\textwidth}
      \includegraphics[width=\linewidth]{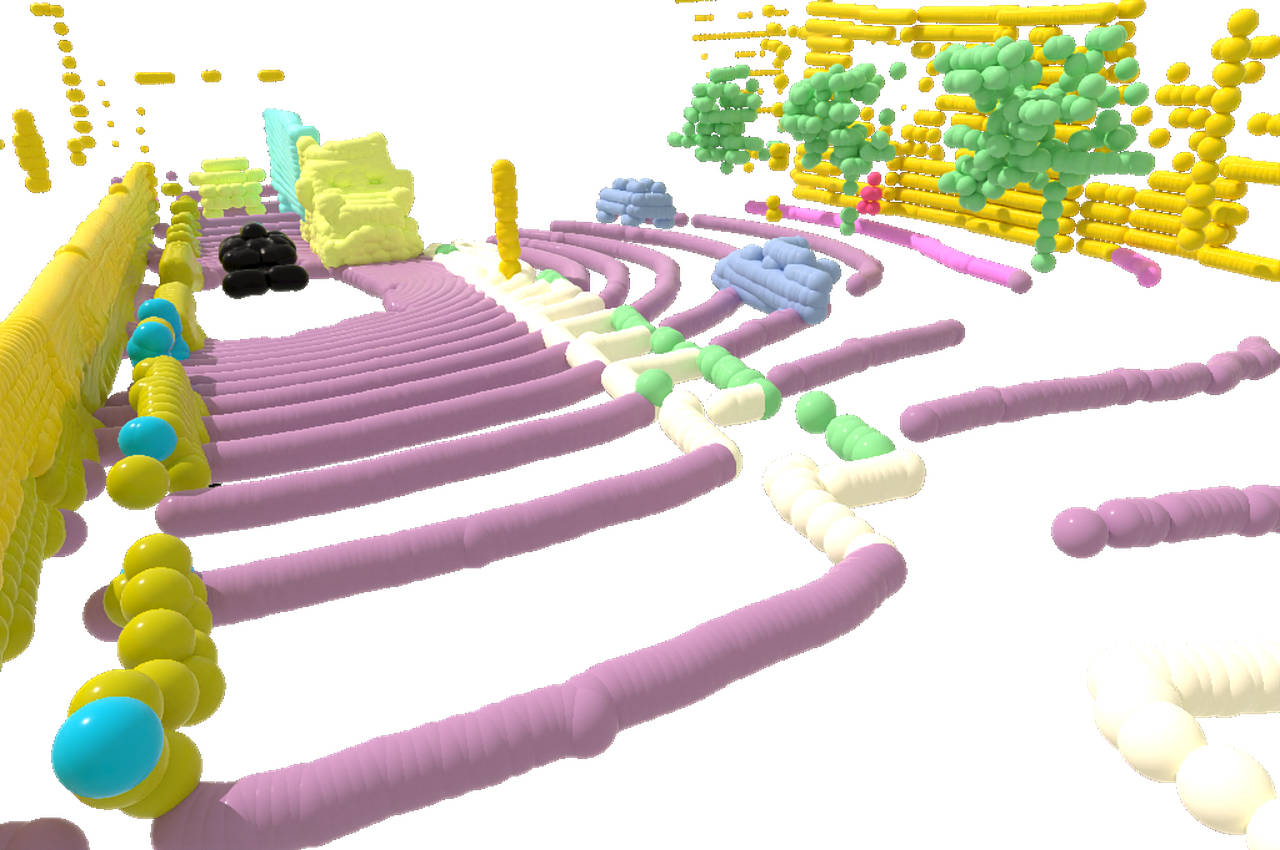}
    \end{subfigure}  \\

    \rotatebox{90}{\scriptsize \texttt{049d115cb992491b...}}
    &
    \begin{subfigure}[b]{0.28\textwidth}
      \includegraphics[width=\linewidth]{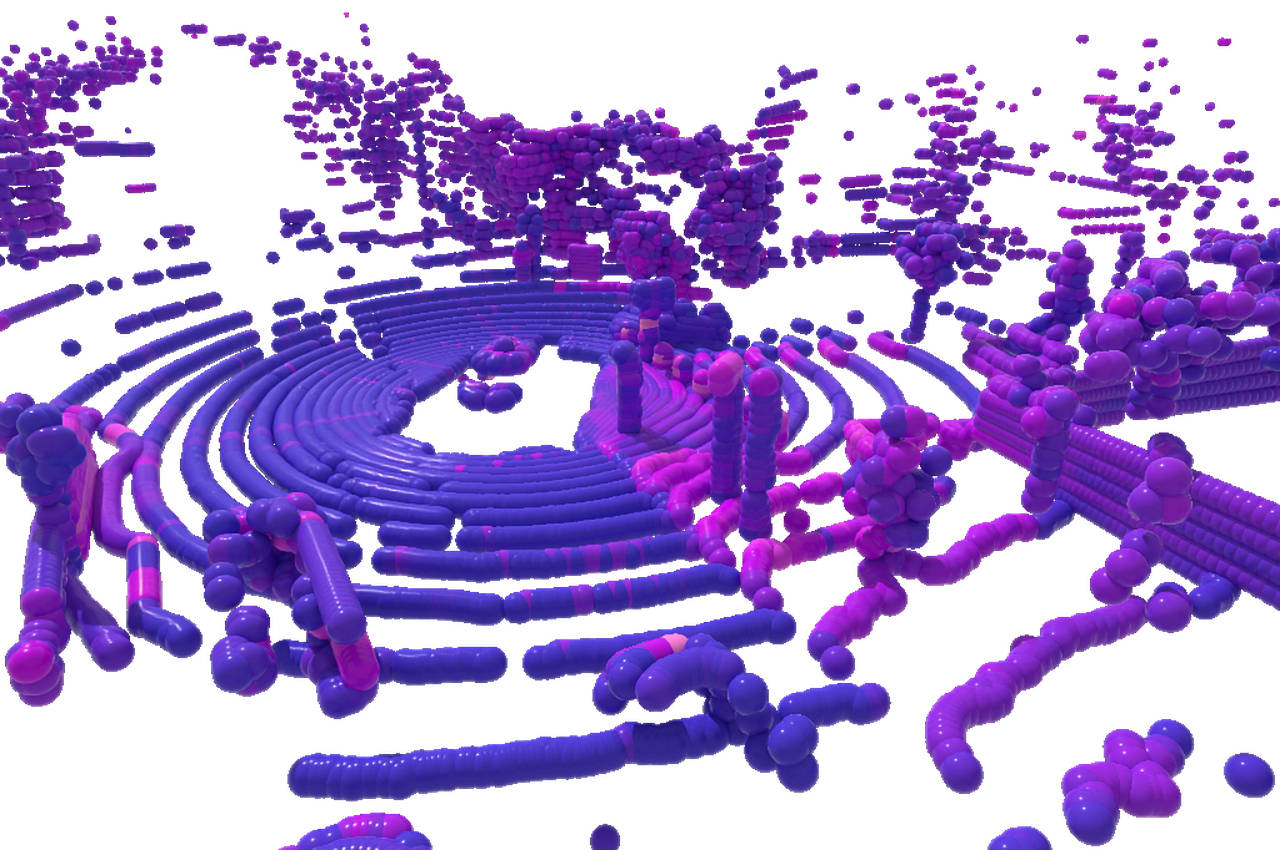}
      \caption{Input}
      \label{fig:quali_nuscenes_semseg:input}
    \end{subfigure}
         & 
    \begin{subfigure}[b]{0.28\textwidth}
      \includegraphics[width=\linewidth]{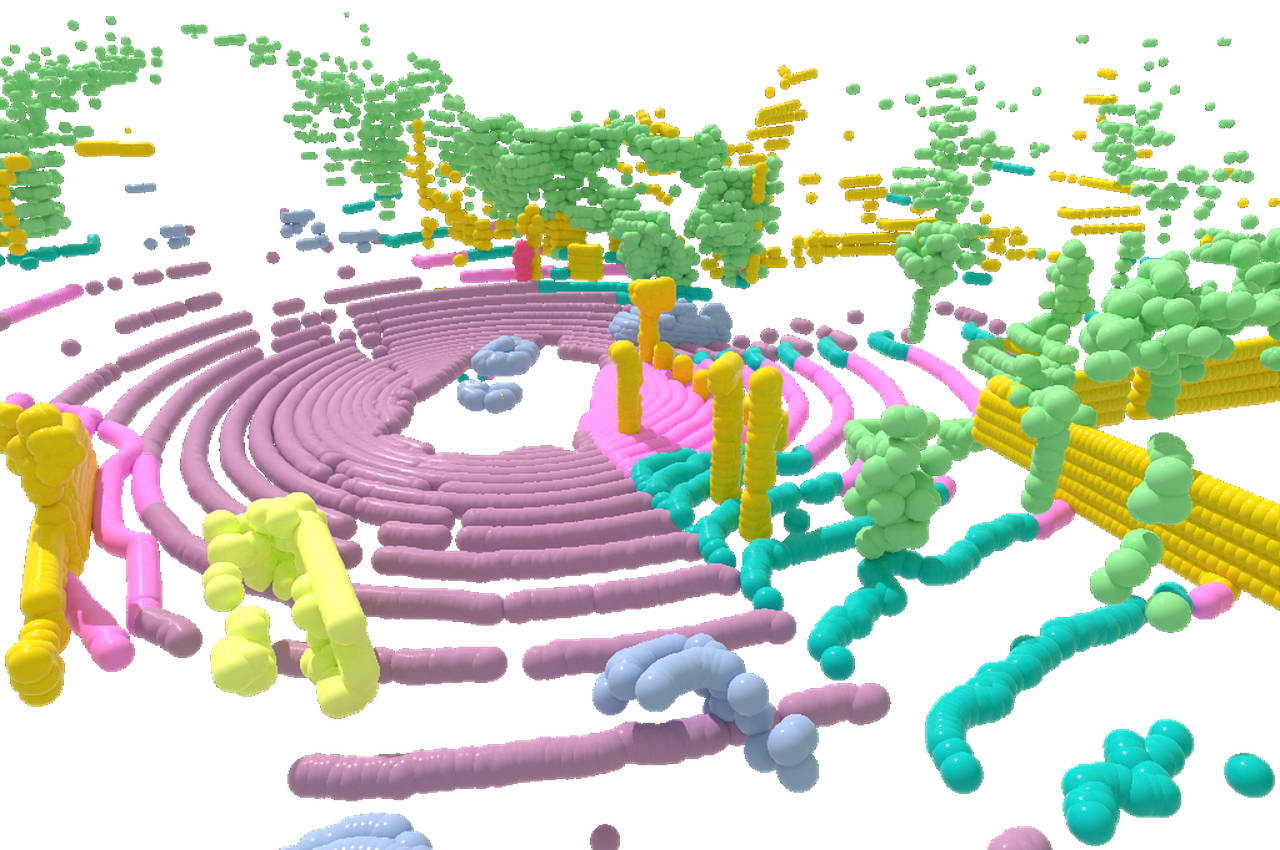}
      \caption{Prediction}
      \label{fig:quali_nuscenes_semseg:pred}
    \end{subfigure}
         & 
    \begin{subfigure}[b]{0.28\textwidth}
      \includegraphics[width=\linewidth]{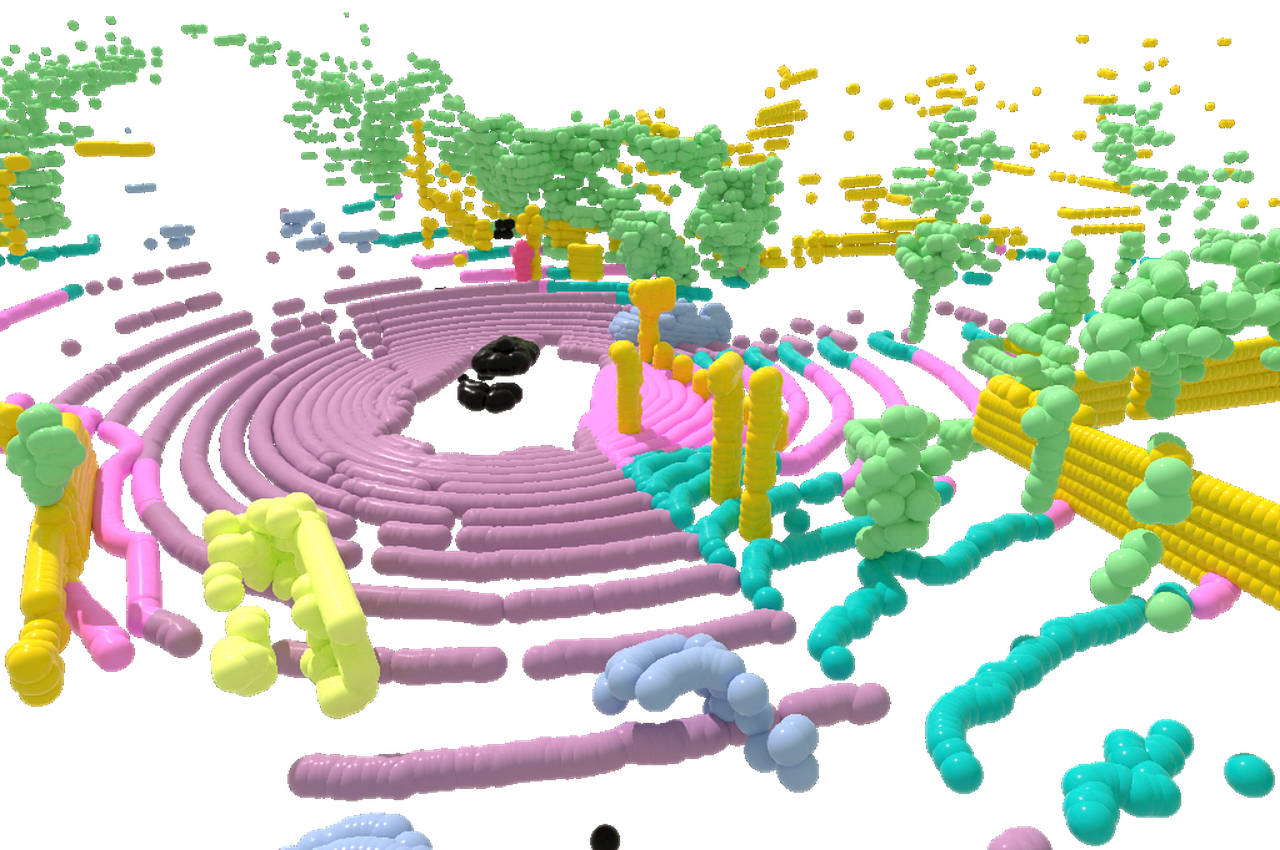}
      \caption{Ground Truth}
      \label{fig:quali_nuscenes_semseg:gt}
    \end{subfigure}  \\

    \end{tabular}
    \caption{
    {\bf nuScenes semantic segmentation.} 
    We present various scenes of the nuScenes validation set: the input point cloud coloured by LiDAR intensity, the semantic segmentation from \name{}-S, and the corresponding ground truth. 
    }
    \label{fig:quali_nuscenes_semseg}
\end{figure*}

\begin{figure*}
    \centering
    \begin{tabular}{@{}lccc@{}}

    \multicolumn{4}{c}{

    \begin{minipage}{\textwidth}
    \centering
        {\textcolor{waymo_car}{\ding{108}}}\,\,car\,\, 
        {\textcolor{waymo_truck}{\ding{108}}}\,\,truck\,\, 
        {\textcolor{waymo_bus}{\ding{108}}}\,\,bus\,\, 
        {\textcolor{waymo_other_veh}{\ding{108}}}\,\,other vehicle\,\, 
        {\textcolor{waymo_motorcyclist}{\ding{108}}}\,\,motorcyclist\,\, 
        {\textcolor{waymo_bicyclist}{\ding{108}}}\,\,bicyclist\,\, 
        {\textcolor{waymo_pedestrian}{\ding{108}}}\,\,pedestrian\,\, 
        {\textcolor{waymo_sign}{\ding{108}}}\,\,sign\,\, 
        {\textcolor{waymo_traffic_light}{\ding{108}}}\,\,traffic light\,\, 
        {\textcolor{waymo_traffic_pole}{\ding{108}}}\,\,traffic pole\,\, 
        {\textcolor{waymo_construction_cone}{\ding{108}}}\,\,construction cone\,\, 
        {\textcolor{waymo_bicycle}{\ding{108}}}\,\,bicycle\,\, 
        {\textcolor{waymo_motorcycle}{\ding{108}}}\,\,motorcycle\,\, 
        {\textcolor{waymo_building}{\ding{108}}}\,\,building\,\, 
        {\textcolor{waymo_vegetation}{\ding{108}}}\,\,vegetation\,\, 
        {\textcolor{waymo_tree_trunk}{\ding{108}}}\,\,tree trunk\,\, 
        {\textcolor{waymo_curb}{\ding{108}}}\,\,curb\,\, 
        {\textcolor{waymo_road}{\ding{108}}}\,\,road\,\, 
        {\textcolor{waymo_lane_marker}{\ding{108}}}\,\,lane marker\,\, 
        {\textcolor{waymo_other_ground}{\ding{108}}}\,\,other ground\,\, 
        {\textcolor{waymo_walkable}{\ding{108}}}\,\,walkable\,\, 
        {\textcolor{waymo_sidewalk}{\ding{108}}}\,\,sidewalk\,\, 
        {\textcolor{scannet_unlabeled}{\ding{108}}}\,\,unlabelled
        
    \end{minipage}
    } 
    \vspace{15pt}
    
    \\ 

    \rotatebox{90}{\scriptsize \texttt{3077229433993844...}}
    &
    \begin{subfigure}[b]{0.28\textwidth}
      \includegraphics[width=\linewidth]{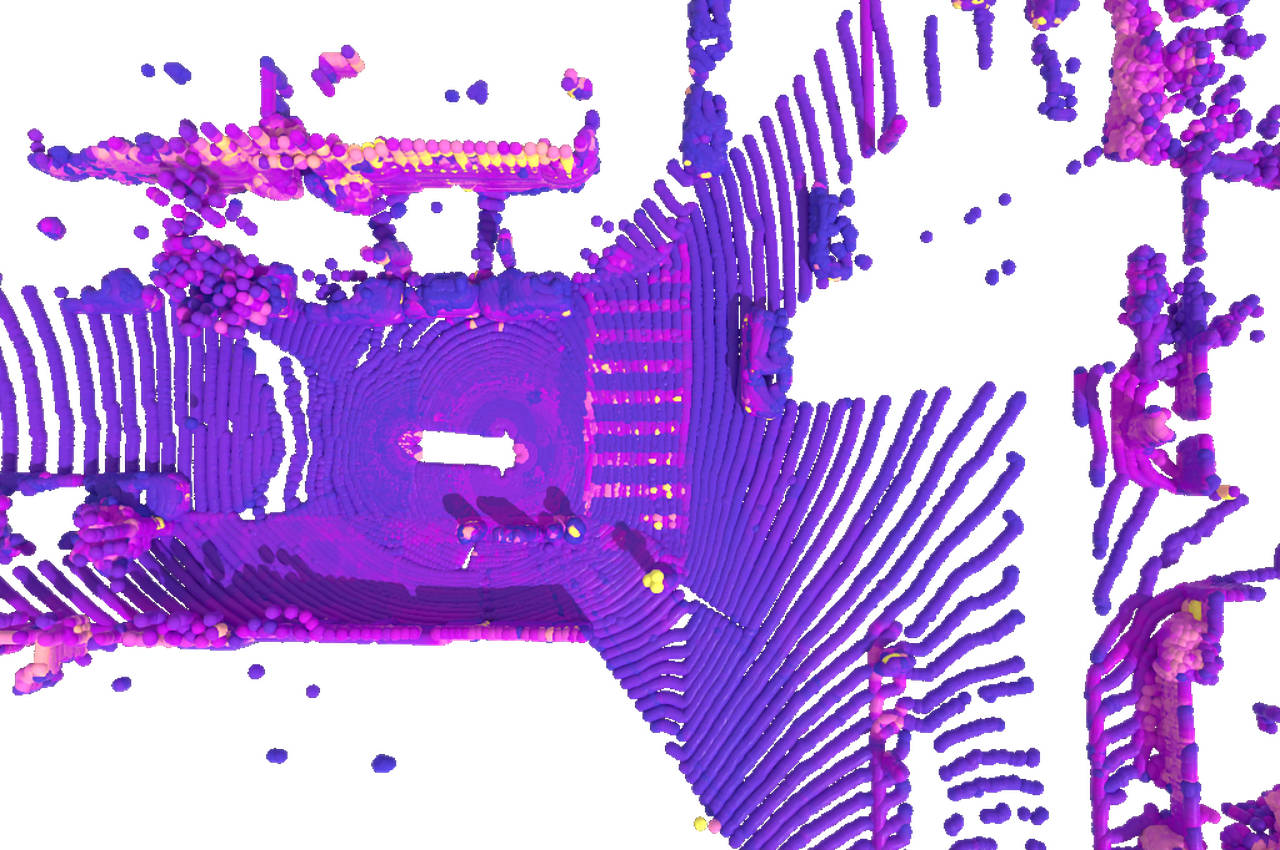}
    \end{subfigure}
         & 
    \begin{subfigure}[b]{0.28\textwidth}
      \includegraphics[width=\linewidth]{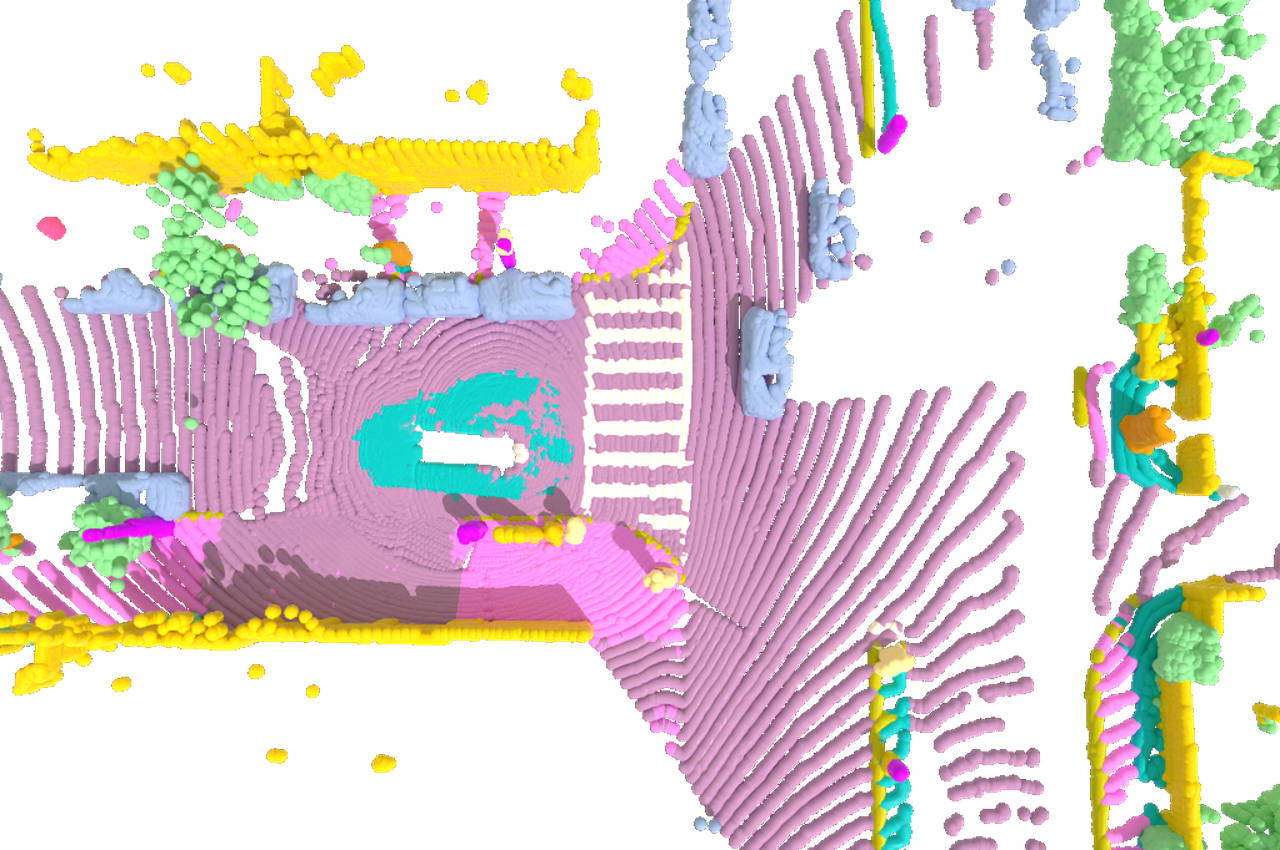}
    \end{subfigure}
         & 
    \begin{subfigure}[b]{0.28\textwidth}
      \includegraphics[width=\linewidth]{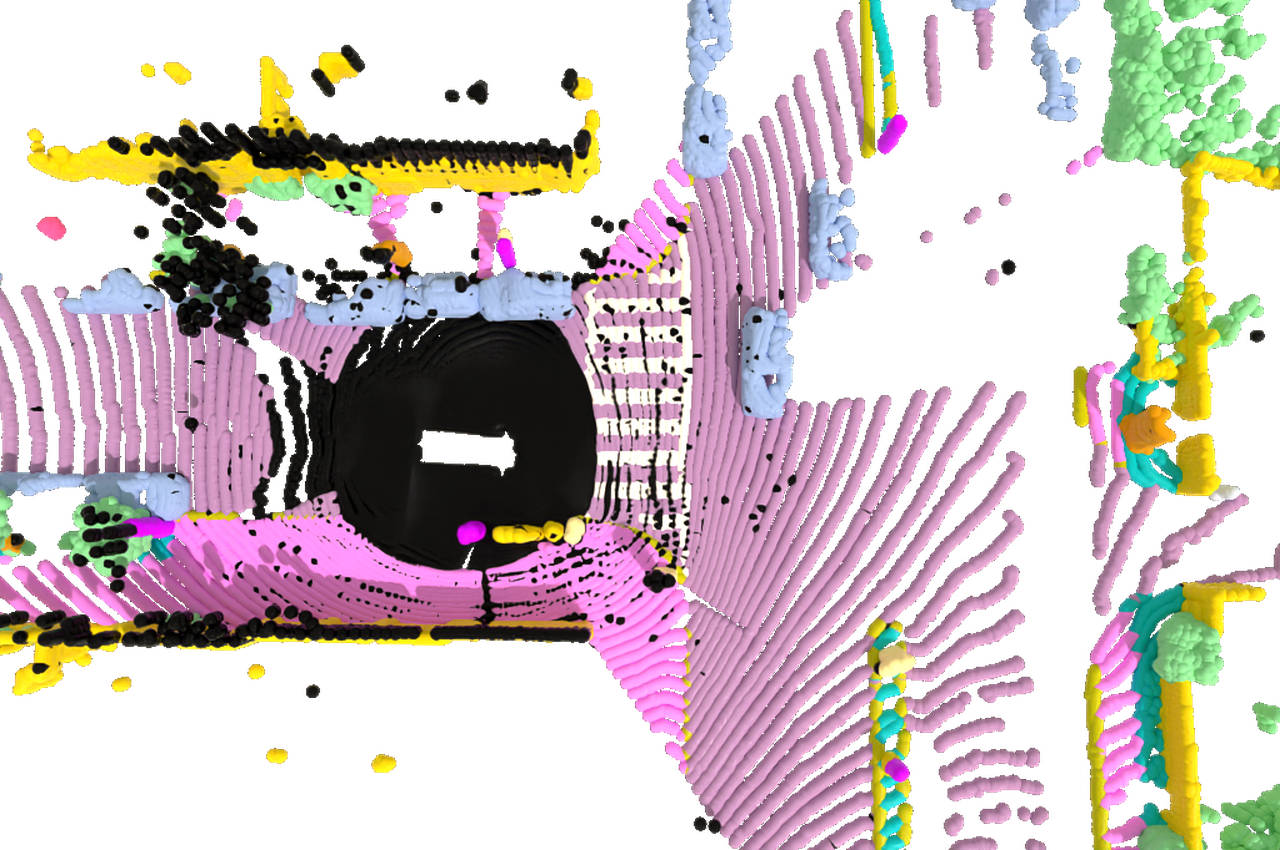}
    \end{subfigure}  \\
    \vspace{-2pt}
    \rotatebox{90}{\scriptsize \texttt{8956556778987472...}}
    &
    \begin{subfigure}[b]{0.28\textwidth}
      \includegraphics[width=\linewidth]{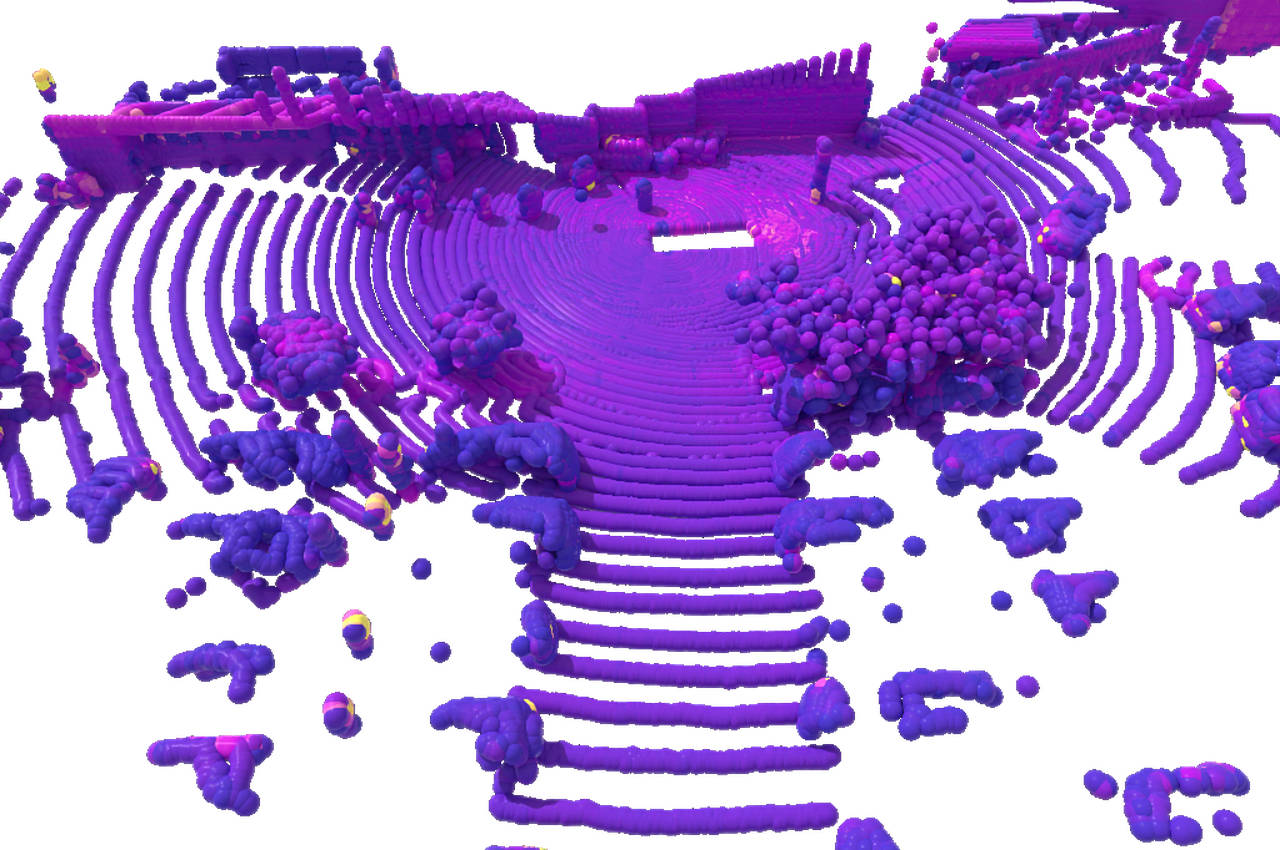}
    \end{subfigure}
         & 
    \begin{subfigure}[b]{0.28\textwidth}
      \includegraphics[width=\linewidth]{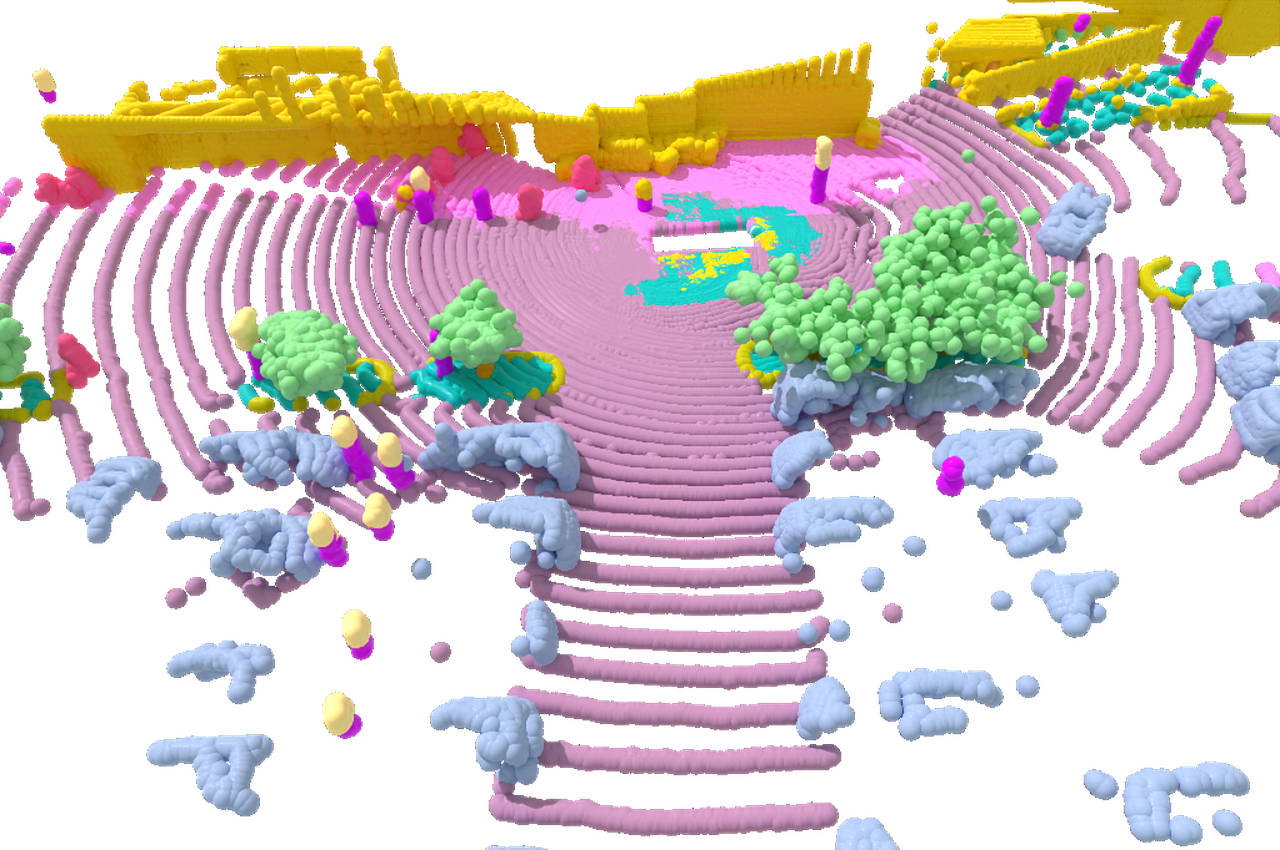}
    \end{subfigure}
         & 
    \begin{subfigure}[b]{0.28\textwidth}
      \includegraphics[width=\linewidth]{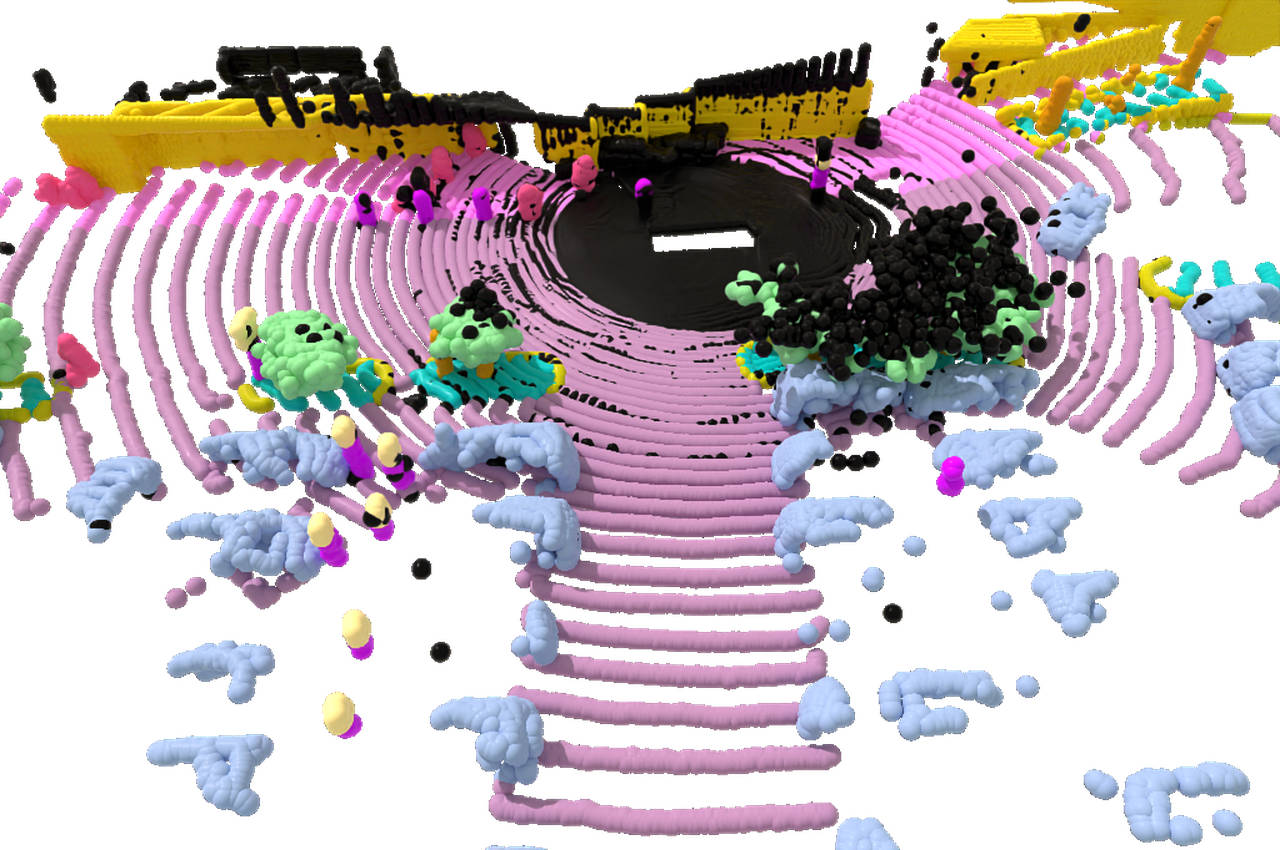}
    \end{subfigure}  \\
    \vspace{-2pt}
    \rotatebox{90}{\scriptsize \texttt{9041488218266405...}}
    &
    \begin{subfigure}[b]{0.28\textwidth}
      \includegraphics[width=\linewidth]{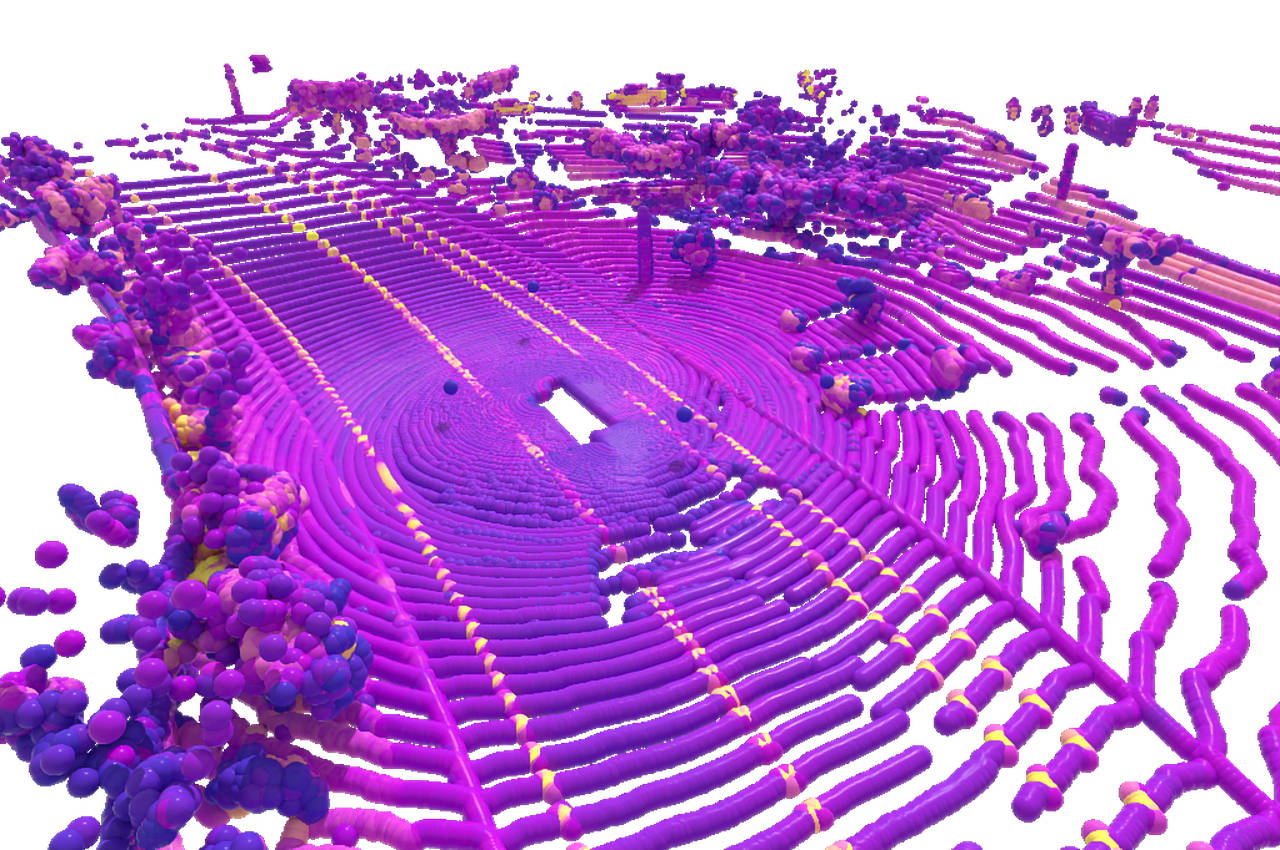}
    \end{subfigure}
         & 
    \begin{subfigure}[b]{0.28\textwidth}
      \includegraphics[width=\linewidth]{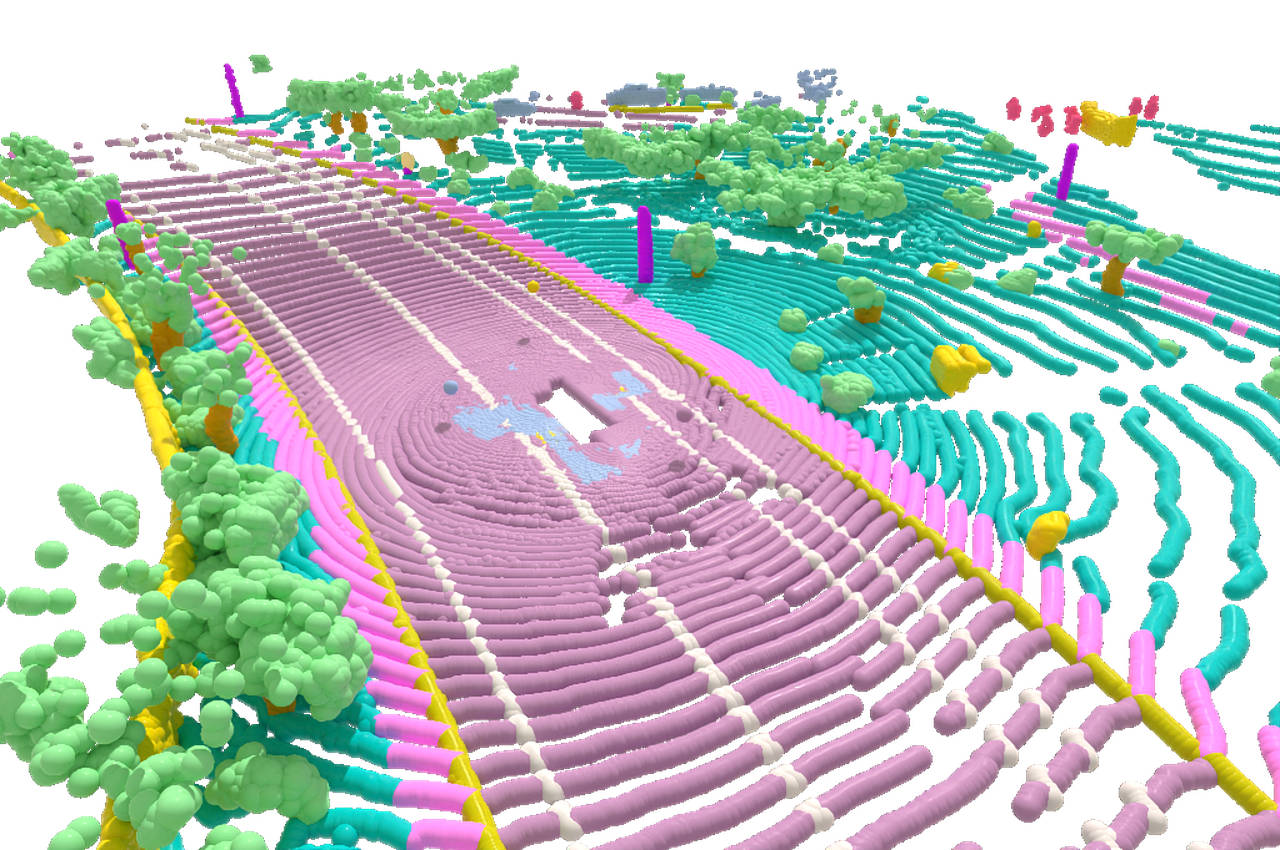}
    \end{subfigure}
         & 
    \begin{subfigure}[b]{0.28\textwidth}
      \includegraphics[width=\linewidth]{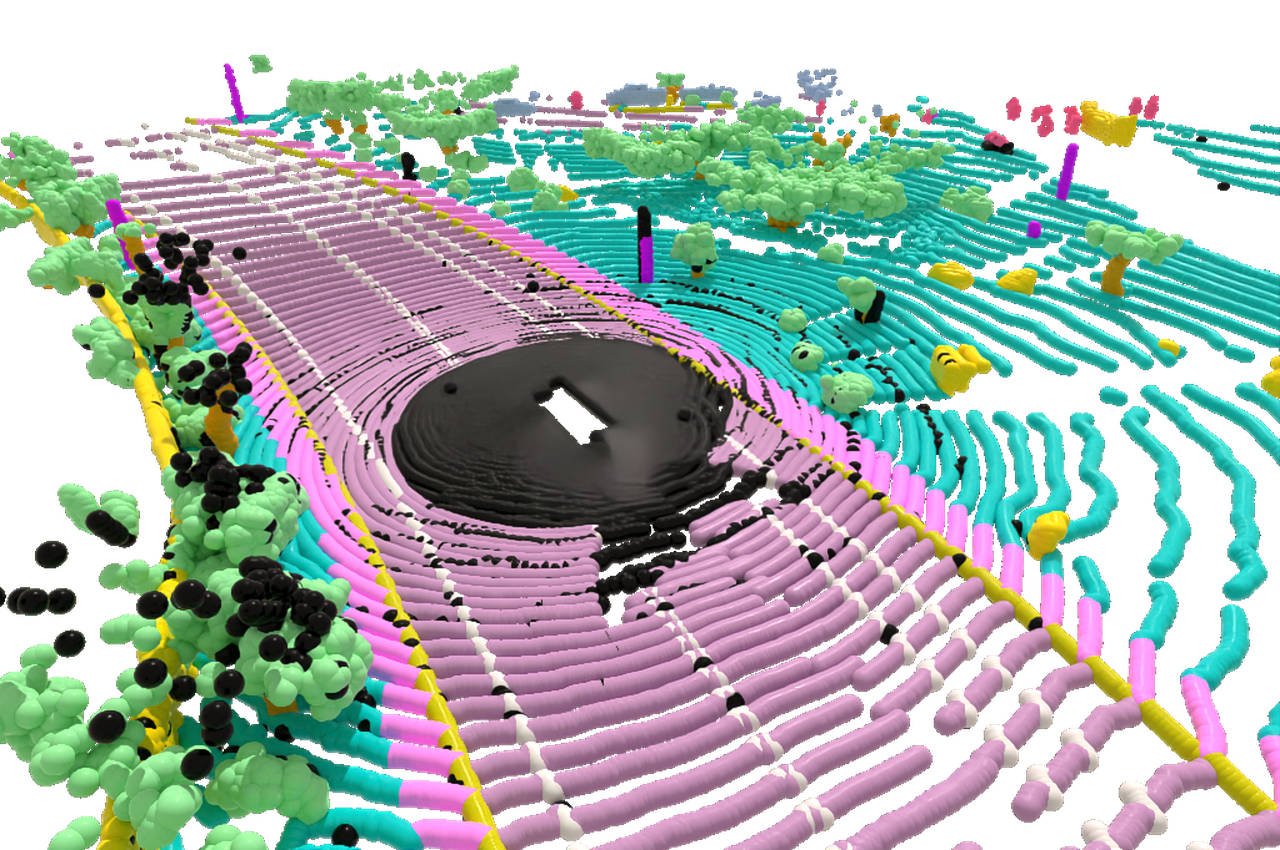}
    \end{subfigure}  \\
    \vspace{-2pt}
    \rotatebox{90}{\scriptsize \texttt{110376513715...}}
    &
    \begin{subfigure}[b]{0.28\textwidth}
      \includegraphics[width=\linewidth]{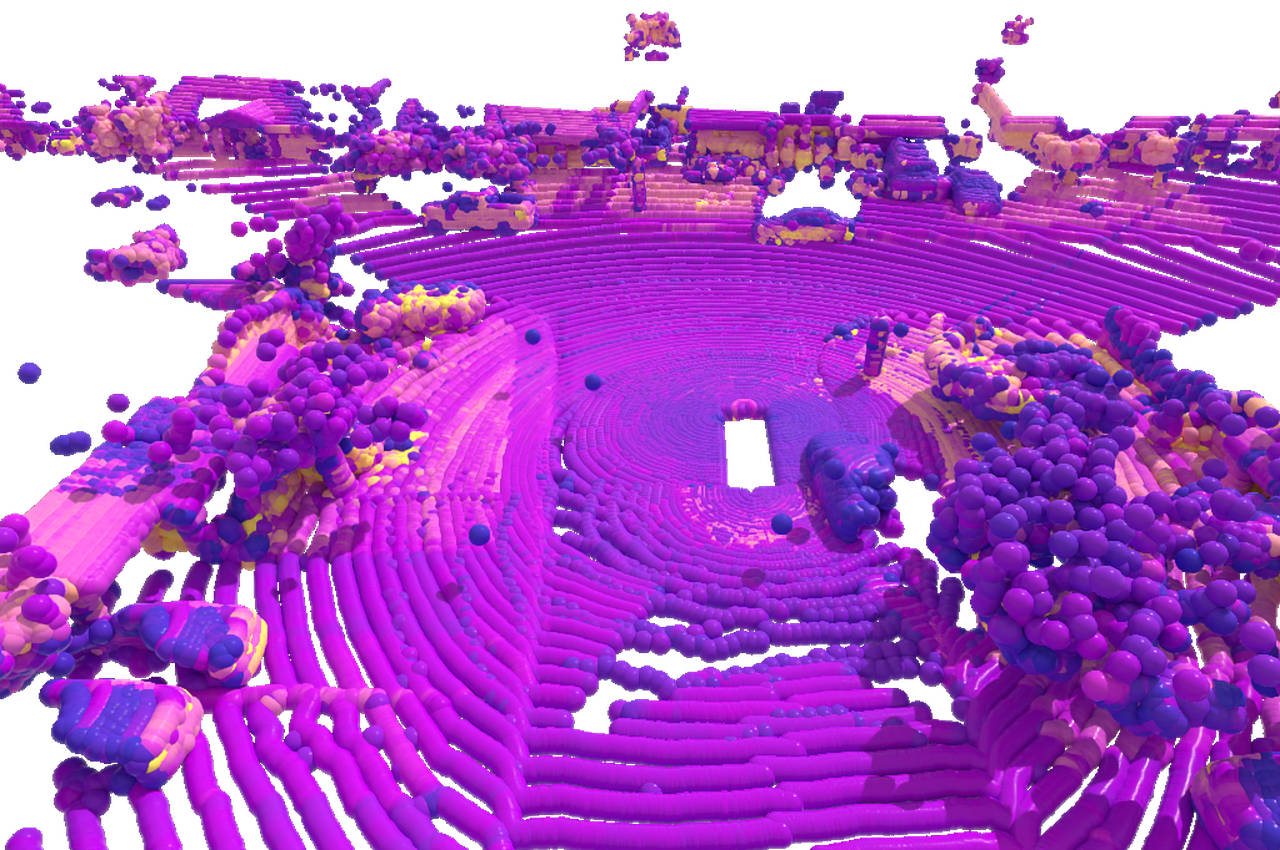}
    \end{subfigure}
         & 
    \begin{subfigure}[b]{0.28\textwidth}
      \includegraphics[width=\linewidth]{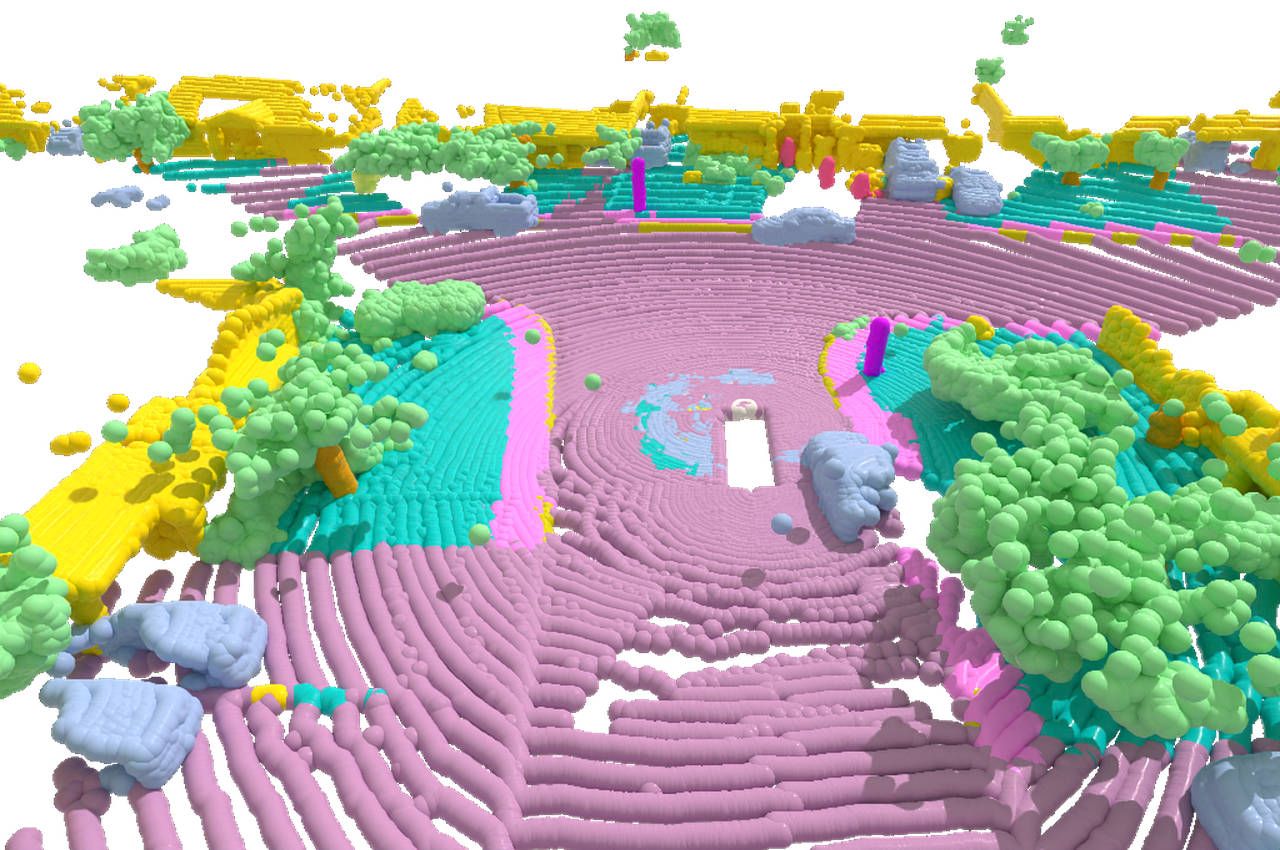}
    \end{subfigure}
         & 
    \begin{subfigure}[b]{0.28\textwidth}
      \includegraphics[width=\linewidth]{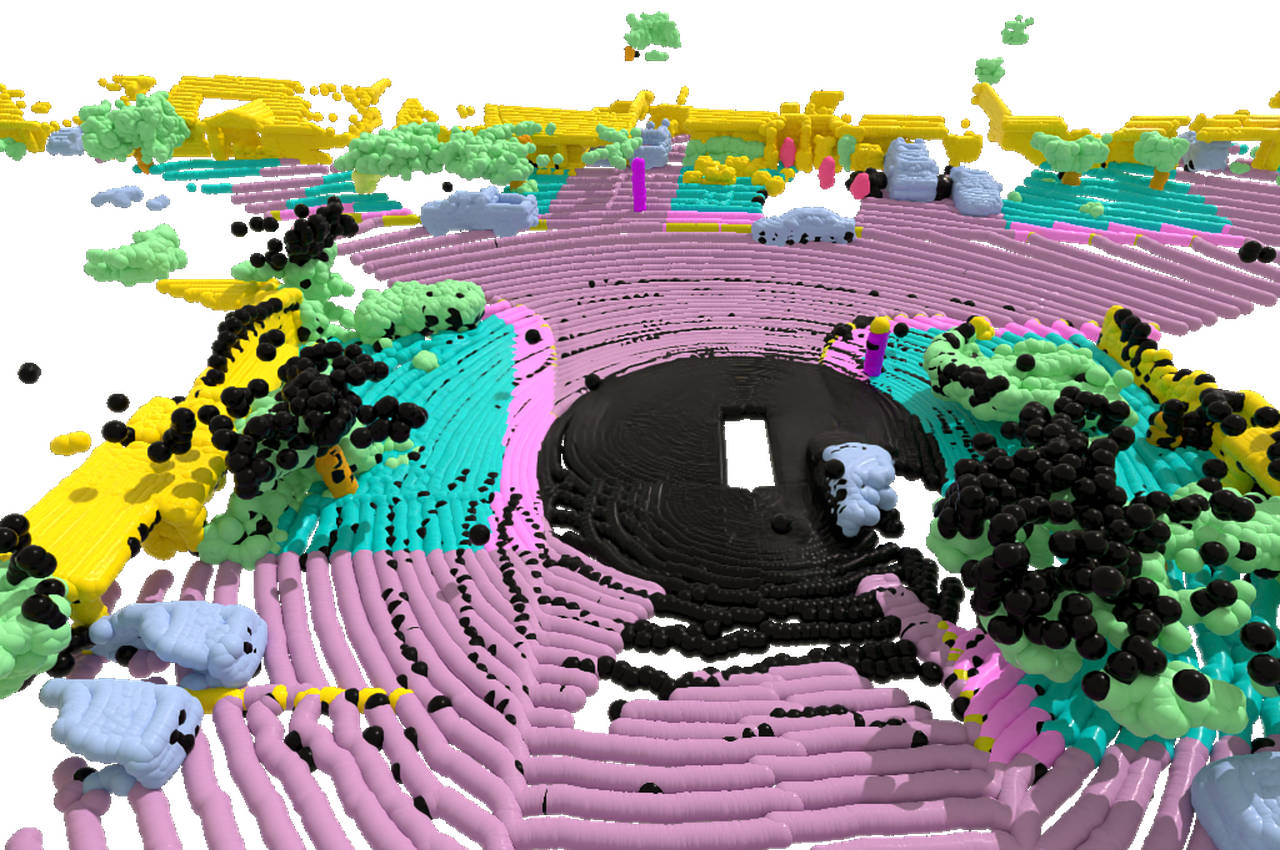}
    \end{subfigure}  \\
    \vspace{-2pt}
    \rotatebox{90}{\scriptsize \texttt{1825211188287550...}}
    &
    \begin{subfigure}[b]{0.28\textwidth}
      \includegraphics[width=\linewidth]{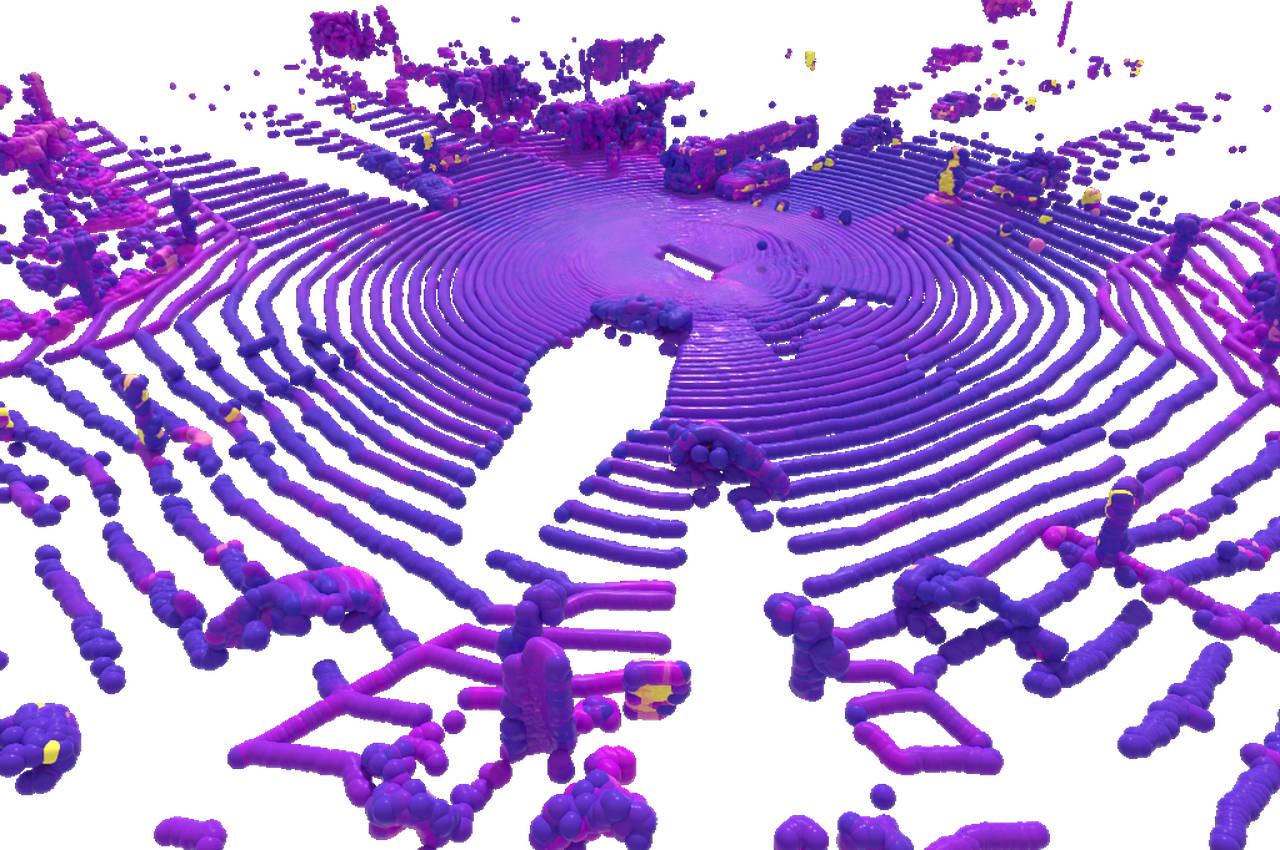}
    \end{subfigure}
         & 
    \begin{subfigure}[b]{0.28\textwidth}
      \includegraphics[width=\linewidth]{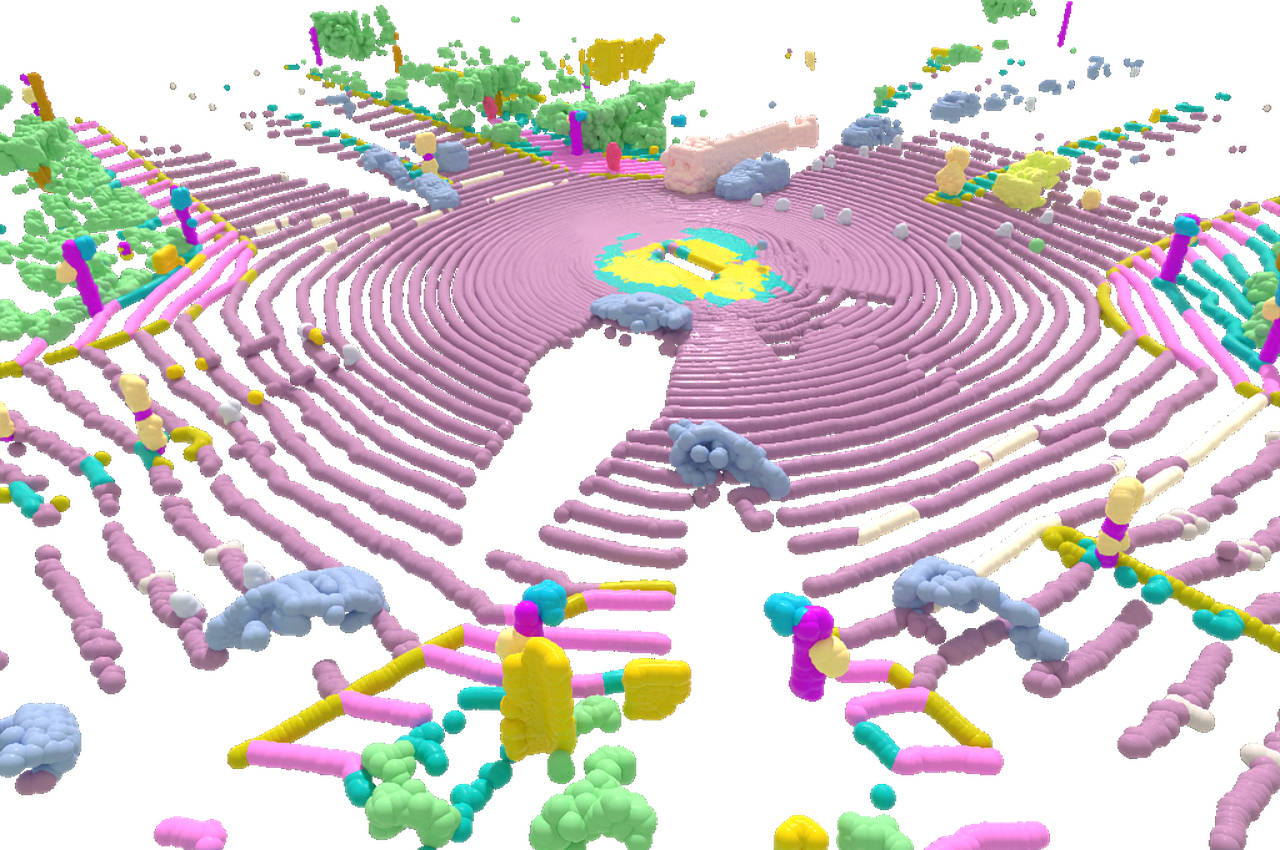}
    \end{subfigure}
         & 
    \begin{subfigure}[b]{0.28\textwidth}
      \includegraphics[width=\linewidth]{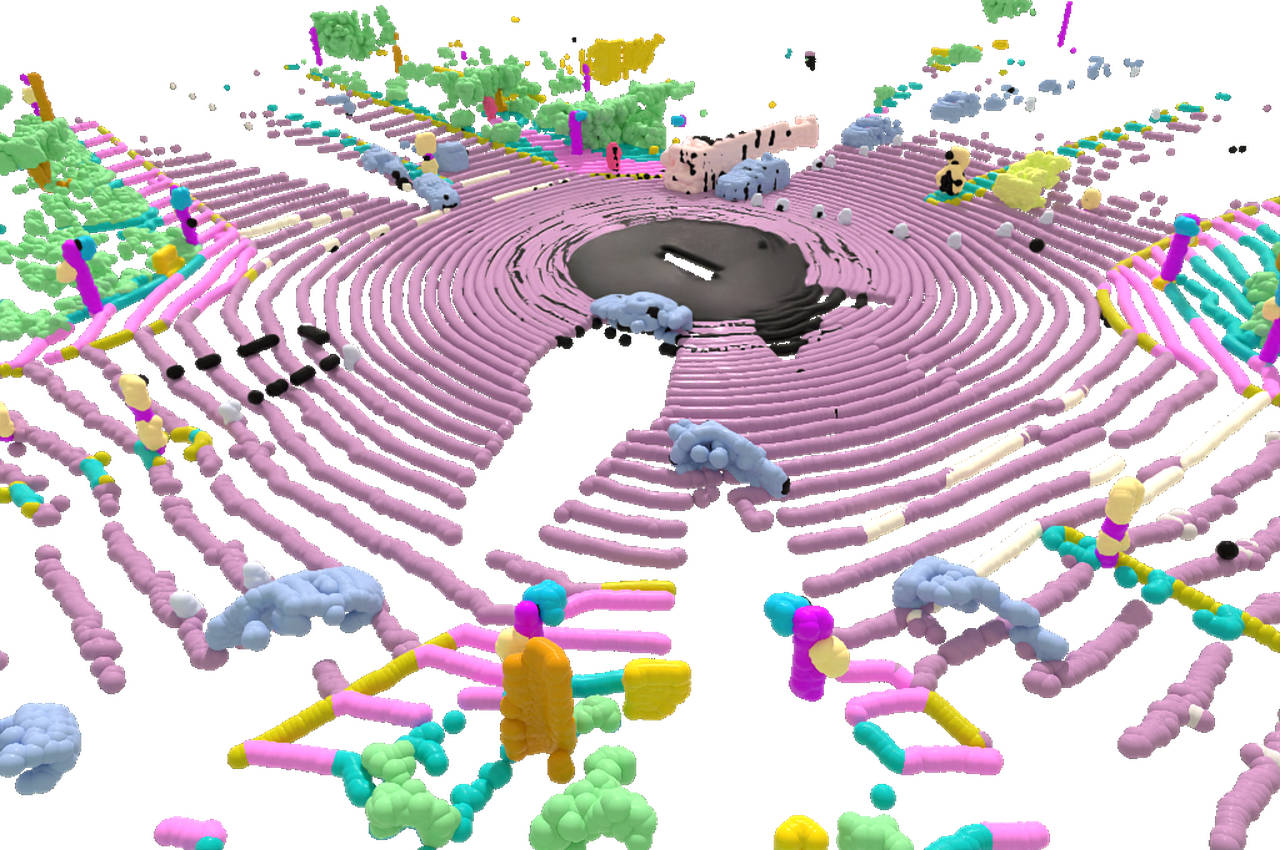}
    \end{subfigure}  \\
    \vspace{-2pt}
    \rotatebox{90}{\scriptsize \texttt{1833392207058224...}}
    &
    \begin{subfigure}[b]{0.28\textwidth}
      \includegraphics[width=\linewidth]{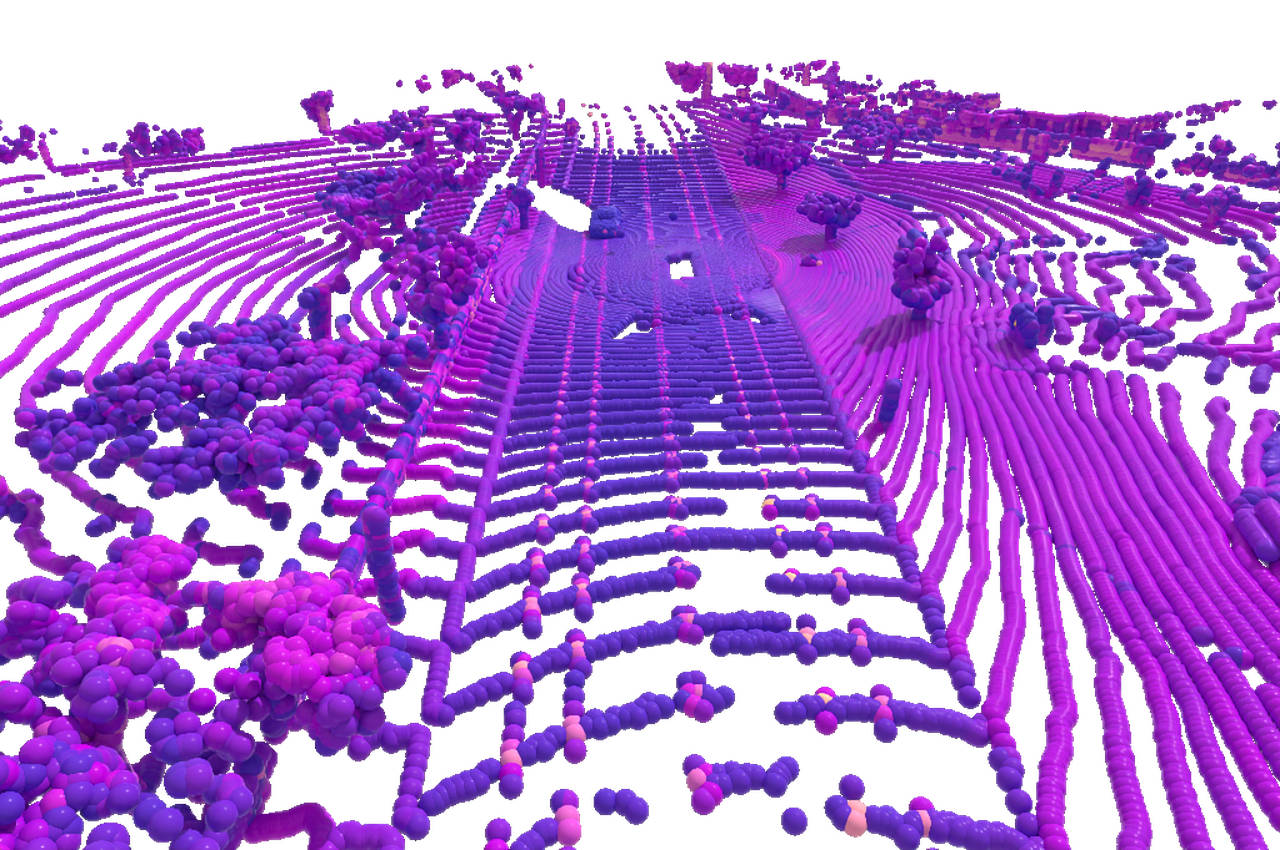}
      \caption{Input}
      \label{fig:quali_waymo_semseg:input}
    \end{subfigure}
         & 
    \begin{subfigure}[b]{0.28\textwidth}
      \includegraphics[width=\linewidth]{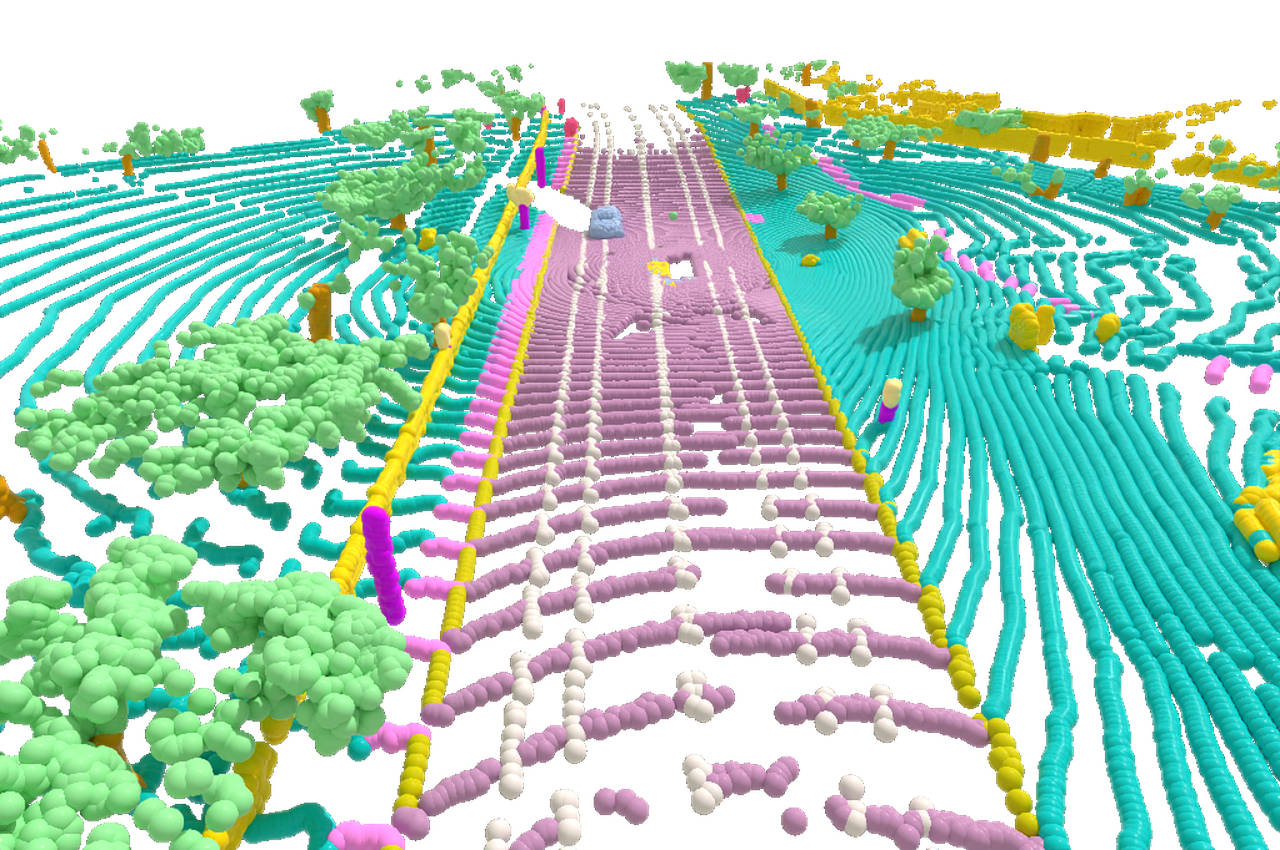}
      \caption{Prediction}
      \label{fig:quali_waymo_semseg:pred}
    \end{subfigure}
         & 
    \begin{subfigure}[b]{0.28\textwidth}
      \includegraphics[width=\linewidth]{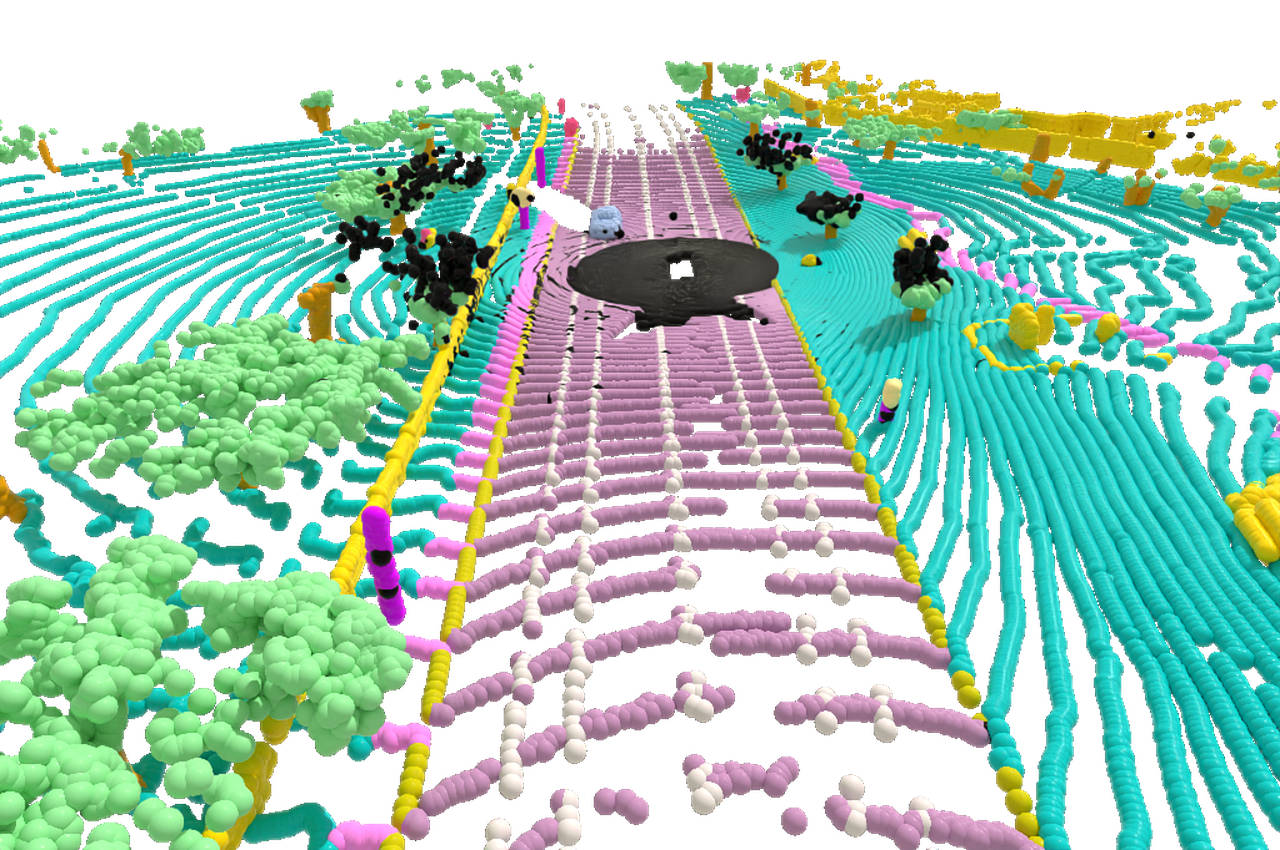}
      \caption{Ground Truth}
      \label{fig:quali_waymo_semseg:gt}
    \end{subfigure}  \\

    \end{tabular}
    \caption{
    {\bf Waymo semantic segmentation.} 
    We present various scenes of the Waymo validation set: the input point cloud coloured by LiDAR intensity, the semantic segmentation from \name{}-S, and the corresponding ground truth. 
    }
    \label{fig:quali_waymo_semseg}
\end{figure*}

\begin{figure*}
    \centering
    \begin{tabular}{@{}lccc@{}}

    \multicolumn{4}{c}{

    \begin{minipage}{0.98\textwidth}
    \centering
        {\textcolor{scannet_wall}{\ding{108}}}\,\,wall \,\, {\textcolor{scannet_floor}{\ding{108}}}\,\,floor\,\,
        {\textcolor{scannet_cabinet}{\ding{108}}}\,\,cabinet\,\,
        {\textcolor{scannet_bed}{\ding{108}}}\,\,bed\,\,
        {\textcolor{scannet_chair}{\ding{108}}}\,\,chair\,\,
        {\textcolor{scannet_sofa}{\ding{108}}}\,\,sofa\,\,
        {\textcolor{scannet_table}{\ding{108}}}\,\,table\,\,
        {\textcolor{scannet_door}{\ding{108}}}\,\,door\,\,
        {\textcolor{scannet_window}{\ding{108}}}\,\,window\,\,
        {\textcolor{scannet_bookshelf}{\ding{108}}}\,\,bookshelf\,\,
        {\textcolor{scannet_picture}{\ding{108}}}\,\,picture\,\,
        {\textcolor{scannet_counter}{\ding{108}}}\,\,counter\,\,
        {\textcolor{scannet_desk}{\ding{108}}}\,\,desk\,\,
        {\textcolor{scannet_curtain}{\ding{108}}}\,\,curtain\,\,
        {\textcolor{scannet_refrigerator}{\ding{108}}}\,\,refrigerator\,\,
        {\textcolor{scannet_shower}{\ding{108}}}\,\,shower\,\,
        {\textcolor{scannet_toilet}{\ding{108}}}\,\,toilet\,\,
        {\textcolor{scannet_sink}{\ding{108}}}\,\,sink\,\,
        {\textcolor{scannet_bathtub}{\ding{108}}}\,\,bathtub\,\,
        {\textcolor{scannet_otherfurniture}{\ding{108}}}\,\,other furniture\,\,
        {\textcolor{scannet_unlabeled}{\ding{108}}}\,\,unlabelled
    \end{minipage}
    } 
    \vspace{15pt}
    
    \\ 
    \rotatebox{90}{ \quad \texttt{scene0030\_00}}
    &
    \begin{subfigure}[b]{0.28\textwidth}
      \includegraphics[width=\linewidth]{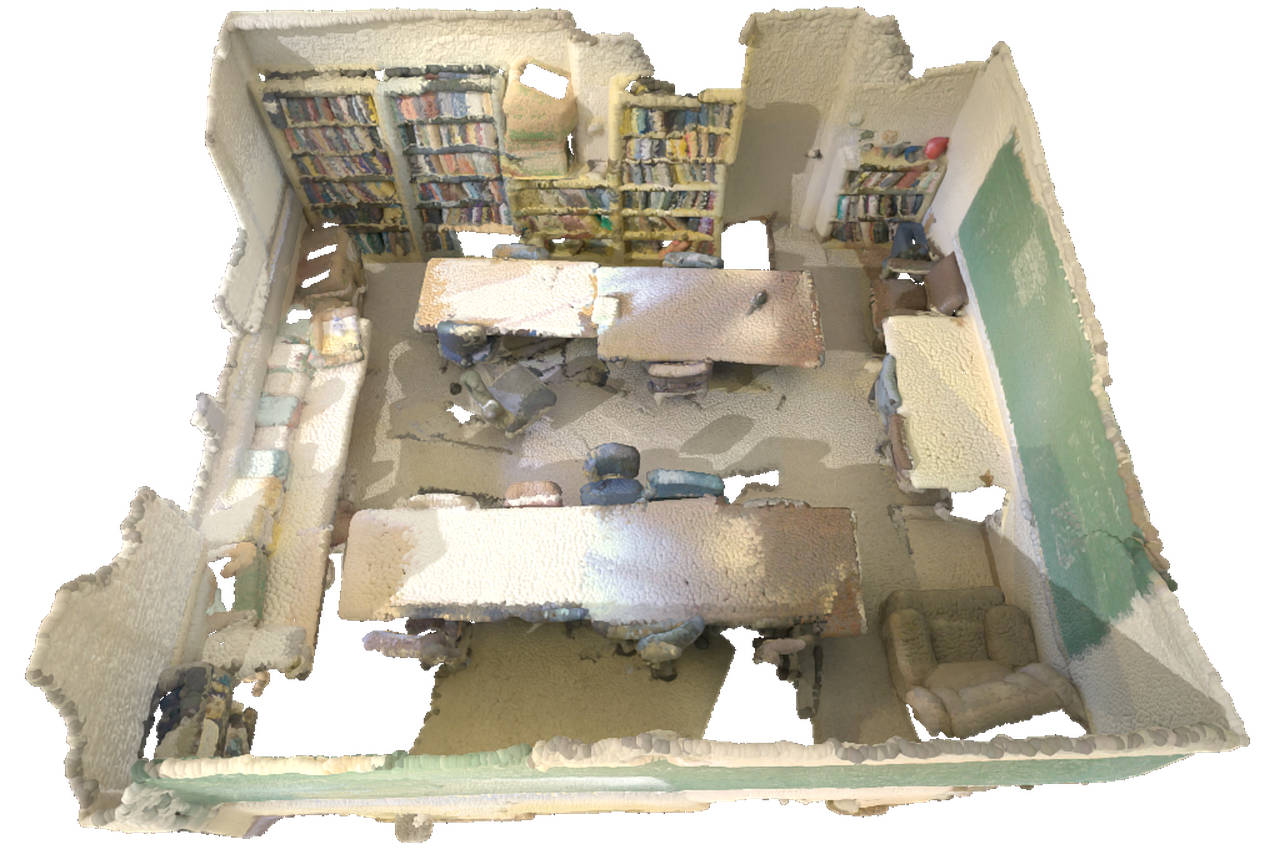}
    \end{subfigure}
         & 
    \begin{subfigure}[b]{0.28\textwidth}
      \includegraphics[width=\linewidth]{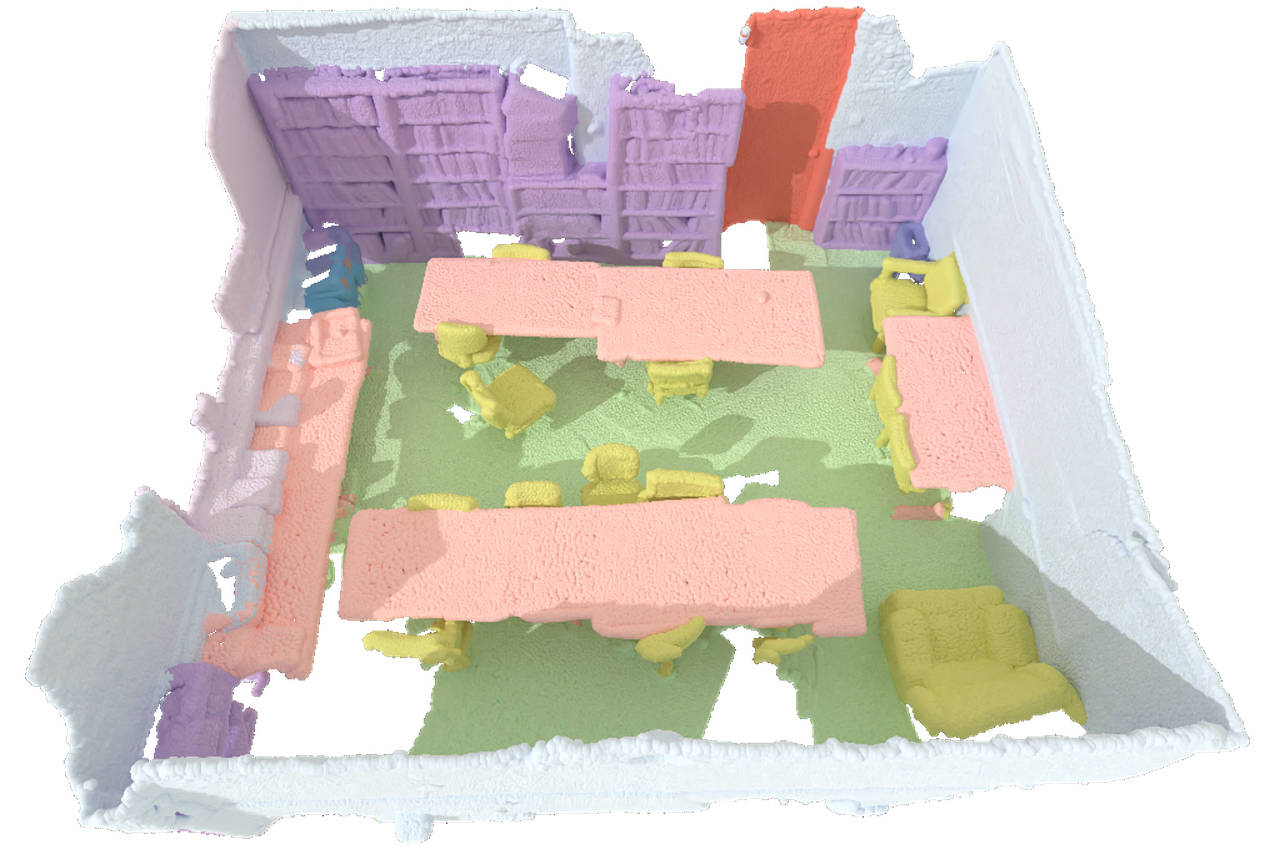}
    \end{subfigure}
         & 
    \begin{subfigure}[b]{0.28\textwidth}
      \includegraphics[width=\linewidth]{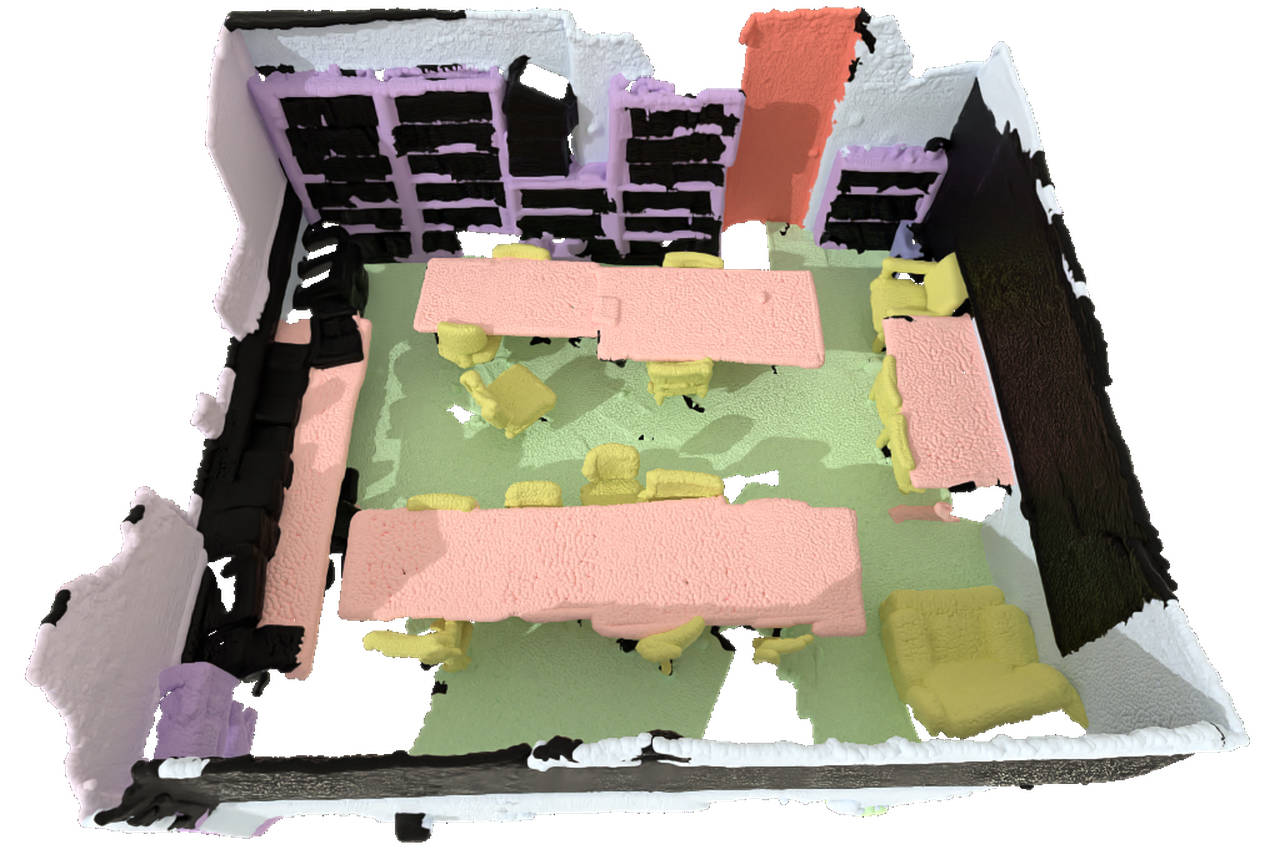}
    \end{subfigure}  \\

    \rotatebox{90}{ \quad \texttt{scene0169\_00}}
    &
    \begin{subfigure}[b]{0.28\textwidth}
      \includegraphics[width=\linewidth]{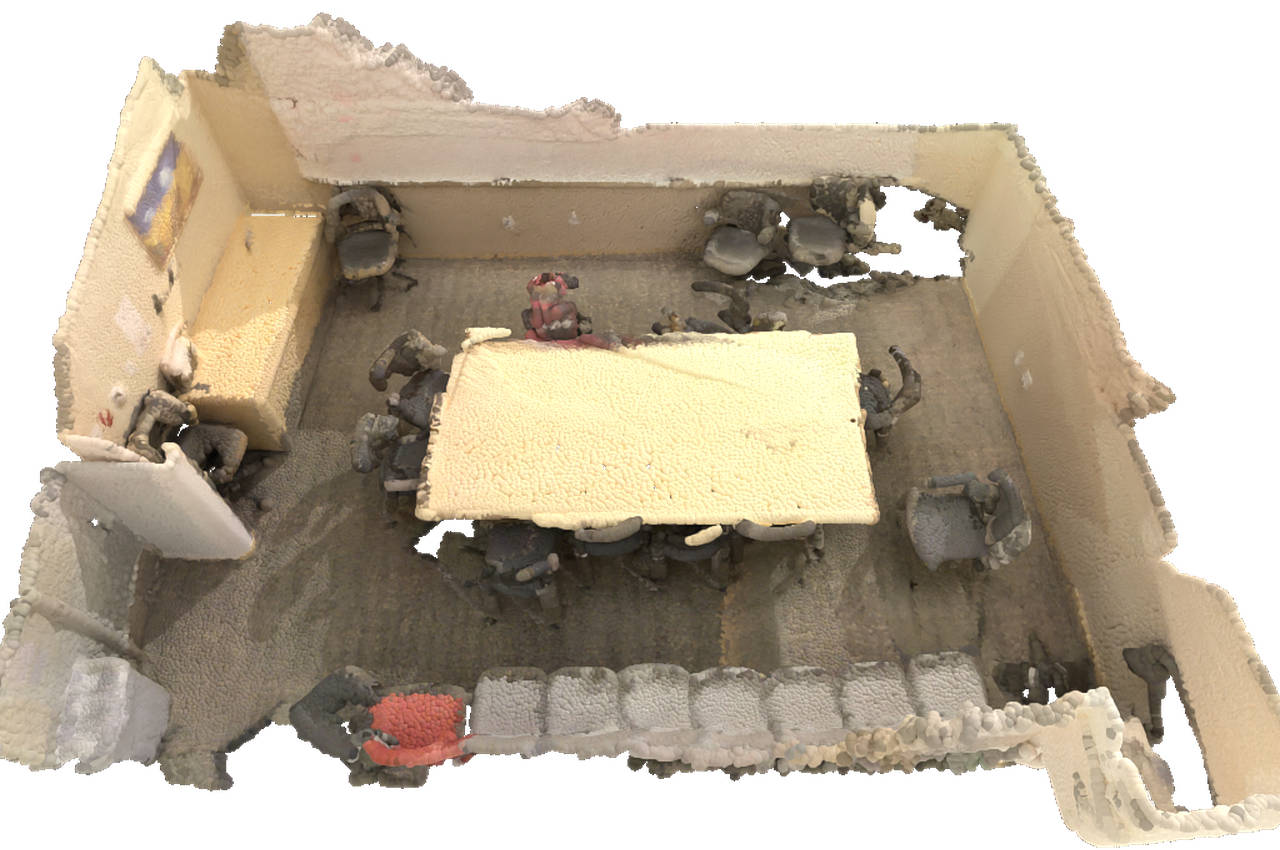}
    \end{subfigure}
         & 
    \begin{subfigure}[b]{0.28\textwidth}
      \includegraphics[width=\linewidth]{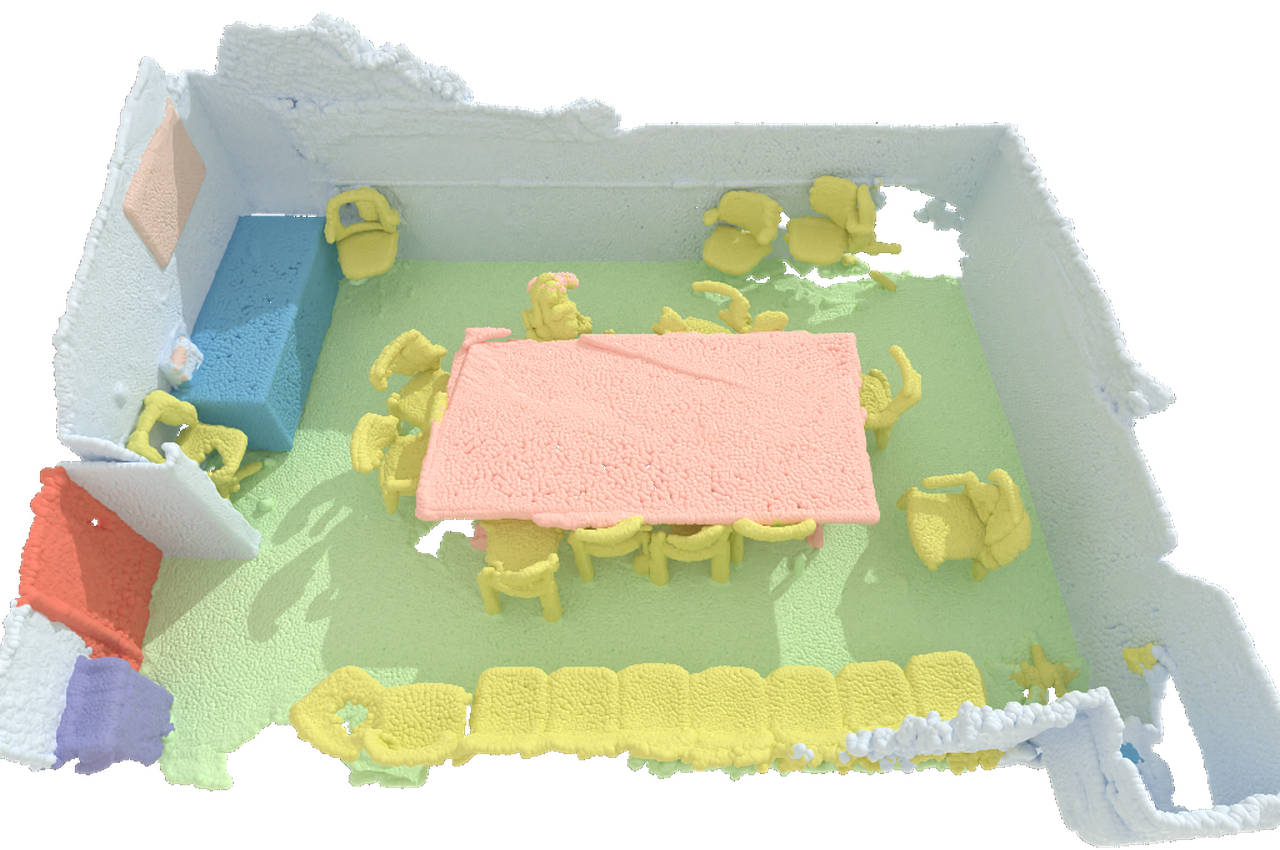}
    \end{subfigure}
         & 
    \begin{subfigure}[b]{0.28\textwidth}
      \includegraphics[width=\linewidth]{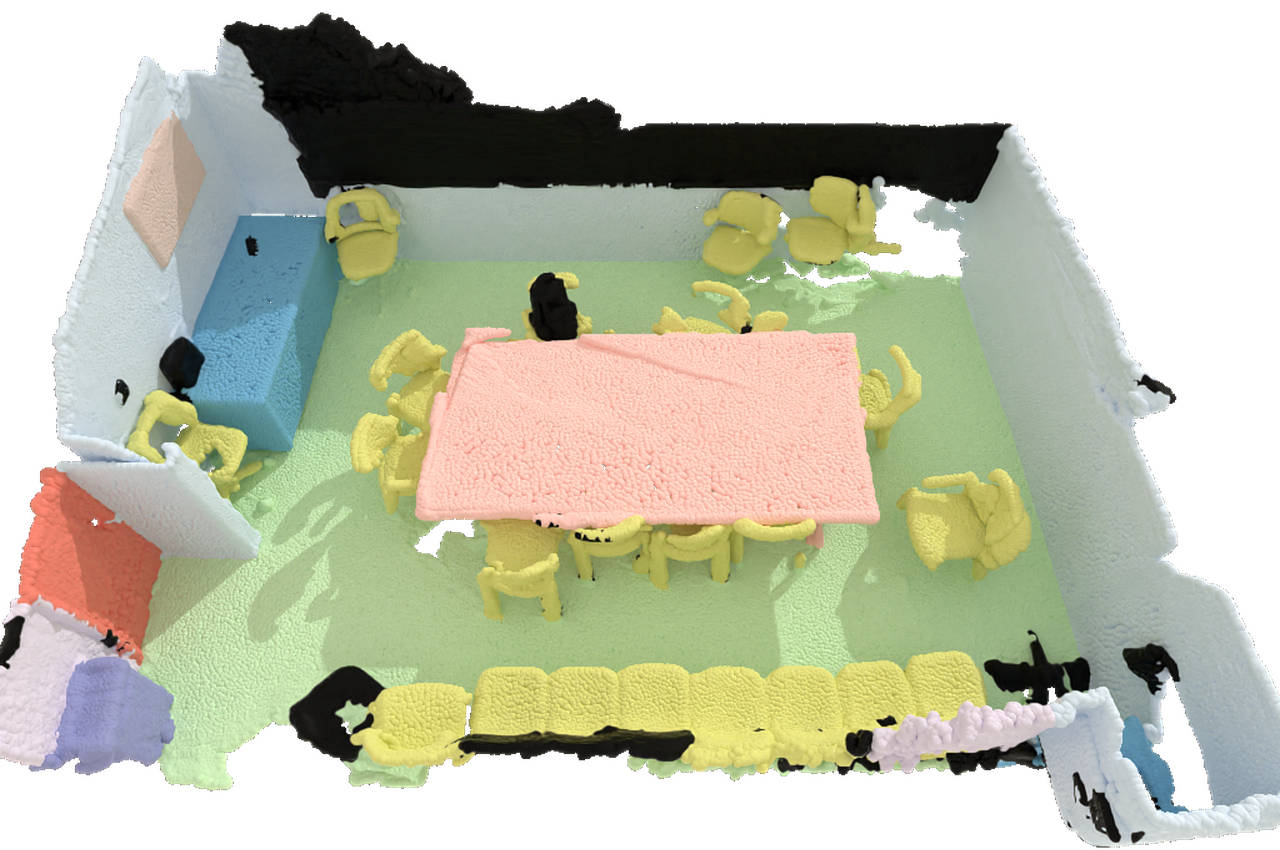}
    \end{subfigure}  \\

    \rotatebox{90}{ \qquad \texttt{scene0378\_02}}
    &
    \begin{subfigure}[b]{0.28\textwidth}
      \includegraphics[width=\linewidth]{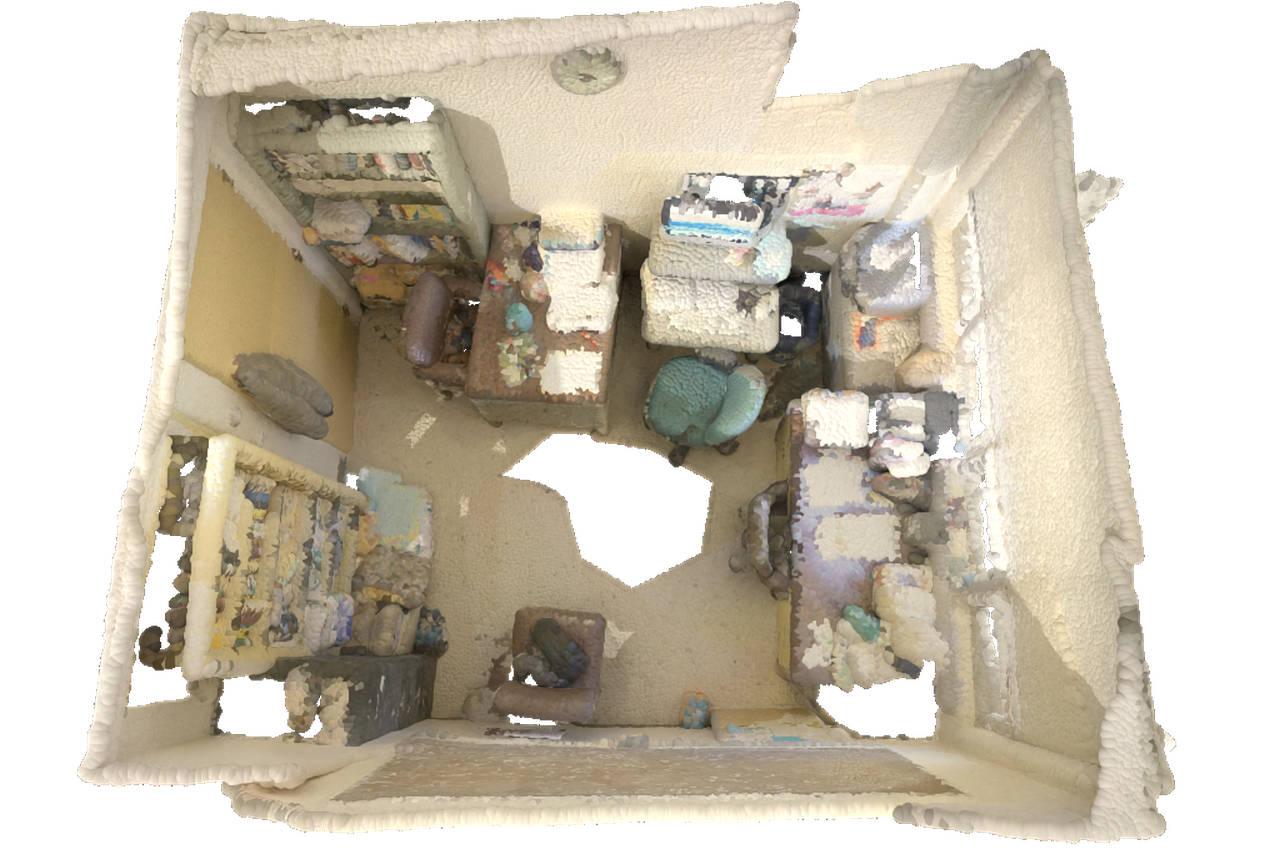}
    \end{subfigure}
         & 
    \begin{subfigure}[b]{0.28\textwidth}
      \includegraphics[width=\linewidth]{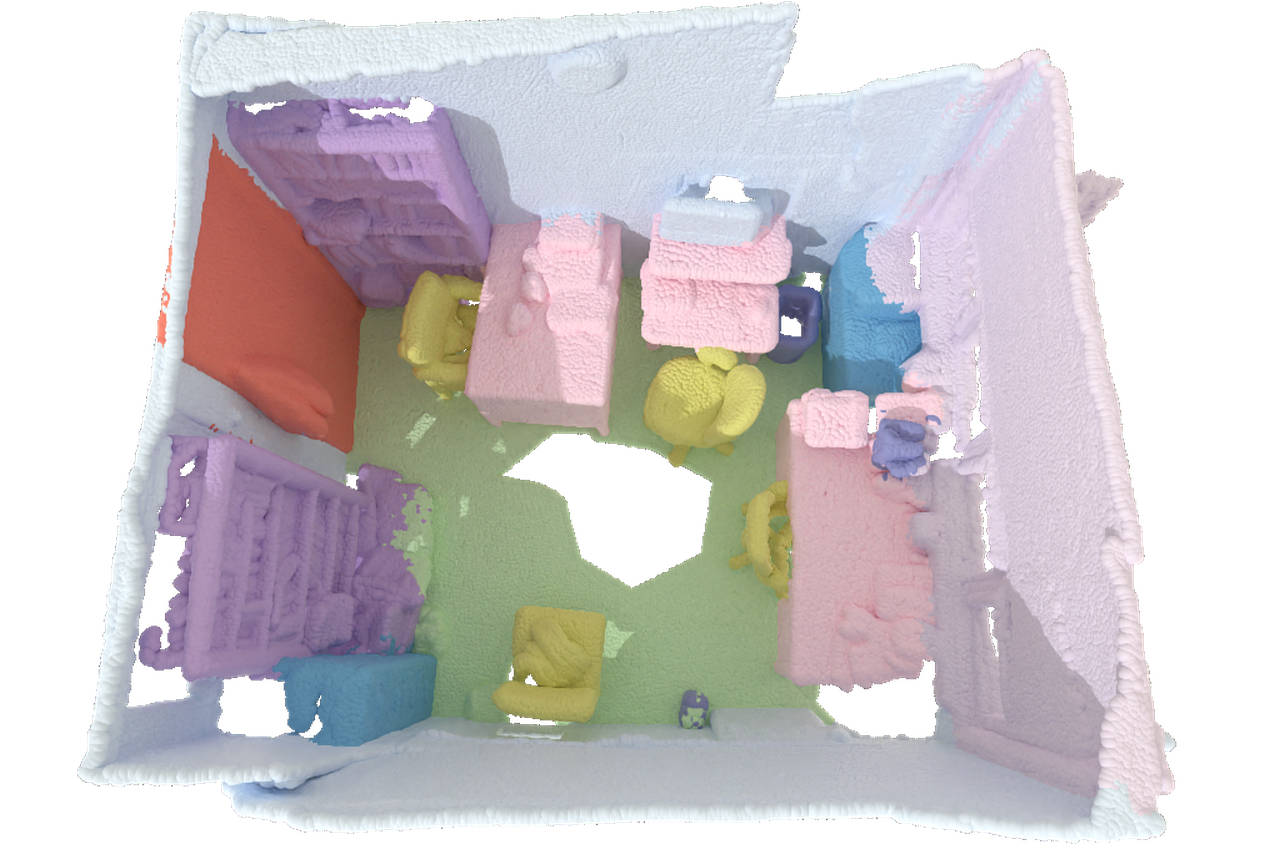}
    \end{subfigure}
         & 
    \begin{subfigure}[b]{0.28\textwidth}
      \includegraphics[width=\linewidth]{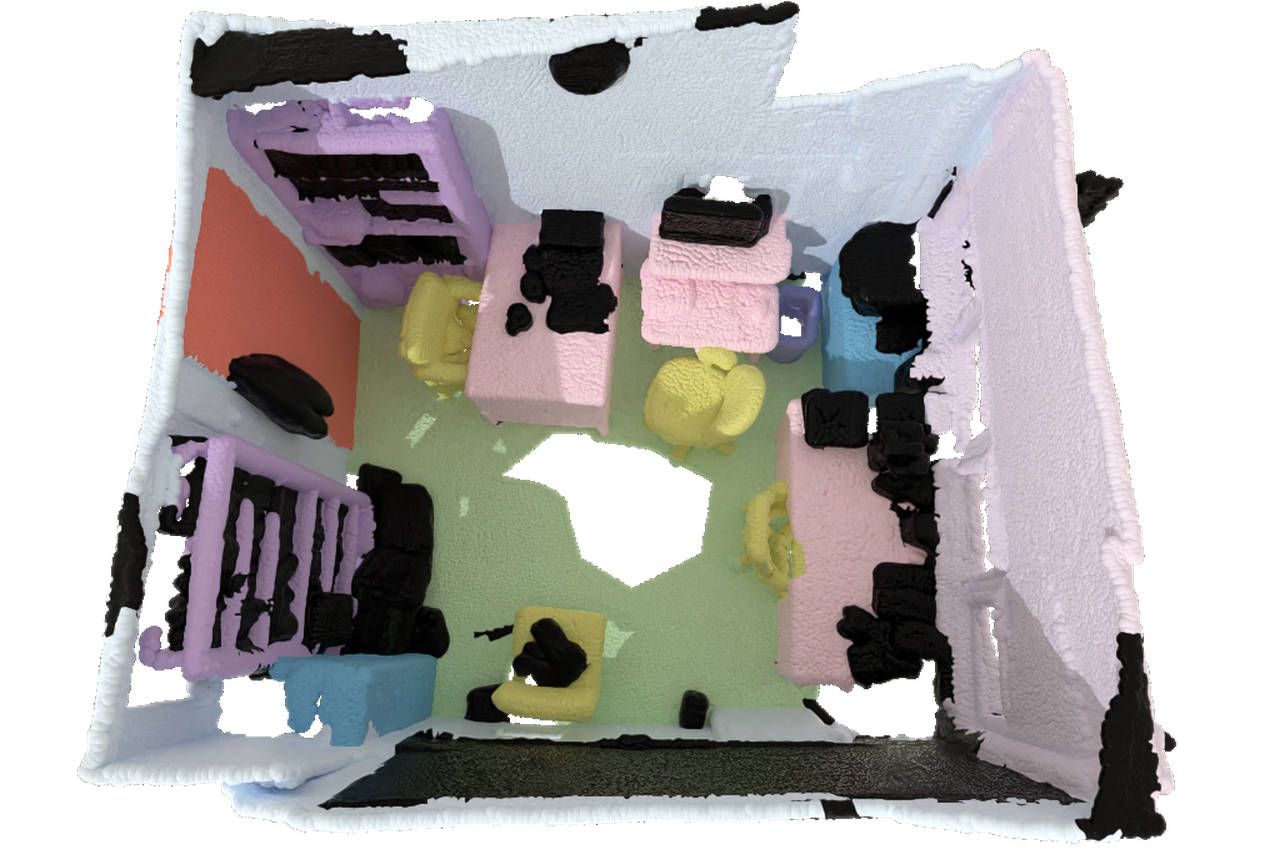}
    \end{subfigure}  \\

    \rotatebox{90}{ \qquad \texttt{scene0406\_02}}
    &
    \begin{subfigure}[b]{0.28\textwidth}
      \includegraphics[width=\linewidth]{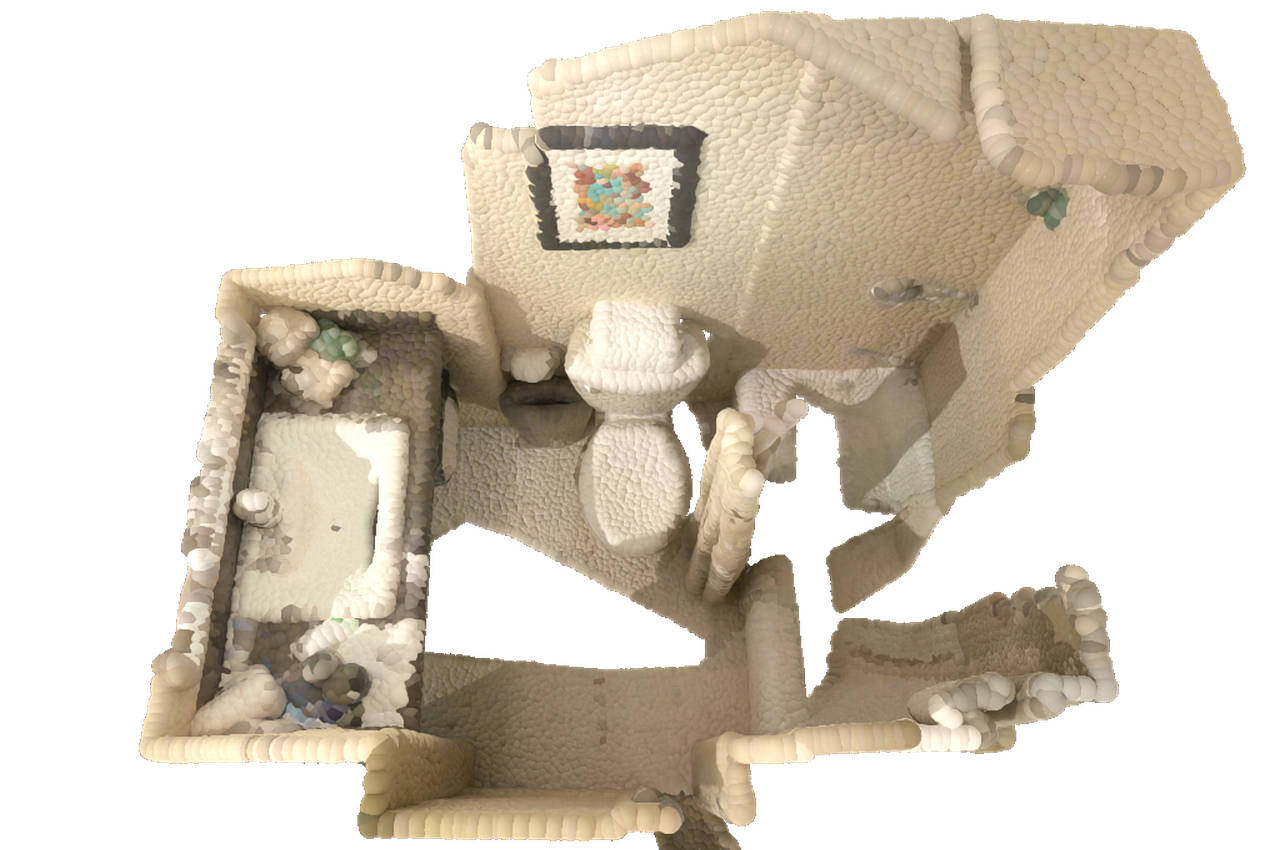}
    \end{subfigure}
         & 
    \begin{subfigure}[b]{0.28\textwidth}
      \includegraphics[width=\linewidth]{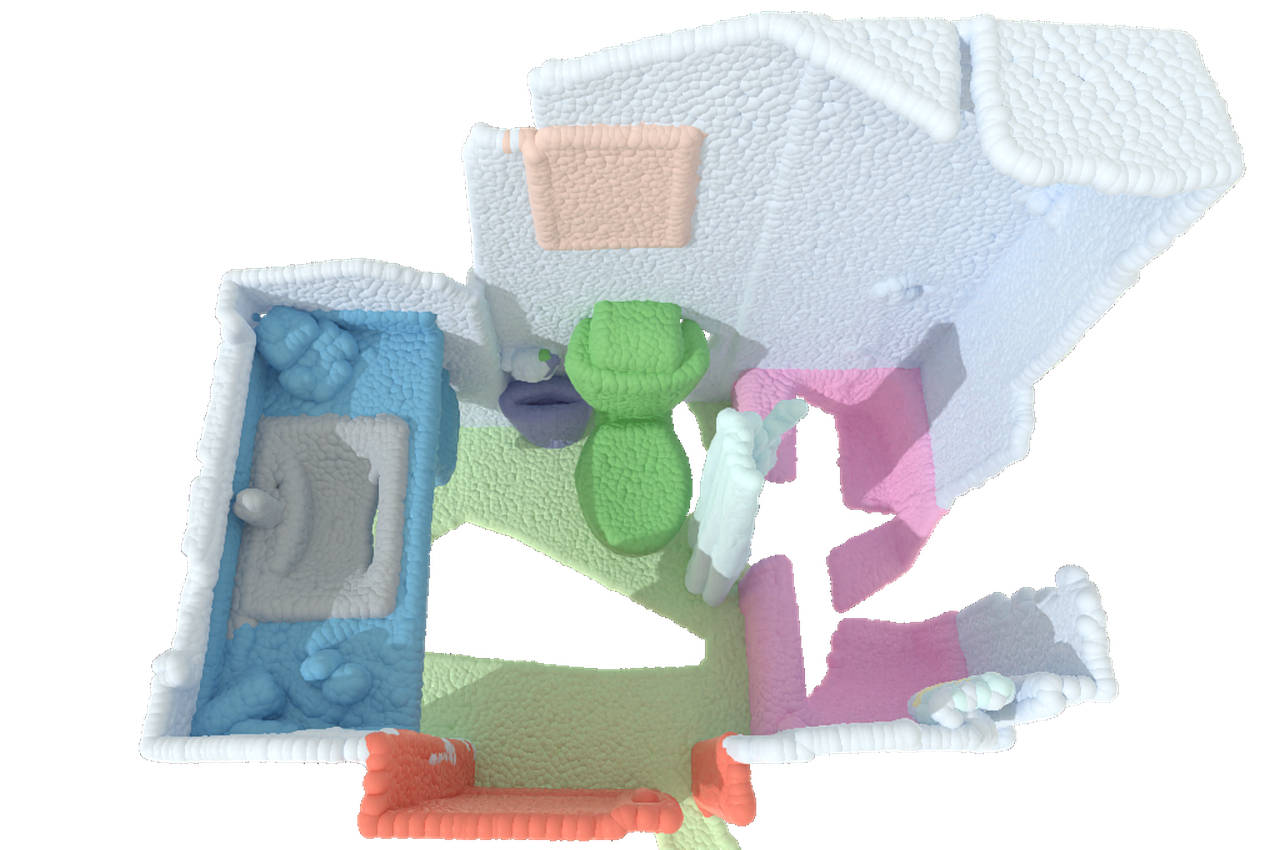}
    \end{subfigure}
         & 
    \begin{subfigure}[b]{0.28\textwidth}
      \includegraphics[width=\linewidth]{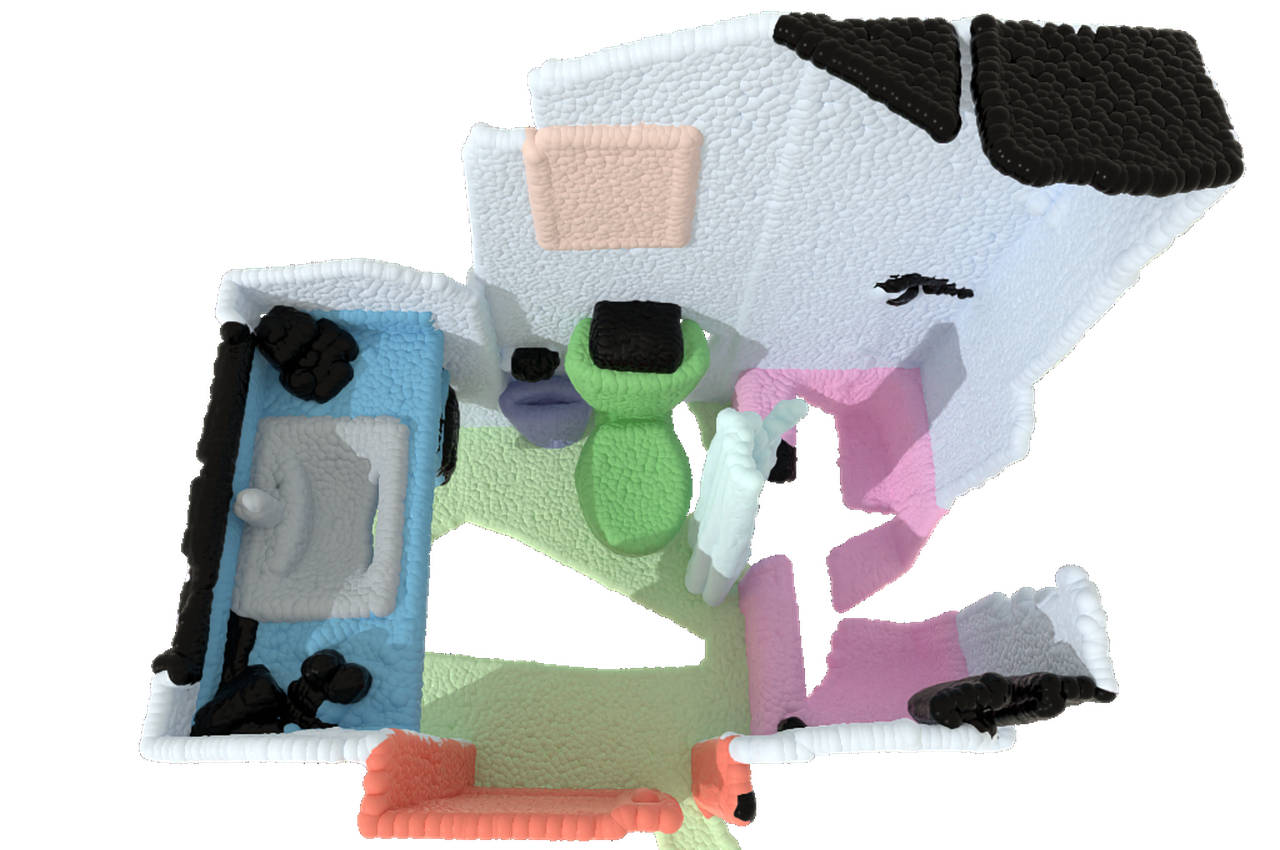}
    \end{subfigure}  \\

    \rotatebox{90}{ \quad \texttt{scene0645\_01}}
    &
    \begin{subfigure}[b]{0.28\textwidth}
      \includegraphics[width=\linewidth]{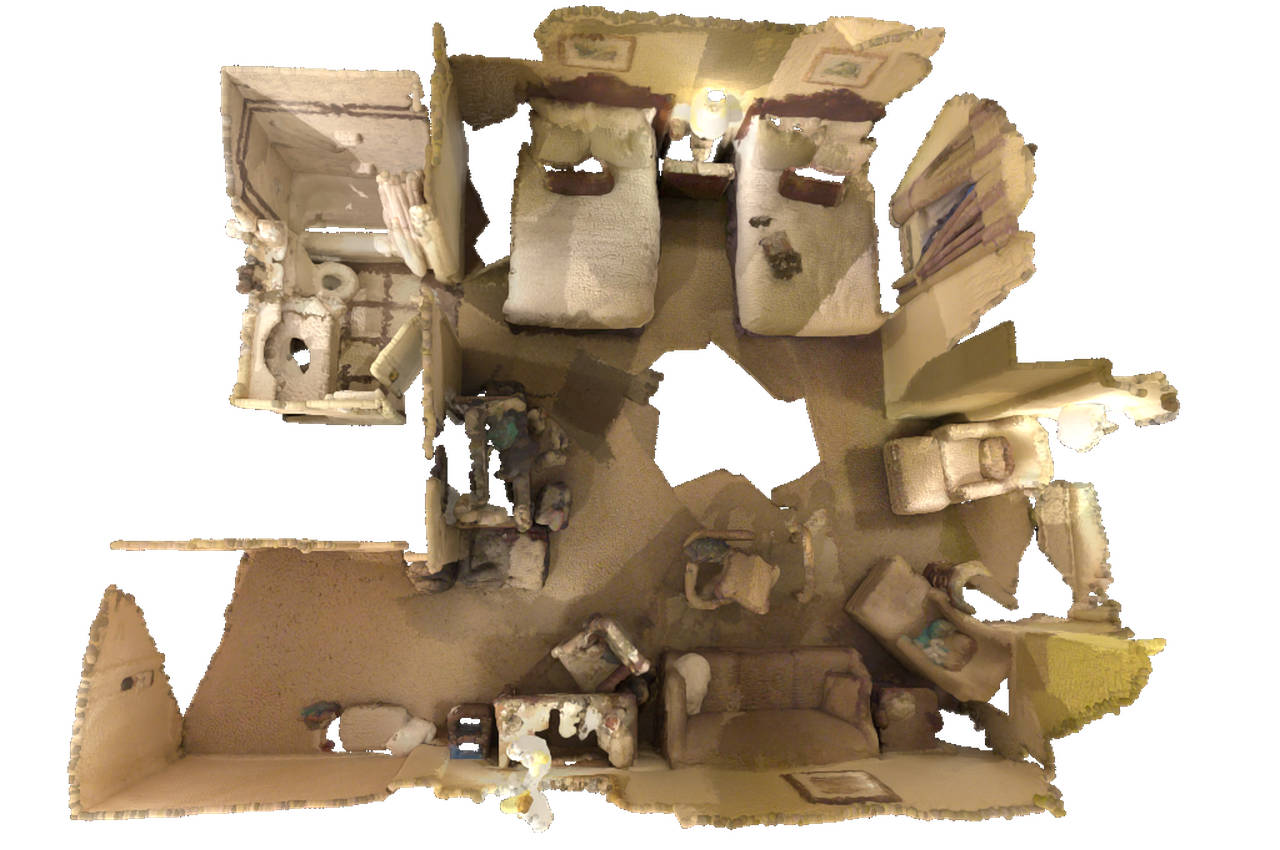}
    \end{subfigure}
         & 
    \begin{subfigure}[b]{0.28\textwidth}
      \includegraphics[width=\linewidth]{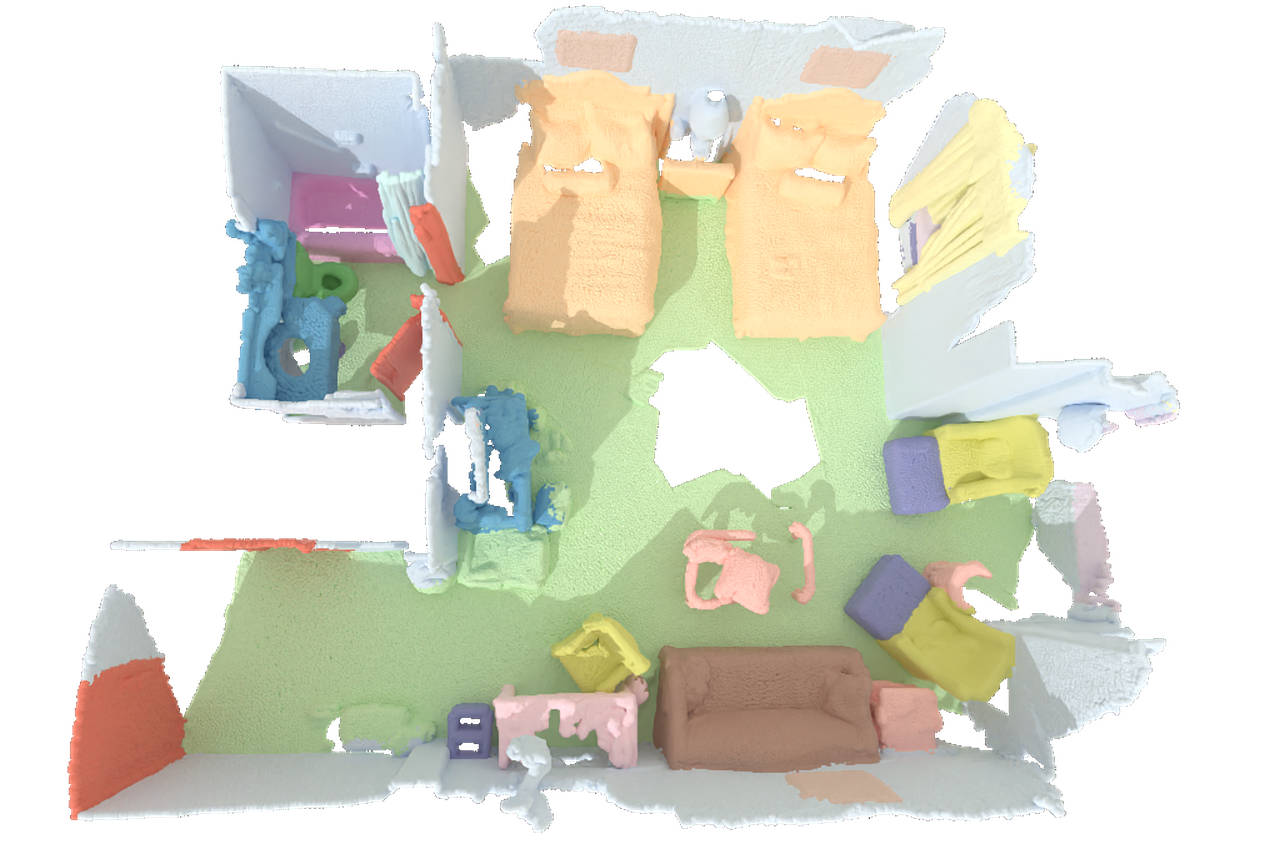}
    \end{subfigure}
         & 
    \begin{subfigure}[b]{0.28\textwidth}
      \includegraphics[width=\linewidth]{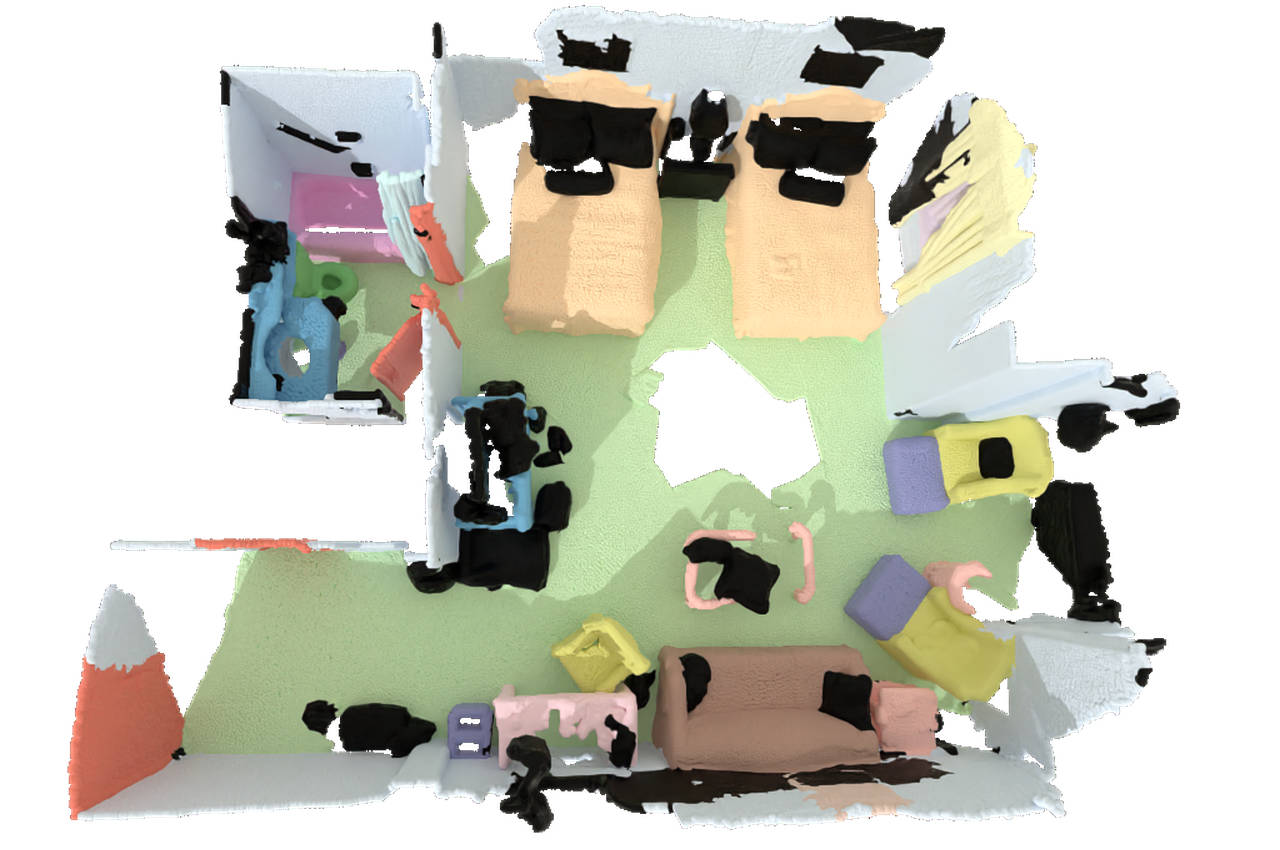}
    \end{subfigure}  \\

    \rotatebox{90}{ \qquad \texttt{scene0651\_00}}
    &
    \begin{subfigure}[b]{0.28\textwidth}
      \includegraphics[width=\linewidth]{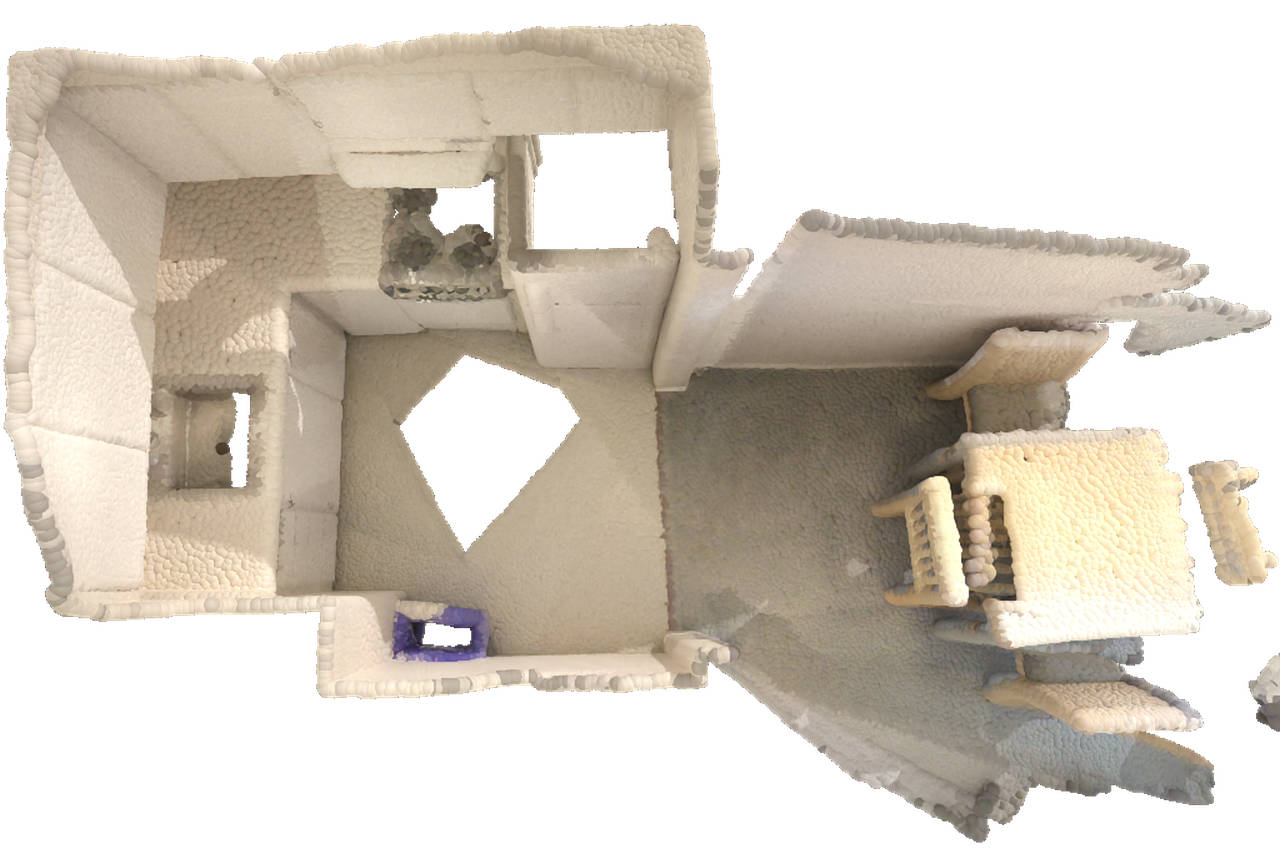}
      \caption{Input}
      \label{fig:quali_scannet_semseg:input}
    \end{subfigure}
         & 
    \begin{subfigure}[b]{0.28\textwidth}
      \includegraphics[width=\linewidth]{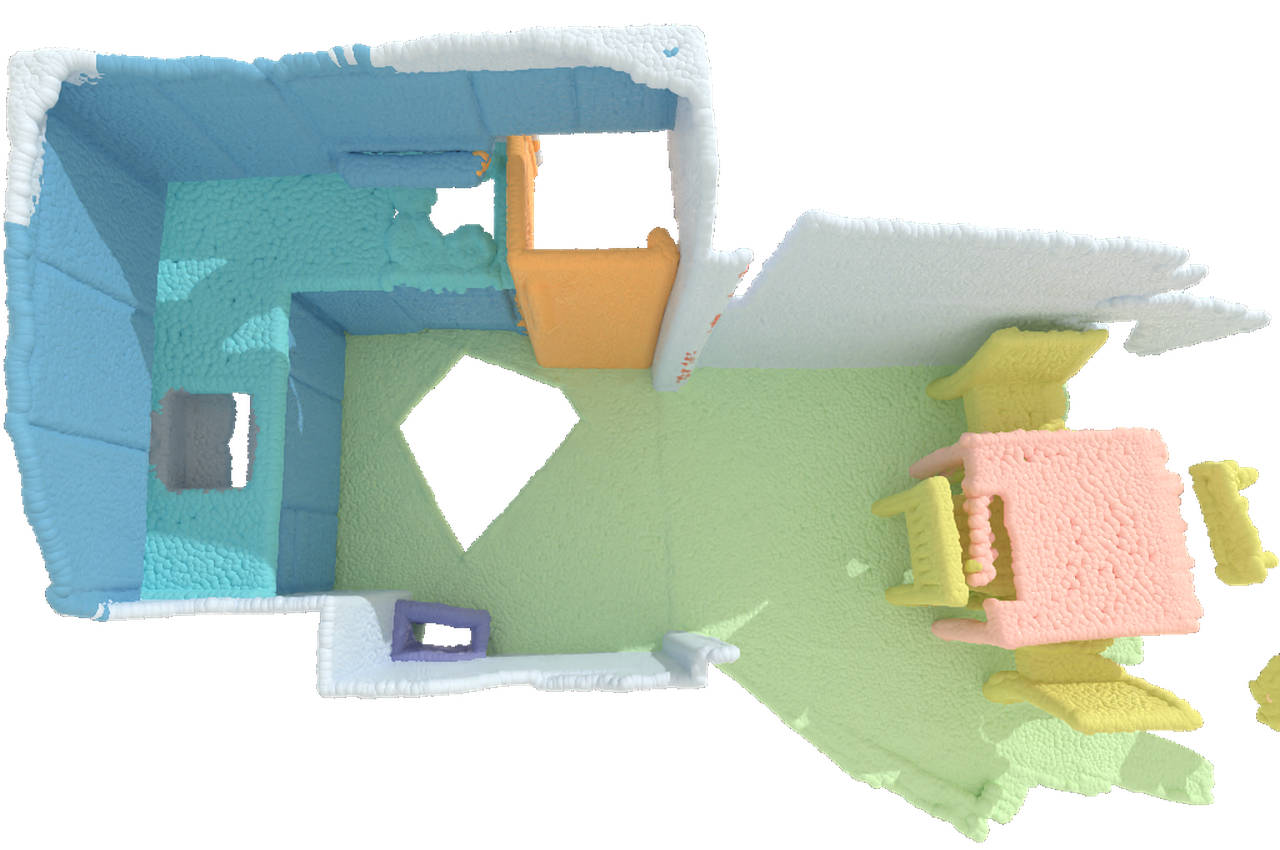}
      \caption{Prediction}
      \label{fig:quali_scannet_semseg:pred}
    \end{subfigure}
         & 
    \begin{subfigure}[b]{0.28\textwidth}
      \includegraphics[width=\linewidth]{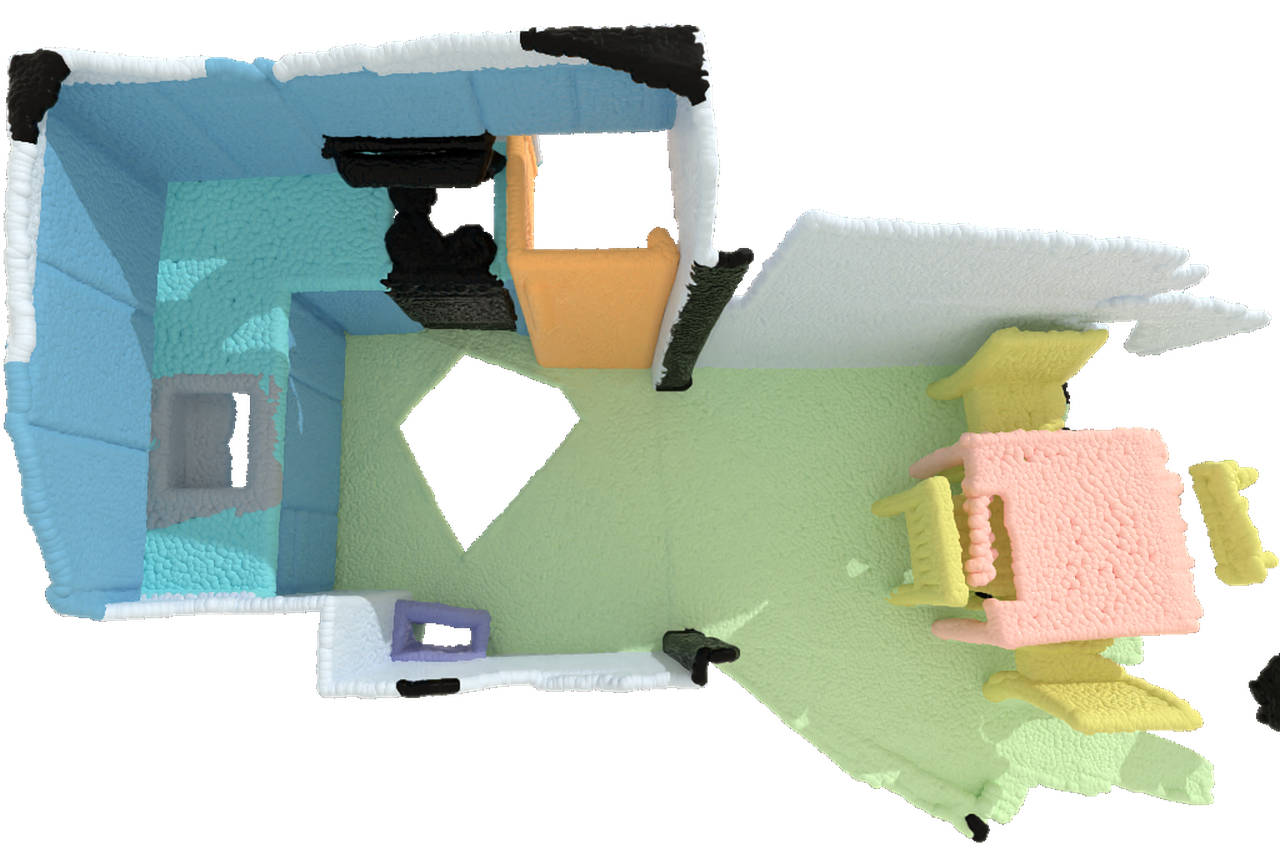}
      \caption{Ground Truth}
      \label{fig:quali_scannet_semseg:gt}
    \end{subfigure}  \\

    \end{tabular}

    \caption{
    {\bf ScanNet semantic segmentation.} 
    We present various scenes of the ScanNet validation set: the input point cloud, the semantic segmentation from \name{}-S, and the corresponding ground truth. 
    }
    \label{fig:quali_scannet_semseg}
\end{figure*}

\begin{figure*}
    \centering
    \begin{tabular}{@{}lccc@{}}

    \multicolumn{4}{c}{

    \begin{minipage}{\textwidth}
    \centering
        {\textcolor{stru3d_wall}{\ding{108}}}\,\,wall\,\, 
        {\textcolor{stru3d_floor}{\ding{108}}}\,\,floor\,\,
        {\textcolor{stru3d_cabinet}{\ding{108}}}\,\,cabinet\,\, 
        {\textcolor{stru3d_bed}{\ding{108}}}\,\,bed\,\,
        {\textcolor{stru3d_chair}{\ding{108}}}\,\,chair\,\, 
        {\textcolor{stru3d_sofa}{\ding{108}}}\,\,sofa\,\,
        {\textcolor{stru3d_table}{\ding{108}}}\,\,table\,\, 
        {\textcolor{stru3d_door}{\ding{108}}}\,\,door\,\,
        {\textcolor{stru3d_window}{\ding{108}}}\,\,window\,\, 
        {\textcolor{stru3d_picture}{\ding{108}}}\,\,picture\,\,
        {\textcolor{stru3d_desk}{\ding{108}}}\,\,desk\,\, 
        {\textcolor{stru3d_shelves}{\ding{108}}}\,\,shelves\,\,
        {\textcolor{stru3d_curtain}{\ding{108}}}\,\,curtain\,\, 
        {\textcolor{stru3d_dresser}{\ding{108}}}\,\,dresser\,\,
        {\textcolor{stru3d_pillow}{\ding{108}}}\,\,pillow\,\, 
        {\textcolor{stru3d_mirror}{\ding{108}}}\,\,mirror\,\,
        {\textcolor{stru3d_ceiling}{\ding{108}}}\,\,ceiling\,\, 
        {\textcolor{stru3d_refrigerator}{\ding{108}}}\,\,refrigerator\,\,
        {\textcolor{stru3d_television}{\ding{108}}}\,\,television\,\, 
        {\textcolor{stru3d_nightstand}{\ding{108}}}\,\,nightstand\,\,
        {\textcolor{stru3d_sink}{\ding{108}}}\,\,sink\,\, 
        {\textcolor{stru3d_lamp}{\ding{108}}}\,\,lamp\,\,
        {\textcolor{stru3d_otherstructure}{\ding{108}}}\,\,other structure\,\, 
        {\textcolor{stru3d_otherfurniture}{\ding{108}}}\,\,other furniture\,\,
        {\textcolor{stru3d_otherprop}{\ding{108}}}\,\,other properties\,\,
        
    \end{minipage}
    } 
    \vspace{15pt}
    
    \\ 
    \rotatebox{90}{\scriptsize \texttt{scene\_03022\_room\_8765}}
    &
    \begin{subfigure}[b]{0.28\textwidth}
      \includegraphics[width=\linewidth]{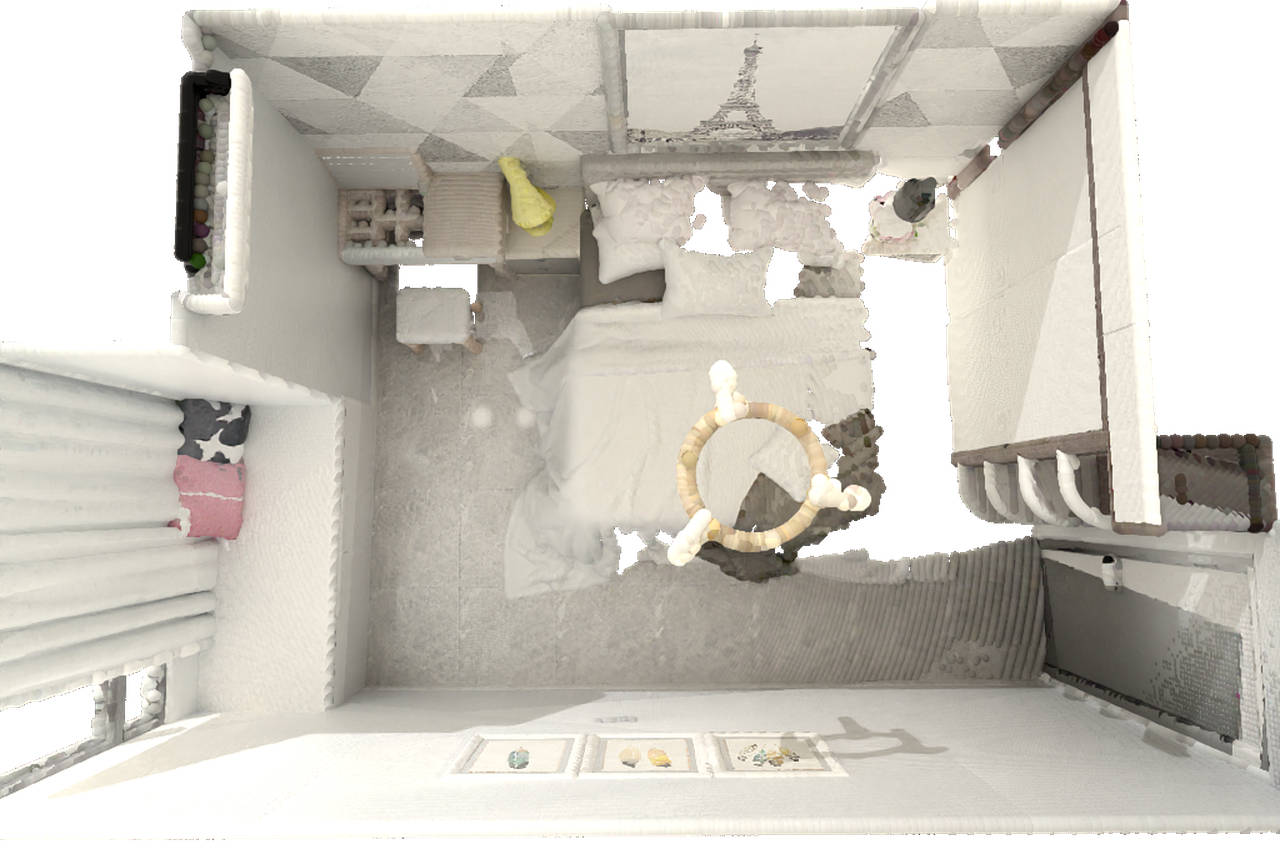}
    \end{subfigure}
         & 
    \begin{subfigure}[b]{0.28\textwidth}
      \includegraphics[width=\linewidth]{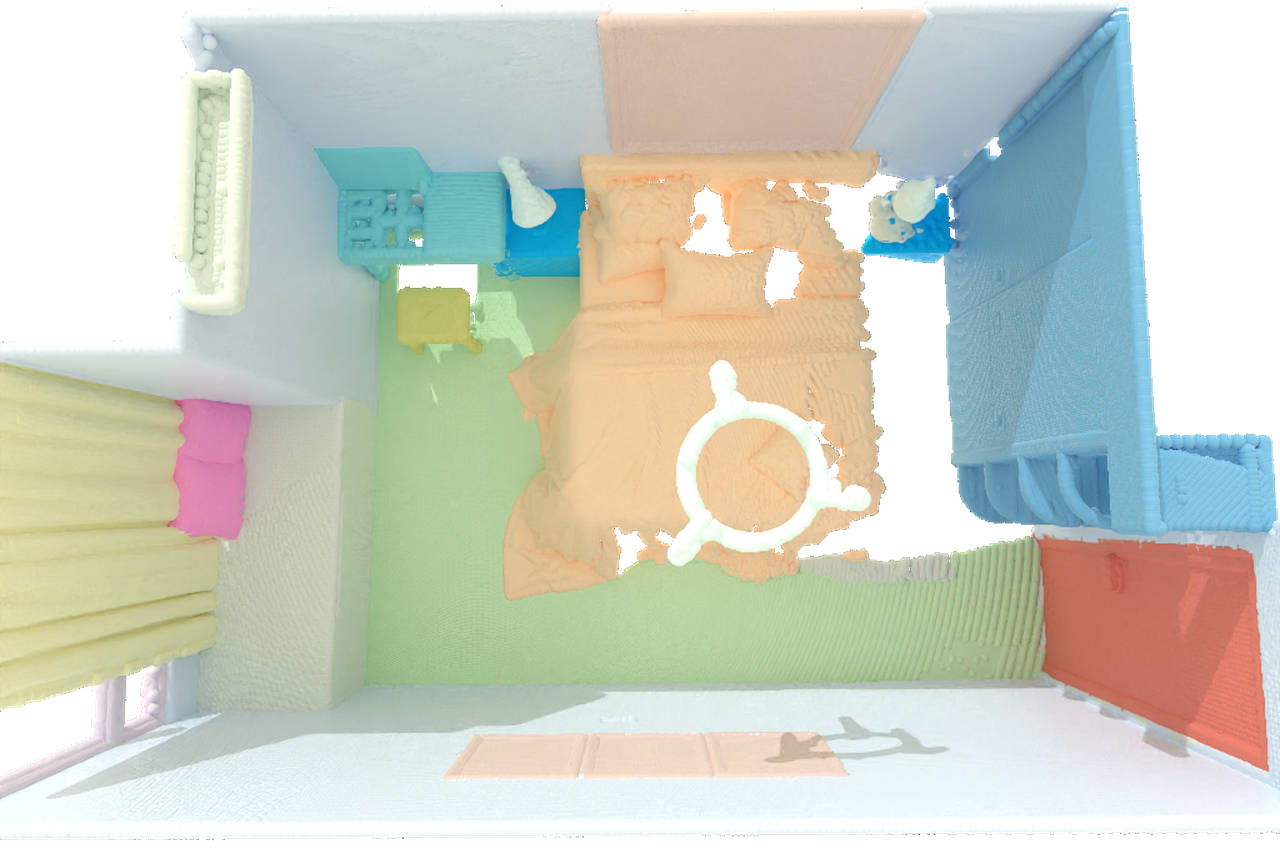}
    \end{subfigure}
         & 
    \begin{subfigure}[b]{0.28\textwidth}
      \includegraphics[width=\linewidth]{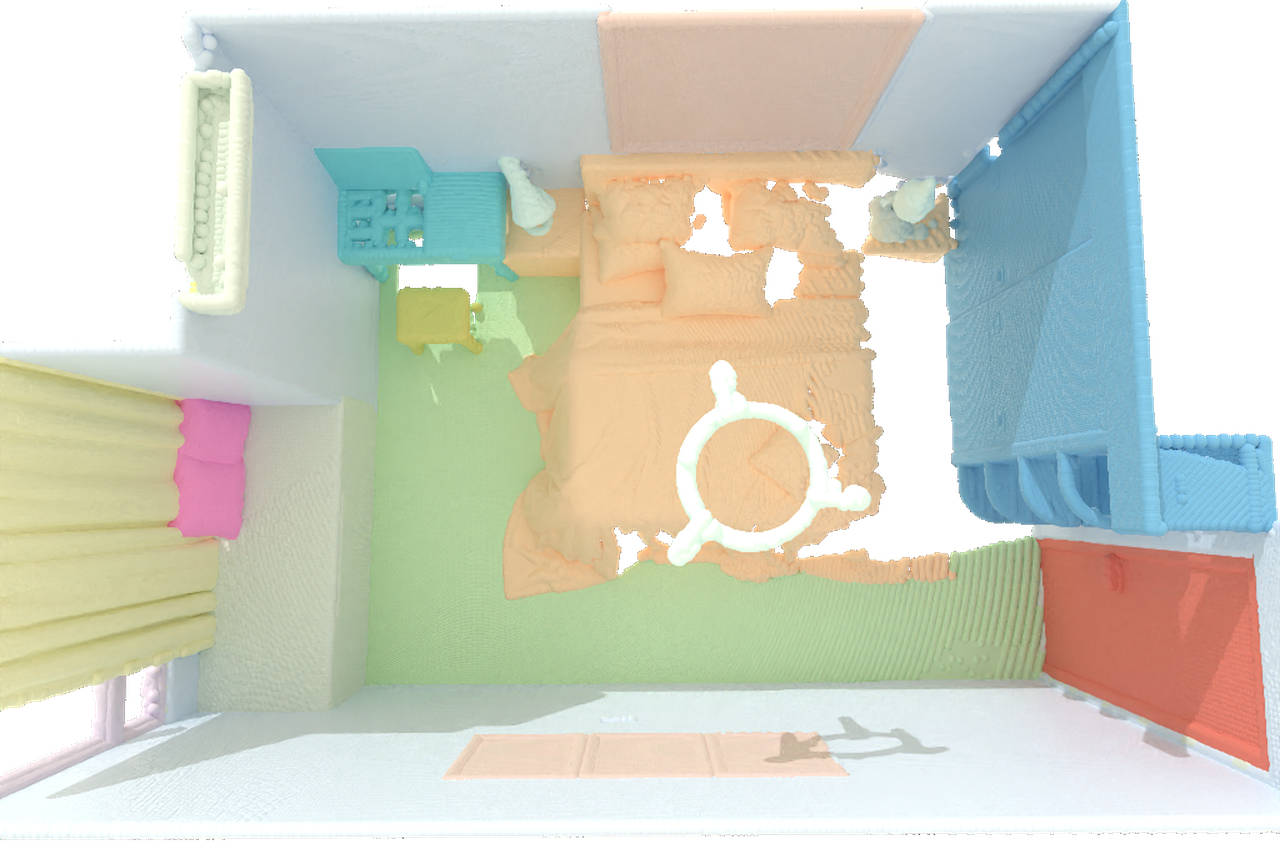}
    \end{subfigure}  \\
    \vspace{-2pt}
    \rotatebox{90}{\scriptsize \texttt{scene\_03034\_room\_401}}
    &
    \begin{subfigure}[b]{0.28\textwidth}
      \includegraphics[width=\linewidth]{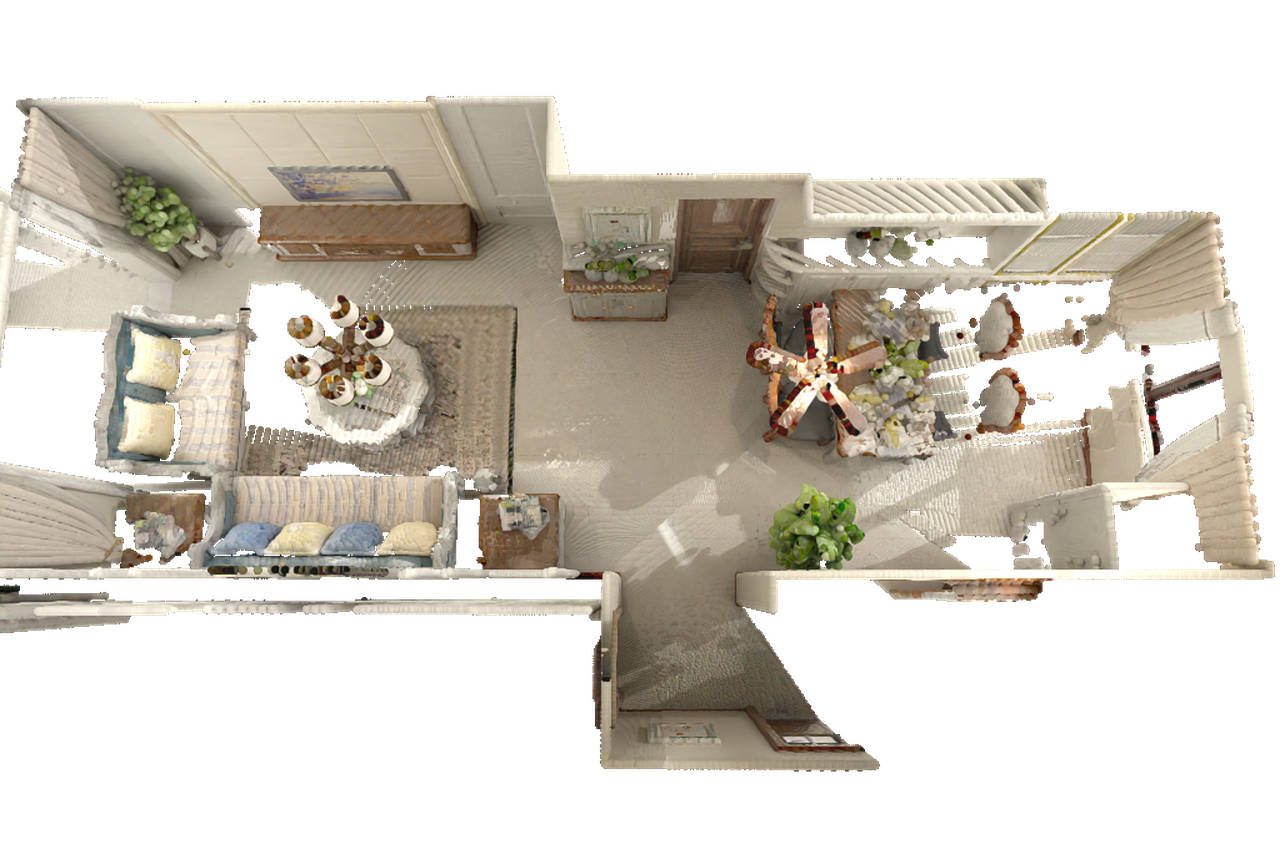}
    \end{subfigure}
         & 
    \begin{subfigure}[b]{0.28\textwidth}
      \includegraphics[width=\linewidth]{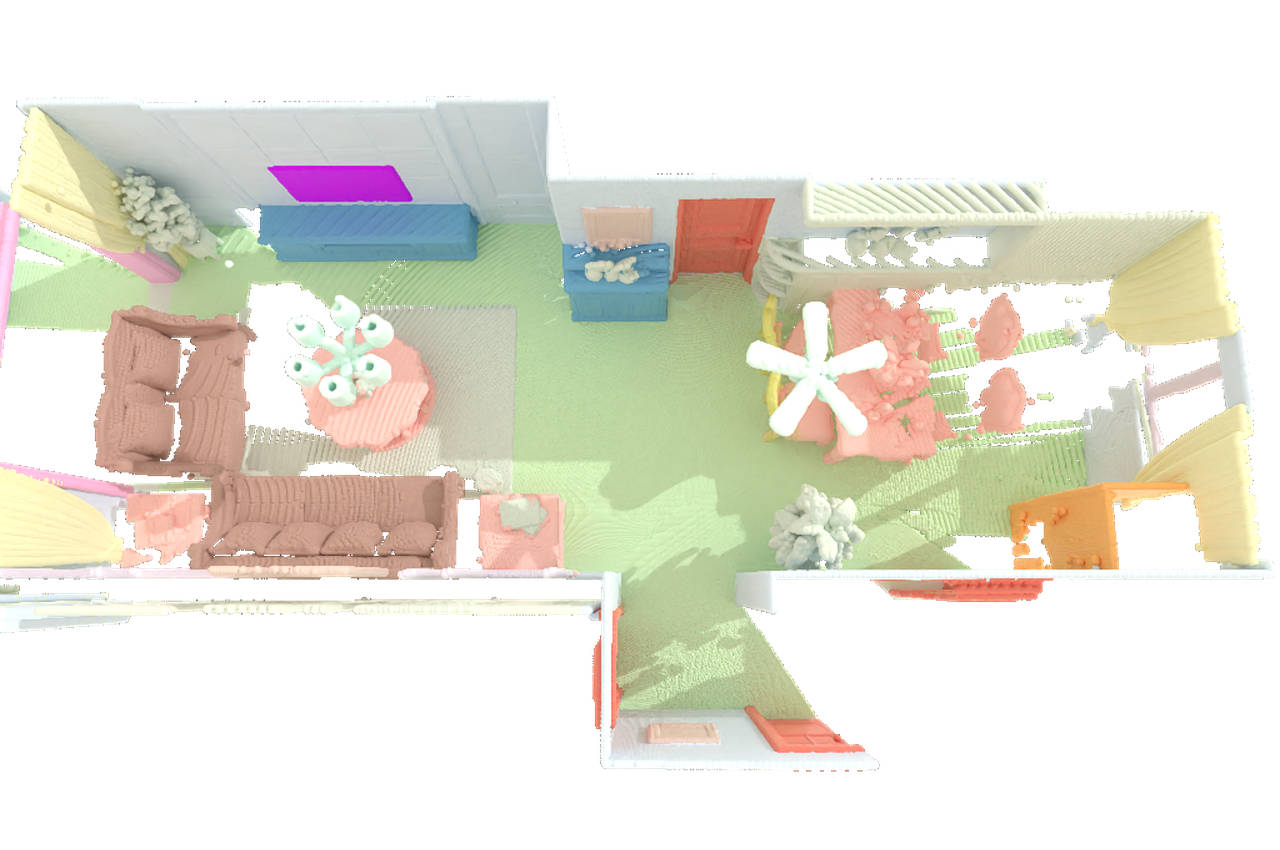}
    \end{subfigure}
         & 
    \begin{subfigure}[b]{0.28\textwidth}
      \includegraphics[width=\linewidth]{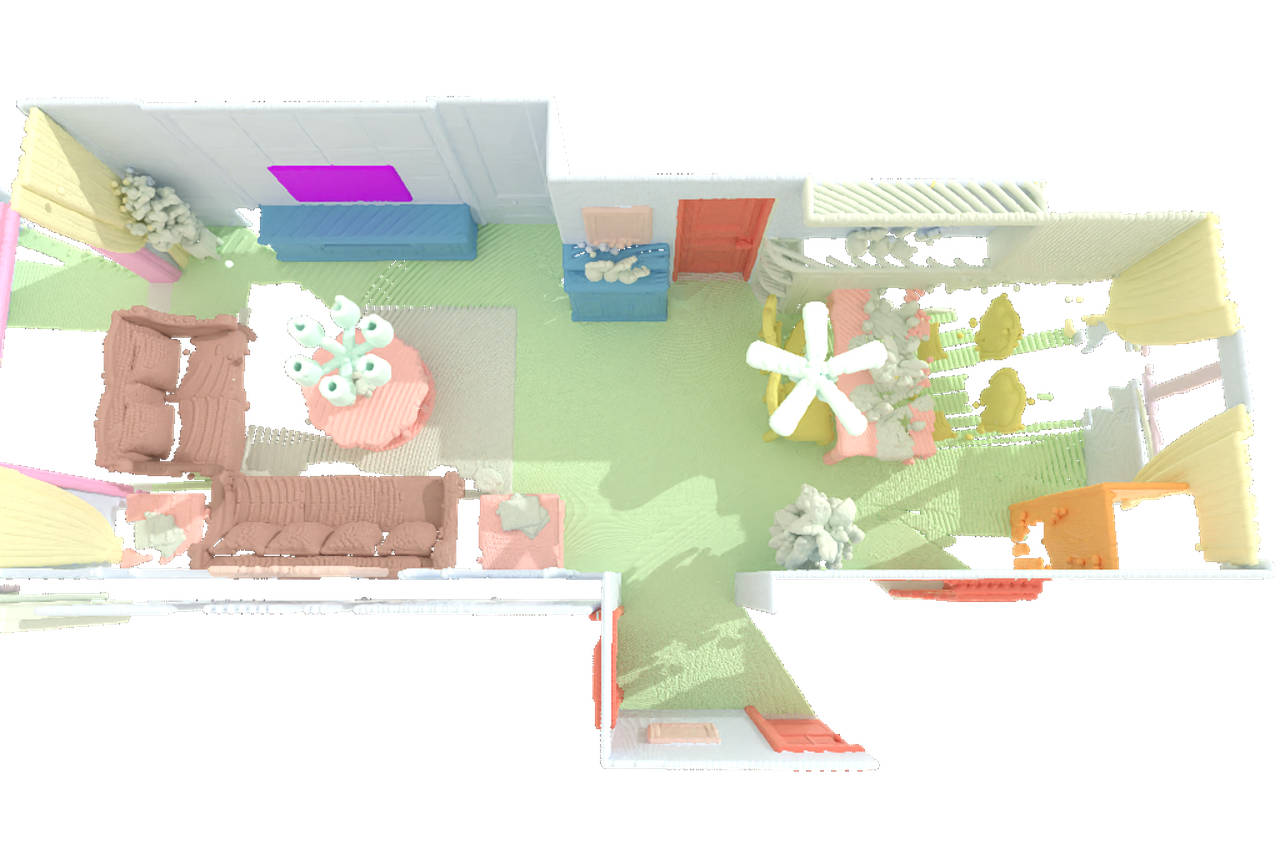}
    \end{subfigure}  \\
    \vspace{-2pt}
    \rotatebox{90}{\scriptsize \texttt{scene\_03113\_room\_560}}
    &
    \begin{subfigure}[b]{0.28\textwidth}
      \includegraphics[width=\linewidth]{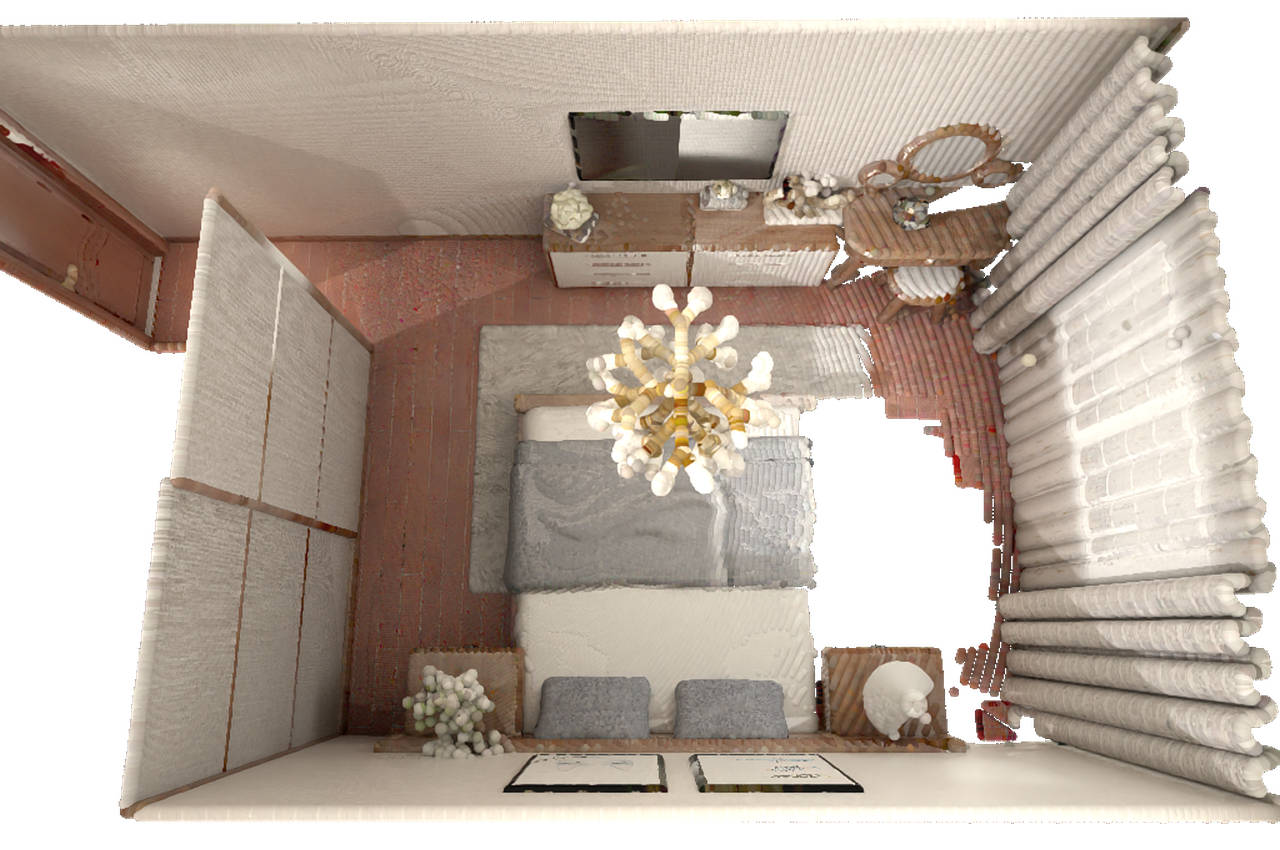}
    \end{subfigure}
         & 
    \begin{subfigure}[b]{0.28\textwidth}
      \includegraphics[width=\linewidth]{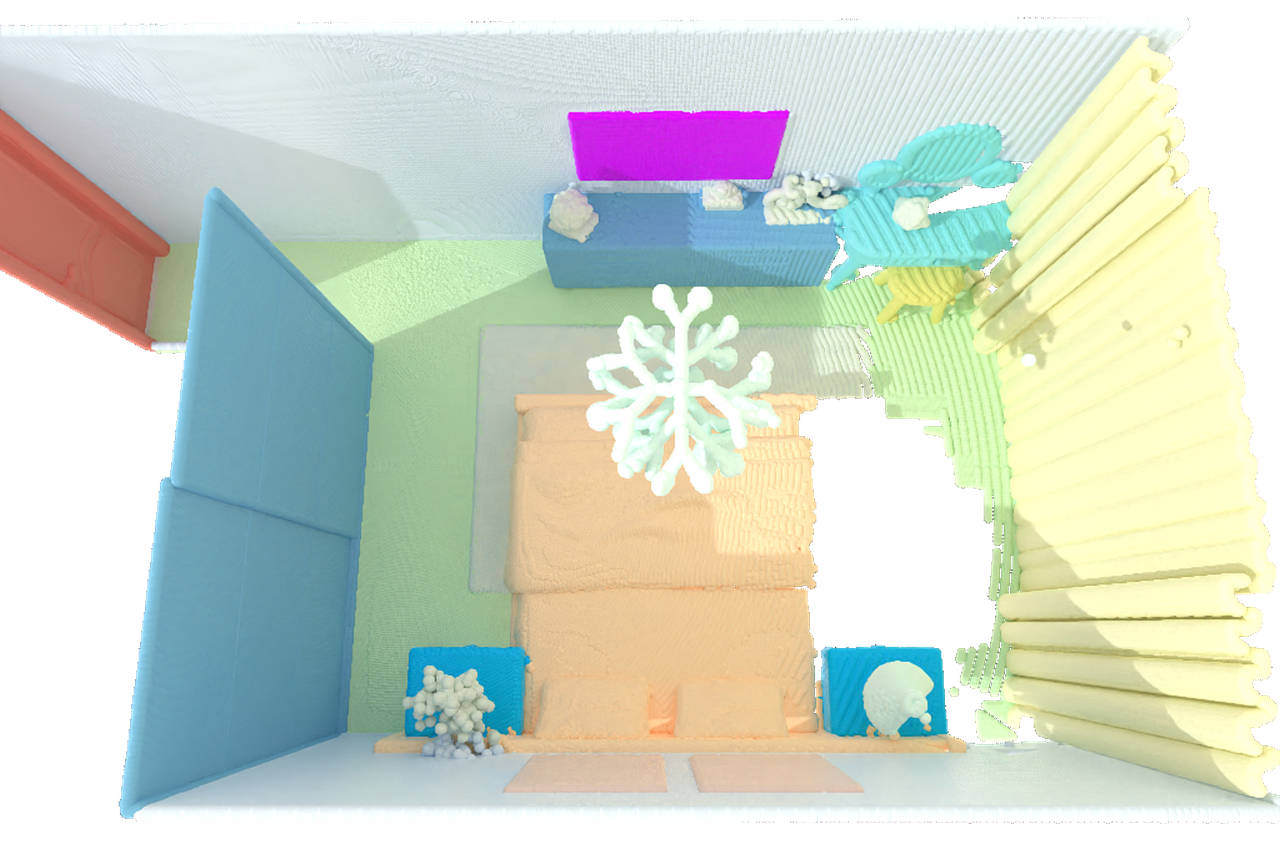}
    \end{subfigure}
         & 
    \begin{subfigure}[b]{0.28\textwidth}
      \includegraphics[width=\linewidth]{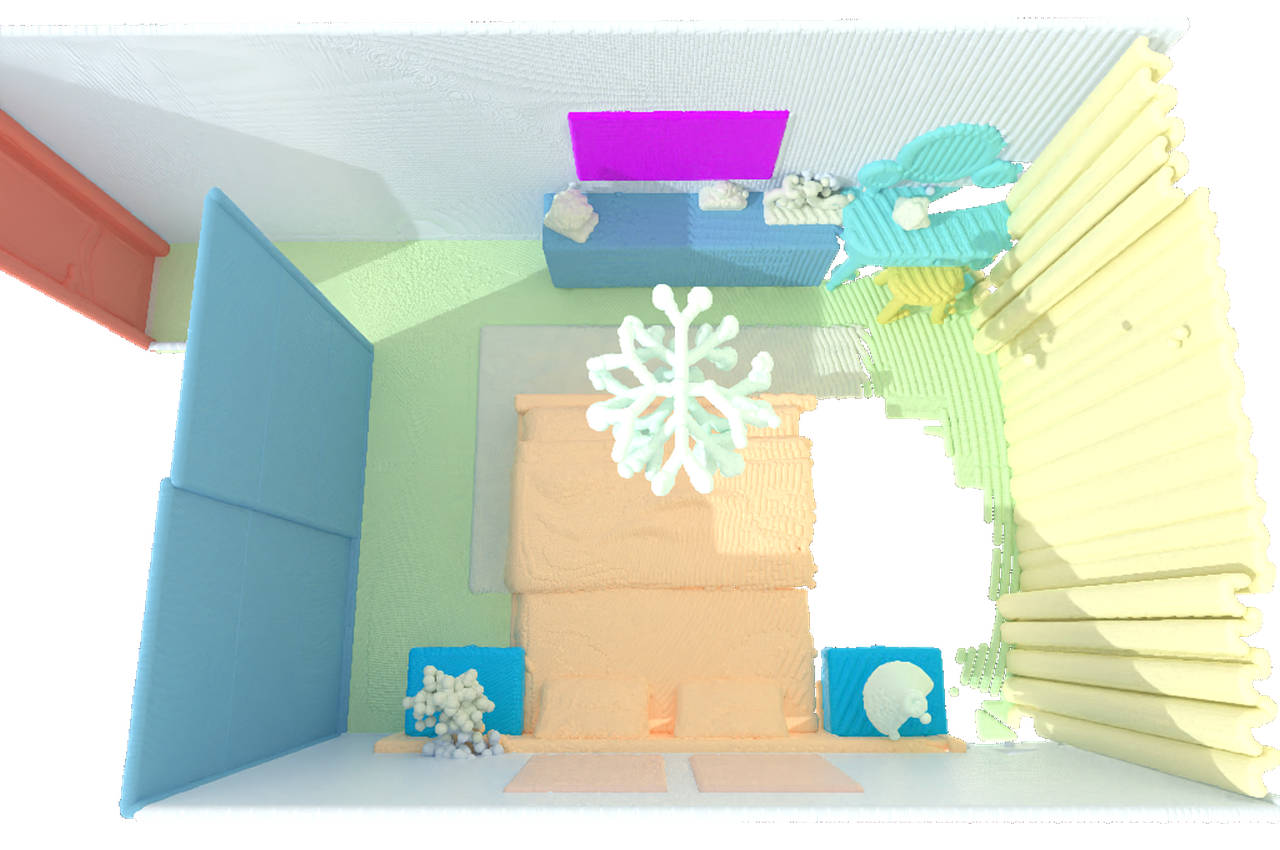}
    \end{subfigure}  \\
    \vspace{-2pt}
    \rotatebox{90}{\scriptsize \texttt{scene\_03195\_room\_1764}}
    &
    \begin{subfigure}[b]{0.28\textwidth}
      \includegraphics[width=\linewidth]{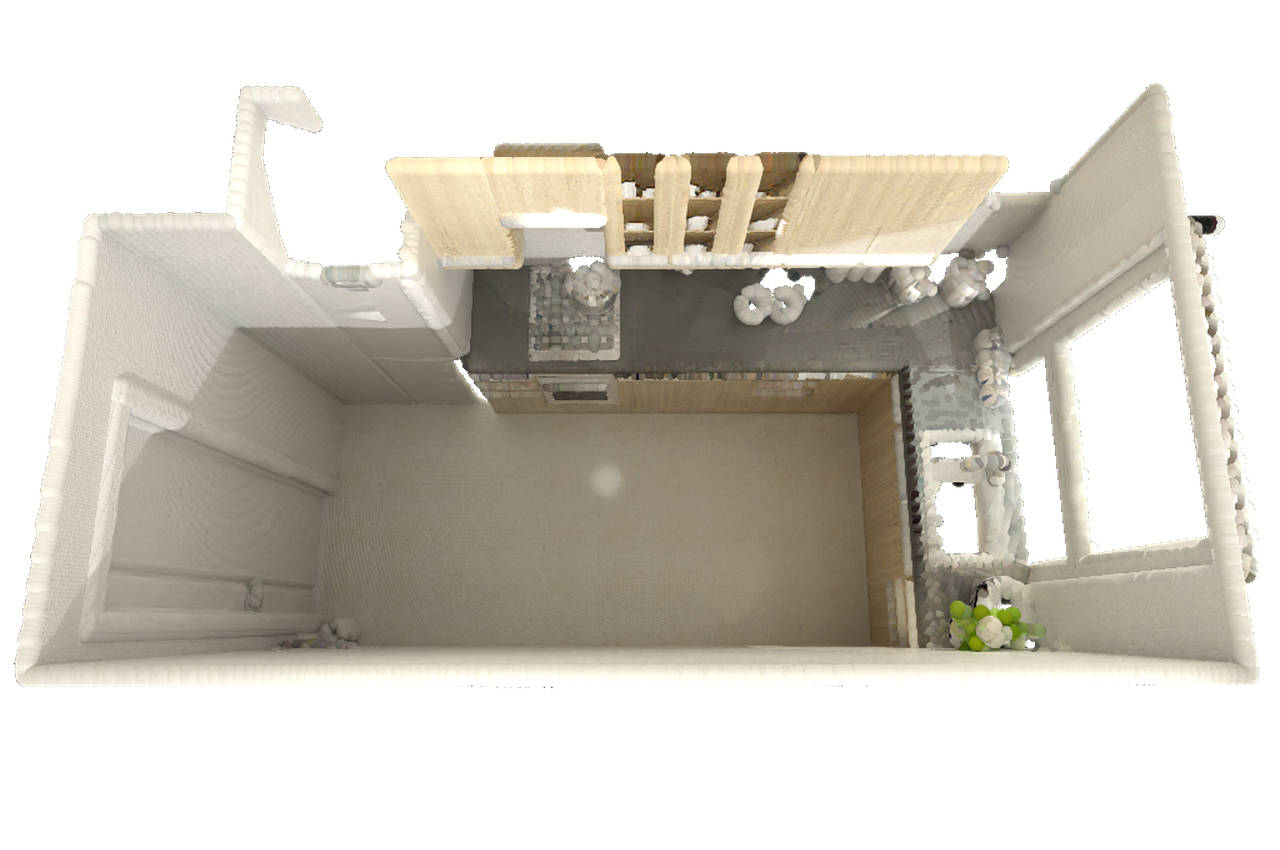}
    \end{subfigure}
         & 
    \begin{subfigure}[b]{0.28\textwidth}
      \includegraphics[width=\linewidth]{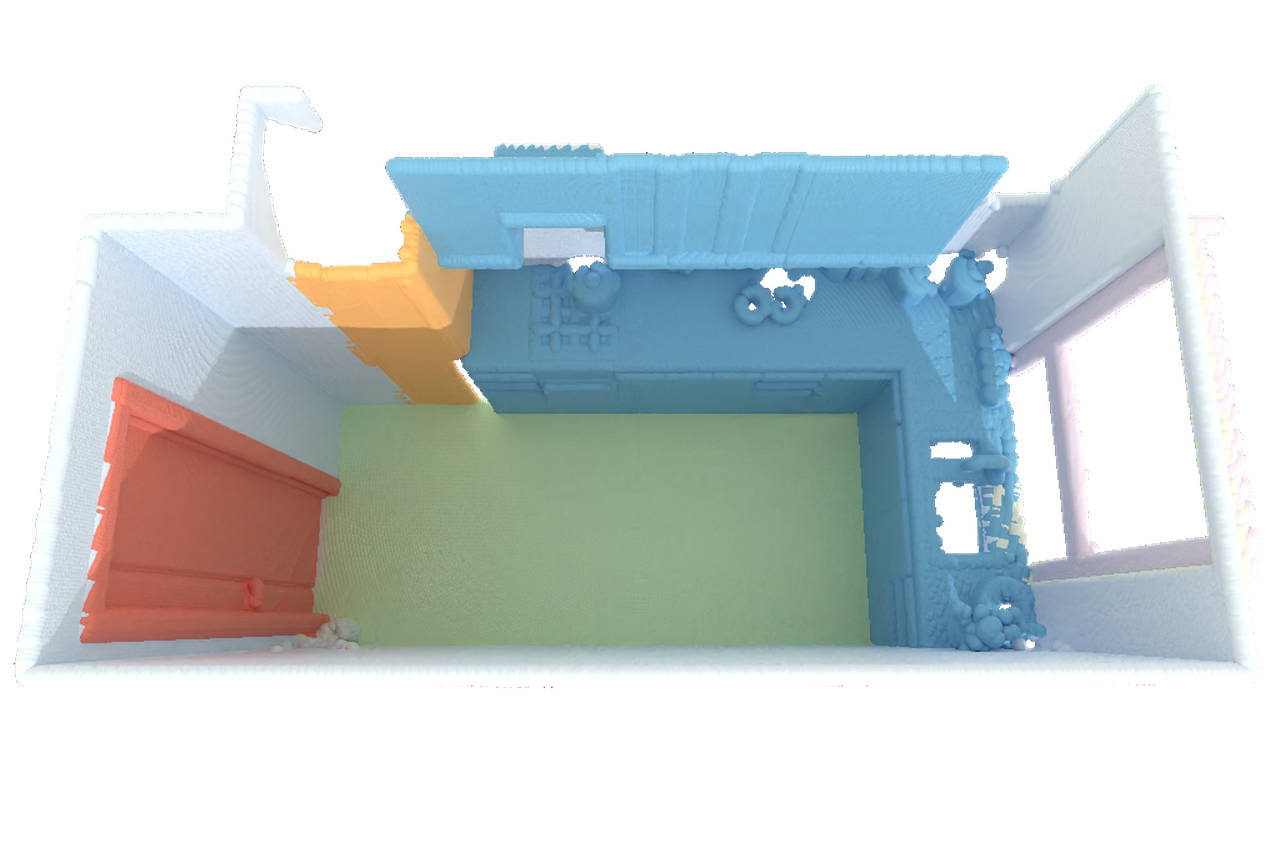}
    \end{subfigure}
         & 
    \begin{subfigure}[b]{0.28\textwidth}
      \includegraphics[width=\linewidth]{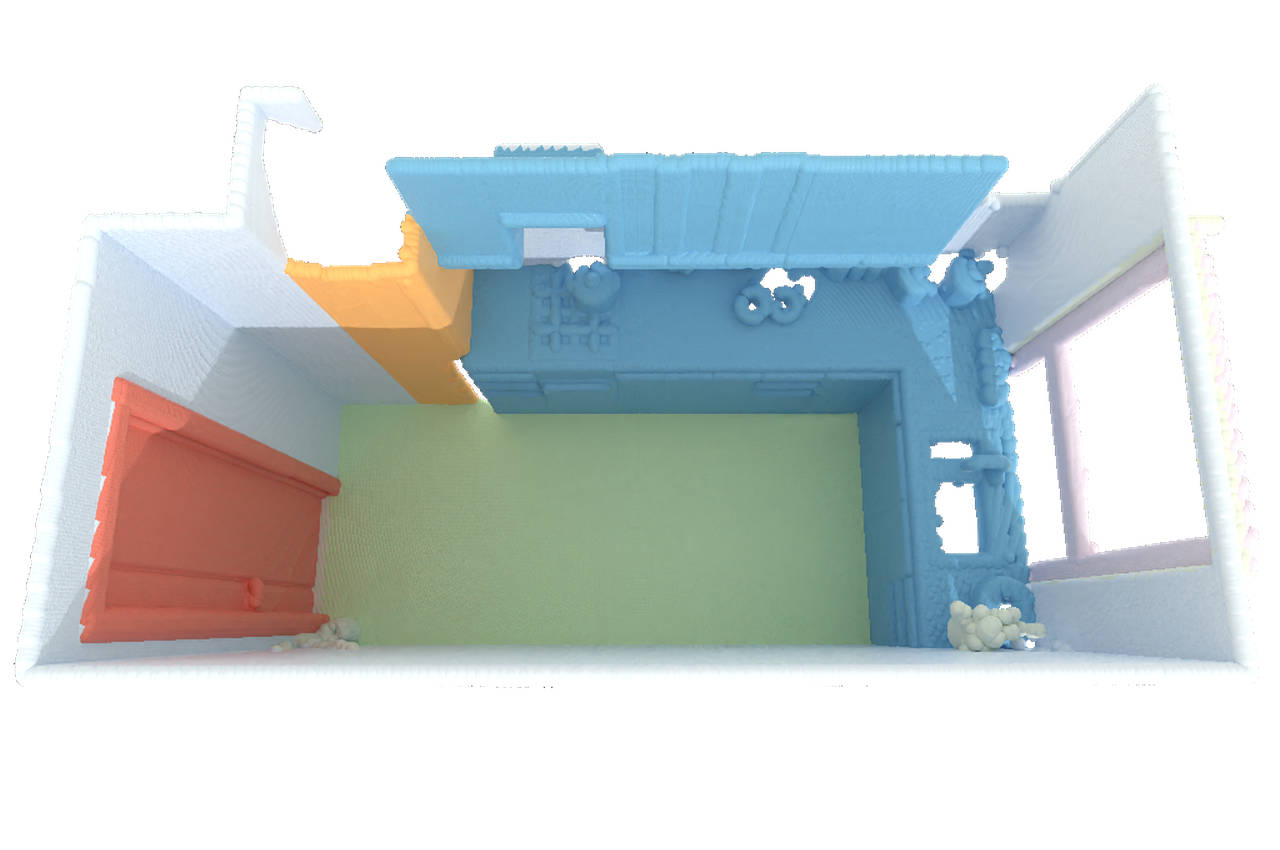}
    \end{subfigure}  \\
    \vspace{-2pt}
    \rotatebox{90}{\scriptsize \texttt{scene\_03223\_room\_4894}}
    &
    \begin{subfigure}[b]{0.28\textwidth}
      \includegraphics[width=\linewidth]{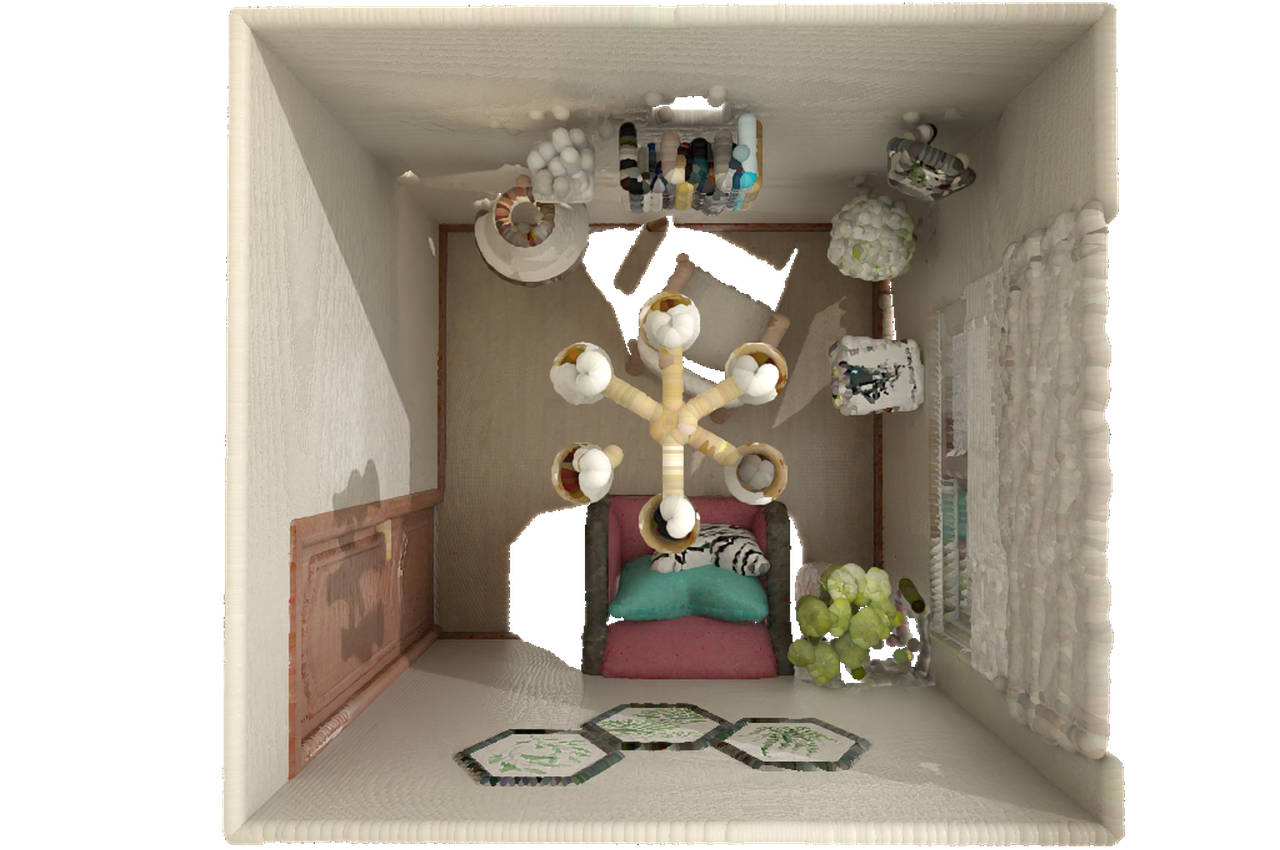}
    \end{subfigure}
         & 
    \begin{subfigure}[b]{0.28\textwidth}
      \includegraphics[width=\linewidth]{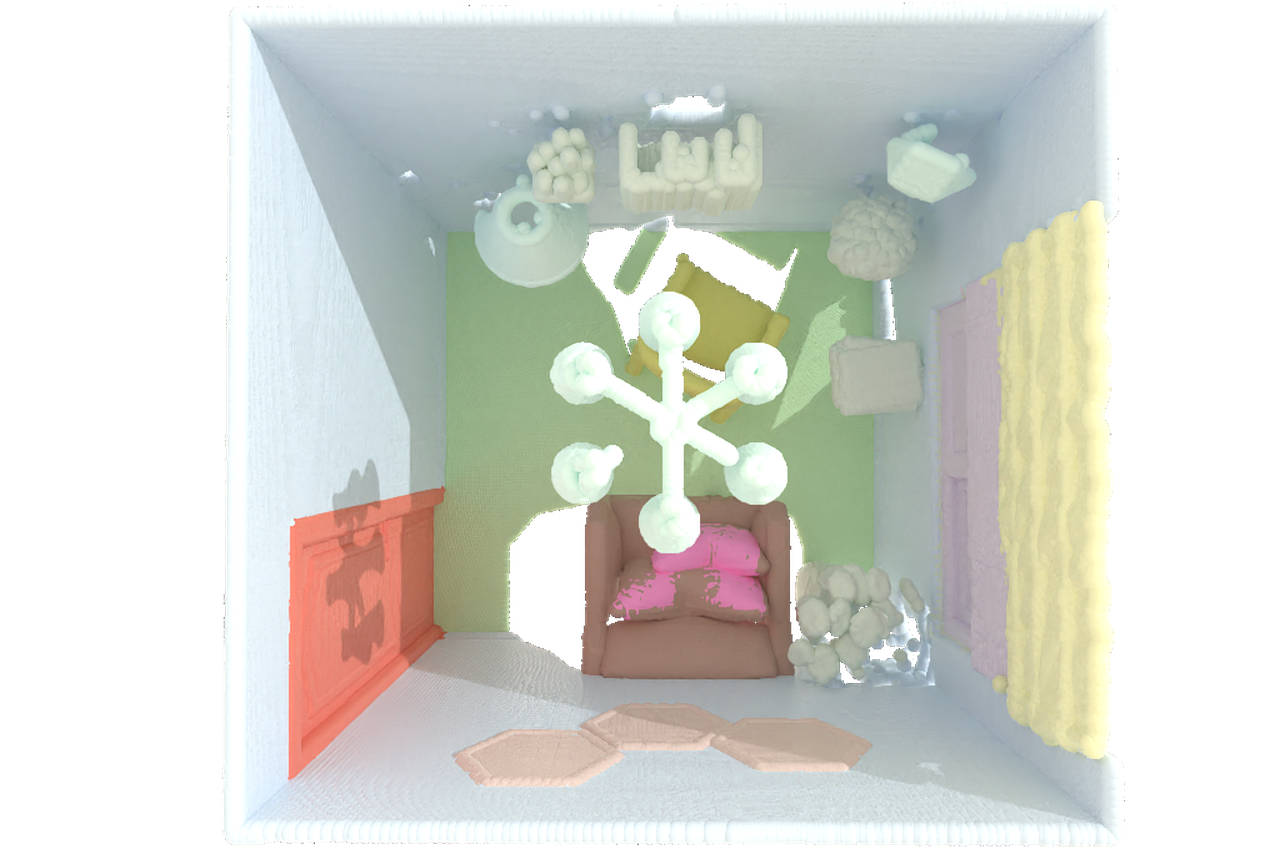}
    \end{subfigure}
         & 
    \begin{subfigure}[b]{0.28\textwidth}
      \includegraphics[width=\linewidth]{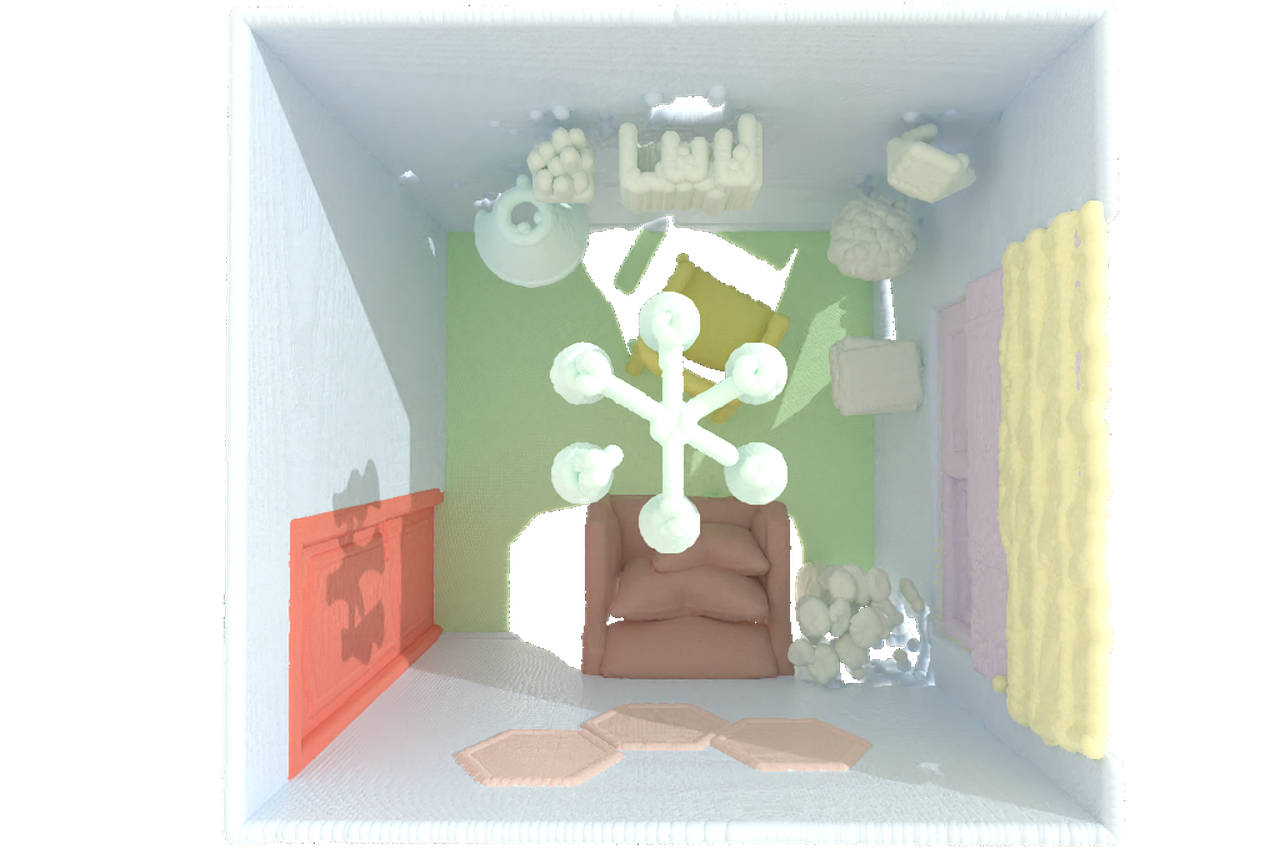}
    \end{subfigure}  \\
    \vspace{-2pt}
    \rotatebox{90}{\scriptsize \texttt{scene\_03237\_room\_2846}}
    &
    \begin{subfigure}[b]{0.28\textwidth}
      \includegraphics[width=\linewidth]{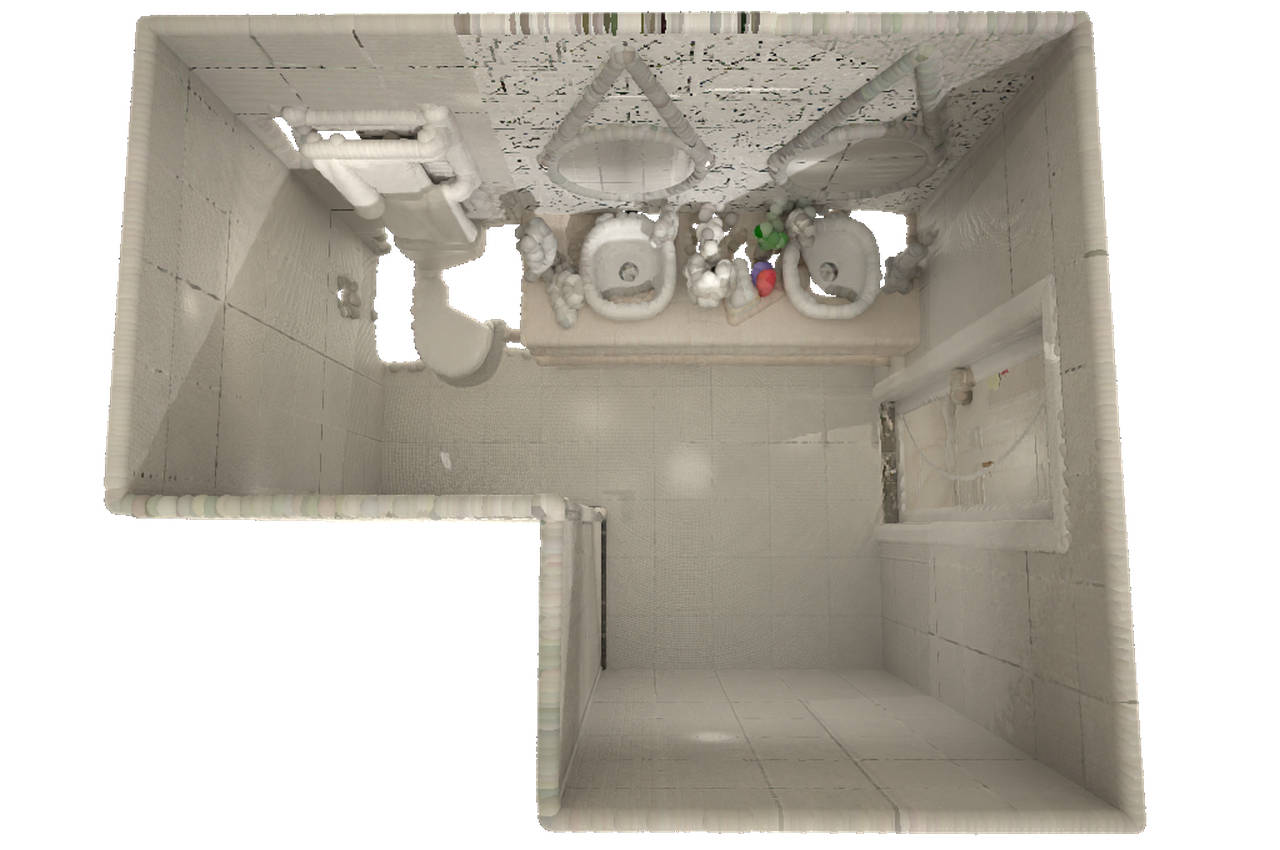}
      \caption{Input}
      \label{fig:quali_structure3d_semseg:input}
    \end{subfigure}
         & 
    \begin{subfigure}[b]{0.28\textwidth}
      \includegraphics[width=\linewidth]{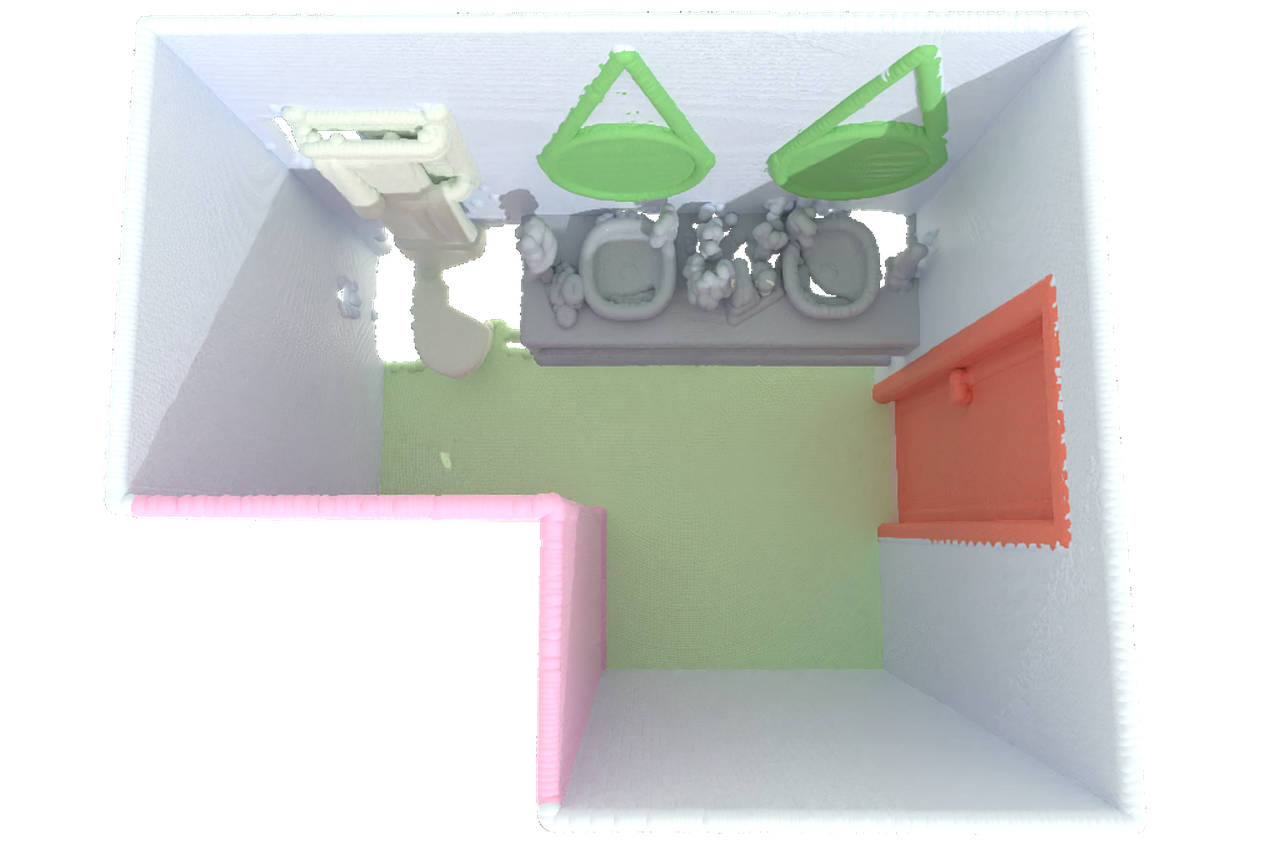}
      \caption{Prediction}
      \label{fig:quali_structure3d_semseg:pred}
    \end{subfigure}
         & 
    \begin{subfigure}[b]{0.28\textwidth}
      \includegraphics[width=\linewidth]{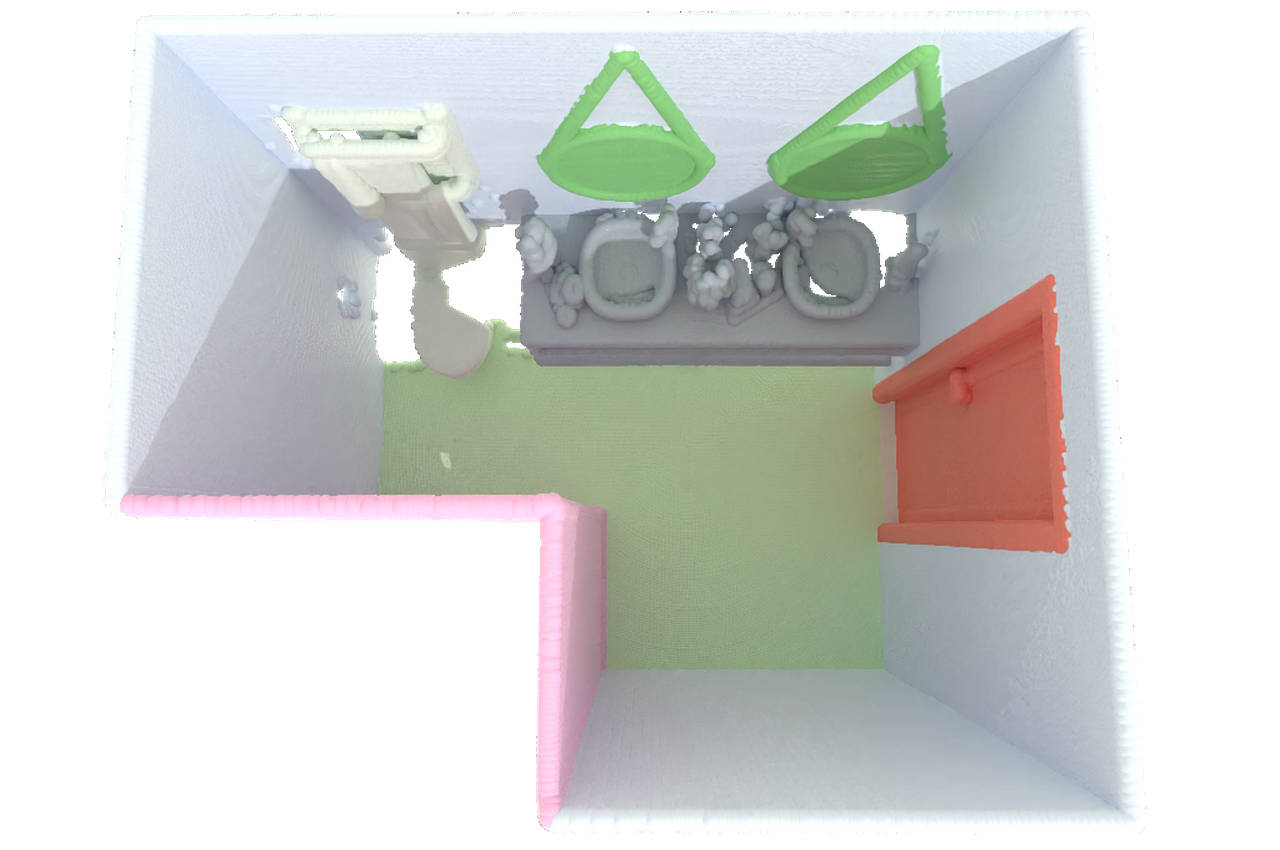}
      \caption{Ground Truth}
      \label{fig:quali_structure3d_semseg:gt}
    \end{subfigure}  \\

    \end{tabular}
    \caption{
    {\bf Structured3D semantic segmentation.} 
    We present various scenes of the Structured3D validation set: the input point cloud, the semantic segmentation from \name{}-S, and the corresponding ground truth. 
    }
    \label{fig:quali_structure3d_semseg}
\end{figure*}

\begin{figure*}
    \centering
    \begin{tabular}{@{}lccc@{}}

    \rotatebox{90}{ \quad \texttt{scene0011\_01}}
    &
    \begin{subfigure}[b]{0.28\textwidth}
      \includegraphics[width=\linewidth]{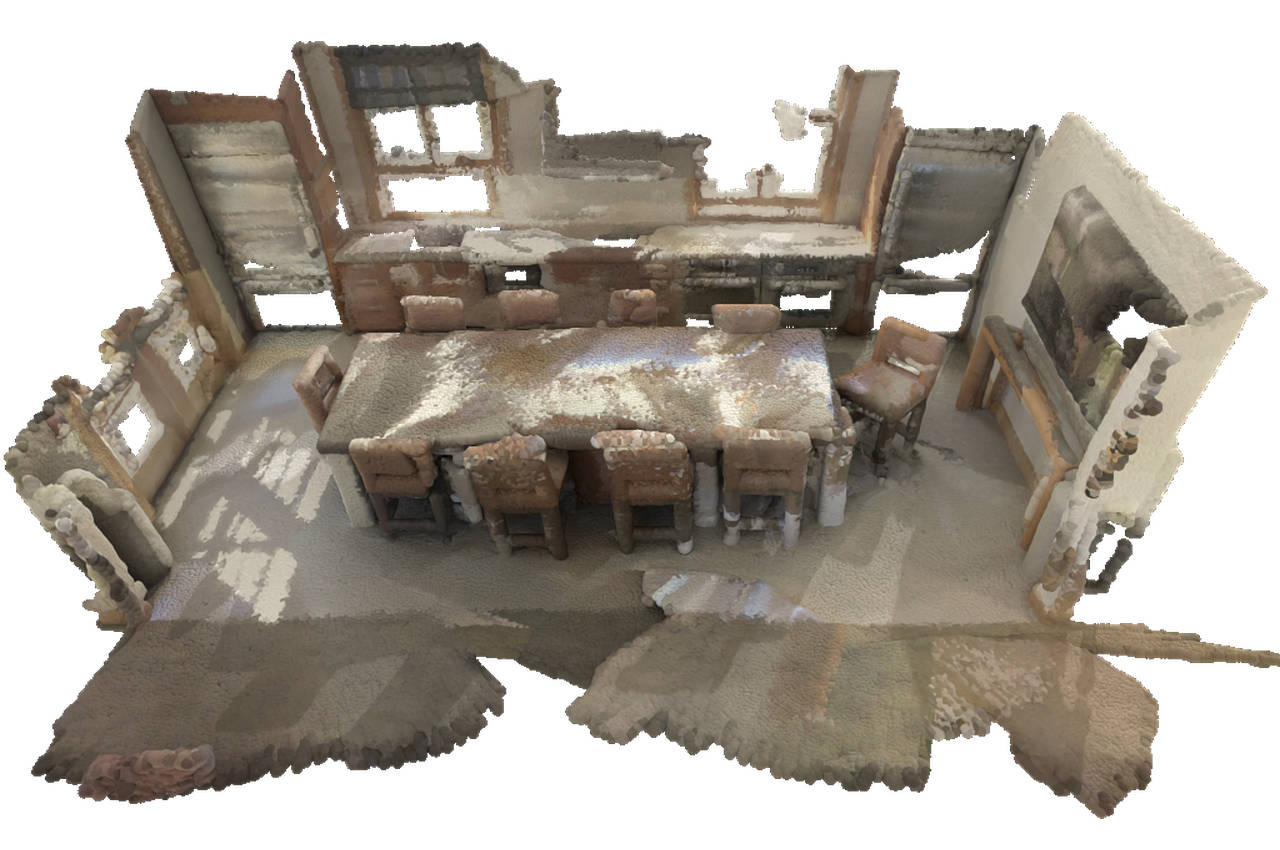}
    \end{subfigure}
         & 
    \begin{subfigure}[b]{0.28\textwidth}
      \includegraphics[width=\linewidth]{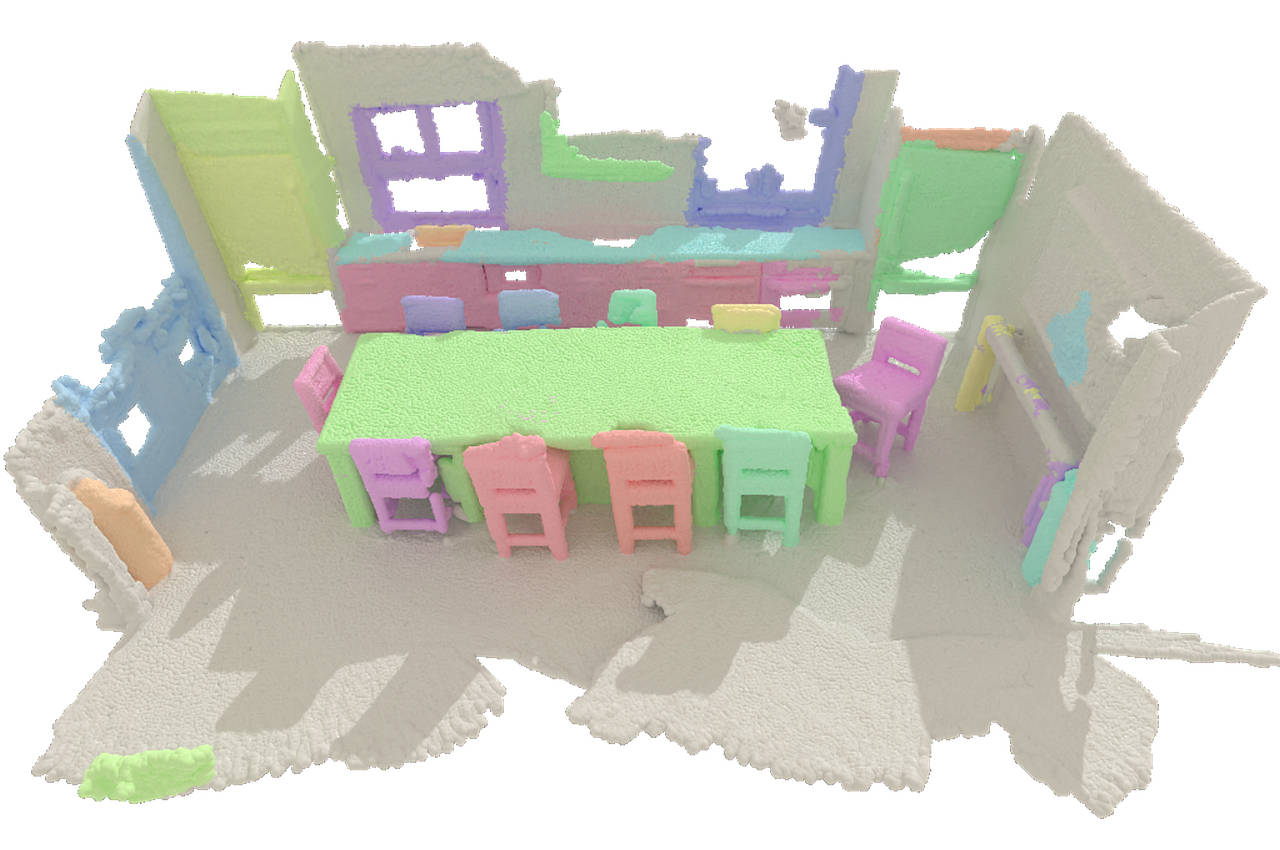}
    \end{subfigure}
         & 
    \begin{subfigure}[b]{0.28\textwidth}
      \includegraphics[width=\linewidth]{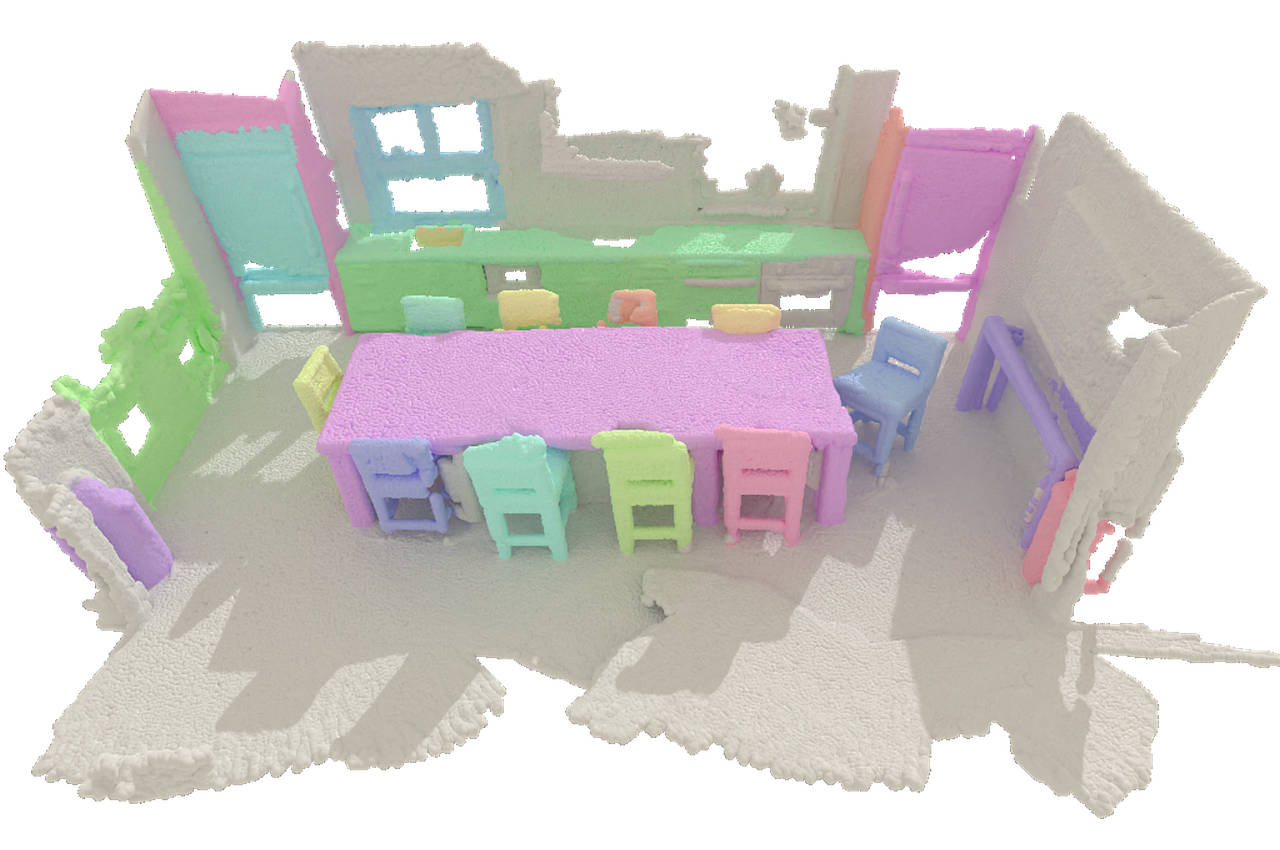}
    \end{subfigure}  \\

    \rotatebox{90}{ \quad \texttt{scene0164\_00}}
    &
    \begin{subfigure}[b]{0.28\textwidth}
      \includegraphics[width=\linewidth]{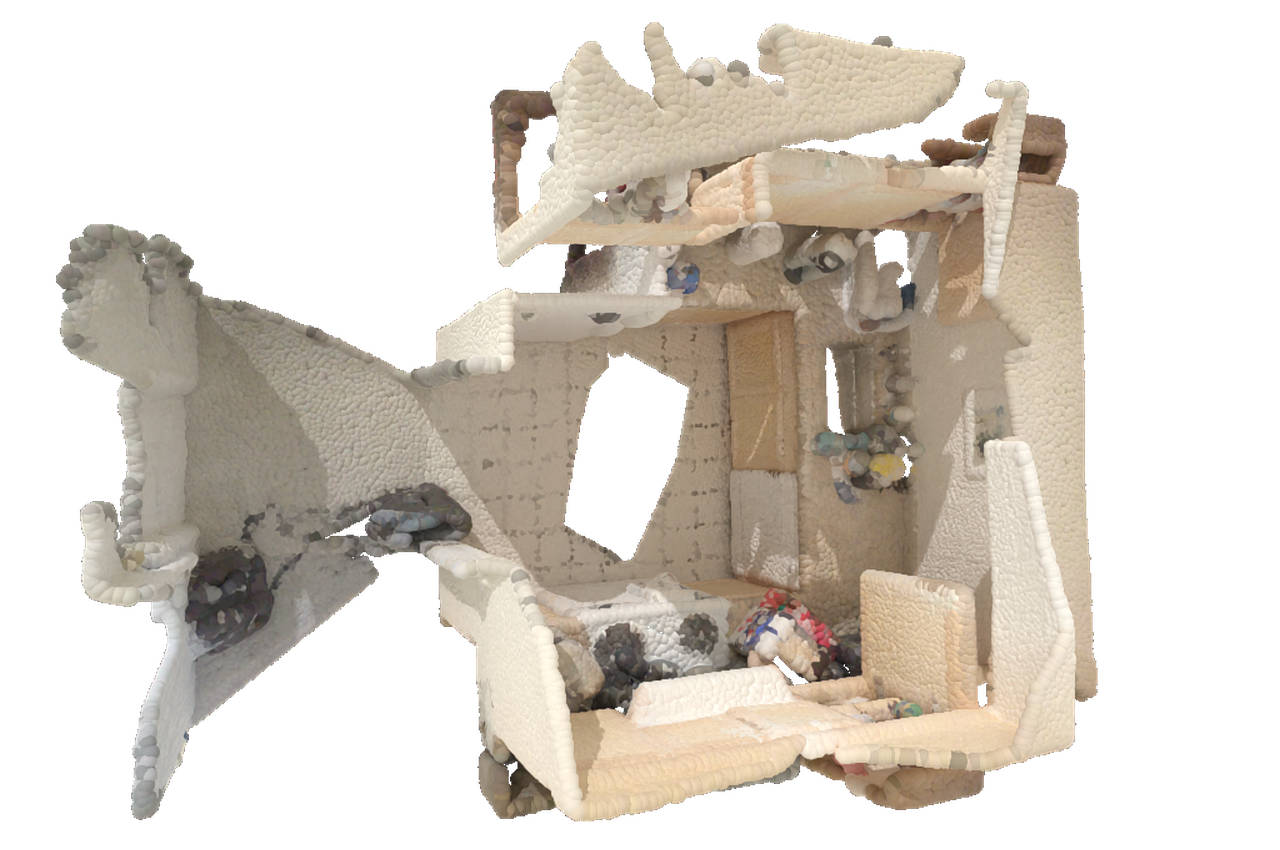}
    \end{subfigure}
         & 
    \begin{subfigure}[b]{0.28\textwidth}
      \includegraphics[width=\linewidth]{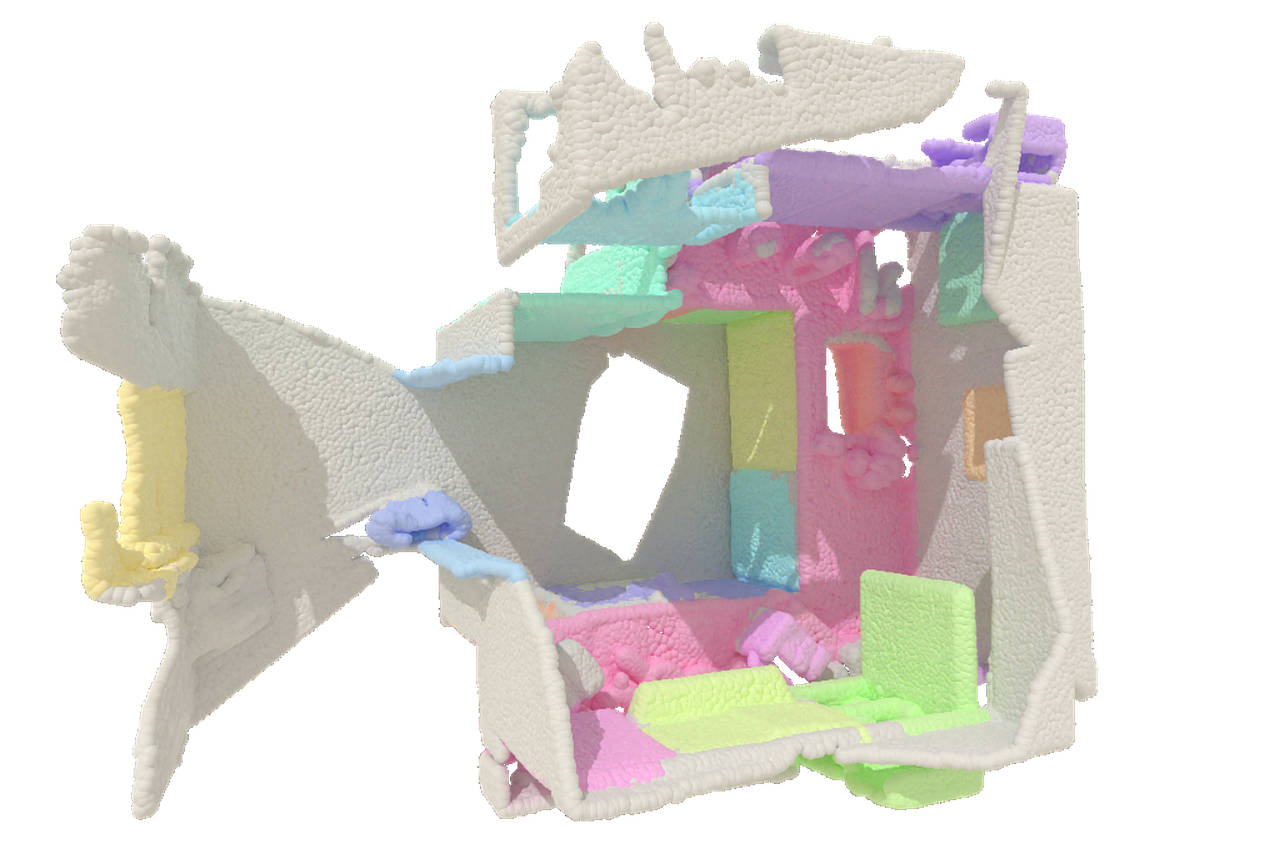}
    \end{subfigure}
         & 
    \begin{subfigure}[b]{0.28\textwidth}
      \includegraphics[width=\linewidth]{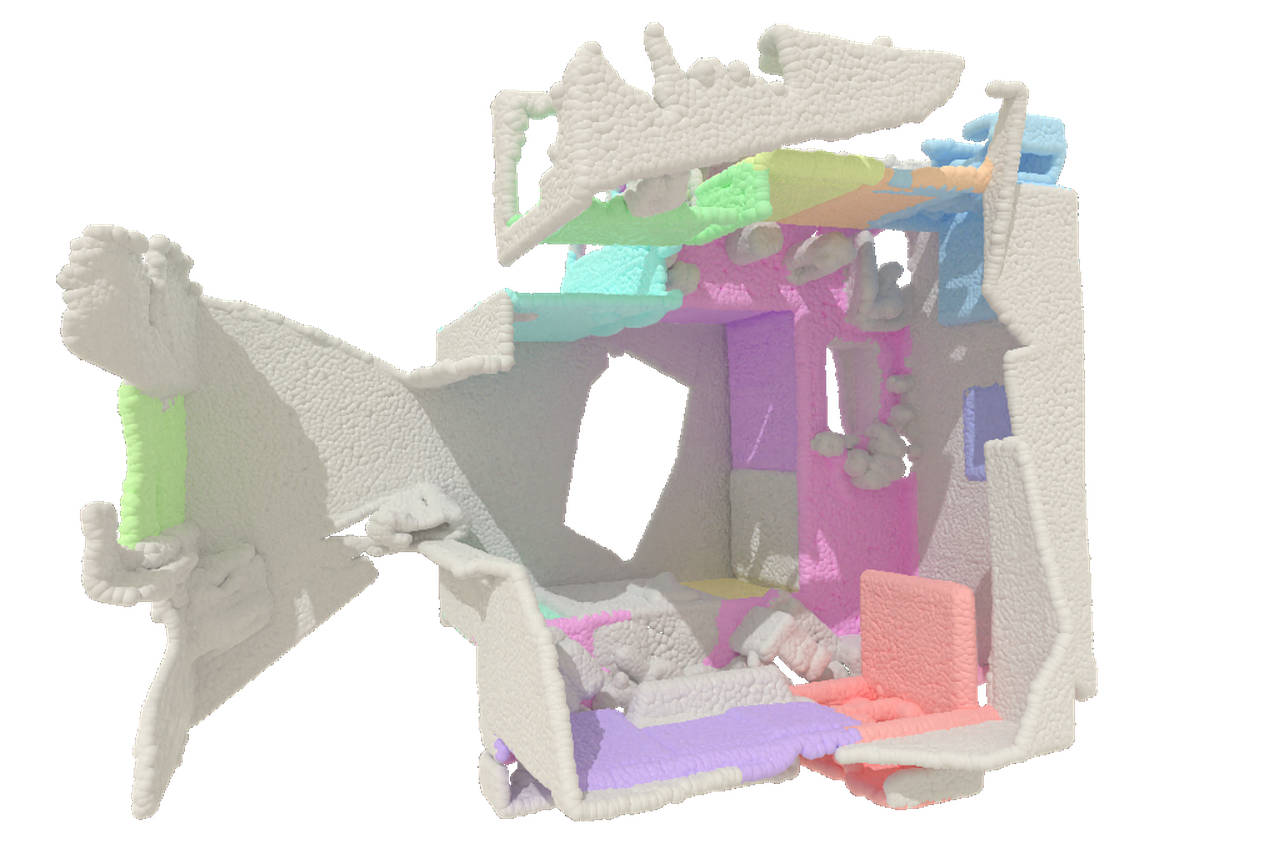}
    \end{subfigure}  \\

    \rotatebox{90}{ \quad \texttt{scene0591\_02}}
    &
    \begin{subfigure}[b]{0.28\textwidth}
      \includegraphics[width=\linewidth]{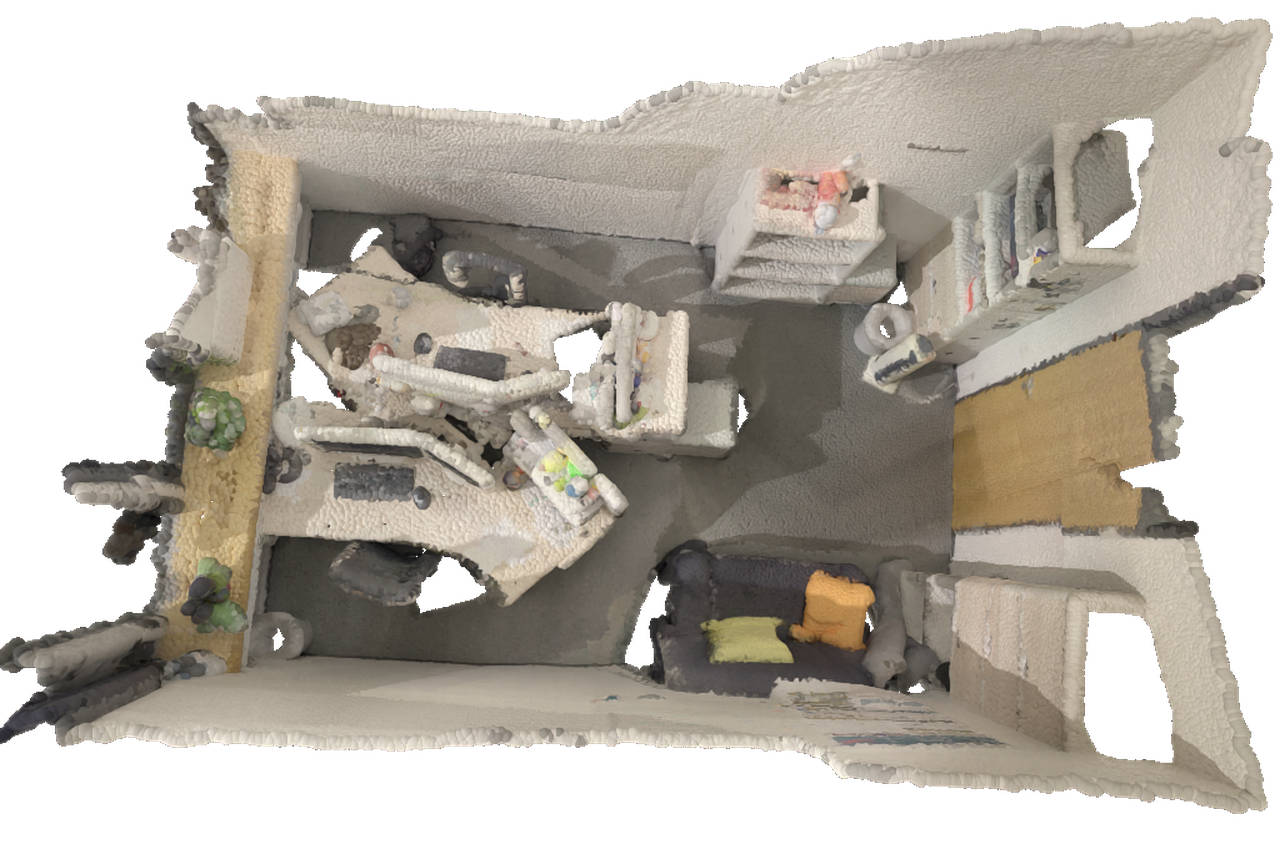}
    \end{subfigure}
         & 
    \begin{subfigure}[b]{0.28\textwidth}
      \includegraphics[width=\linewidth]{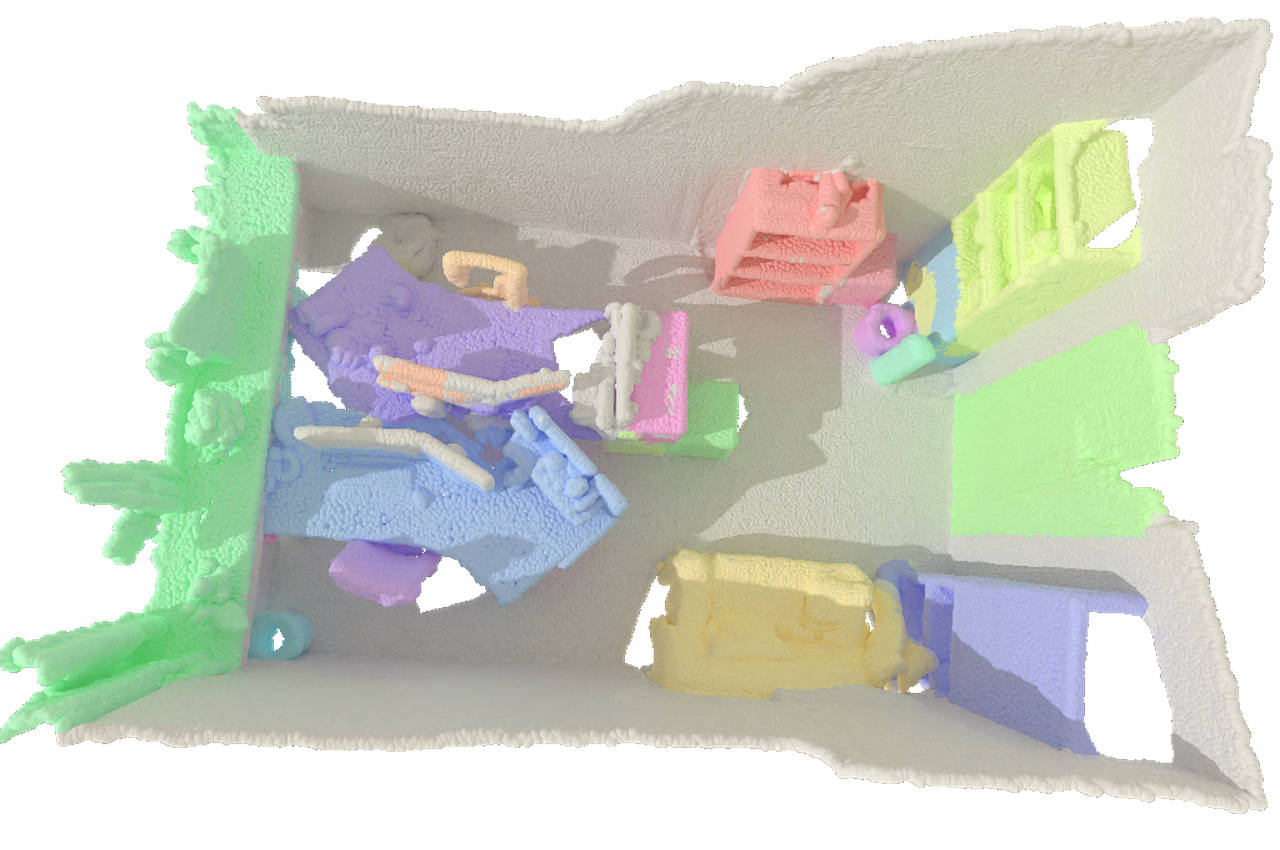}
    \end{subfigure}
         & 
    \begin{subfigure}[b]{0.28\textwidth}
      \includegraphics[width=\linewidth]{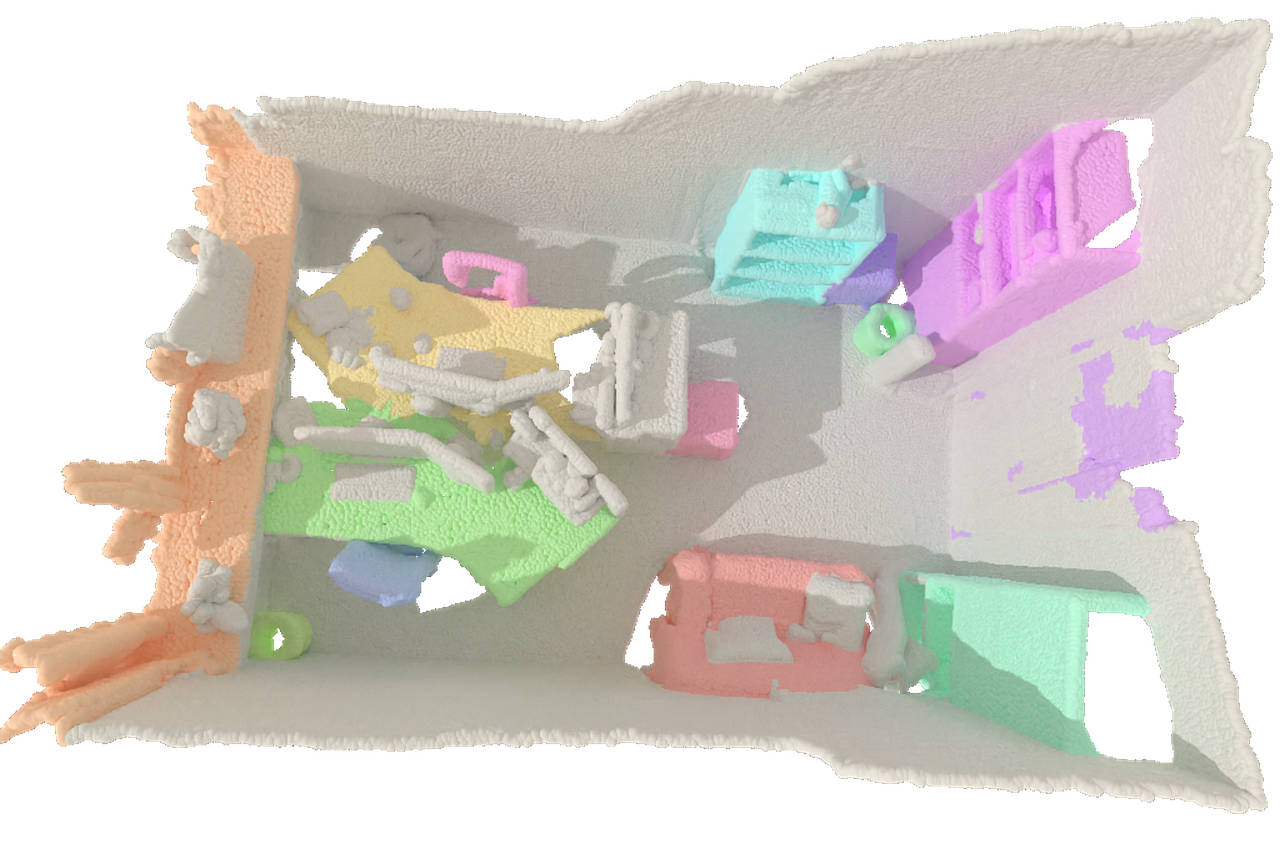}
    \end{subfigure}  \\

    \rotatebox{90}{ \quad \texttt{scene0621\_00}}
    &
    \begin{subfigure}[b]{0.28\textwidth}
      \includegraphics[width=\linewidth]{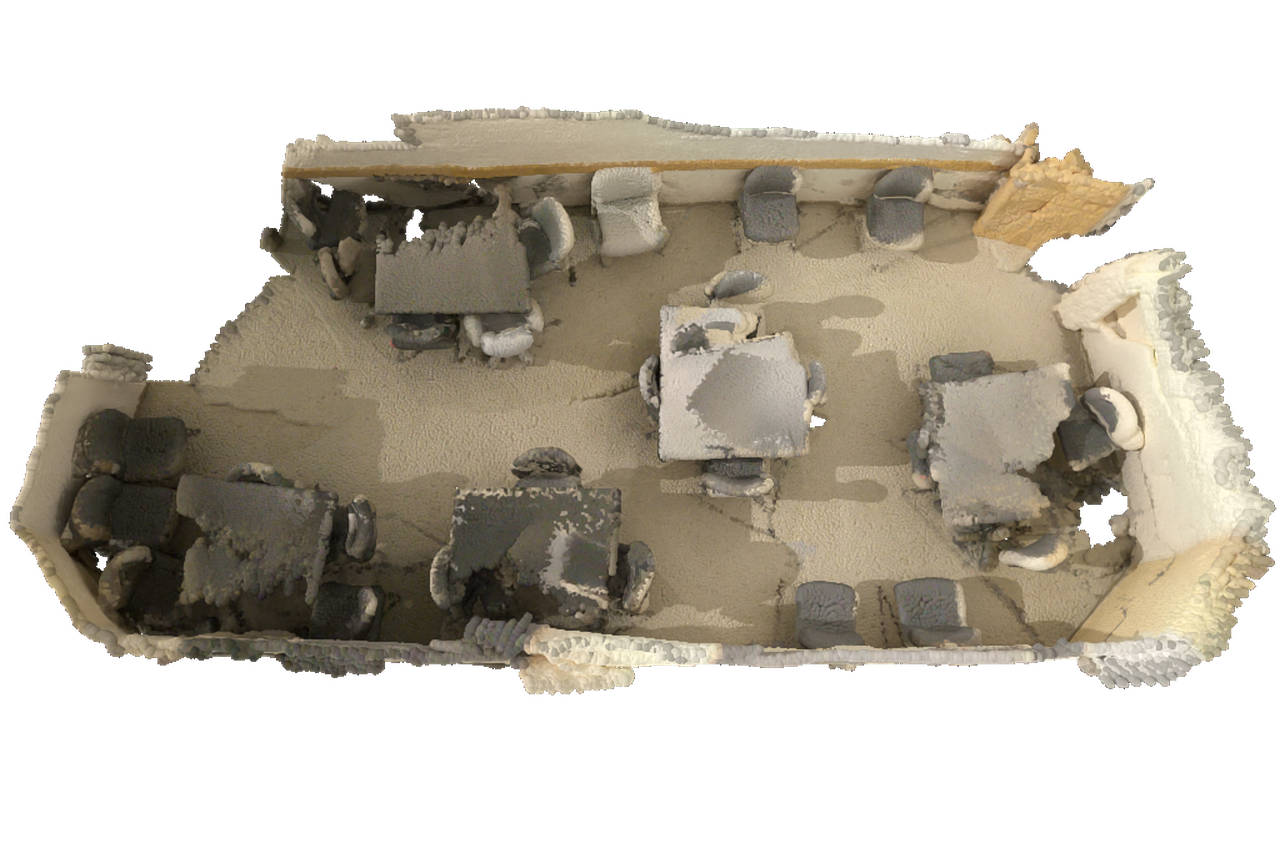}
    \end{subfigure}
         & 
    \begin{subfigure}[b]{0.28\textwidth}
      \includegraphics[width=\linewidth]{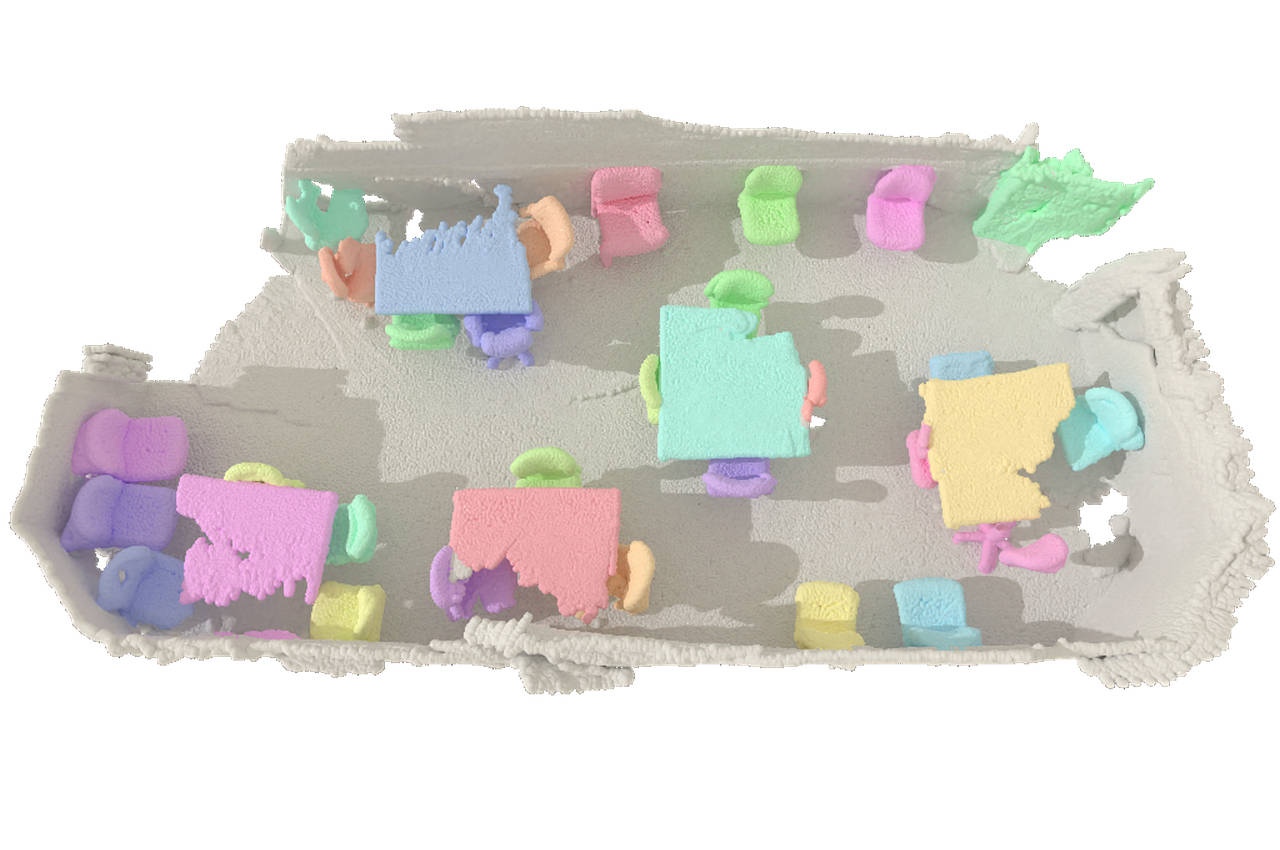}
    \end{subfigure}
         & 
    \begin{subfigure}[b]{0.28\textwidth}
      \includegraphics[width=\linewidth]{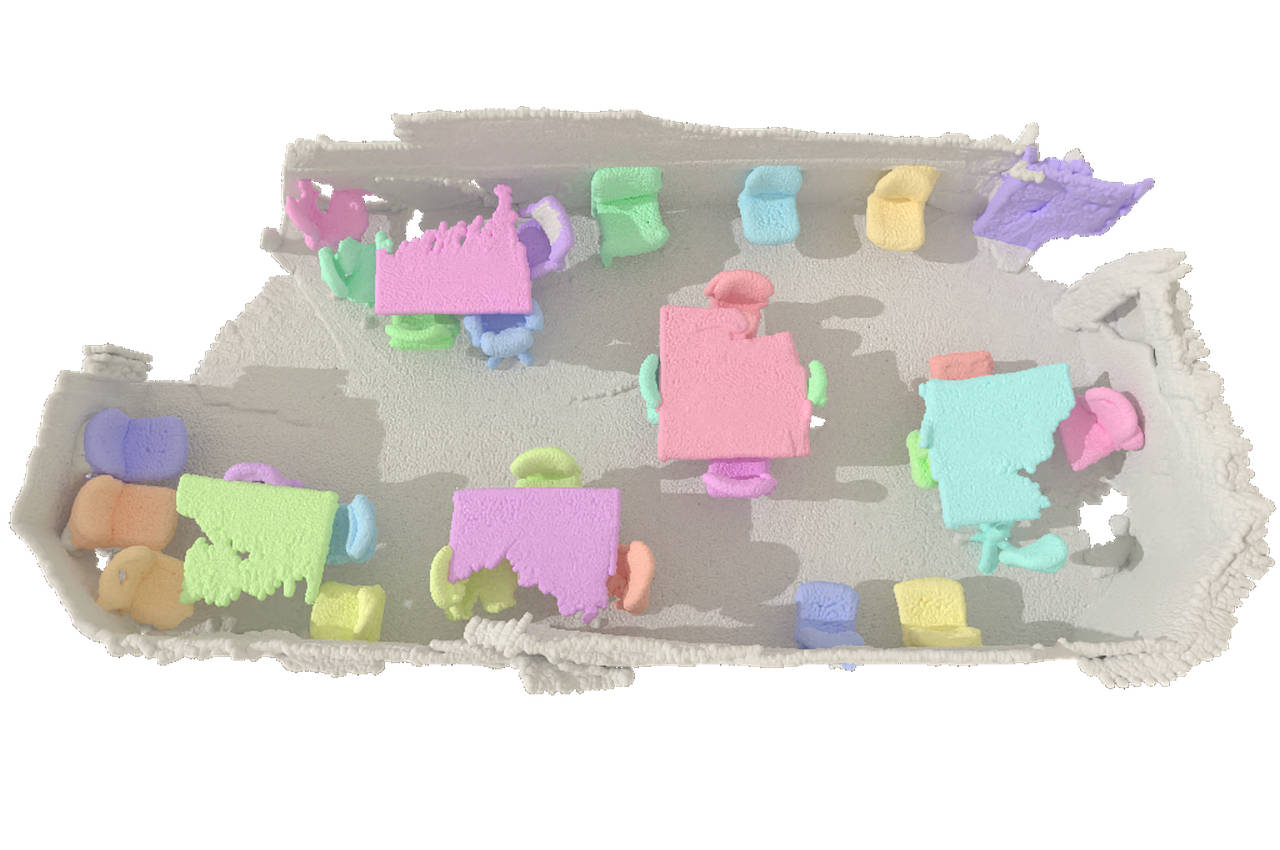}
    \end{subfigure}  \\

    \rotatebox{90}{ \quad \texttt{scene0645\_01}}
    &
    \begin{subfigure}[b]{0.28\textwidth}
      \includegraphics[width=\linewidth]{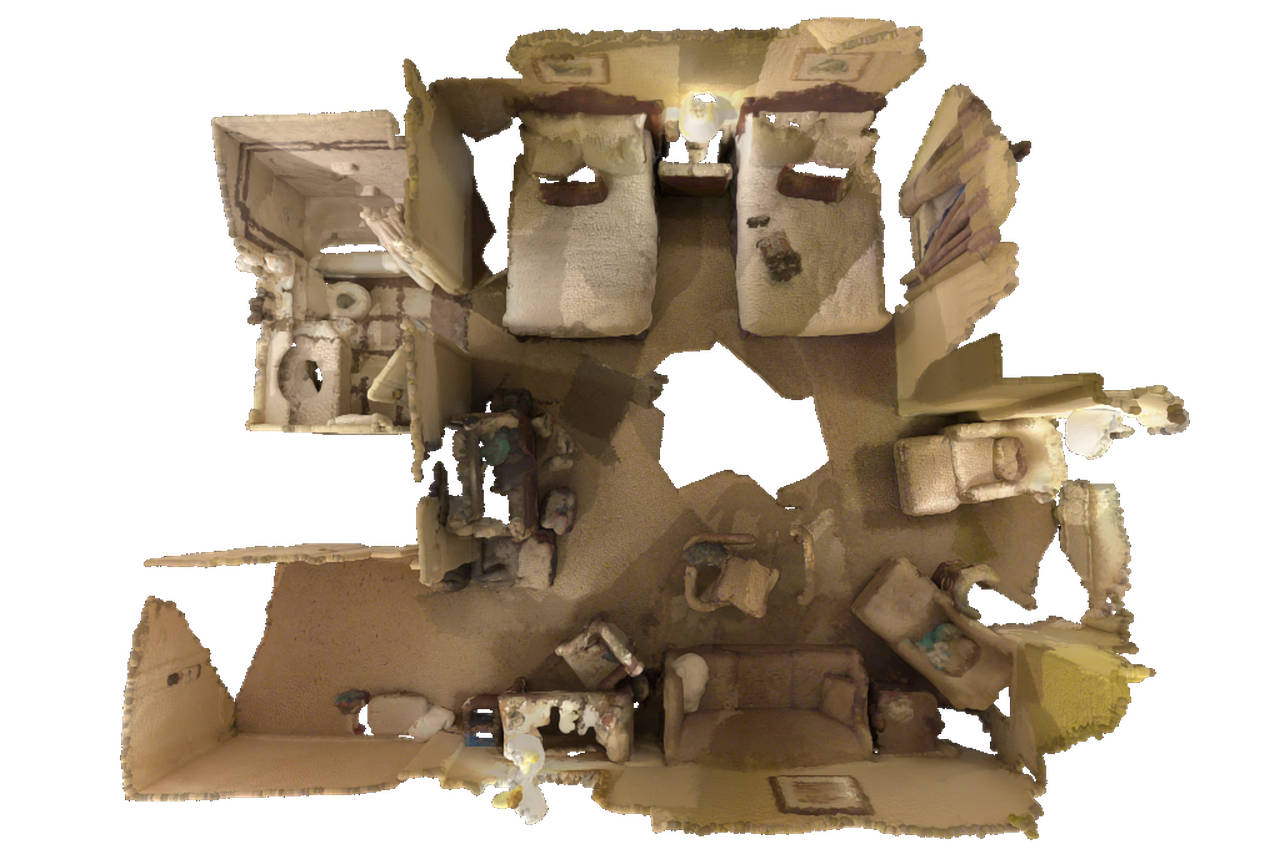}
    \end{subfigure}
         & 
    \begin{subfigure}[b]{0.28\textwidth}
      \includegraphics[width=\linewidth]{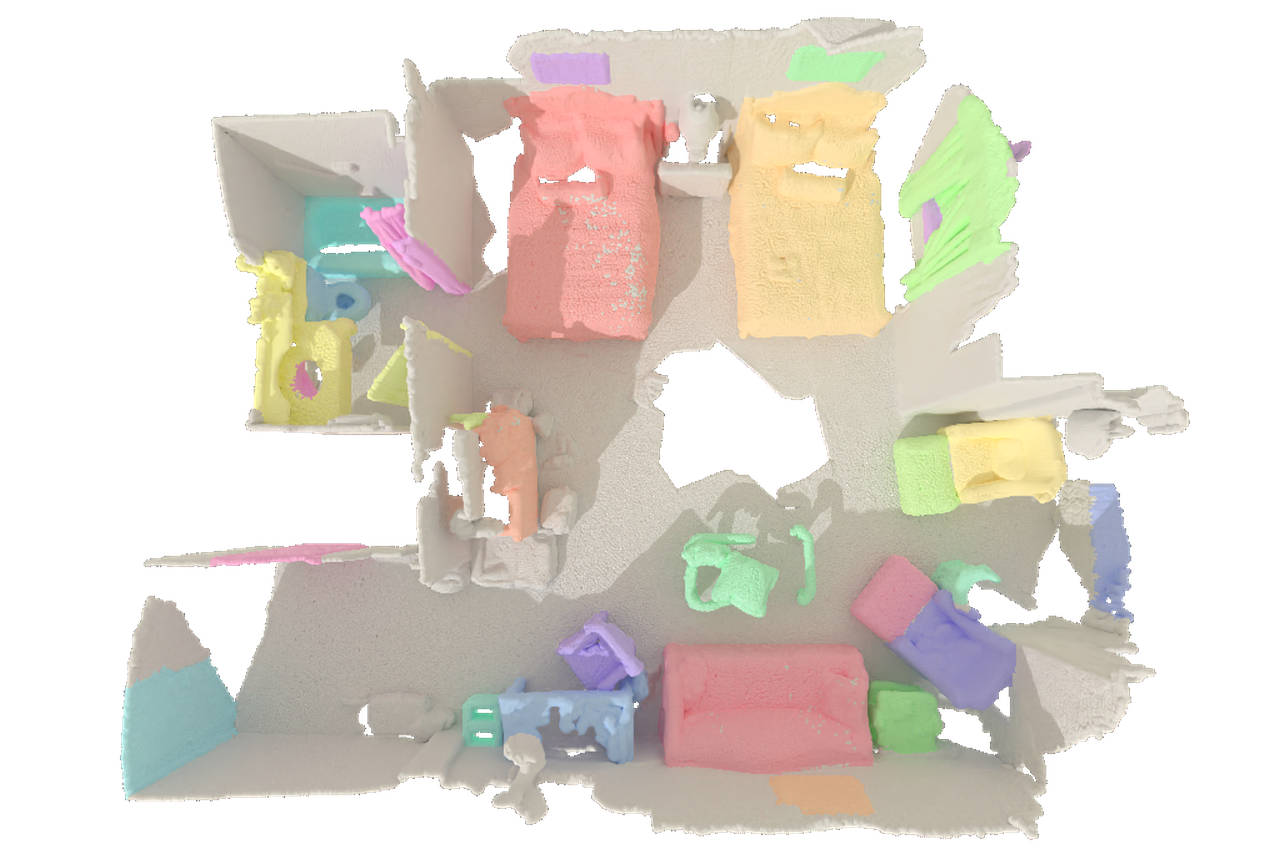}
    \end{subfigure}
         & 
    \begin{subfigure}[b]{0.28\textwidth}
      \includegraphics[width=\linewidth]{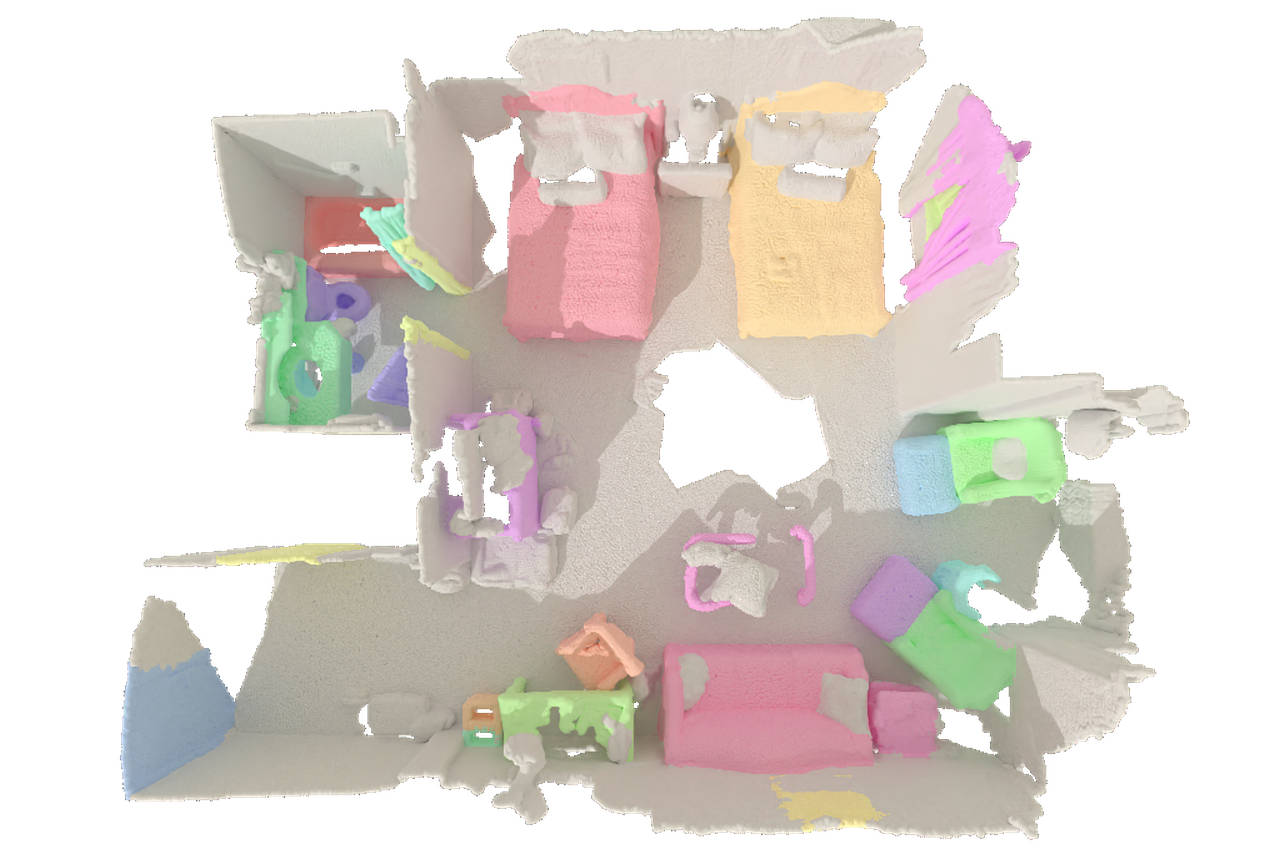}
    \end{subfigure}  \\

    \rotatebox{90}{ \quad \texttt{scene0651\_02}}
    &
    \begin{subfigure}[b]{0.28\textwidth}
      \includegraphics[width=\linewidth]{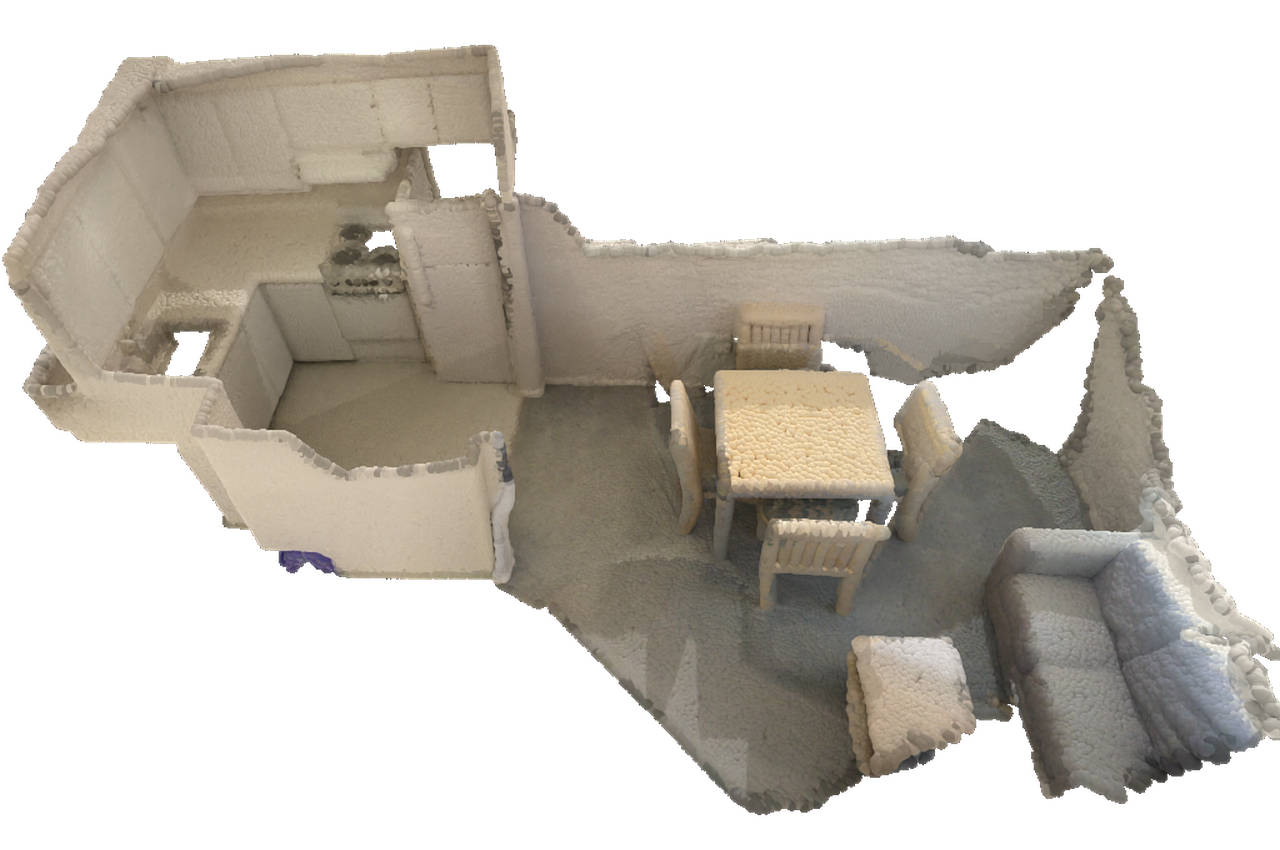}
      \caption{Input}
      \label{fig:quali_scannet_insseg:input}
    \end{subfigure}
         & 
    \begin{subfigure}[b]{0.28\textwidth}
      \includegraphics[width=\linewidth]{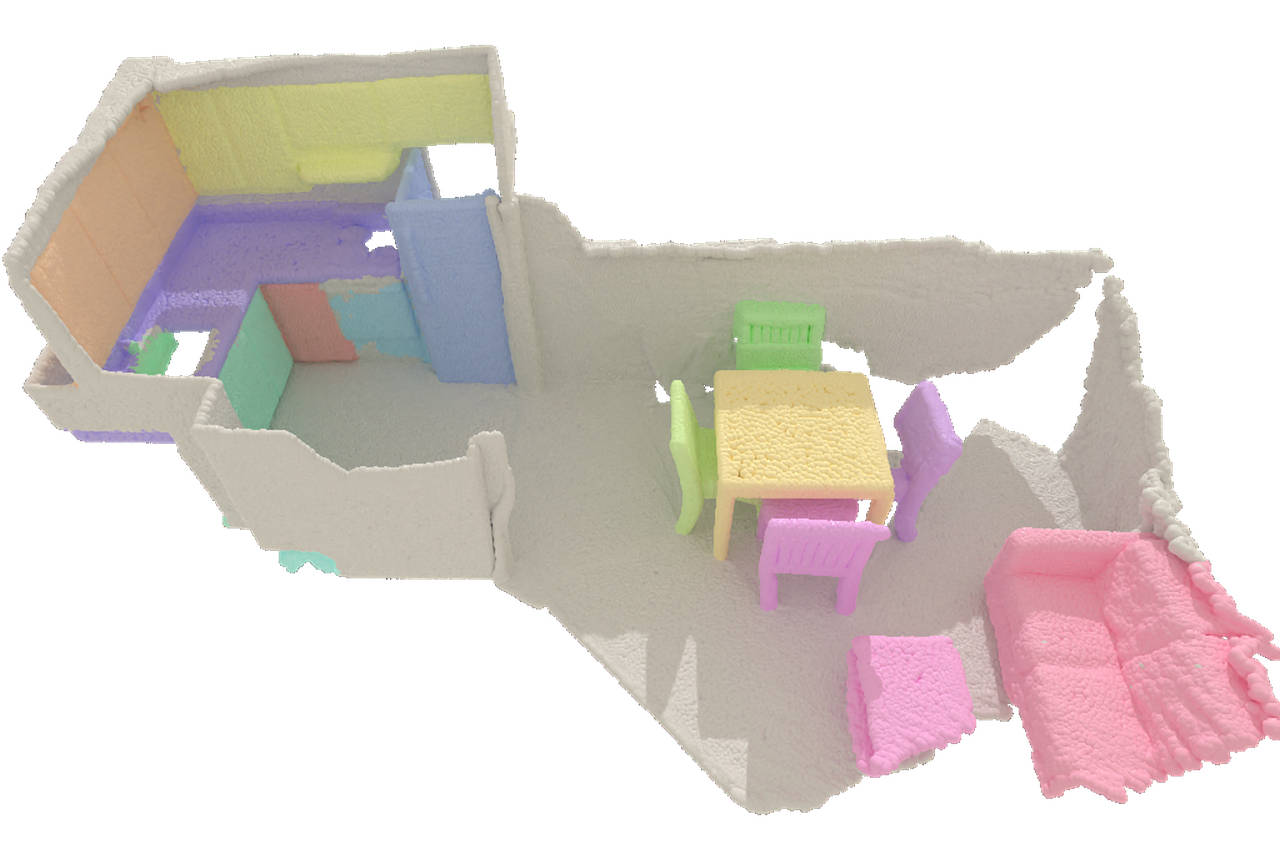}
      \caption{Prediction}
      \label{fig:quali_scannet_insseg:pred}
    \end{subfigure}
         & 
    \begin{subfigure}[b]{0.28\textwidth}
      \includegraphics[width=\linewidth]{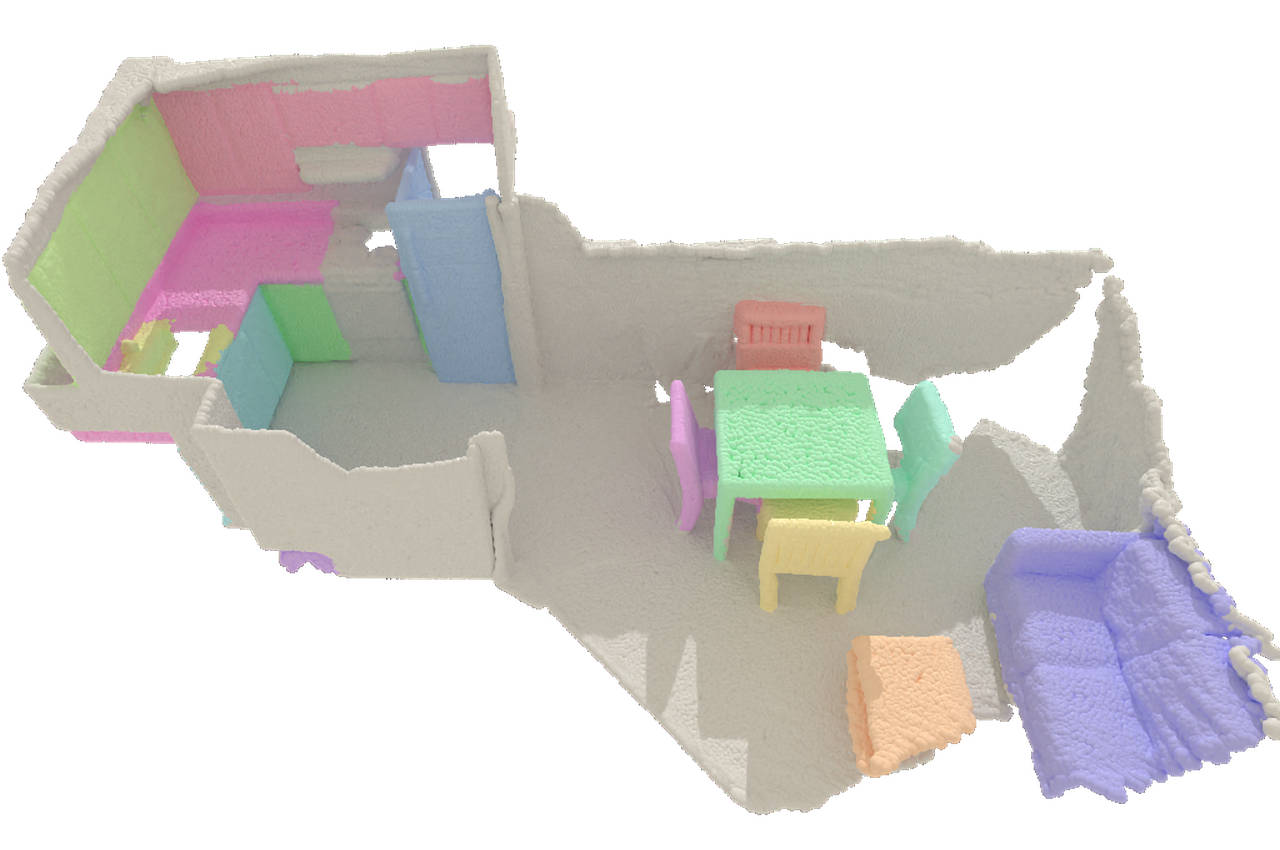}
      \caption{Ground Truth}
      \label{fig:quali_scannet_insseg:gt}
    \end{subfigure}  \\

    \end{tabular}
    \caption{
    {\bf ScanNet instance segmentation.} 
    We present various scenes of the ScanNet validation set: the input point cloud, the instance segmentation from \name{}-S*, and the corresponding ground truth. Colours for each instance are randomly assigned.
    }
    \label{fig:quali_scannet_insseg}
\end{figure*}

\begin{figure*}
    \centering
    \begin{tabular}{@{}lccc@{}}

    \multicolumn{4}{c}{

    \begin{minipage}{\textwidth}
    \centering
        {\textcolor{waymoDetVehicle}{\ding{108}}}\,\,vehicle\,\,\,\,\,\,\,\,\,\,\,\,\,\,\,\, 
        {\textcolor{waymoDetPedestrian}{\ding{108}}}\,\,pedestrian\,\,\,\,\,\,\,\,\,\,\,\,\,\,\,\,  
        {\textcolor{waymoDetCyclist}{\ding{108}}}\,\,cyclist

    \end{minipage}
    } 
    \vspace{15pt}
    
    \\ 
    
    \rotatebox{90}{\scriptsize \texttt{3077939657605416...}}
    &
    \begin{subfigure}[b]{0.28\textwidth}
      \includegraphics[width=\linewidth]{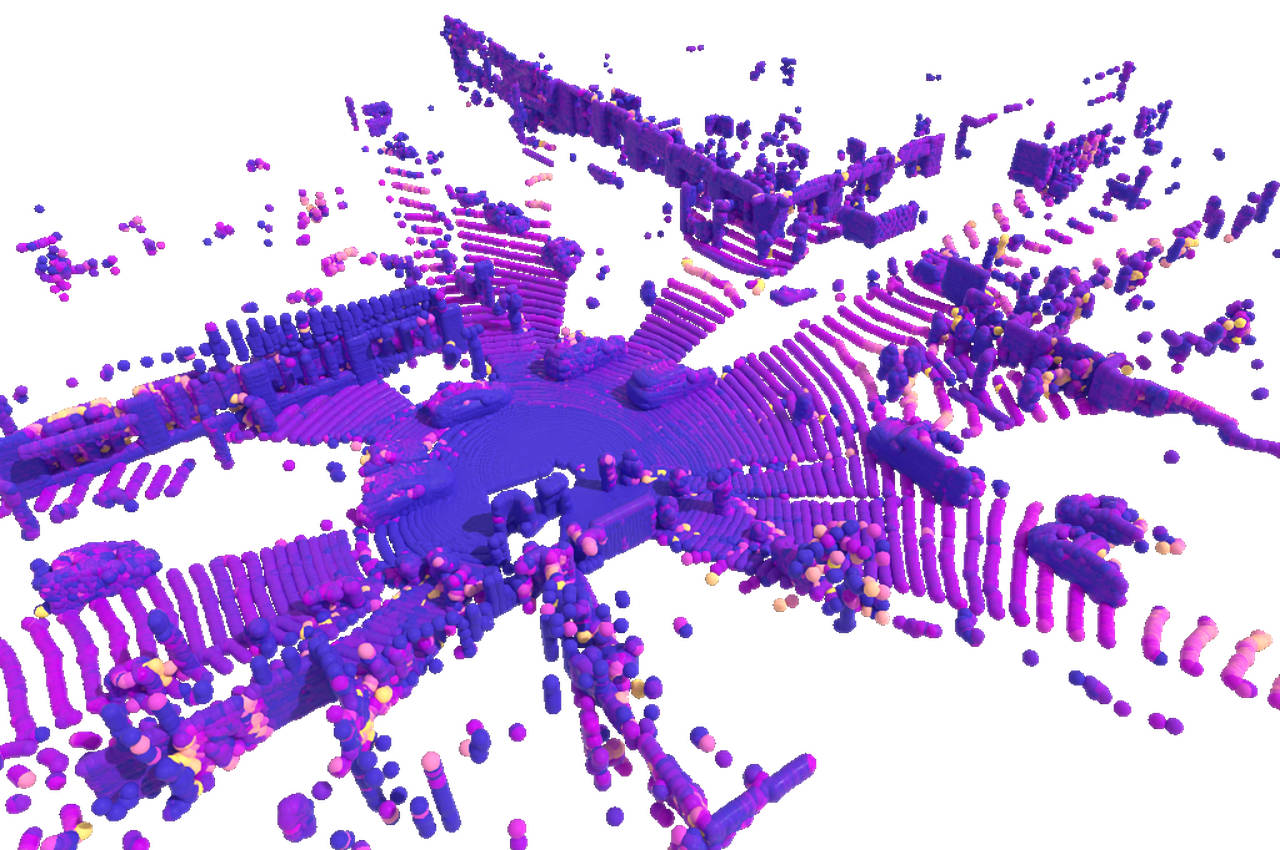}
    \end{subfigure}
         & 
    \begin{subfigure}[b]{0.28\textwidth}
      \includegraphics[width=\linewidth]{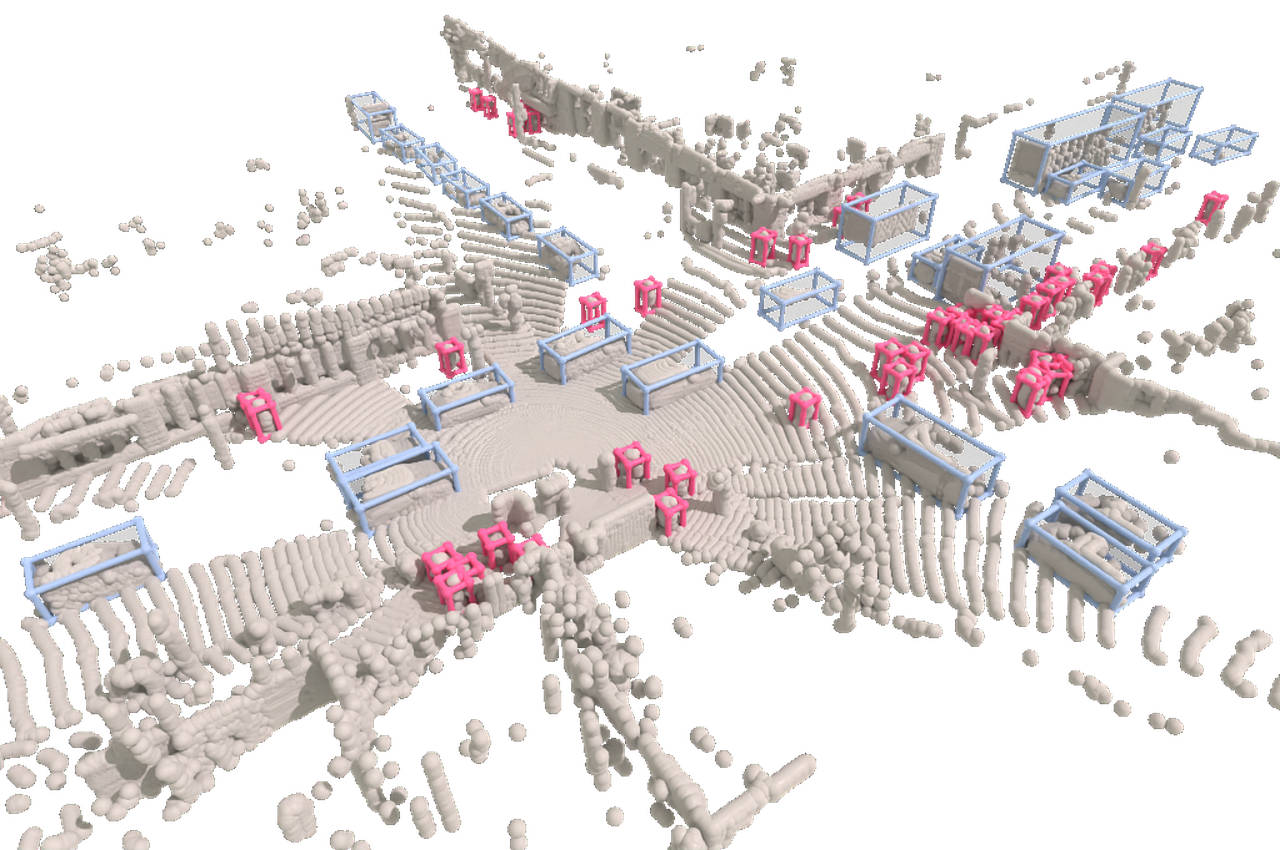}
    \end{subfigure}
         & 
    \begin{subfigure}[b]{0.28\textwidth}
      \includegraphics[width=\linewidth]{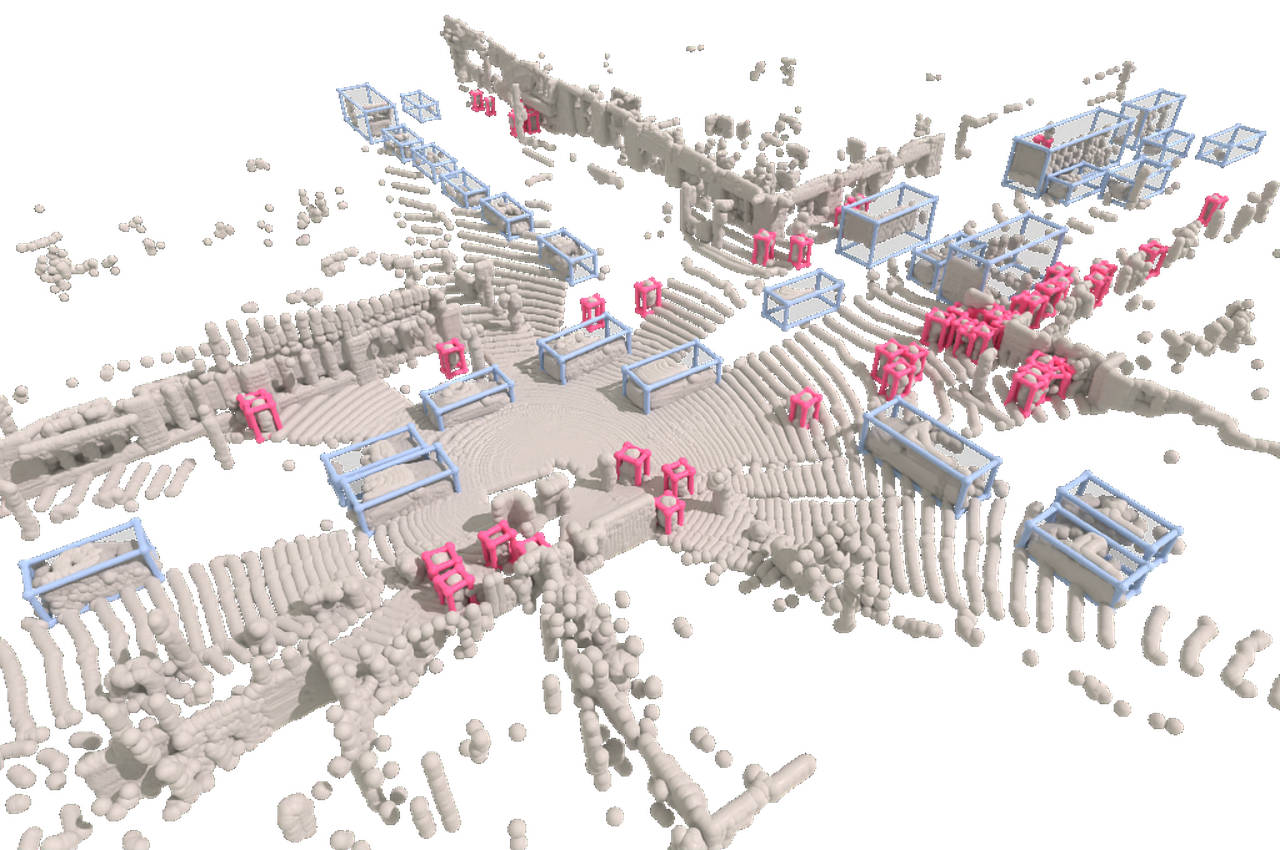}
    \end{subfigure}  \\

    \rotatebox{90}{\scriptsize \texttt{6621886863973648...}}
    &
    \begin{subfigure}[b]{0.28\textwidth}
      \includegraphics[width=\linewidth]{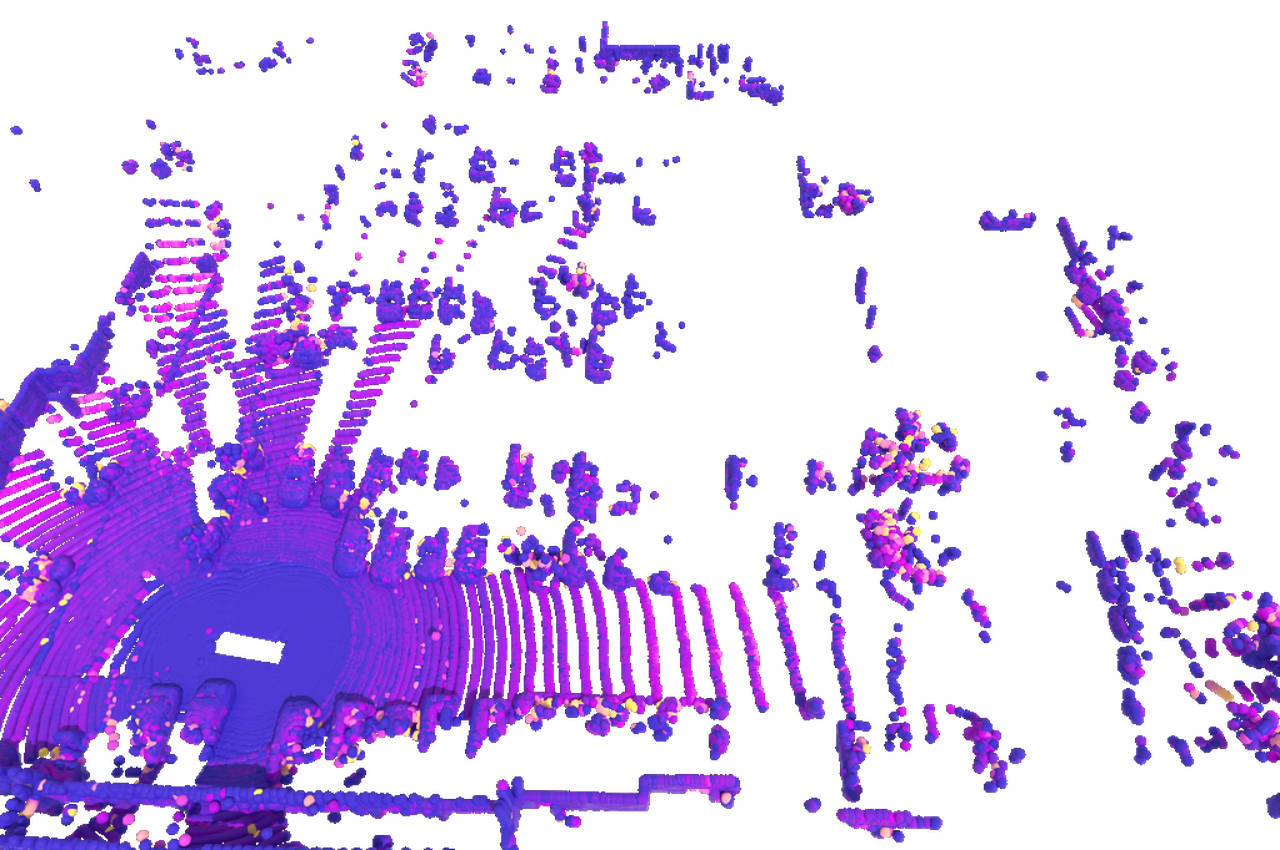}
    \end{subfigure}
         & 
    \begin{subfigure}[b]{0.28\textwidth}
      \includegraphics[width=\linewidth]{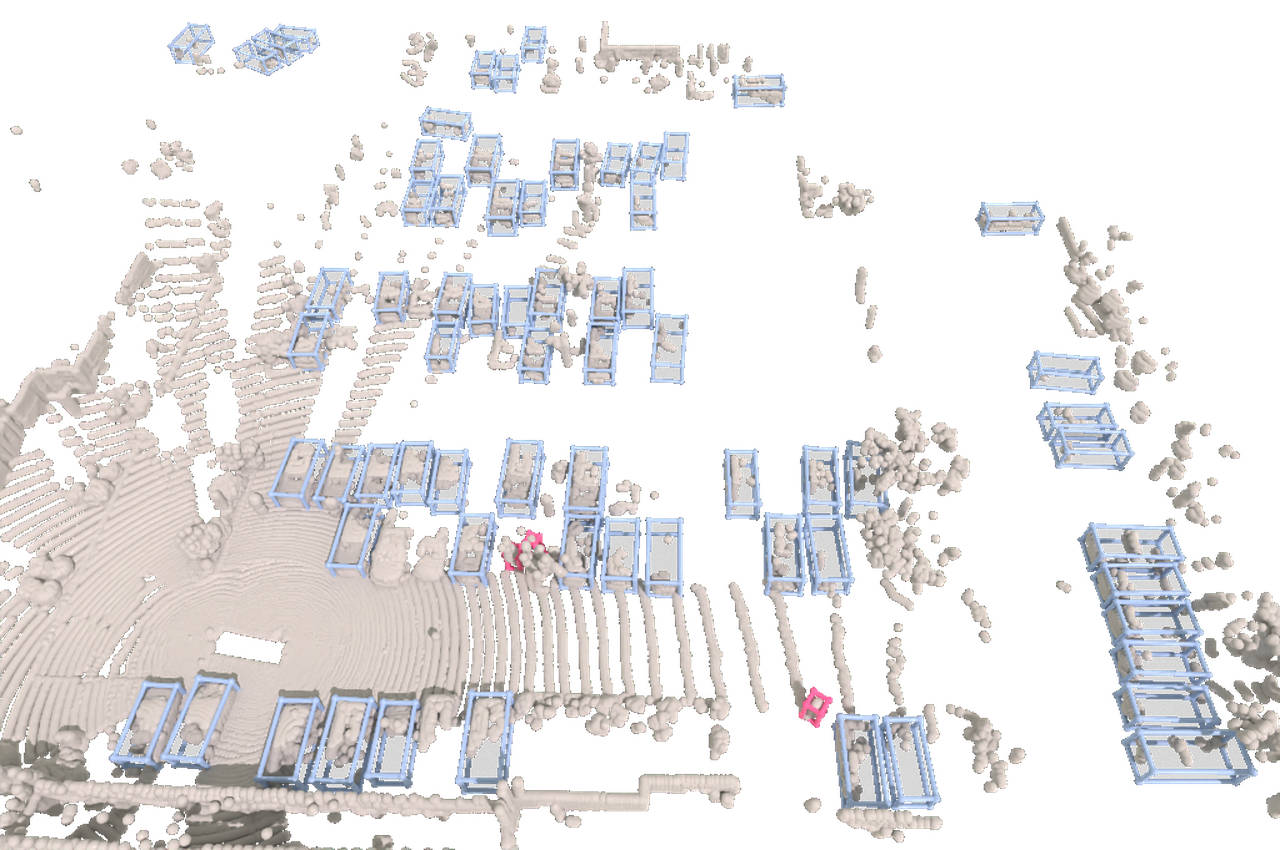}
    \end{subfigure}
         & 
    \begin{subfigure}[b]{0.28\textwidth}
      \includegraphics[width=\linewidth]{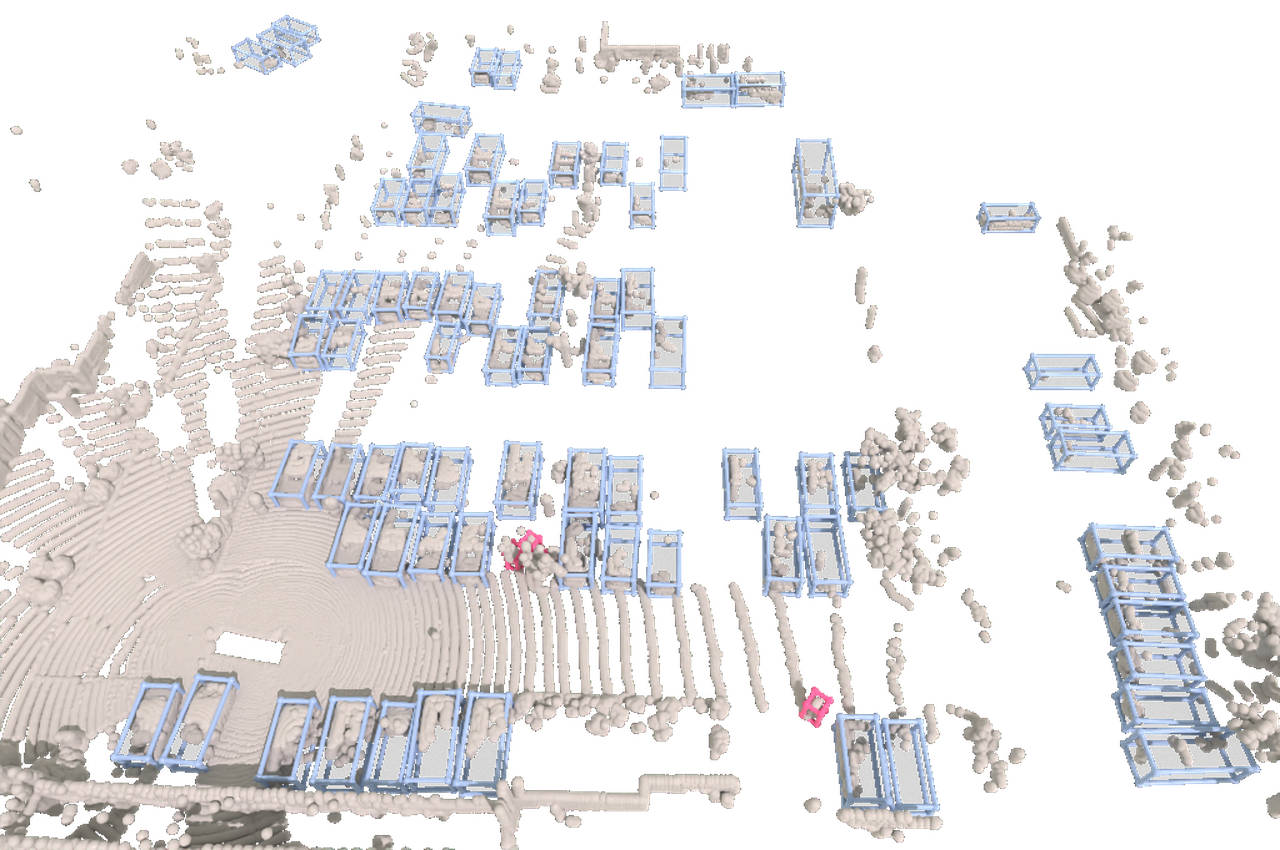}
    \end{subfigure}  \\

    \rotatebox{90}{\scriptsize \texttt{8956556778987472...}}
    &
    \begin{subfigure}[b]{0.28\textwidth}
      \includegraphics[width=\linewidth]{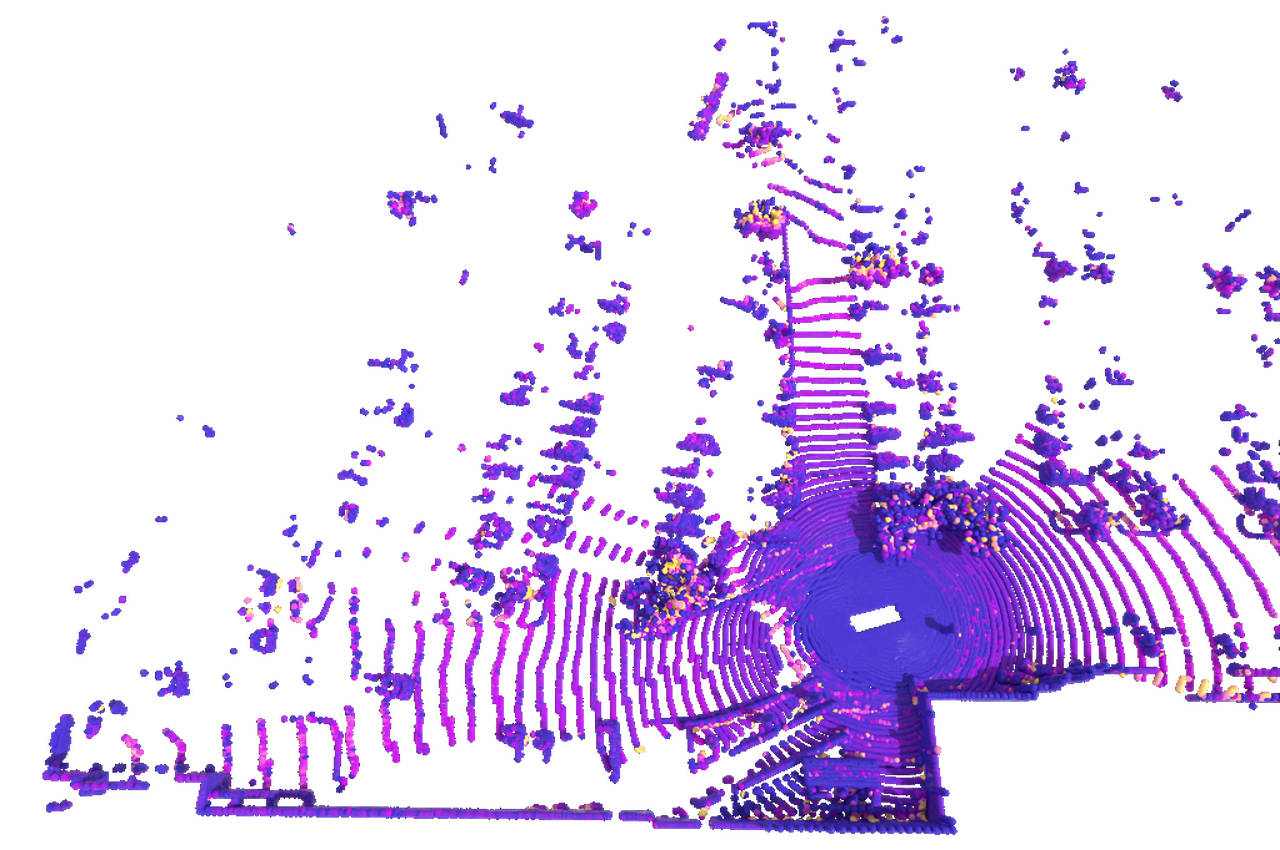}
    \end{subfigure}
         & 
    \begin{subfigure}[b]{0.28\textwidth}
      \includegraphics[width=\linewidth]{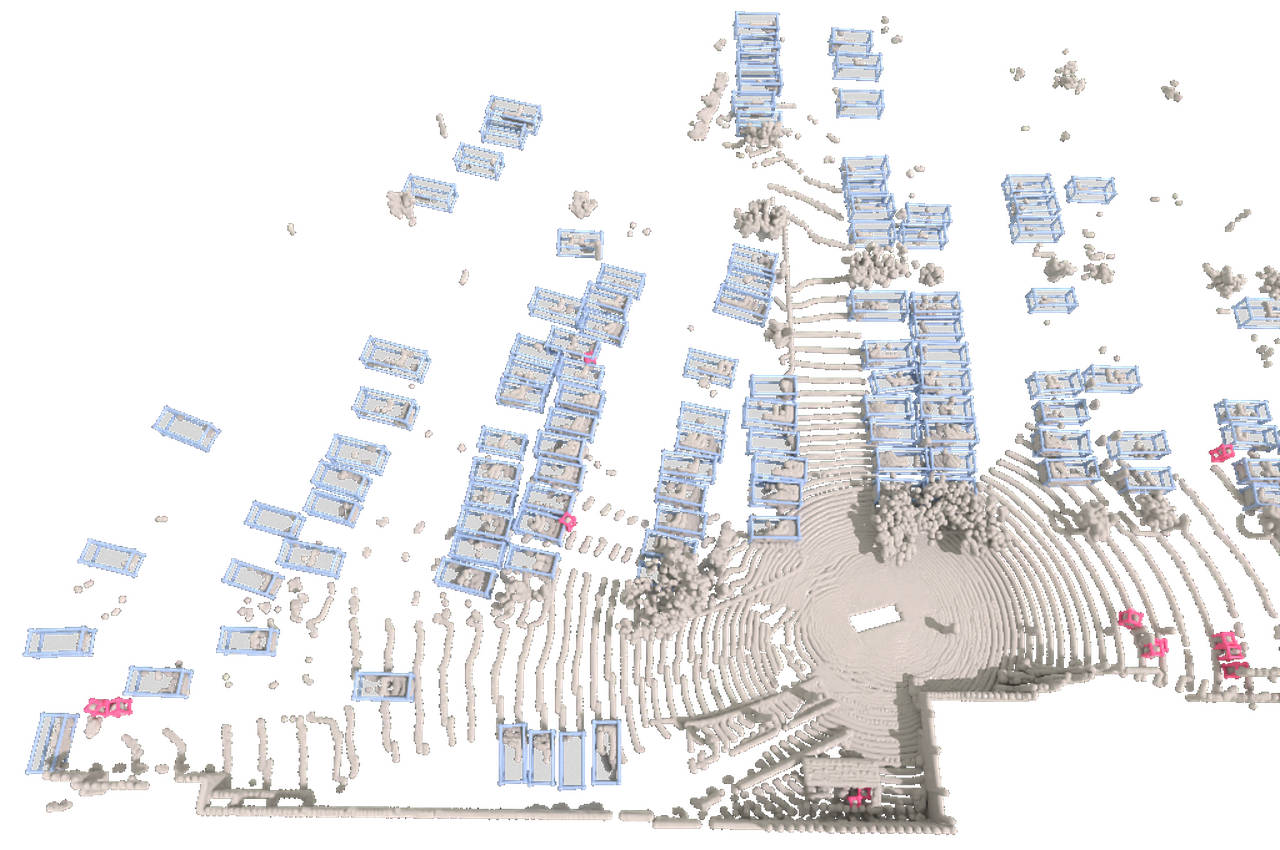}
    \end{subfigure}
         & 
    \begin{subfigure}[b]{0.28\textwidth}
      \includegraphics[width=\linewidth]{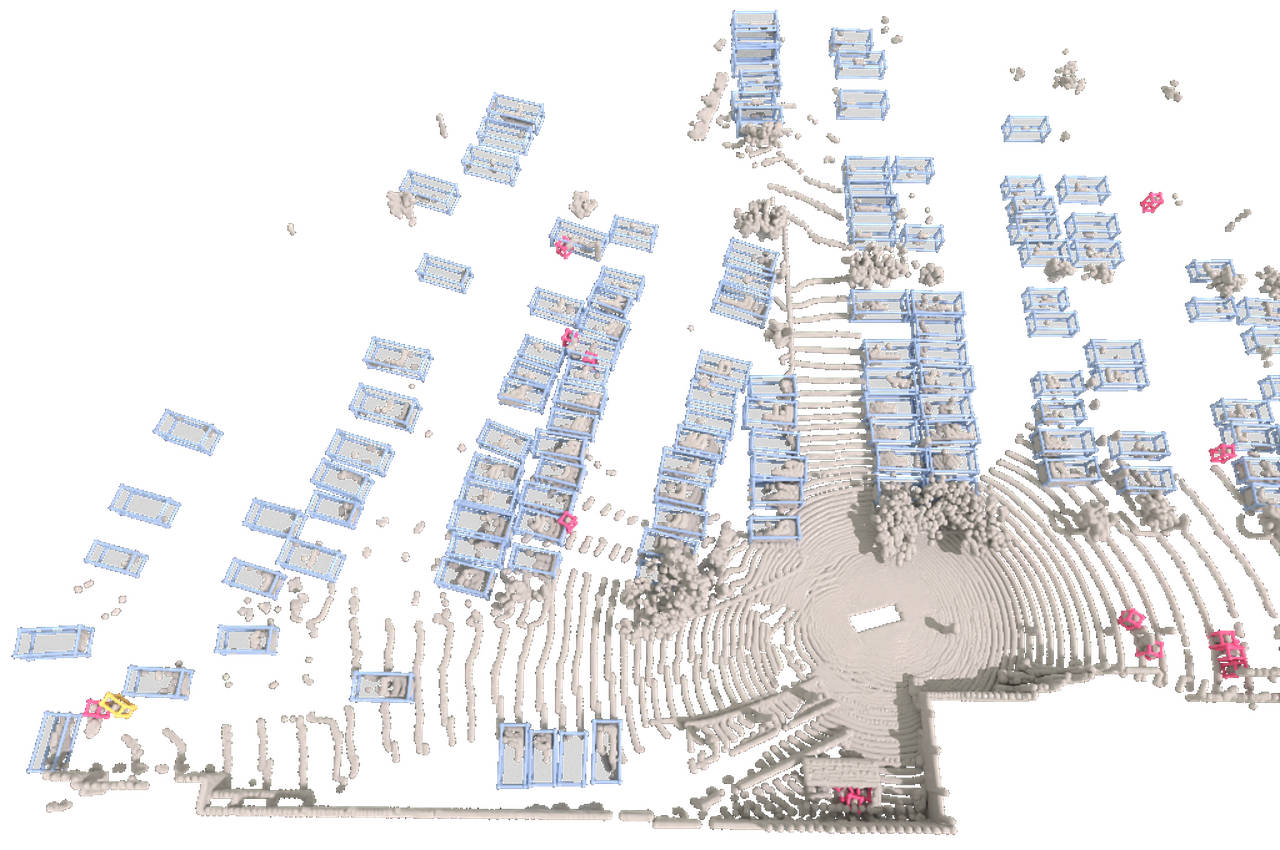}
    \end{subfigure}  \\

    \rotatebox{90}{\scriptsize \texttt{1333688303428388...}}
    &
    \begin{subfigure}[b]{0.28\textwidth}
      \includegraphics[width=\linewidth]{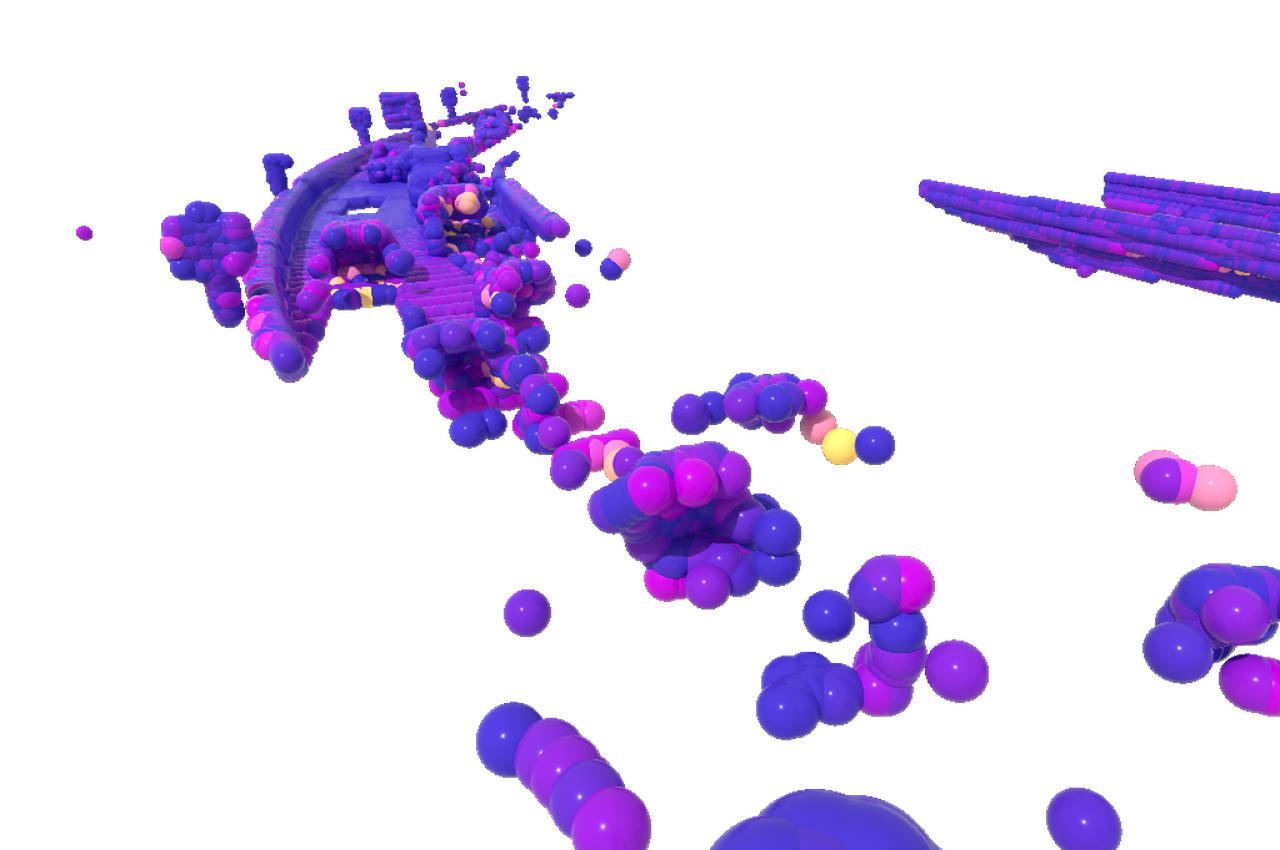}
    \end{subfigure}
         & 
    \begin{subfigure}[b]{0.28\textwidth}
      \includegraphics[width=\linewidth]{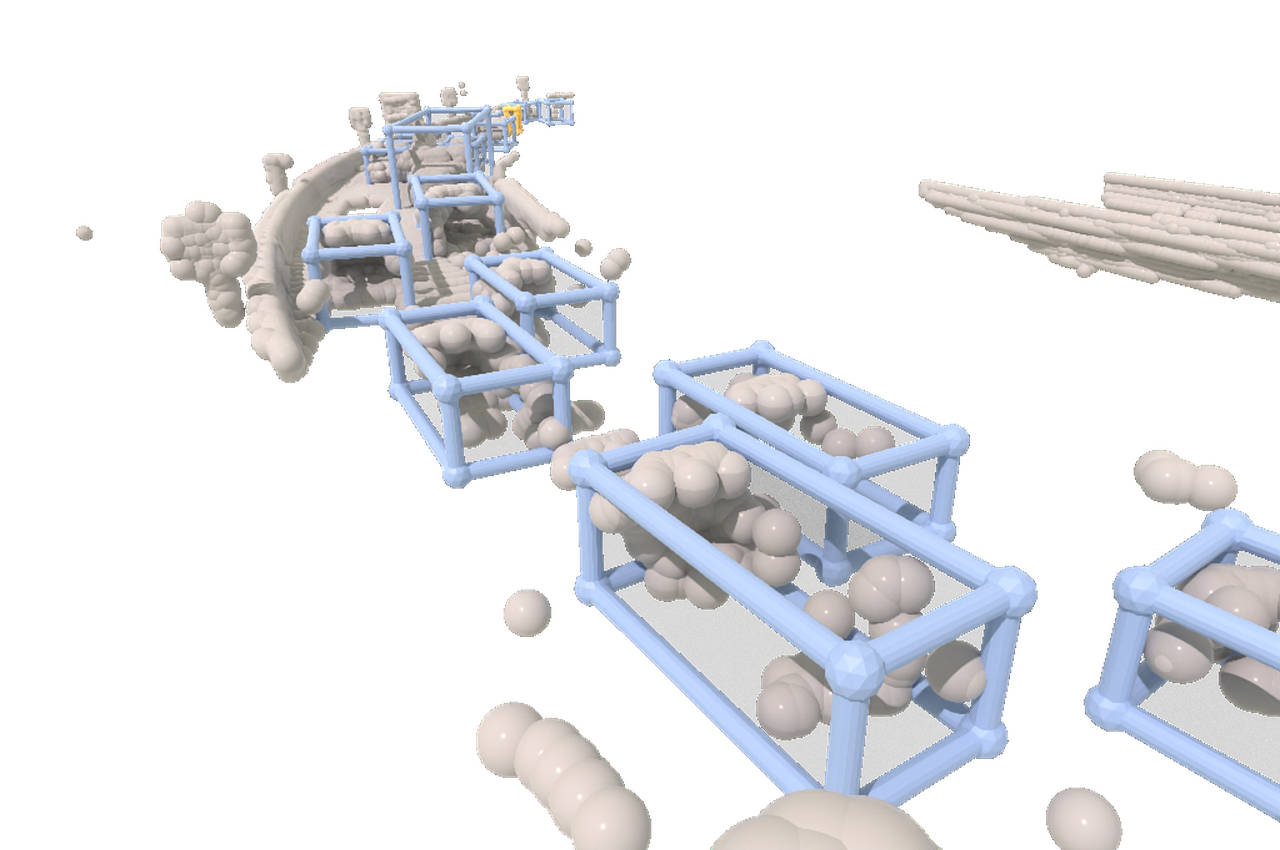}
    \end{subfigure}
         & 
    \begin{subfigure}[b]{0.28\textwidth}
      \includegraphics[width=\linewidth]{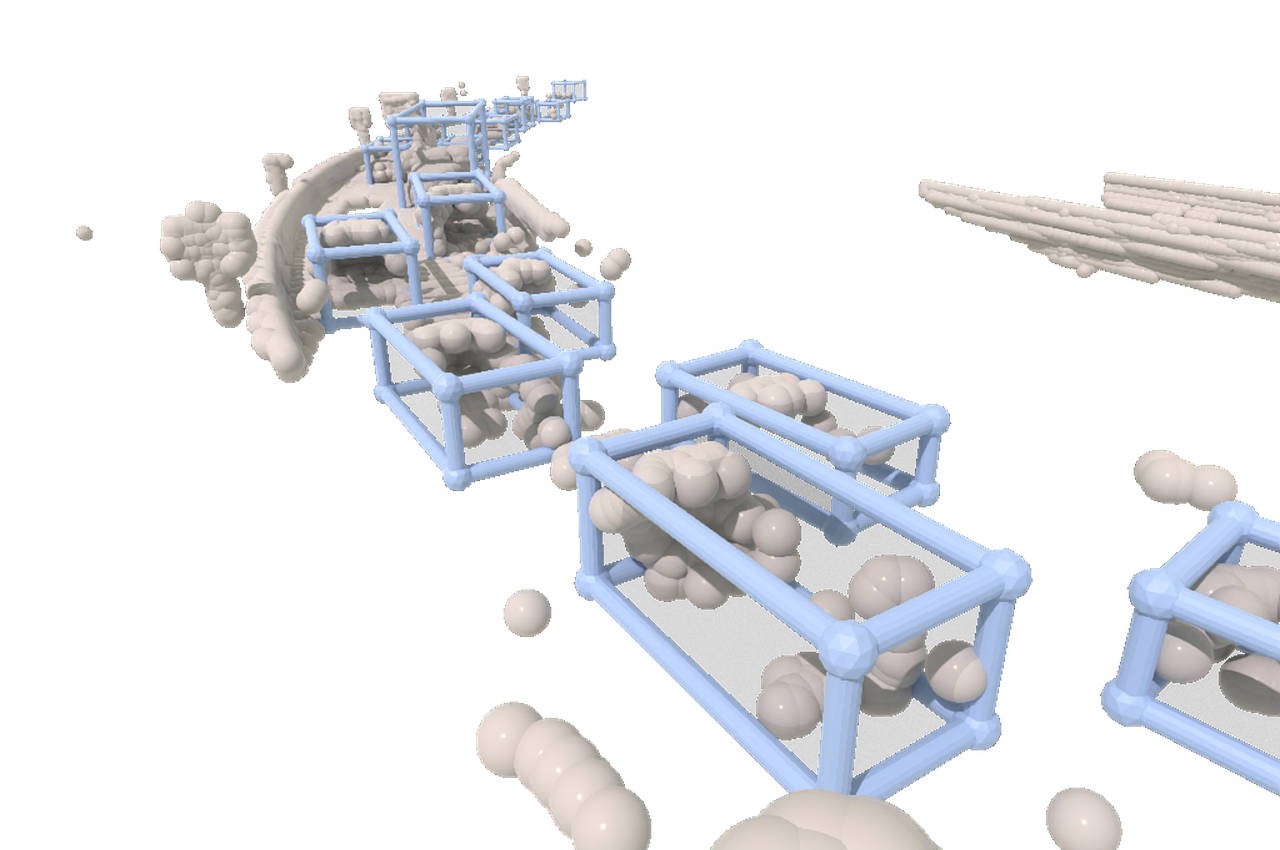}
    \end{subfigure}  \\

    \rotatebox{90}{\scriptsize \texttt{1335699760417784...}}
    &
    \begin{subfigure}[b]{0.28\textwidth}
      \includegraphics[width=\linewidth]{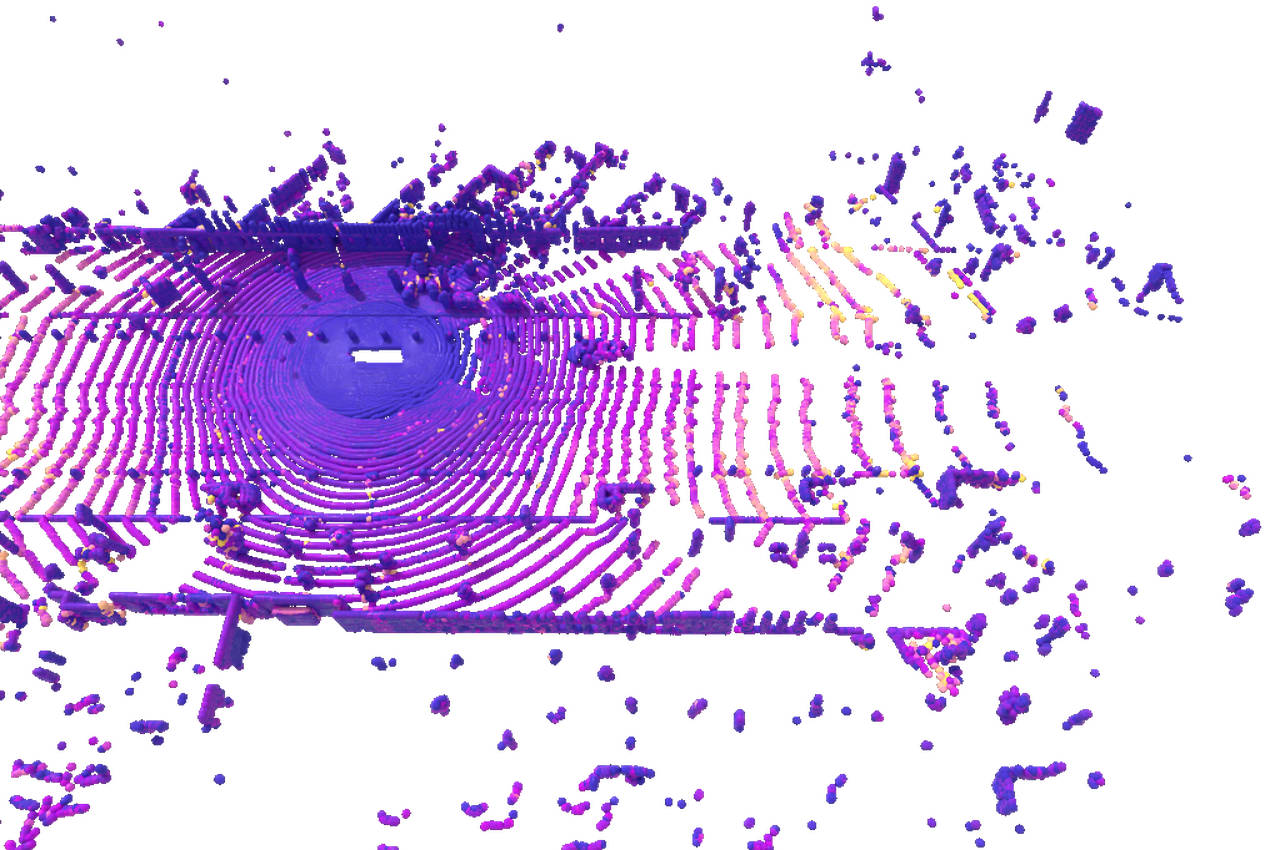}
    \end{subfigure}
         & 
    \begin{subfigure}[b]{0.28\textwidth}
      \includegraphics[width=\linewidth]{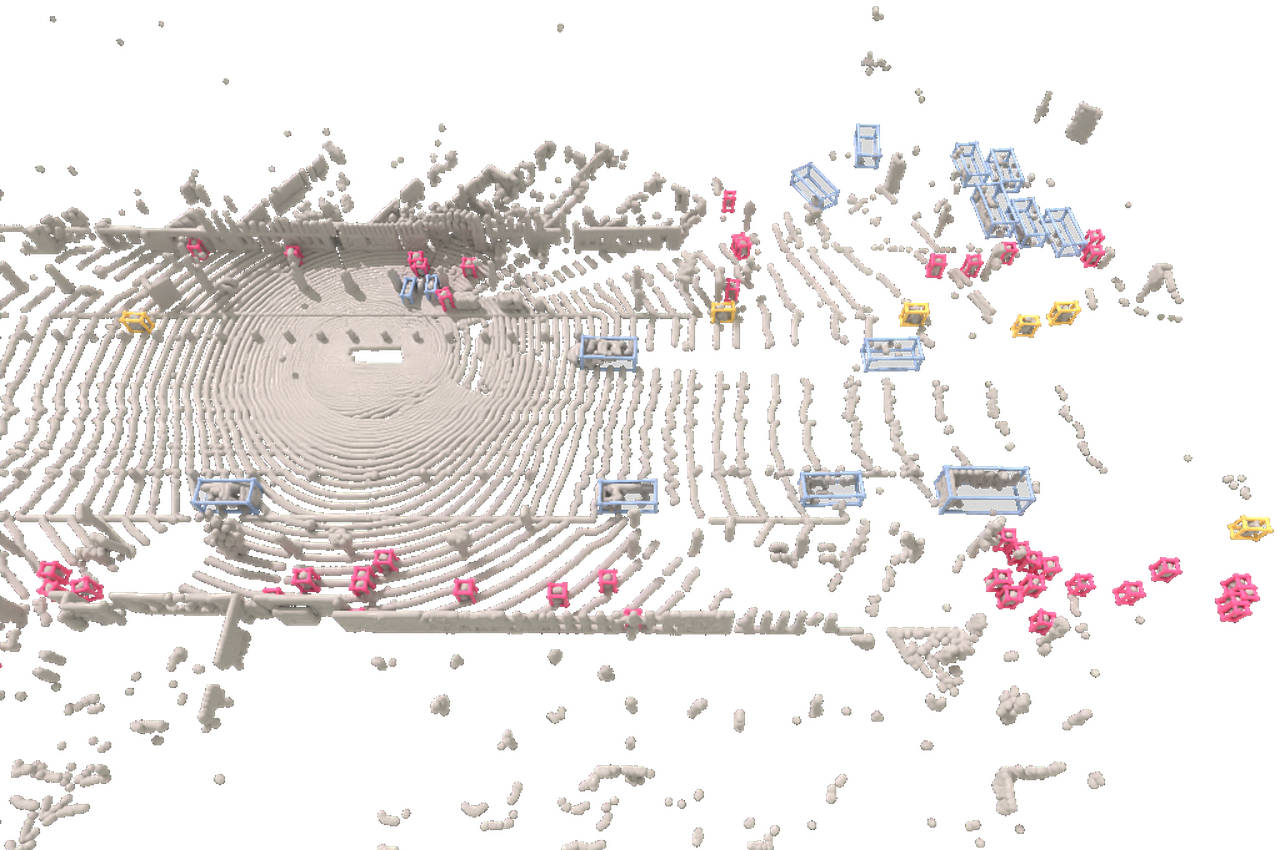}
    \end{subfigure}
         & 
    \begin{subfigure}[b]{0.28\textwidth}
      \includegraphics[width=\linewidth]{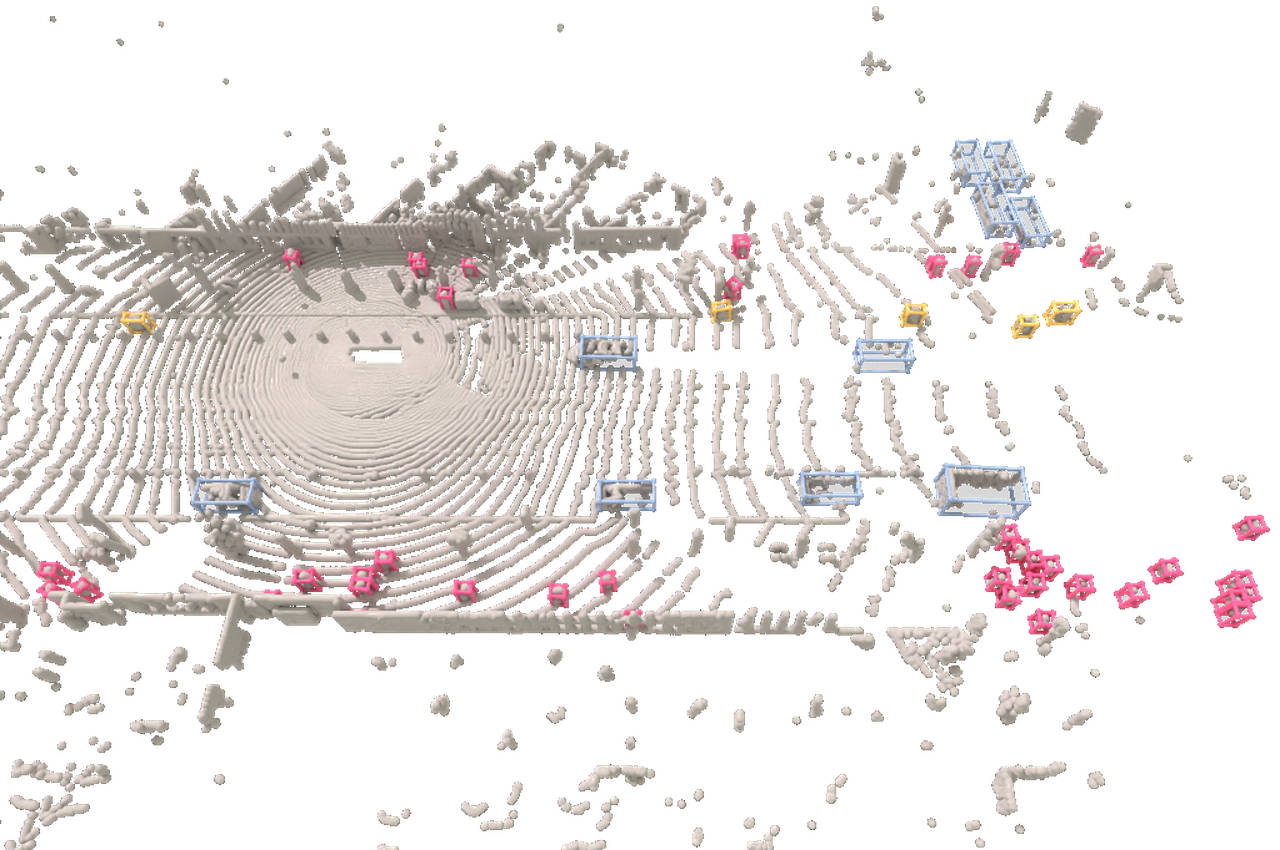}
    \end{subfigure}  \\

    \rotatebox{90}{\scriptsize \texttt{1430000760420586...}}
    &
    \begin{subfigure}[b]{0.28\textwidth}
      \includegraphics[width=\linewidth]{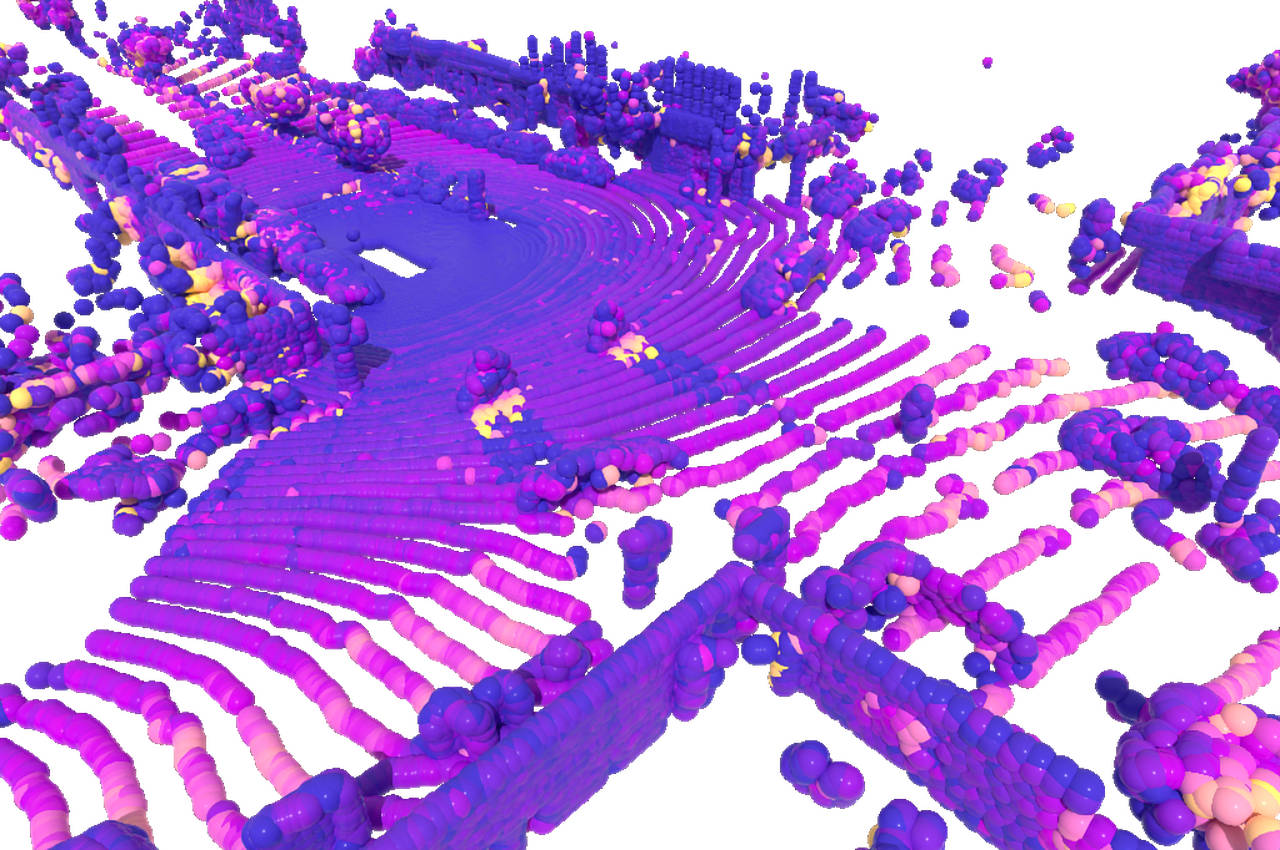}
      \caption{Input}
      \label{fig:quali_waymo_det:input}
    \end{subfigure}
         & 
    \begin{subfigure}[b]{0.28\textwidth}
      \includegraphics[width=\linewidth]{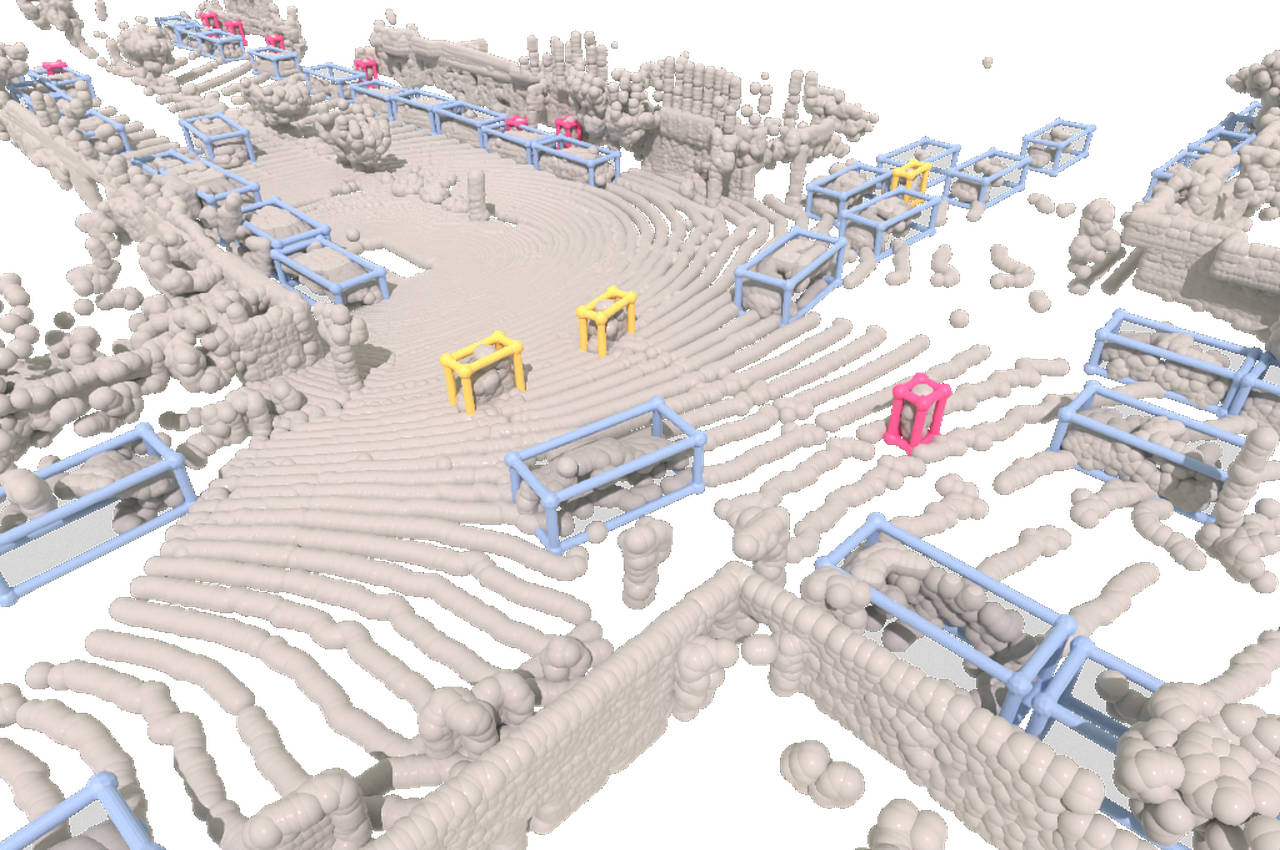}
      \caption{Prediction}
      \label{fig:quali_waymo_det:pred}
    \end{subfigure}
         & 
    \begin{subfigure}[b]{0.28\textwidth}
      \includegraphics[width=\linewidth]{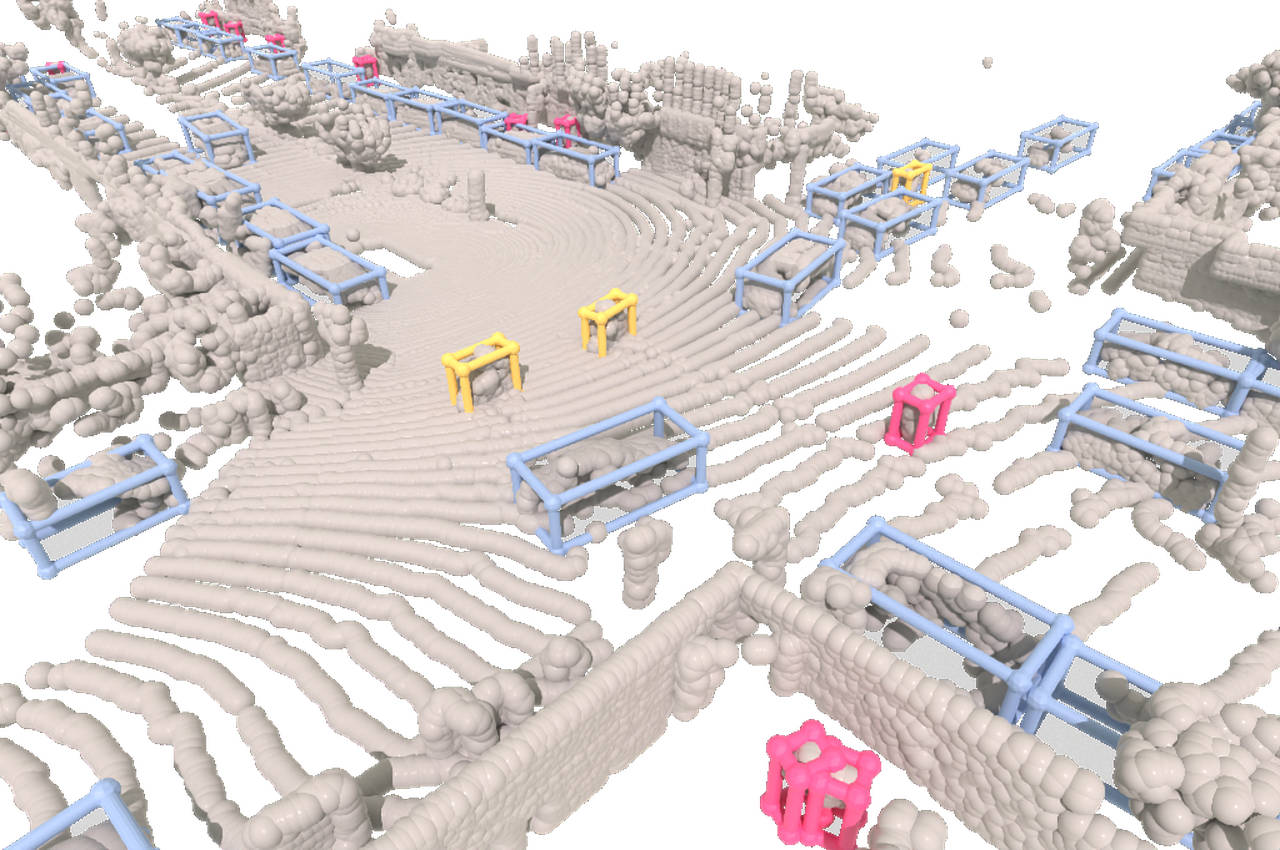}
      \caption{Ground Truth}
      \label{fig:quali_waymo_det:gt}
    \end{subfigure}  \\

    \end{tabular}
    \caption{
    {\bf Waymo object detection.} 
    We present various scenes of the Waymo validation set: the input point cloud, the object detections from \name{}, and the corresponding ground truth.
    }
    \label{fig:quali_waymo_det}
\end{figure*}

We visualise sample predictions of \name{} on three tasks: 3D semantic segmentation (\cref{fig:quali_scannet_semseg,fig:quali_structure3d_semseg,fig:quali_nuscenes_semseg,fig:quali_waymo_semseg}), 3D instance segmentation (\cref{fig:quali_scannet_insseg}), and 3D object detection (\cref{fig:quali_waymo_det}).

{
    \small
    \bibliographystyle{ieeenat_fullname}
    \bibliography{main}

\begin{thebibliography}{105}
\providecommand{\natexlab}[1]{#1}
\providecommand{\url}[1]{\texttt{#1}}
\expandafter\ifx\csname urlstyle\endcsname\relax
  \providecommand{\doi}[1]{doi: #1}\else
  \providecommand{\doi}{doi: \begingroup \urlstyle{rm}\Url}\fi

\bibitem[Atzmon et~al.(2018)Atzmon, Maron, and Lipman]{atzmon2018point}
Matan Atzmon, Haggai Maron, and Yaron Lipman.
\newblock {Point Convolutional Neural Networks by Extension Operators}.
\newblock \emph{ACM Transactions on Graphics (TOG)}, 2018.

\bibitem[Ba et~al.(2016)Ba, Kiros, and Hinton]{ba2016layer}
Jimmy~Lei Ba, Jamie~Ryan Kiros, and Geoffrey~E Hinton.
\newblock {Layer Normalization}.
\newblock \emph{arXiv preprint arXiv:1607.06450}, 2016.

\bibitem[Berman et~al.(2018)Berman, Triki, and Blaschko]{berman2018lovasz}
Maxim Berman, Amal~Rannen Triki, and Matthew~B Blaschko.
\newblock {The Lovász-Softmax Loss: A Tractable Surrogate for the Optimization of the Intersection-Over-Union Measure in Neural Networks}.
\newblock In \emph{IEEE/CVF Conference on Computer Vision and Pattern Recognition (CVPR)}, 2018.

\bibitem[Boulch et~al.(2018)Boulch, Guerry, Le~Saux, and Audebert]{boulch2018snapnet}
Alexandre Boulch, Joris Guerry, Bertrand Le~Saux, and Nicolas Audebert.
\newblock {SnapNet: 3D Point Cloud Semantic Labeling with 2D Deep Segmentation Networks}.
\newblock \emph{Computers \& Graphics}, 2018.

\bibitem[Busch et~al.(2025)Busch, Homberger, Ortega-Peimbert, Yang, and Andersson]{busch2025one}
Finn~Lukas Busch, Timon Homberger, Jes{\'u}s Ortega-Peimbert, Quantao Yang, and Olov Andersson.
\newblock {One Map to Find them All: Real-time Open-vocabulary Mapping for Zero-shot Multi-object Navigation}.
\newblock In \emph{International Conference on Robotics and Automation (ICRA)}, 2025.

\bibitem[Caesar et~al.(2020)Caesar, Bankiti, Lang, Vora, Liong, Xu, Krishnan, Pan, Baldan, and Beijbom]{caesar2020nuscenes}
Holger Caesar, Varun Bankiti, Alex~H Lang, Sourabh Vora, Venice~Erin Liong, Qiang Xu, Anush Krishnan, Yu Pan, Giancarlo Baldan, and Oscar Beijbom.
\newblock {nuScenes: A Multimodal Dataset for Autonomous Driving}.
\newblock In \emph{IEEE/CVF Conference on Computer Vision and Pattern Recognition (CVPR)}, 2020.

\bibitem[Chen et~al.(2025)Chen, Meyer, Zhang, Wolff, and Vernaza]{chen2025flash3d}
Liyan Chen, Gregory~P Meyer, Zaiwei Zhang, Eric~M Wolff, and Paul Vernaza.
\newblock {Flash3D: Super-scaling Point Transformers through Joint Hardware-Geometry Locality}.
\newblock In \emph{IEEE/CVF Conference on Computer Vision and Pattern Recognition (CVPR)}, 2025.

\bibitem[Chen et~al.(2022)Chen, Zhu, Chen, and Yu]{chen2022efficient}
Wanli Chen, Xinge Zhu, Guojin Chen, and Bei Yu.
\newblock {Efficient Point Cloud Analysis Using Hilbert Curve}.
\newblock In \emph{European Conference on Computer Vision (ECCV)}, 2022.

\bibitem[Chen et~al.(2017)Chen, Ma, Wan, Li, and Xia]{chen2017multi}
Xiaozhi Chen, Huimin Ma, Ji Wan, Bo Li, and Tian Xia.
\newblock {Multi-View 3D Object Detection Network for Autonomous Driving}.
\newblock In \emph{IEEE/CVF Conference on Computer Vision and Pattern Recognition (CVPR)}, 2017.

\bibitem[Chen et~al.(2023)Chen, Liu, Zhang, Qi, and Jia]{chen2023largekernel3d}
Yukang Chen, Jianhui Liu, Xiangyu Zhang, Xiaojuan Qi, and Jiaya Jia.
\newblock {LargeKernel3D: Scaling Up Kernels in 3D Sparse CNNs}.
\newblock In \emph{IEEE/CVF Conference on Computer Vision and Pattern Recognition (CVPR)}, 2023.

\bibitem[Cheng et~al.(2021)Cheng, Razani, Taghavi, Li, and Liu]{cheng20212}
Ran Cheng, Ryan Razani, Ehsan Taghavi, Enxu Li, and Bingbing Liu.
\newblock {(AF)2-S3Net: Attentive Feature Fusion with Adaptive Feature Selection for Sparse Semantic Segmentation Network}.
\newblock In \emph{IEEE/CVF Conference on Computer Vision and Pattern Recognition (CVPR)}, 2021.

\bibitem[Choy et~al.(2019)Choy, Gwak, and Savarese]{choy20194d}
Christopher Choy, JunYoung Gwak, and Silvio Savarese.
\newblock {4D Spatio-Temporal ConvNets: Minkowski Convolutional Neural Networks}.
\newblock In \emph{IEEE/CVF Conference on Computer Vision and Pattern Recognition (CVPR)}, 2019.

\bibitem[Chu et~al.(2023)Chu, Tian, Zhang, Wang, and Shen]{chu2021conditional}
Xiangxiang Chu, Zhi Tian, Bo Zhang, Xinlong Wang, and Chunhua Shen.
\newblock {Conditional Positional Encodings for Vision Transformers}.
\newblock \emph{International Conference on Learning Representations (ICLR)}, 2023.

\bibitem[Dai et~al.(2017)Dai, Chang, Savva, Halber, Funkhouser, and Nie{\ss}ner]{dai2017scannet}
Angela Dai, Angel~X Chang, Manolis Savva, Maciej Halber, Thomas Funkhouser, and Matthias Nie{\ss}ner.
\newblock {ScanNet: Richly-Annotated 3D Reconstructions of Indoor Scenes}.
\newblock In \emph{IEEE/CVF Conference on Computer Vision and Pattern Recognition (CVPR)}, 2017.

\bibitem[Dai et~al.(2021)Dai, Liu, Le, and Tan]{dai2021coatnet}
Zihang Dai, Hanxiao Liu, Quoc~V Le, and Mingxing Tan.
\newblock {CoAtNet: Marrying Convolution and Attention for All Data Sizes}.
\newblock \emph{Advances in Neural Information Processing Systems (NeurIPS)}, 2021.

\bibitem[Dao(2024)]{dao2023flashattention}
Tri Dao.
\newblock {FlashAttention-2: Faster Attention with Better Parallelism and Work Partitioning}.
\newblock \emph{International Conference on Learning Representations (ICLR)}, 2024.

\bibitem[Dao et~al.(2022)Dao, Fu, Ermon, Rudra, and R{\'e}]{dao2022flashattention}
Tri Dao, Dan Fu, Stefano Ermon, Atri Rudra, and Christopher R{\'e}.
\newblock {FlashAttention: Fast and Memory-Efficient Exact Attention with IO-Awareness}.
\newblock \emph{Advances in Neural Information Processing Systems (NeurIPS)}, 2022.

\bibitem[Deng et~al.(2024)Deng, Jing, Cheng, Liu, Ru, Bo, and Wang]{deng2024linnet}
Hao Deng, Kunlei Jing, Shengmei Cheng, Cheng Liu, Jiawei Ru, Jiang Bo, and Lin Wang.
\newblock {LinNet: Linear Network for Efficient Point Cloud Representation Learning}.
\newblock \emph{Advances in Neural Information Processing Systems (NeurIPS)}, 2024.

\bibitem[Dosovitskiy et~al.(2021)Dosovitskiy, Beyer, Kolesnikov, Weissenborn, Zhai, Unterthiner, Dehghani, Minderer, Heigold, Gelly, et~al.]{dosovitskiy2020image}
Alexey Dosovitskiy, Lucas Beyer, Alexander Kolesnikov, Dirk Weissenborn, Xiaohua Zhai, Thomas Unterthiner, Mostafa Dehghani, Matthias Minderer, Georg Heigold, Sylvain Gelly, et~al.
\newblock {An Image is Worth 16$\times$16 Words: Transformers for Image Recognition at Scale}.
\newblock \emph{International Conference on Learning Representations (ICLR)}, 2021.

\bibitem[Duan et~al.(2023)Duan, Zhao, Xue, Gong, Xia, and Tao]{duan2023condaformer}
Lunhao Duan, Shanshan Zhao, Nan Xue, Mingming Gong, Gui-Song Xia, and Dacheng Tao.
\newblock {ConDaFormer: Disassembled Transformer with Local Structure Enhancement for 3D Point Cloud Understanding}.
\newblock \emph{Advances in Neural Information Processing Systems (NeurIPS)}, 2023.

\bibitem[Fan et~al.(2022)Fan, Pang, Zhang, Wang, Zhao, Wang, Wang, and Zhang]{fan2022embracing}
Lue Fan, Ziqi Pang, Tianyuan Zhang, Yu-Xiong Wang, Hang Zhao, Feng Wang, Naiyan Wang, and Zhaoxiang Zhang.
\newblock {Embracing Single Stride 3D Object Detector with Sparse Transformer}.
\newblock In \emph{IEEE/CVF Conference on Computer Vision and Pattern Recognition (CVPR)}, 2022.

\bibitem[Geiger et~al.(2012)Geiger, Lenz, and Urtasun]{geiger2012kitti}
Andreas Geiger, Philip Lenz, and Raquel Urtasun.
\newblock {Are we ready for Autonomous Driving? The KITTI Vision Benchmark Suite}.
\newblock In \emph{IEEE/CVF Conference on Computer Vision and Pattern Recognition (CVPR)}, 2012.

\bibitem[Graham et~al.(2018)Graham, Engelcke, and Van Der~Maaten]{graham20183d}
Benjamin Graham, Martin Engelcke, and Laurens Van Der~Maaten.
\newblock {3D Semantic Segmentation with Submanifold Sparse Convolutional Networks}.
\newblock In \emph{IEEE/CVF Conference on Computer Vision and Pattern Recognition (CVPR)}, 2018.

\bibitem[Groh et~al.(2018)Groh, Wieschollek, and Lensch]{groh2018flex}
Fabian Groh, Patrick Wieschollek, and Hendrik~PA Lensch.
\newblock {Flex-Convolution}.
\newblock In \emph{Asian Conference on Computer Vision (ACCV)}, 2018.

\bibitem[Guo et~al.(2022)Guo, Han, Wu, Tang, Chen, Wang, and Xu]{guo2022cmt}
Jianyuan Guo, Kai Han, Han Wu, Yehui Tang, Xinghao Chen, Yunhe Wang, and Chang Xu.
\newblock {CMT: Convolutional Neural Networks Meet Vision Transformers }.
\newblock In \emph{IEEE/CVF Conference on Computer Vision and Pattern Recognition (CVPR)}, 2022.

\bibitem[Guo et~al.(2021)Guo, Cai, Liu, Mu, Martin, and Hu]{guo2021pct}
Meng-Hao Guo, Jun-Xiong Cai, Zheng-Ning Liu, Tai-Jiang Mu, Ralph~R Martin, and Shi-Min Hu.
\newblock {PCT: Point Cloud Transformer}.
\newblock \emph{Computational Visual Media}, 2021.

\bibitem[Han et~al.(2020)Han, Zheng, Xu, and Fang]{han2020occuseg}
Lei Han, Tian Zheng, Lan Xu, and Lu Fang.
\newblock {OccuSeg: Occupancy-aware 3D Instance Segmentation}.
\newblock In \emph{IEEE/CVF Conference on Computer Vision and Pattern Recognition (CVPR)}, 2020.

\bibitem[He et~al.(2022)He, Li, Li, and Zhang]{he2022voxel}
Chenhang He, Ruihuang Li, Shuai Li, and Lei Zhang.
\newblock {Voxel Set Transformer: A Set-to-Set Approach to 3D Object Detection From Point Clouds}.
\newblock In \emph{IEEE/CVF Conference on Computer Vision and Pattern Recognition (CVPR)}, 2022.

\bibitem[He et~al.(2016)He, Zhang, Ren, and Sun]{he2016deep}
Kaiming He, Xiangyu Zhang, Shaoqing Ren, and Jian Sun.
\newblock {Deep Residual Learning for Image Recognition}.
\newblock In \emph{IEEE/CVF Conference on Computer Vision and Pattern Recognition (CVPR)}, 2016.

\bibitem[He et~al.(2021)He, Shen, and Van Den~Hengel]{he2021dyco3d}
Tong He, Chunhua Shen, and Anton Van Den~Hengel.
\newblock {DyCo3D: Robust Instance Segmentation of 3D Point Clouds Through Dynamic Convolution}.
\newblock In \emph{IEEE/CVF Conference on Computer Vision and Pattern Recognition (CVPR)}, 2021.

\bibitem[Hendrycks and Gimpel(2016)]{hendrycks2016gaussian}
Dan Hendrycks and Kevin Gimpel.
\newblock {Gaussian Error Linear Units (GELUs)}.
\newblock \emph{arXiv preprint arXiv:1606.08415}, 2016.

\bibitem[Hou et~al.(2021)Hou, Graham, Nie{\ss}ner, and Xie]{hou2021exploring}
Ji Hou, Benjamin Graham, Matthias Nie{\ss}ner, and Saining Xie.
\newblock {Exploring Data-Efficient 3D Scene Understanding with Contrastive Scene Contexts}.
\newblock In \emph{IEEE/CVF Conference on Computer Vision and Pattern Recognition (CVPR)}, 2021.

\bibitem[Hua et~al.(2018)Hua, Tran, and Yeung]{hua2018pointwise}
Binh-Son Hua, Minh-Khoi Tran, and Sai-Kit Yeung.
\newblock {Pointwise Convolutional Neural Networks}.
\newblock In \emph{IEEE/CVF Conference on Computer Vision and Pattern Recognition (CVPR)}, 2018.

\bibitem[Huang and You(2016)]{huang2016point}
Jing Huang and Suya You.
\newblock {Point Cloud Labeling Using 3D Convolutional Neural Network}.
\newblock In \emph{International Conference on Pattern Recognition (ICPR)}, 2016.

\bibitem[Iglhaut et~al.(2019)Iglhaut, Cabo, Puliti, Piermattei, O’Connor, and Rosette]{iglhaut2019structure}
Jakob Iglhaut, Carlos Cabo, Stefano Puliti, Livia Piermattei, James O’Connor, and Jacqueline Rosette.
\newblock {Structure from Motion Photogrammetry in Forestry: A Review}.
\newblock \emph{Current Forestry Reports}, 2019.

\bibitem[Ioffe and Szegedy(2015)]{ioffe2015batch}
Sergey Ioffe and Christian Szegedy.
\newblock {Batch Normalization: Accelerating Deep Network Training by Reducing Internal Covariate Shift}.
\newblock In \emph{International Conference on Machine Learning (ICML)}, 2015.

\bibitem[Jiang et~al.(2020)Jiang, Zhao, Shi, Liu, Fu, and Jia]{jiang2020pointgroup}
Li Jiang, Hengshuang Zhao, Shaoshuai Shi, Shu Liu, Chi-Wing Fu, and Jiaya Jia.
\newblock {PointGroup: Dual-Set Point Grouping for 3D Instance Segmentation}.
\newblock In \emph{IEEE/CVF Conference on Computer Vision and Pattern Recognition (CVPR)}, 2020.

\bibitem[Kalogerakis et~al.(2017)Kalogerakis, Averkiou, Maji, and Chaudhuri]{kalogerakis20173d}
Evangelos Kalogerakis, Melinos Averkiou, Subhransu Maji, and Siddhartha Chaudhuri.
\newblock {3D Shape Segmentation with Projective Convolutional Networks}.
\newblock In \emph{IEEE/CVF Conference on Computer Vision and Pattern Recognition (CVPR)}, 2017.

\bibitem[Kingma and Ba(2015)]{kingma2014adam}
Diederik~P Kingma and Jimmy Ba.
\newblock {Adam: A Method for Stochastic Optimization}.
\newblock \emph{International Conference on Learning Representations (ICLR)}, 2015.

\bibitem[Lai et~al.(2022)Lai, Liu, Jiang, Wang, Zhao, Liu, Qi, and Jia]{lai2022stratified}
Xin Lai, Jianhui Liu, Li Jiang, Liwei Wang, Hengshuang Zhao, Shu Liu, Xiaojuan Qi, and Jiaya Jia.
\newblock {Stratified Transformer for 3D Point Cloud Segmentation}.
\newblock In \emph{IEEE/CVF Conference on Computer Vision and Pattern Recognition (CVPR)}, 2022.

\bibitem[Lai et~al.(2023)Lai, Chen, Lu, Liu, and Jia]{lai2023spherical}
Xin Lai, Yukang Chen, Fanbin Lu, Jianhui Liu, and Jiaya Jia.
\newblock {Spherical Transformer for LiDAR-Based 3D Recognition}.
\newblock In \emph{IEEE/CVF Conference on Computer Vision and Pattern Recognition (CVPR)}, 2023.

\bibitem[Landrieu and Simonovsky(2018)]{landrieu2018large}
Loic Landrieu and Martin Simonovsky.
\newblock {Large-Scale Point Cloud Semantic Segmentation with Superpoint Graphs}.
\newblock In \emph{IEEE/CVF Conference on Computer Vision and Pattern Recognition (CVPR)}, 2018.

\bibitem[Lang et~al.(2019)Lang, Vora, Caesar, Zhou, Yang, and Beijbom]{lang2019pointpillars}
Alex~H Lang, Sourabh Vora, Holger Caesar, Lubing Zhou, Jiong Yang, and Oscar Beijbom.
\newblock {PointPillars: Fast Encoders for Object Detection from Point Clouds}.
\newblock In \emph{IEEE/CVF Conference on Computer Vision and Pattern Recognition (CVPR)}, 2019.

\bibitem[Li et~al.(2018)Li, Bu, Sun, Wu, Di, and Chen]{li2018pointcnn}
Yangyan Li, Rui Bu, Mingchao Sun, Wei Wu, Xinhan Di, and Baoquan Chen.
\newblock {PointCNN: Convolution On X-Transformed Points}.
\newblock \emph{Advances in Neural Information Processing Systems (NeurIPS)}, 2018.

\bibitem[Liang et~al.(2024)Liang, Zhou, Xu, Zhu, Zou, Ye, Tan, and Bai]{liang2024pointmamba}
Dingkang Liang, Xin Zhou, Wei Xu, Xingkui Zhu, Zhikang Zou, Xiaoqing Ye, Xiao Tan, and Xiang Bai.
\newblock {PointMamba: A Simple State Space Model for Point Cloud Analysis}.
\newblock \emph{Advances in Neural Information Processing Systems (NeurIPS)}, 2024.

\bibitem[Liu et~al.(2019)Liu, Tang, Lin, and Han]{liu2019point}
Zhijian Liu, Haotian Tang, Yujun Lin, and Song Han.
\newblock {Point-Voxel CNN for Efficient 3D Deep Learning}.
\newblock \emph{Advances in Neural Information Processing Systems (NeurIPS)}, 2019.

\bibitem[Liu et~al.(2023)Liu, Yang, Tang, Yang, and Han]{liu2023flatformer}
Zhijian Liu, Xinyu Yang, Haotian Tang, Shang Yang, and Song Han.
\newblock {FlatFormer: Flattened Window Attention for Efficient Point Cloud Transformer}.
\newblock In \emph{IEEE/CVF Conference on Computer Vision and Pattern Recognition (CVPR)}, 2023.

\bibitem[Loshchilov and Hutter(2019)]{loshchilov2017decoupled}
Ilya Loshchilov and Frank Hutter.
\newblock {Decoupled Weight Decay Regularization}.
\newblock \emph{International Conference on Learning Representations (ICLR)}, 2019.

\bibitem[Luo et~al.(2024)Luo, Yu, Chen, Yang, Wang, Cheng, and Mian]{luo20243d}
Kan Luo, Hongshan Yu, Xieyuanli Chen, Zhengeng Yang, Jingwen Wang, Panfei Cheng, and Ajmal Mian.
\newblock {3D Point Cloud-based Place Recognition: A Survey}.
\newblock \emph{Artificial Intelligence Review}, 2024.

\bibitem[Ma et~al.(2022)Ma, Qin, You, Ran, and Fu]{ma2022rethinking}
Xu Ma, Can Qin, Haoxuan You, Haoxi Ran, and Yun Fu.
\newblock {Rethinking Network Design and Local Geometry in Point Cloud: A Simple Residual MLP Framework}.
\newblock \emph{International Conference on Learning Representations (ICLR)}, 2022.

\bibitem[Maturana and Scherer(2015)]{maturana2015voxnet}
Daniel Maturana and Sebastian Scherer.
\newblock {VoxNet: A 3D Convolutional Neural Network for Real-Time Object Recognition}.
\newblock In \emph{IEEE/RSJ International Conference on Intelligent Robots and Systems (IROS)}, 2015.

\bibitem[Mehta and Rastegari(2022)]{mehta2021mobilevit}
Sachin Mehta and Mohammad Rastegari.
\newblock {MobileViT: Light-Weight, General-Purpose, and Mobile-Friendly Vision Transformer}.
\newblock \emph{International Conference on Learning Representations (ICLR)}, 2022.

\bibitem[Park et~al.(2022)Park, Jeong, Cho, and Park]{park2022fast}
Chunghyun Park, Yoonwoo Jeong, Minsu Cho, and Jaesik Park.
\newblock {Fast Point Transformer}.
\newblock In \emph{IEEE/CVF Conference on Computer Vision and Pattern Recognition (CVPR)}, 2022.

\bibitem[Peng et~al.(2024)Peng, Wu, Jiang, Chen, Zhao, Tian, and Jia]{peng2024oa}
Bohao Peng, Xiaoyang Wu, Li Jiang, Yukang Chen, Hengshuang Zhao, Zhuotao Tian, and Jiaya Jia.
\newblock {OA-CNNs: Omni-Adaptive Sparse CNNs for 3D Semantic Segmentation}.
\newblock In \emph{IEEE/CVF Conference on Computer Vision and Pattern Recognition (CVPR)}, 2024.

\bibitem[Pfaff et~al.(2007)Pfaff, Triebel, Stachniss, Lamon, Burgard, and Siegwart]{pfaff2007towards}
Patrick Pfaff, Rudolph Triebel, Cyrill Stachniss, Pierre Lamon, Wolfram Burgard, and Roland Siegwart.
\newblock {Towards Mapping of Cities}.
\newblock In \emph{International Conference on Robotics and Automation (ICRA)}, 2007.

\bibitem[Qi et~al.(2017{\natexlab{a}})Qi, Su, Mo, and Guibas]{qi2017pointnet}
Charles~R Qi, Hao Su, Kaichun Mo, and Leonidas~J Guibas.
\newblock {PointNet: Deep Learning on Point Sets for 3D Classification and Segmentation}.
\newblock In \emph{IEEE/CVF Conference on Computer Vision and Pattern Recognition (CVPR)}, 2017{\natexlab{a}}.

\bibitem[Qi et~al.(2017{\natexlab{b}})Qi, Yi, Su, and Guibas]{qi2017pointnet++}
Charles~Ruizhongtai Qi, Li Yi, Hao Su, and Leonidas~J Guibas.
\newblock {PointNet++: Deep Hierarchical Feature Learning on Point Sets in a Metric Space}.
\newblock \emph{Advances in Neural Information Processing Systems (NeurIPS)}, 2017{\natexlab{b}}.

\bibitem[Qian et~al.(2022)Qian, Li, Peng, Mai, Hammoud, Elhoseiny, and Ghanem]{qian2022pointnext}
Guocheng Qian, Yuchen Li, Houwen Peng, Jinjie Mai, Hasan Hammoud, Mohamed Elhoseiny, and Bernard Ghanem.
\newblock {PointNeXt: Revisiting PointNet++ with Improved Training and Scaling Strategies}.
\newblock \emph{Advances in Neural Information Processing Systems (NeurIPS)}, 2022.

\bibitem[Robert et~al.(2023)Robert, Raguet, and Landrieu]{robert2023efficient}
Damien Robert, Hugo Raguet, and Loic Landrieu.
\newblock {Efficient 3D Semantic Segmentation with Superpoint Transformer}.
\newblock In \emph{IEEE/CVF Conference on Computer Vision and Pattern Recognition (CVPR)}, 2023.

\bibitem[Robert et~al.(2024)Robert, Raguet, and Landrieu]{robert2024scalable}
Damien Robert, Hugo Raguet, and Loic Landrieu.
\newblock {Scalable 3D Panoptic Segmentation As Superpoint Graph Clustering}.
\newblock In \emph{International Conference on 3D Vision (3DV)}, 2024.

\bibitem[Ronneberger et~al.(2015)Ronneberger, Fischer, and Brox]{ronneberger2015u}
Olaf Ronneberger, Philipp Fischer, and Thomas Brox.
\newblock {U-Net: Convolutional Networks for Biomedical Image Segmentation}.
\newblock In \emph{International Conference on Medical Image Computing and Computer-Assisted Intervention ( MICCAI)}, 2015.

\bibitem[Rozenberszki et~al.(2022)Rozenberszki, Litany, and Dai]{rozenberszki2022language}
David Rozenberszki, Or Litany, and Angela Dai.
\newblock {Language-Grounded Indoor 3D Semantic Segmentation in the Wild}.
\newblock In \emph{European Conference on Computer Vision (ECCV)}, 2022.

\bibitem[Schult et~al.(2023)Schult, Engelmann, Hermans, Litany, Tang, and Leibe]{schult2023mask3d}
Jonas Schult, Francis Engelmann, Alexander Hermans, Or Litany, Siyu Tang, and Bastian Leibe.
\newblock {Mask3D: Mask Transformer for 3D Semantic Instance Segmentation}.
\newblock In \emph{International Conference on Robotics and Automation (ICRA)}, 2023.

\bibitem[Shah et~al.(2024)Shah, Bikshandi, Zhang, Thakkar, Ramani, and Dao]{shah2024flashattention}
Jay Shah, Ganesh Bikshandi, Ying Zhang, Vijay Thakkar, Pradeep Ramani, and Tri Dao.
\newblock {FlashAttention-3: Fast and Accurate Attention with Asynchrony and Low-precision}.
\newblock \emph{Advances in Neural Information Processing Systems (NeurIPS)}, 2024.

\bibitem[Shi et~al.(2022)Shi, Li, and Ma]{shi2022pillarnet}
Guangsheng Shi, Ruifeng Li, and Chao Ma.
\newblock {PillarNet: Real-Time and High-Performance Pillar-Based 3D Object Detection}.
\newblock In \emph{European Conference on Computer Vision (ECCV)}, 2022.

\bibitem[Simonyan and Zisserman(2015)]{simonyan2014very}
Karen Simonyan and Andrew Zisserman.
\newblock {Very Deep Convolutional Networks for Large-Scale Image Recognition}.
\newblock \emph{International Conference on Learning Representations (ICLR)}, 2015.

\bibitem[Smith and Topin(2019)]{smith2019super}
Leslie~N Smith and Nicholay Topin.
\newblock {Super-Convergence: Very Fast Training of Neural Networks Using Large Learning Rates}.
\newblock In \emph{Artificial Intelligence and Machine Learning for Multi-Domain Operations Applications}, 2019.

\bibitem[Song et~al.(2025)Song, Wen, Wu, and Guo]{song2025comprehensive}
Hongli Song, Weiliang Wen, Sheng Wu, and Xinyu Guo.
\newblock {Comprehensive Review on 3D Point Cloud Segmentation in Plants}.
\newblock \emph{Artificial Intelligence in Agriculture}, 2025.

\bibitem[Song et~al.(2017)Song, Yu, Zeng, Chang, Savva, and Funkhouser]{song2017semantic}
Shuran Song, Fisher Yu, Andy Zeng, Angel~X Chang, Manolis Savva, and Thomas Funkhouser.
\newblock {Semantic Scene Completion from a Single Depth Image}.
\newblock In \emph{IEEE/CVF Conference on Computer Vision and Pattern Recognition (CVPR)}, 2017.

\bibitem[Su et~al.(2015)Su, Maji, Kalogerakis, and Learned-Miller]{su2015multi}
Hang Su, Subhransu Maji, Evangelos Kalogerakis, and Erik Learned-Miller.
\newblock {Multi-View Convolutional Neural Networks for 3D Shape Recognition}.
\newblock In \emph{International Conference on Computer Vision (ICCV)}, 2015.

\bibitem[Su et~al.(2024)Su, Ahmed, Lu, Pan, Bo, and Liu]{su2024roformer}
Jianlin Su, Murtadha Ahmed, Yu Lu, Shengfeng Pan, Wen Bo, and Yunfeng Liu.
\newblock {RoFormer: Enhanced Transformer with Rotary Position Embedding}.
\newblock \emph{Neurocomputing}, 2024.

\bibitem[Sun et~al.(2020)Sun, Kretzschmar, Dotiwalla, Chouard, Patnaik, Tsui, Guo, Zhou, Chai, Caine, et~al.]{sun2020scalability}
Pei Sun, Henrik Kretzschmar, Xerxes Dotiwalla, Aurelien Chouard, Vijaysai Patnaik, Paul Tsui, James Guo, Yin Zhou, Yuning Chai, Benjamin Caine, et~al.
\newblock {Scalability in Perception for Autonomous Driving: Waymo Open Dataset}.
\newblock In \emph{IEEE/CVF Conference on Computer Vision and Pattern Recognition (CVPR)}, 2020.

\bibitem[Sun et~al.(2022)Sun, Tan, Wang, Liu, Xia, Leng, and Anguelov]{sun2022swformer}
Pei Sun, Mingxing Tan, Weiyue Wang, Chenxi Liu, Fei Xia, Zhaoqi Leng, and Dragomir Anguelov.
\newblock {SWFormer: Sparse Window Transformer for 3D Object Detection in Point Clouds}.
\newblock In \emph{European Conference on Computer Vision (ECCV)}, 2022.

\bibitem[Tang et~al.(2020)Tang, Liu, Zhao, Lin, Lin, Wang, and Han]{tang2020searching}
Haotian Tang, Zhijian Liu, Shengyu Zhao, Yujun Lin, Ji Lin, Hanrui Wang, and Song Han.
\newblock {Searching Efficient 3D Architectures with Sparse Point-Voxel Convolution}.
\newblock In \emph{European Conference on Computer Vision (ECCV)}, 2020.

\bibitem[Thomas et~al.(2019)Thomas, Qi, Deschaud, Marcotegui, Goulette, and Guibas]{thomas2019kpconv}
Hugues Thomas, Charles~R Qi, Jean-Emmanuel Deschaud, Beatriz Marcotegui, Fran{\c{c}}ois Goulette, and Leonidas~J Guibas.
\newblock {KPConv: Flexible and Deformable Convolution for Point Clouds}.
\newblock In \emph{International Conference on Computer Vision (ICCV)}, 2019.

\bibitem[Thomas et~al.(2024)Thomas, Tsai, Barfoot, and Zhang]{thomas2024kpconvx}
Hugues Thomas, Yao-Hung~Hubert Tsai, Timothy~D Barfoot, and Jian Zhang.
\newblock {KPConvX: Modernizing Kernel Point Convolution with Kernel Attention}.
\newblock In \emph{IEEE/CVF Conference on Computer Vision and Pattern Recognition (CVPR)}, 2024.

\bibitem[Tran et~al.(2025)Tran, Nguyen, Tran, Barz, Doan, Wattenhofer, Ngo, Niepert, Sonntag, and Swoboda]{gitmerge3d2025}
Tuan~Anh Tran, Duy Minh~Ho Nguyen, Hoai-Chau Tran, Michael Barz, Khoa~D. Doan, Roger Wattenhofer, Vien~Anh Ngo, Mathias Niepert, Daniel Sonntag, and Paul Swoboda.
\newblock {How Many Tokens Do 3D Point Cloud Transformer Architectures Really Need?}
\newblock \emph{Advances in Neural Information Processing Systems (NeurIPS)}, 2025.

\bibitem[Tu et~al.(2022)Tu, Talebi, Zhang, Yang, Milanfar, Bovik, and Li]{tu2022maxvit}
Zhengzhong Tu, Hossein Talebi, Han Zhang, Feng Yang, Peyman Milanfar, Alan Bovik, and Yinxiao Li.
\newblock {MaxViT: Multi-axis Vision Transformer}.
\newblock In \emph{European Conference on Computer Vision (ECCV)}, 2022.

\bibitem[Varney et~al.(2020)Varney, Asari, and Graehling]{varney2020dales}
Nina Varney, Vijayan~K Asari, and Quinn Graehling.
\newblock {DALES: A Large-scale Aerial LiDAR Data Set for Semantic Segmentation}.
\newblock In \emph{IEEE/CVF Conference on Computer Vision and Pattern Recognition (CVPR) Workshops}, 2020.

\bibitem[Wang(2023)]{wang2023octformer}
Peng-Shuai Wang.
\newblock {OctFormer: Octree-based Transformers for 3D Point Clouds}.
\newblock \emph{ACM Transactions on Graphics (TOG)}, 2023.

\bibitem[Wang et~al.(2018)Wang, Peethambaran, and Chen]{wang2018lidar}
Ruisheng Wang, Jiju Peethambaran, and Dong Chen.
\newblock {Lidar Point Clouds to 3-D Urban Models: A Review}.
\newblock \emph{IEEE Journal of Selected Topics in Applied Earth Observations and Remote Sensing}, 2018.

\bibitem[Wang et~al.(2019)Wang, Sun, Liu, Sarma, Bronstein, and Solomon]{wang2019dynamic}
Yue Wang, Yongbin Sun, Ziwei Liu, Sanjay~E Sarma, Michael~M Bronstein, and Justin~M Solomon.
\newblock {Dynamic Graph CNN for Learning on Point Clouds}.
\newblock \emph{ACM Transactions on Graphics (TOG)}, 2019.

\bibitem[Wu et~al.(2018)Wu, Wan, Yue, and Keutzer]{wu2018squeezeseg}
Bichen Wu, Alvin Wan, Xiangyu Yue, and Kurt Keutzer.
\newblock {SqueezeSeg: Convolutional Neural Nets with Recurrent CRF for Real-Time Road-Object Segmentation from 3D Lidar Point Cloud}.
\newblock In \emph{International Conference on Robotics and Automation (ICRA)}, 2018.

\bibitem[Wu et~al.(2021)Wu, Xiao, Codella, Liu, Dai, Yuan, and Zhang]{wu2021cvt}
Haiping Wu, Bin Xiao, Noel Codella, Mengchen Liu, Xiyang Dai, Lu Yuan, and Lei Zhang.
\newblock {CvT: Introducing Convolutions to Vision Transformers}.
\newblock In \emph{International Conference on Computer Vision (ICCV)}, 2021.

\bibitem[Wu et~al.(2019)Wu, Qi, and Fuxin]{wu2019pointconv}
Wenxuan Wu, Zhongang Qi, and Li Fuxin.
\newblock {PointConv: Deep Convolutional Networks on 3D Point Clouds}.
\newblock In \emph{IEEE/CVF Conference on Computer Vision and Pattern Recognition (CVPR)}, 2019.

\bibitem[Wu et~al.(2023)Wu, Fuxin, and Shan]{wu2023pointconvformer}
Wenxuan Wu, Li Fuxin, and Qi Shan.
\newblock {PointConvFormer: Revenge of the Point-based Convolution}.
\newblock In \emph{IEEE/CVF Conference on Computer Vision and Pattern Recognition (CVPR)}, 2023.

\bibitem[Wu et~al.(2022)Wu, Lao, Jiang, Liu, and Zhao]{wu2022point}
Xiaoyang Wu, Yixing Lao, Li Jiang, Xihui Liu, and Hengshuang Zhao.
\newblock {Point Transformer V2: Grouped Vector Attention and Partition-Based Pooling}.
\newblock \emph{Advances in Neural Information Processing Systems (NeurIPS)}, 2022.

\bibitem[Wu et~al.(2024)Wu, Jiang, Wang, Liu, Liu, Qiao, Ouyang, He, and Zhao]{wu2024point}
Xiaoyang Wu, Li Jiang, Peng-Shuai Wang, Zhijian Liu, Xihui Liu, Yu Qiao, Wanli Ouyang, Tong He, and Hengshuang Zhao.
\newblock {Point Transformer V3: Simpler, Faster, Stronger}.
\newblock In \emph{IEEE/CVF Conference on Computer Vision and Pattern Recognition (CVPR)}, 2024.

\bibitem[Wu et~al.(2025)Wu, DeTone, Frost, Shen, Xie, Yang, Engel, Newcombe, Zhao, and Straub]{wu2025sonata}
Xiaoyang Wu, Daniel DeTone, Duncan Frost, Tianwei Shen, Chris Xie, Nan Yang, Jakob Engel, Richard Newcombe, Hengshuang Zhao, and Julian Straub.
\newblock {Sonata: Self-Supervised Learning of Reliable Point Representations}.
\newblock In \emph{IEEE/CVF Conference on Computer Vision and Pattern Recognition (CVPR)}, 2025.

\bibitem[Wurm et~al.(2010)Wurm, Hornung, Bennewitz, Stachniss, and Burgard]{wurm2010occupancy}
Kai~M Wurm, Armin Hornung, Maren Bennewitz, Cyrill Stachniss, and Wolfram Burgard.
\newblock {OctoMap: A Probabilistic, Flexible, and Compact 3D Map Representation for Robotic Systems}.
\newblock In \emph{International Conference on Robotics and Automation (ICRA)}, 2010.

\bibitem[Xiong et~al.(2020)Xiong, Yang, He, Zheng, Zheng, Xing, Zhang, Lan, Wang, and Liu]{xiong2020layer}
Ruibin Xiong, Yunchang Yang, Di He, Kai Zheng, Shuxin Zheng, Chen Xing, Huishuai Zhang, Yanyan Lan, Liwei Wang, and Tieyan Liu.
\newblock {On Layer Normalization in the Transformer Architecture}.
\newblock In \emph{International Conference on Machine Learning (ICML)}, 2020.

\bibitem[Xu et~al.(2015)Xu, Honegger, Pollefeys, and Heng]{xu2015real}
Shengdong Xu, Dominik Honegger, Marc Pollefeys, and Lionel Heng.
\newblock {Real-Time 3D Navigation for Autonomous Vision-Guided MAVs}.
\newblock In \emph{IEEE/RSJ International Conference on Intelligent Robots and Systems (IROS)}, 2015.

\bibitem[Xu et~al.(2018)Xu, Fan, Xu, Zeng, and Qiao]{xu2018spidercnn}
Yifan Xu, Tianqi Fan, Mingye Xu, Long Zeng, and Yu Qiao.
\newblock {SpiderCNN: Deep Learning on Point Sets with Parameterized Convolutional Filter}.
\newblock In \emph{European Conference on Computer Vision (ECCV)}, 2018.

\bibitem[Xu et~al.(2021)Xu, Zhang, Zhang, and Tao]{xu2021vitae}
Yufei Xu, Qiming Zhang, Jing Zhang, and Dacheng Tao.
\newblock {ViTAE: Vision Transformer Advanced by Exploring Intrinsic Inductive Bias}.
\newblock \emph{Advances in Neural Information Processing Systems (NeurIPS)}, 2021.

\bibitem[Yang et~al.(2023)Yang, Wang, Chen, Lin, He, Chen, He, and Ouyang]{yang2023pvt}
Honghui Yang, Wenxiao Wang, Minghao Chen, Binbin Lin, Tong He, Hua Chen, Xiaofei He, and Wanli Ouyang.
\newblock {PVT-SSD: Single-Stage 3D Object Detector With Point-Voxel Transformer}.
\newblock In \emph{IEEE/CVF Conference on Computer Vision and Pattern Recognition (CVPR)}, 2023.

\bibitem[Yang et~al.(2025{\natexlab{a}})Yang, Guo, and Liu]{yang2025swin3d++}
Yu-Qi Yang, Yu-Xiao Guo, and Yang Liu.
\newblock {Swin3D++: Effective Multi-Source Pretraining for 3D Indoor Scene Understanding}.
\newblock \emph{Computational Visual Media}, 2025{\natexlab{a}}.

\bibitem[Yang et~al.(2025{\natexlab{b}})Yang, Guo, Xiong, Liu, Pan, Wang, Tong, and Guo]{yang2025swin3d}
Yu-Qi Yang, Yu-Xiao Guo, Jian-Yu Xiong, Yang Liu, Hao Pan, Peng-Shuai Wang, Xin Tong, and Baining Guo.
\newblock {Swin3D: A Pretrained Transformer Backbone for 3D Indoor Scene Understanding}.
\newblock \emph{Computational Visual Media}, 2025{\natexlab{b}}.

\bibitem[Yang et~al.(2022)Yang, Jiang, Sun, Schiele, and Jia]{yang2022unified}
Zetong Yang, Li Jiang, Yanan Sun, Bernt Schiele, and Jiaya Jia.
\newblock {A Unified Query-Based Paradigm for Point Cloud Understanding}.
\newblock In \emph{IEEE/CVF Conference on Computer Vision and Pattern Recognition (CVPR)}, 2022.

\bibitem[Yin et~al.(2021)Yin, Zhou, and Krahenbuhl]{yin2021center}
Tianwei Yin, Xingyi Zhou, and Philipp Krahenbuhl.
\newblock {Center-Based 3D Object Detection and Tracking}.
\newblock In \emph{IEEE/CVF Conference on Computer Vision and Pattern Recognition (CVPR)}, 2021.

\bibitem[Zeng et~al.(2025)Zeng, Dong, Zhou, Qiu, Dong, Luo, and Li]{zeng2025deepla}
Ziyin Zeng, Mingyue Dong, Jian Zhou, Huan Qiu, Zhen Dong, Man Luo, and Bijun Li.
\newblock {DeepLA-Net: Very Deep Local Aggregation Networks for Point Cloud Analysis}.
\newblock In \emph{IEEE/CVF Conference on Computer Vision and Pattern Recognition (CVPR)}, 2025.

\bibitem[Zhang et~al.(2022)Zhang, Wan, Shen, and Wu]{zhang2022patchformer}
Cheng Zhang, Haocheng Wan, Xinyi Shen, and Zizhao Wu.
\newblock {PatchFormer: An Efficient Point Transformer with Patch Attention}.
\newblock In \emph{IEEE/CVF Conference on Computer Vision and Pattern Recognition (CVPR)}, 2022.

\bibitem[Zhang and Singh(2014)]{zhang2014loam}
Ji Zhang and Sanjiv Singh.
\newblock {LOAM}: Lidar odometry and mapping in real-time.
\newblock In \emph{Robotics: Science and Systems}, 2014.

\bibitem[Zhao et~al.(2021)Zhao, Jiang, Jia, Torr, and Koltun]{zhao2021point}
Hengshuang Zhao, Li Jiang, Jiaya Jia, Philip~HS Torr, and Vladlen Koltun.
\newblock {Point Transformer}.
\newblock In \emph{International Conference on Computer Vision (ICCV)}, 2021.

\bibitem[Zheng et~al.(2020)Zheng, Zhang, Li, Tang, Gao, and Zhou]{zheng2020structured3d}
Jia Zheng, Junfei Zhang, Jing Li, Rui Tang, Shenghua Gao, and Zihan Zhou.
\newblock {Structured3D: A Large Photo-Realistic Dataset for Structured 3D Modeling}.
\newblock In \emph{European Conference on Computer Vision (ECCV)}, 2020.

\bibitem[Zhu et~al.(2021)Zhu, Zhou, Wang, Hong, Ma, Li, Li, and Lin]{zhu2021cylindrical}
Xinge Zhu, Hui Zhou, Tai Wang, Fangzhou Hong, Yuexin Ma, Wei Li, Hongsheng Li, and Dahua Lin.
\newblock {Cylindrical and Asymmetrical 3D Convolution Networks for LiDAR Segmentation}.
\newblock In \emph{IEEE/CVF Conference on Computer Vision and Pattern Recognition (CVPR)}, 2021.

\end{thebibliography}
}

\clearpage

\end{appendices}
\clearpage

\end{document}